\documentclass{article}

\usepackage[main,final]{neurips_2026}

\usepackage[utf8]{inputenc}
\usepackage[T1]{fontenc}

\usepackage{url}
\usepackage{booktabs}
\usepackage{array}
\usepackage{amsfonts}
\usepackage{amsmath}
\usepackage{amssymb}
\usepackage{amsthm}
\usepackage{microtype}
\usepackage{xcolor}
\usepackage{graphicx}
\usepackage{placeins}
\usepackage{float}
\usepackage{flafter}
\usepackage{algorithm}
\usepackage{algpseudocode}
\usepackage{xspace}
\usepackage[strings]{underscore}
\usepackage{hyperref}
\definecolor{citeblue}{HTML}{0072B2}
\definecolor{refvermillion}{HTML}{A64B00}
\definecolor{urlgreen}{HTML}{00785A}
\hypersetup{colorlinks=true, citecolor=citeblue, linkcolor=refvermillion, urlcolor=urlgreen}

\newcommand{\nAcctLargestScore}{12.99\xspace}

\newcommand{\nAcctWindows}{60,000\xspace}
\newcommand{\nAciFloor}{0.01\xspace}

\newcommand{\nAddCountMax}{200\xspace}
\newcommand{\nAddCutoffHalf}{2\xspace}
\newcommand{\nAddCutoffNineTenths}{10\xspace}
\newcommand{\nAddCutoffNineteenTwentieths}{20\xspace}
\newcommand{\nAddCutoffSixTenths}{3\xspace}
\newcommand{\nAddCutoffTenth}{2\xspace}
\newcommand{\nAddCutoffThreeQuarters}{4\xspace}

\newcommand{\nAddInductionPoints}{36\xspace}
\newcommand{\nAddLevelHalf}{0.50\xspace}
\newcommand{\nAddLevelNineTenths}{0.90\xspace}
\newcommand{\nAddLevelNineteenTwentieths}{0.95\xspace}
\newcommand{\nAddLevelSixTenths}{0.60\xspace}
\newcommand{\nAddLevelTenth}{0.10\xspace}
\newcommand{\nAddLevelThreeQuarters}{0.75\xspace}
\newcommand{\nAddLevels}{6\xspace}

\newcommand{\nAddObjElevenB}{1\xspace}
\newcommand{\nAddObjElevenBound}{2\xspace}
\newcommand{\nAddObjElevenFloor}{10\xspace}
\newcommand{\nAddObjElevenInitial}{0.9050\xspace}
\newcommand{\nAddObjElevenRange}{9.0966\xspace}
\newcommand{\nAddObjElevenReset}{0.9034\xspace}
\newcommand{\nAddObjElevenStateMax}{10.00\xspace}
\newcommand{\nAddRealizations}{20,000\xspace}
\newcommand{\nAddStateInside}{72\xspace}
\newcommand{\nAddStateRuns}{72\xspace}

\newcommand{\nAgaciEta}{0.5\xspace}
\newcommand{\nAgaciSignedAgree}{386\xspace}

\newcommand{\nAgaciWidthWeight}{0.001\xspace}

\newcommand{\nAlpha}{0.1\xspace}

\newcommand{\nAnnErcotCapAddress}{\url{https://ftp.puc.texas.gov/public/puct-info/agency/rulesnlaws/subrules/electric/25.505/52631adt.pdf}\xspace}
\newcommand{\nAnnErcotCapFirstAffected}{2022-01-01\xspace}
\newcommand{\nAnnErcotCapIssuer}{Public Utility Commission of Texas\xspace}
\newcommand{\nAnnErcotCapOwnDate}{December 2, 2021\xspace}
\newcommand{\nAnnErcotCapOwnDateKind}{approval at the commission's open meeting, as the document's heading states\xspace}
\newcommand{\nAnnErcotCapPassage}{The system-wide offer cap is set to the LCAP for the remainder of 2021, but it will revert to the HCAP on January 1, 2022.\xspace}
\newcommand{\nAnnErcotCapTitle}{Project No. 52631, proposal for adoption for amendments to 16 TAC 25.505 as approved at the December 2, 2021 open meeting\xspace}
\newcommand{\nAnnErcotRtcbAddress}{\url{https://www.ercot.com/services/comm/mkt_notices/M-F110525-01}\xspace}
\newcommand{\nAnnErcotRtcbFirstAffected}{2025-12-05\xspace}
\newcommand{\nAnnErcotRtcbIssuer}{ERCOT\xspace}
\newcommand{\nAnnErcotRtcbOwnDate}{November 5, 2025\xspace}
\newcommand{\nAnnErcotRtcbOwnDateKind}{the notice date the notice states\xspace}
\newcommand{\nAnnErcotRtcbPassage}{This implementation will occur during the late evening hours of December 4, 2025, leading up to and through the midnight cutover going into December 5, 2025.\xspace}
\newcommand{\nAnnErcotRtcbTitle}{Market notice M-F110525-01, the go-live plan of real-time co-optimization plus batteries for operating day December 5, 2025\xspace}
\newcommand{\nAnnFomcAddressFive}{\url{https://web.archive.org/web/20241202030119id_/https://www.federalreserve.gov/monetarypolicy/fomccalendars.htm}\xspace}
\newcommand{\nAnnFomcAddressFour}{\url{https://web.archive.org/web/20231201060312id_/https://www.federalreserve.gov/monetarypolicy/fomccalendars.htm}\xspace}
\newcommand{\nAnnFomcAddressOne}{\url{https://web.archive.org/web/20201205052640id_/https://www.federalreserve.gov/monetarypolicy/fomccalendars.htm}\xspace}
\newcommand{\nAnnFomcAddressThree}{\url{https://web.archive.org/web/20221201063808id_/https://www.federalreserve.gov/monetarypolicy/fomccalendars.htm}\xspace}
\newcommand{\nAnnFomcAddressTwo}{\url{https://web.archive.org/web/20211201182151id_/https://www.federalreserve.gov/monetarypolicy/fomccalendars.htm}\xspace}
\newcommand{\nAnnFomcCaptureFive}{December 2, 2024\xspace}
\newcommand{\nAnnFomcCaptureFour}{December 1, 2023\xspace}
\newcommand{\nAnnFomcCaptureOne}{December 5, 2020\xspace}
\newcommand{\nAnnFomcCaptureThree}{December 1, 2022\xspace}
\newcommand{\nAnnFomcCaptureTwo}{December 1, 2021\xspace}

\newcommand{\nAnnFomcEvaluatedAccepted}{38\xspace}
\newcommand{\nAnnFomcFirstDecisionFive}{2025-01-29\xspace}
\newcommand{\nAnnFomcFirstDecisionFour}{2024-01-31\xspace}
\newcommand{\nAnnFomcFirstDecisionOne}{2021-01-27\xspace}
\newcommand{\nAnnFomcFirstDecisionThree}{2023-02-01\xspace}
\newcommand{\nAnnFomcFirstDecisionTwo}{2022-01-26\xspace}
\newcommand{\nAnnFomcLastUpdateFive}{November 26, 2024\xspace}
\newcommand{\nAnnFomcLastUpdateFour}{November 21, 2023\xspace}
\newcommand{\nAnnFomcLastUpdateOne}{November 25, 2020\xspace}
\newcommand{\nAnnFomcLastUpdateThree}{November 23, 2022\xspace}
\newcommand{\nAnnFomcLastUpdateTwo}{November 24, 2021\xspace}
\newcommand{\nAnnFomcPassageFive}{2025 FOMC Meetings January 28-29 March 18-19* May 6-7 June 17-18* July 29-30 September 16-17* October 28-29 December 9-10*\xspace}
\newcommand{\nAnnFomcPassageFour}{2024 FOMC Meetings January 30-31 March 19-20* Apr/May 30-1 June 11-12* July 30-31 September 17-18* November 6-7 December 17-18*\xspace}
\newcommand{\nAnnFomcPassageOne}{2021 FOMC Meetings January 26-27 March 16-17* April 27-28 June 15-16* July 27-28 September 21-22* November 2-3 December 14-15*\xspace}
\newcommand{\nAnnFomcPassageThree}{2023 FOMC Meetings Jan/Feb 31-1 March 21-22* May 2-3 June 13-14* July 25-26 September 19-20* Oct/Nov 31-1 December 12-13*\xspace}
\newcommand{\nAnnFomcPassageTwo}{2022 FOMC Meetings January 25-26 March 15-16* May 3-4 June 14-15* July 26-27 September 20-21* November 1-2 December 13-14*\xspace}

\newcommand{\nAnnFomcSingleAddress}{\url{https://web.archive.org/web/20211201182151id_/https://www.federalreserve.gov/monetarypolicy/fomccalendars.htm}\xspace}
\newcommand{\nAnnFomcSingleFirstAffected}{2022-03-16\xspace}
\newcommand{\nAnnFomcSingleIssuer}{Board of Governors of the Federal Reserve System\xspace}
\newcommand{\nAnnFomcSingleOwnDate}{November 24, 2021\xspace}
\newcommand{\nAnnFomcSingleOwnDateKind}{the page's last update date the page states, beside the capture's timestamp\xspace}
\newcommand{\nAnnFomcSinglePassage}{2022 FOMC Meetings January 25-26 March 15-16* May 3-4 June 14-15* July 26-27 September 20-21* November 1-2 December 13-14*\xspace}
\newcommand{\nAnnFomcSingleTitle}{Meeting calendars and information, the 2022 schedule\xspace}
\newcommand{\nAnnFomcTentative}{Each meeting date is tentative until confirmed at the meeting immediately preceding it\xspace}
\newcommand{\nAnnFomcYearFive}{2025\xspace}
\newcommand{\nAnnFomcYearFour}{2024\xspace}
\newcommand{\nAnnFomcYearOne}{2021\xspace}
\newcommand{\nAnnFomcYearThree}{2023\xspace}
\newcommand{\nAnnFomcYearTwo}{2022\xspace}

\newcommand{\nAnnLibAccepted}{9\xspace}

\newcommand{\nAnnNemApcAddress}{\url{https://www.aemc.gov.au/sites/default/files/2022-11/Amending\%20the\%20administered\%20price\%20cap\%20-\%20Final\%20Determination.pdf}\xspace}
\newcommand{\nAnnNemApcFirstAffected}{2022-12-01\xspace}
\newcommand{\nAnnNemApcIssuer}{Australian Energy Market Commission\xspace}
\newcommand{\nAnnNemApcOwnDate}{November 17, 2022\xspace}
\newcommand{\nAnnNemApcOwnDateKind}{the determination's date its cover states\xspace}
\newcommand{\nAnnNemApcPassage}{The final rule commences on 1 December 2022, two weeks after the publication of this final determination.\xspace}
\newcommand{\nAnnNemApcTitle}{Rule determination, National Electricity Amendment, Amending the administered price cap, Rule 2022\xspace}
\newcommand{\nAnnNemFiveAddress}{\url{https://www.aemc.gov.au/news-centre/media-releases/commission-confirms-five-minute-settlement-commence-1-october-2021}\xspace}
\newcommand{\nAnnNemFiveFirstAffected}{2021-10-01\xspace}
\newcommand{\nAnnNemFiveIssuer}{Australian Energy Market Commission\xspace}
\newcommand{\nAnnNemFiveOwnDate}{September 9, 2021\xspace}
\newcommand{\nAnnNemFiveOwnDateKind}{the release date the release states\xspace}
\newcommand{\nAnnNemFivePassage}{Commission confirms five-minute settlement to commence on 1 October 2021\xspace}
\newcommand{\nAnnNemFiveTitle}{Media release, Commission confirms five-minute settlement to commence on 1 October 2021\xspace}
\newcommand{\nAnnNemMpcAddress}{\url{https://www.aemc.gov.au/news-centre/media-releases/aemc-updates-market-price-cap-2025-26}\xspace}
\newcommand{\nAnnNemMpcFirstAffected}{2025-07-01\xspace}
\newcommand{\nAnnNemMpcIssuer}{Australian Energy Market Commission\xspace}
\newcommand{\nAnnNemMpcOwnDate}{February 27, 2025\xspace}
\newcommand{\nAnnNemMpcOwnDateKind}{the release date the release states\xspace}
\newcommand{\nAnnNemMpcPassage}{which will progressively increase from 1 July 2025 through to 1 July 2027\xspace}
\newcommand{\nAnnNemMpcTitle}{Media release, AEMC updates market price cap for 2025-26\xspace}
\newcommand{\nAnnOrdcOneAddress}{\url{https://www.ercot.com/services/comm/mkt_notices/W-A021319-01}\xspace}
\newcommand{\nAnnOrdcOneFirstAffected}{2019-03-01\xspace}
\newcommand{\nAnnOrdcOneIssuer}{ERCOT\xspace}
\newcommand{\nAnnOrdcOneOwnDate}{February 13, 2019\xspace}
\newcommand{\nAnnOrdcOneOwnDateKind}{the notice date the notice states\xspace}
\newcommand{\nAnnOrdcOnePassage}{On March 1, 2019, ERCOT will implement the changes described in OBDRR011.\xspace}
\newcommand{\nAnnOrdcOneTitle}{Market notice W-A021319-01, implementation details for OBDRR011\xspace}
\newcommand{\nAnnOrdcRevision}{OBDRR011\xspace}
\newcommand{\nAnnOrdcTwoAddress}{\url{https://www.ercot.com/services/comm/mkt_notices/archives/4442}\xspace}
\newcommand{\nAnnOrdcTwoFirstAffected}{2020-03-01\xspace}
\newcommand{\nAnnOrdcTwoIssuer}{ERCOT\xspace}
\newcommand{\nAnnOrdcTwoOwnDate}{February 27, 2020\xspace}
\newcommand{\nAnnOrdcTwoOwnDateKind}{the notice date the notice states\xspace}
\newcommand{\nAnnOrdcTwoPassage}{Specifically, ERCOT will implement the second of two rightward shifts of 0.25 standard deviations to the Loss of Load Probability\xspace}
\newcommand{\nAnnOrdcTwoTitle}{Market notice W-A021319-02, implementation details for OBDRR011\xspace}
\newcommand{\nAnnPjmRpfAddress}{\url{https://www.pjm.com/-/media/committees-groups/committees/mic/2022/20220713/item-09a---reserve-price-formation-manual-revisions.ashx}\xspace}
\newcommand{\nAnnPjmRpfFirstAffected}{2022-10-01\xspace}
\newcommand{\nAnnPjmRpfIssuer}{PJM Interconnection\xspace}
\newcommand{\nAnnPjmRpfOwnDate}{July 13, 2022\xspace}
\newcommand{\nAnnPjmRpfOwnDateKind}{the committee meeting's date the document's first page states\xspace}
\newcommand{\nAnnPjmRpfPassage}{Go-live for Reserve Price Formation changes is October 1\xspace}
\newcommand{\nAnnPjmRpfTitle}{Reserve Price Formation Manual Revisions, Market Implementation Committee, July 13, 2022\xspace}
\newcommand{\nAnnReadDate}{September 29, 2026\xspace}
\newcommand{\nArchBandBothShiftFormAFreqFiftyMacDiff}{-0.0626\xspace}
\newcommand{\nArchBandBothShiftFormAFreqFiftyMacHi}{-0.0609\xspace}
\newcommand{\nArchBandBothShiftFormAFreqFiftyMacLo}{-0.0645\xspace}
\newcommand{\nArchBandBothShiftFormAFreqFiftyWorkersDiff}{-0.0622\xspace}
\newcommand{\nArchBandBothShiftFormAFreqFiftyWorkersHi}{-0.0604\xspace}
\newcommand{\nArchBandBothShiftFormAFreqFiftyWorkersLo}{-0.0640\xspace}
\newcommand{\nArchBandBothShiftResetPtFiftyMacDiff}{-0.035\xspace}
\newcommand{\nArchBandBothShiftResetPtFiftyMacHi}{-0.031\xspace}
\newcommand{\nArchBandBothShiftResetPtFiftyMacLo}{-0.040\xspace}
\newcommand{\nArchBandBothShiftResetPtFiftyWorkersDiff}{-0.034\xspace}
\newcommand{\nArchBandBothShiftResetPtFiftyWorkersHi}{-0.030\xspace}
\newcommand{\nArchBandBothShiftResetPtFiftyWorkersLo}{-0.039\xspace}
\newcommand{\nArchBandBothShiftResetPtTenMacDiff}{-0.023\xspace}
\newcommand{\nArchBandBothShiftResetPtTenMacHi}{-0.018\xspace}
\newcommand{\nArchBandBothShiftResetPtTenMacLo}{-0.029\xspace}
\newcommand{\nArchBandBothShiftResetPtTenWorkersDiff}{-0.023\xspace}
\newcommand{\nArchBandBothShiftResetPtTenWorkersHi}{-0.017\xspace}
\newcommand{\nArchBandBothShiftResetPtTenWorkersLo}{-0.028\xspace}
\newcommand{\nArchBandBothShiftTwinPtFiftyMacDiff}{-0.248\xspace}
\newcommand{\nArchBandBothShiftTwinPtFiftyMacHi}{-0.243\xspace}
\newcommand{\nArchBandBothShiftTwinPtFiftyMacLo}{-0.254\xspace}
\newcommand{\nArchBandBothShiftTwinPtFiftyWorkersDiff}{-0.249\xspace}
\newcommand{\nArchBandBothShiftTwinPtFiftyWorkersHi}{-0.244\xspace}
\newcommand{\nArchBandBothShiftTwinPtFiftyWorkersLo}{-0.254\xspace}
\newcommand{\nArchBandBothShiftTwinPtTenMacDiff}{-0.209\xspace}
\newcommand{\nArchBandBothShiftTwinPtTenMacHi}{-0.204\xspace}
\newcommand{\nArchBandBothShiftTwinPtTenMacLo}{-0.214\xspace}
\newcommand{\nArchBandBothShiftTwinPtTenWorkersDiff}{-0.209\xspace}
\newcommand{\nArchBandBothShiftTwinPtTenWorkersHi}{-0.204\xspace}
\newcommand{\nArchBandBothShiftTwinPtTenWorkersLo}{-0.214\xspace}
\newcommand{\nArchBandLargeMeanFormAFreqFiftyMacDiff}{-0.0632\xspace}
\newcommand{\nArchBandLargeMeanFormAFreqFiftyMacHi}{-0.0614\xspace}
\newcommand{\nArchBandLargeMeanFormAFreqFiftyMacLo}{-0.0651\xspace}
\newcommand{\nArchBandLargeMeanFormAFreqFiftyWorkersDiff}{-0.0616\xspace}
\newcommand{\nArchBandLargeMeanFormAFreqFiftyWorkersHi}{-0.0597\xspace}
\newcommand{\nArchBandLargeMeanFormAFreqFiftyWorkersLo}{-0.0634\xspace}
\newcommand{\nArchBandLargeMeanResetPtFiftyMacDiff}{-0.043\xspace}
\newcommand{\nArchBandLargeMeanResetPtFiftyMacHi}{-0.037\xspace}
\newcommand{\nArchBandLargeMeanResetPtFiftyMacLo}{-0.049\xspace}
\newcommand{\nArchBandLargeMeanResetPtFiftyWorkersDiff}{-0.045\xspace}
\newcommand{\nArchBandLargeMeanResetPtFiftyWorkersHi}{-0.039\xspace}
\newcommand{\nArchBandLargeMeanResetPtFiftyWorkersLo}{-0.052\xspace}
\newcommand{\nArchBandLargeMeanResetPtTenMacDiff}{-0.010\xspace}
\newcommand{\nArchBandLargeMeanResetPtTenMacHi}{-0.003\xspace}
\newcommand{\nArchBandLargeMeanResetPtTenMacLo}{-0.017\xspace}
\newcommand{\nArchBandLargeMeanResetPtTenWorkersDiff}{-0.010\xspace}
\newcommand{\nArchBandLargeMeanResetPtTenWorkersHi}{-0.003\xspace}
\newcommand{\nArchBandLargeMeanResetPtTenWorkersLo}{-0.017\xspace}
\newcommand{\nArchBandLargeMeanTwinPtFiftyMacDiff}{-0.659\xspace}
\newcommand{\nArchBandLargeMeanTwinPtFiftyMacHi}{-0.651\xspace}
\newcommand{\nArchBandLargeMeanTwinPtFiftyMacLo}{-0.666\xspace}
\newcommand{\nArchBandLargeMeanTwinPtFiftyWorkersDiff}{-0.657\xspace}
\newcommand{\nArchBandLargeMeanTwinPtFiftyWorkersHi}{-0.649\xspace}
\newcommand{\nArchBandLargeMeanTwinPtFiftyWorkersLo}{-0.664\xspace}
\newcommand{\nArchBandLargeMeanTwinPtTenMacDiff}{-0.696\xspace}
\newcommand{\nArchBandLargeMeanTwinPtTenMacHi}{-0.689\xspace}
\newcommand{\nArchBandLargeMeanTwinPtTenMacLo}{-0.703\xspace}
\newcommand{\nArchBandLargeMeanTwinPtTenWorkersDiff}{-0.696\xspace}
\newcommand{\nArchBandLargeMeanTwinPtTenWorkersHi}{-0.689\xspace}
\newcommand{\nArchBandLargeMeanTwinPtTenWorkersLo}{-0.703\xspace}
\newcommand{\nArchBandMeanShiftFormAFreqFiftyMacDiff}{-0.0484\xspace}
\newcommand{\nArchBandMeanShiftFormAFreqFiftyMacHi}{-0.0467\xspace}
\newcommand{\nArchBandMeanShiftFormAFreqFiftyMacLo}{-0.0502\xspace}
\newcommand{\nArchBandMeanShiftFormAFreqFiftyWorkersDiff}{-0.0485\xspace}
\newcommand{\nArchBandMeanShiftFormAFreqFiftyWorkersHi}{-0.0467\xspace}
\newcommand{\nArchBandMeanShiftFormAFreqFiftyWorkersLo}{-0.0503\xspace}
\newcommand{\nArchBandMeanShiftResetPtFiftyMacDiff}{-0.026\xspace}
\newcommand{\nArchBandMeanShiftResetPtFiftyMacHi}{-0.022\xspace}
\newcommand{\nArchBandMeanShiftResetPtFiftyMacLo}{-0.029\xspace}
\newcommand{\nArchBandMeanShiftResetPtFiftyWorkersDiff}{-0.026\xspace}
\newcommand{\nArchBandMeanShiftResetPtFiftyWorkersHi}{-0.023\xspace}
\newcommand{\nArchBandMeanShiftResetPtFiftyWorkersLo}{-0.031\xspace}
\newcommand{\nArchBandMeanShiftResetPtTenMacDiff}{-0.013\xspace}
\newcommand{\nArchBandMeanShiftResetPtTenMacHi}{-0.008\xspace}
\newcommand{\nArchBandMeanShiftResetPtTenMacLo}{-0.018\xspace}
\newcommand{\nArchBandMeanShiftResetPtTenWorkersDiff}{-0.013\xspace}
\newcommand{\nArchBandMeanShiftResetPtTenWorkersHi}{-0.008\xspace}
\newcommand{\nArchBandMeanShiftResetPtTenWorkersLo}{-0.018\xspace}
\newcommand{\nArchBandMeanShiftTwinPtFiftyMacDiff}{-0.251\xspace}
\newcommand{\nArchBandMeanShiftTwinPtFiftyMacHi}{-0.245\xspace}
\newcommand{\nArchBandMeanShiftTwinPtFiftyMacLo}{-0.257\xspace}
\newcommand{\nArchBandMeanShiftTwinPtFiftyWorkersDiff}{-0.250\xspace}
\newcommand{\nArchBandMeanShiftTwinPtFiftyWorkersHi}{-0.244\xspace}
\newcommand{\nArchBandMeanShiftTwinPtFiftyWorkersLo}{-0.256\xspace}
\newcommand{\nArchBandMeanShiftTwinPtTenMacDiff}{-0.191\xspace}
\newcommand{\nArchBandMeanShiftTwinPtTenMacHi}{-0.186\xspace}
\newcommand{\nArchBandMeanShiftTwinPtTenMacLo}{-0.197\xspace}
\newcommand{\nArchBandMeanShiftTwinPtTenWorkersDiff}{-0.192\xspace}
\newcommand{\nArchBandMeanShiftTwinPtTenWorkersHi}{-0.186\xspace}
\newcommand{\nArchBandMeanShiftTwinPtTenWorkersLo}{-0.197\xspace}
\newcommand{\nArchBandNoBreakFormAFreqFiftyMacDiff}{-0.0090\xspace}
\newcommand{\nArchBandNoBreakFormAFreqFiftyMacHi}{-0.0078\xspace}
\newcommand{\nArchBandNoBreakFormAFreqFiftyMacLo}{-0.0102\xspace}
\newcommand{\nArchBandNoBreakFormAFreqFiftyWorkersDiff}{-0.0092\xspace}
\newcommand{\nArchBandNoBreakFormAFreqFiftyWorkersHi}{-0.0080\xspace}
\newcommand{\nArchBandNoBreakFormAFreqFiftyWorkersLo}{-0.0104\xspace}
\newcommand{\nArchBandNoBreakResetPtFiftyMacDiff}{-0.016\xspace}
\newcommand{\nArchBandNoBreakResetPtFiftyMacHi}{-0.014\xspace}
\newcommand{\nArchBandNoBreakResetPtFiftyMacLo}{-0.019\xspace}
\newcommand{\nArchBandNoBreakResetPtFiftyWorkersDiff}{-0.016\xspace}
\newcommand{\nArchBandNoBreakResetPtFiftyWorkersHi}{-0.013\xspace}
\newcommand{\nArchBandNoBreakResetPtFiftyWorkersLo}{-0.019\xspace}
\newcommand{\nArchBandNoBreakResetPtTenMacDiff}{0.038\xspace}
\newcommand{\nArchBandNoBreakResetPtTenMacHi}{0.044\xspace}
\newcommand{\nArchBandNoBreakResetPtTenMacLo}{0.033\xspace}
\newcommand{\nArchBandNoBreakResetPtTenWorkersDiff}{0.039\xspace}
\newcommand{\nArchBandNoBreakResetPtTenWorkersHi}{0.044\xspace}
\newcommand{\nArchBandNoBreakResetPtTenWorkersLo}{0.033\xspace}
\newcommand{\nArchBandNoBreakTwinPtFiftyMacDiff}{0.028\xspace}
\newcommand{\nArchBandNoBreakTwinPtFiftyMacHi}{0.031\xspace}
\newcommand{\nArchBandNoBreakTwinPtFiftyMacLo}{0.026\xspace}
\newcommand{\nArchBandNoBreakTwinPtFiftyWorkersDiff}{0.027\xspace}
\newcommand{\nArchBandNoBreakTwinPtFiftyWorkersHi}{0.030\xspace}
\newcommand{\nArchBandNoBreakTwinPtFiftyWorkersLo}{0.025\xspace}
\newcommand{\nArchBandNoBreakTwinPtTenMacDiff}{0.031\xspace}
\newcommand{\nArchBandNoBreakTwinPtTenMacHi}{0.034\xspace}
\newcommand{\nArchBandNoBreakTwinPtTenMacLo}{0.028\xspace}
\newcommand{\nArchBandNoBreakTwinPtTenWorkersDiff}{0.031\xspace}
\newcommand{\nArchBandNoBreakTwinPtTenWorkersHi}{0.034\xspace}
\newcommand{\nArchBandNoBreakTwinPtTenWorkersLo}{0.028\xspace}
\newcommand{\nArchBandVarShiftFormAFreqFiftyMacDiff}{-0.0422\xspace}
\newcommand{\nArchBandVarShiftFormAFreqFiftyMacHi}{-0.0404\xspace}
\newcommand{\nArchBandVarShiftFormAFreqFiftyMacLo}{-0.0440\xspace}
\newcommand{\nArchBandVarShiftFormAFreqFiftyWorkersDiff}{-0.0417\xspace}
\newcommand{\nArchBandVarShiftFormAFreqFiftyWorkersHi}{-0.0400\xspace}
\newcommand{\nArchBandVarShiftFormAFreqFiftyWorkersLo}{-0.0435\xspace}
\newcommand{\nArchBandVarShiftResetPtFiftyMacDiff}{-0.017\xspace}
\newcommand{\nArchBandVarShiftResetPtFiftyMacHi}{-0.015\xspace}
\newcommand{\nArchBandVarShiftResetPtFiftyMacLo}{-0.020\xspace}
\newcommand{\nArchBandVarShiftResetPtFiftyWorkersDiff}{-0.018\xspace}
\newcommand{\nArchBandVarShiftResetPtFiftyWorkersHi}{-0.015\xspace}
\newcommand{\nArchBandVarShiftResetPtFiftyWorkersLo}{-0.021\xspace}
\newcommand{\nArchBandVarShiftResetPtTenMacDiff}{-0.001\xspace}
\newcommand{\nArchBandVarShiftResetPtTenMacHi}{0.003\xspace}
\newcommand{\nArchBandVarShiftResetPtTenMacLo}{-0.005\xspace}
\newcommand{\nArchBandVarShiftResetPtTenWorkersDiff}{-0.001\xspace}
\newcommand{\nArchBandVarShiftResetPtTenWorkersHi}{0.003\xspace}
\newcommand{\nArchBandVarShiftResetPtTenWorkersLo}{-0.005\xspace}
\newcommand{\nArchBandVarShiftTwinPtFiftyMacDiff}{-0.150\xspace}
\newcommand{\nArchBandVarShiftTwinPtFiftyMacHi}{-0.146\xspace}
\newcommand{\nArchBandVarShiftTwinPtFiftyMacLo}{-0.154\xspace}
\newcommand{\nArchBandVarShiftTwinPtFiftyWorkersDiff}{-0.149\xspace}
\newcommand{\nArchBandVarShiftTwinPtFiftyWorkersHi}{-0.146\xspace}
\newcommand{\nArchBandVarShiftTwinPtFiftyWorkersLo}{-0.153\xspace}
\newcommand{\nArchBandVarShiftTwinPtTenMacDiff}{-0.101\xspace}
\newcommand{\nArchBandVarShiftTwinPtTenMacHi}{-0.097\xspace}
\newcommand{\nArchBandVarShiftTwinPtTenMacLo}{-0.105\xspace}
\newcommand{\nArchBandVarShiftTwinPtTenWorkersDiff}{-0.101\xspace}
\newcommand{\nArchBandVarShiftTwinPtTenWorkersHi}{-0.098\xspace}
\newcommand{\nArchBandVarShiftTwinPtTenWorkersLo}{-0.105\xspace}

\newcommand{\nArchBothShiftFreqFiftyDiff}{0.0005\xspace}
\newcommand{\nArchBothShiftFreqFiftyDiffering}{296\xspace}
\newcommand{\nArchBothShiftFreqFiftyHi}{0.0013\xspace}

\newcommand{\nArchBothShiftFreqFiftyLo}{-0.0004\xspace}

\newcommand{\nArchBothShiftFreqFiveThousandDiff}{0.0000\xspace}
\newcommand{\nArchBothShiftFreqFiveThousandDiffering}{909\xspace}
\newcommand{\nArchBothShiftFreqFiveThousandHi}{0.0001\xspace}

\newcommand{\nArchBothShiftFreqFiveThousandLo}{-0.0001\xspace}

\newcommand{\nArchBothShiftFreqTenDiff}{-0.0005\xspace}
\newcommand{\nArchBothShiftFreqTenDiffering}{66\xspace}
\newcommand{\nArchBothShiftFreqTenHi}{0.0012\xspace}

\newcommand{\nArchBothShiftFreqTenLo}{-0.0023\xspace}

\newcommand{\nArchBothShiftFreqThousandDiff}{0.0001\xspace}
\newcommand{\nArchBothShiftFreqThousandDiffering}{826\xspace}
\newcommand{\nArchBothShiftFreqThousandHi}{0.0002\xspace}

\newcommand{\nArchBothShiftFreqThousandLo}{-0.0001\xspace}

\newcommand{\nArchBothShiftFreqTwentyFiveDiff}{0.0006\xspace}
\newcommand{\nArchBothShiftFreqTwentyFiveDiffering}{188\xspace}
\newcommand{\nArchBothShiftFreqTwentyFiveHi}{0.0018\xspace}

\newcommand{\nArchBothShiftFreqTwentyFiveLo}{-0.0006\xspace}

\newcommand{\nArchBothShiftFreqTwoHundredDiff}{0.0000\xspace}
\newcommand{\nArchBothShiftFreqTwoHundredDiffering}{589\xspace}
\newcommand{\nArchBothShiftFreqTwoHundredHi}{0.0004\xspace}

\newcommand{\nArchBothShiftFreqTwoHundredLo}{-0.0004\xspace}

\newcommand{\nArchBothShiftPtFiftyDiff}{0.001\xspace}
\newcommand{\nArchBothShiftPtFiftyDiffering}{995\xspace}
\newcommand{\nArchBothShiftPtFiftyHi}{0.003\xspace}

\newcommand{\nArchBothShiftPtFiftyLo}{-0.001\xspace}
\newcommand{\nArchBothShiftPtFiftyMac}{0.073\xspace}

\newcommand{\nArchBothShiftPtFiftyWorkers}{0.072\xspace}
\newcommand{\nArchBothShiftPtFiveThousandDiff}{0.000\xspace}
\newcommand{\nArchBothShiftPtFiveThousandDiffering}{1,000\xspace}
\newcommand{\nArchBothShiftPtFiveThousandHi}{0.000\xspace}

\newcommand{\nArchBothShiftPtFiveThousandLo}{-0.001\xspace}
\newcommand{\nArchBothShiftPtFiveThousandMac}{0.042\xspace}

\newcommand{\nArchBothShiftPtFiveThousandWorkers}{0.042\xspace}
\newcommand{\nArchBothShiftPtTenDiff}{0.000\xspace}
\newcommand{\nArchBothShiftPtTenDiffering}{994\xspace}
\newcommand{\nArchBothShiftPtTenHi}{0.002\xspace}

\newcommand{\nArchBothShiftPtTenLo}{-0.001\xspace}
\newcommand{\nArchBothShiftPtTenMac}{0.122\xspace}

\newcommand{\nArchBothShiftPtTenWorkers}{0.121\xspace}
\newcommand{\nArchBothShiftPtThousandDiff}{0.001\xspace}
\newcommand{\nArchBothShiftPtThousandDiffering}{1,000\xspace}
\newcommand{\nArchBothShiftPtThousandHi}{0.001\xspace}

\newcommand{\nArchBothShiftPtThousandLo}{0.000\xspace}
\newcommand{\nArchBothShiftPtThousandMac}{0.048\xspace}

\newcommand{\nArchBothShiftPtThousandWorkers}{0.048\xspace}
\newcommand{\nArchBothShiftPtTwentyFiveDiff}{-0.001\xspace}
\newcommand{\nArchBothShiftPtTwentyFiveDiffering}{994\xspace}
\newcommand{\nArchBothShiftPtTwentyFiveHi}{0.001\xspace}

\newcommand{\nArchBothShiftPtTwentyFiveLo}{-0.003\xspace}
\newcommand{\nArchBothShiftPtTwentyFiveMac}{0.090\xspace}

\newcommand{\nArchBothShiftPtTwentyFiveWorkers}{0.091\xspace}
\newcommand{\nArchBothShiftPtTwoHundredDiff}{0.000\xspace}
\newcommand{\nArchBothShiftPtTwoHundredDiffering}{1,000\xspace}
\newcommand{\nArchBothShiftPtTwoHundredHi}{0.000\xspace}

\newcommand{\nArchBothShiftPtTwoHundredLo}{-0.001\xspace}
\newcommand{\nArchBothShiftPtTwoHundredMac}{0.050\xspace}

\newcommand{\nArchBothShiftPtTwoHundredWorkers}{0.050\xspace}

\newcommand{\nArchControlReps}{125\xspace}
\newcommand{\nArchIsaNew}{AVX-512\xspace}
\newcommand{\nArchIsaOld}{AVX2\xspace}

\newcommand{\nArchLargeMeanFreqFiftyDiff}{0.0017\xspace}
\newcommand{\nArchLargeMeanFreqFiftyDiffering}{525\xspace}
\newcommand{\nArchLargeMeanFreqFiftyHi}{0.0029\xspace}

\newcommand{\nArchLargeMeanFreqFiftyLo}{0.0004\xspace}

\newcommand{\nArchLargeMeanFreqFiveThousandDiff}{0.0001\xspace}
\newcommand{\nArchLargeMeanFreqFiveThousandDiffering}{942\xspace}
\newcommand{\nArchLargeMeanFreqFiveThousandHi}{0.0002\xspace}

\newcommand{\nArchLargeMeanFreqFiveThousandLo}{0.0000\xspace}

\newcommand{\nArchLargeMeanFreqTenDiff}{-0.0001\xspace}
\newcommand{\nArchLargeMeanFreqTenDiffering}{113\xspace}
\newcommand{\nArchLargeMeanFreqTenHi}{0.0022\xspace}

\newcommand{\nArchLargeMeanFreqTenLo}{-0.0024\xspace}

\newcommand{\nArchLargeMeanFreqThousandDiff}{0.0002\xspace}
\newcommand{\nArchLargeMeanFreqThousandDiffering}{861\xspace}
\newcommand{\nArchLargeMeanFreqThousandHi}{0.0004\xspace}

\newcommand{\nArchLargeMeanFreqThousandLo}{0.0000\xspace}

\newcommand{\nArchLargeMeanFreqTwentyFiveDiff}{0.0017\xspace}
\newcommand{\nArchLargeMeanFreqTwentyFiveDiffering}{325\xspace}
\newcommand{\nArchLargeMeanFreqTwentyFiveHi}{0.0034\xspace}

\newcommand{\nArchLargeMeanFreqTwentyFiveLo}{0.0000\xspace}

\newcommand{\nArchLargeMeanFreqTwoHundredDiff}{0.0007\xspace}
\newcommand{\nArchLargeMeanFreqTwoHundredDiffering}{763\xspace}
\newcommand{\nArchLargeMeanFreqTwoHundredHi}{0.0013\xspace}

\newcommand{\nArchLargeMeanFreqTwoHundredLo}{0.0001\xspace}

\newcommand{\nArchLargeMeanFreqTwoHundredPlaceMac}{4\xspace}
\newcommand{\nArchLargeMeanFreqTwoHundredPlaceWorkers}{3\xspace}

\newcommand{\nArchLargeMeanPtFiftyDiff}{-0.002\xspace}
\newcommand{\nArchLargeMeanPtFiftyDiffering}{1,000\xspace}
\newcommand{\nArchLargeMeanPtFiftyHi}{0.001\xspace}

\newcommand{\nArchLargeMeanPtFiftyLo}{-0.005\xspace}
\newcommand{\nArchLargeMeanPtFiftyMac}{0.078\xspace}

\newcommand{\nArchLargeMeanPtFiftyWorkers}{0.080\xspace}
\newcommand{\nArchLargeMeanPtFiveThousandDiff}{0.001\xspace}
\newcommand{\nArchLargeMeanPtFiveThousandDiffering}{1,000\xspace}
\newcommand{\nArchLargeMeanPtFiveThousandHi}{0.002\xspace}

\newcommand{\nArchLargeMeanPtFiveThousandLo}{0.000\xspace}
\newcommand{\nArchLargeMeanPtFiveThousandMac}{0.047\xspace}

\newcommand{\nArchLargeMeanPtFiveThousandWorkers}{0.047\xspace}
\newcommand{\nArchLargeMeanPtTenDiff}{0.000\xspace}
\newcommand{\nArchLargeMeanPtTenDiffering}{1,000\xspace}
\newcommand{\nArchLargeMeanPtTenHi}{0.003\xspace}

\newcommand{\nArchLargeMeanPtTenLo}{-0.004\xspace}
\newcommand{\nArchLargeMeanPtTenMac}{0.105\xspace}

\newcommand{\nArchLargeMeanPtTenWorkers}{0.105\xspace}
\newcommand{\nArchLargeMeanPtThousandDiff}{0.001\xspace}
\newcommand{\nArchLargeMeanPtThousandDiffering}{1,000\xspace}
\newcommand{\nArchLargeMeanPtThousandHi}{0.002\xspace}

\newcommand{\nArchLargeMeanPtThousandLo}{0.000\xspace}
\newcommand{\nArchLargeMeanPtThousandMac}{0.056\xspace}

\newcommand{\nArchLargeMeanPtThousandWorkers}{0.055\xspace}
\newcommand{\nArchLargeMeanPtTwentyFiveDiff}{0.000\xspace}
\newcommand{\nArchLargeMeanPtTwentyFiveDiffering}{1,000\xspace}
\newcommand{\nArchLargeMeanPtTwentyFiveHi}{0.005\xspace}

\newcommand{\nArchLargeMeanPtTwentyFiveLo}{-0.004\xspace}
\newcommand{\nArchLargeMeanPtTwentyFiveMac}{0.105\xspace}

\newcommand{\nArchLargeMeanPtTwentyFiveWorkers}{0.105\xspace}
\newcommand{\nArchLargeMeanPtTwoHundredDiff}{-0.001\xspace}
\newcommand{\nArchLargeMeanPtTwoHundredDiffering}{1,000\xspace}
\newcommand{\nArchLargeMeanPtTwoHundredHi}{0.001\xspace}

\newcommand{\nArchLargeMeanPtTwoHundredLo}{-0.002\xspace}
\newcommand{\nArchLargeMeanPtTwoHundredMac}{0.061\xspace}

\newcommand{\nArchLargeMeanPtTwoHundredWorkers}{0.062\xspace}

\newcommand{\nArchMacCpu}{Apple M4\xspace}
\newcommand{\nArchMacMachine}{arm64\xspace}
\newcommand{\nArchMacNumpy}{2.1.3\xspace}
\newcommand{\nArchMacOs}{macOS 26.5\xspace}
\newcommand{\nArchMacPython}{3.13.5\xspace}

\newcommand{\nArchMeanShiftFreqFiftyDiff}{0.0000\xspace}
\newcommand{\nArchMeanShiftFreqFiftyDiffering}{352\xspace}
\newcommand{\nArchMeanShiftFreqFiftyHi}{0.0009\xspace}

\newcommand{\nArchMeanShiftFreqFiftyLo}{-0.0009\xspace}

\newcommand{\nArchMeanShiftFreqFiveThousandDiff}{0.0000\xspace}
\newcommand{\nArchMeanShiftFreqFiveThousandDiffering}{963\xspace}
\newcommand{\nArchMeanShiftFreqFiveThousandHi}{0.0001\xspace}

\newcommand{\nArchMeanShiftFreqFiveThousandLo}{-0.0001\xspace}

\newcommand{\nArchMeanShiftFreqTenDiff}{0.0006\xspace}
\newcommand{\nArchMeanShiftFreqTenDiffering}{90\xspace}
\newcommand{\nArchMeanShiftFreqTenHi}{0.0027\xspace}

\newcommand{\nArchMeanShiftFreqTenLo}{-0.0015\xspace}

\newcommand{\nArchMeanShiftFreqThousandDiff}{0.0001\xspace}
\newcommand{\nArchMeanShiftFreqThousandDiffering}{878\xspace}
\newcommand{\nArchMeanShiftFreqThousandHi}{0.0003\xspace}

\newcommand{\nArchMeanShiftFreqThousandLo}{-0.0001\xspace}

\newcommand{\nArchMeanShiftFreqTwentyFiveDiff}{0.0006\xspace}
\newcommand{\nArchMeanShiftFreqTwentyFiveDiffering}{211\xspace}
\newcommand{\nArchMeanShiftFreqTwentyFiveHi}{0.0019\xspace}

\newcommand{\nArchMeanShiftFreqTwentyFiveLo}{-0.0008\xspace}

\newcommand{\nArchMeanShiftFreqTwoHundredDiff}{-0.0002\xspace}
\newcommand{\nArchMeanShiftFreqTwoHundredDiffering}{716\xspace}
\newcommand{\nArchMeanShiftFreqTwoHundredHi}{0.0003\xspace}

\newcommand{\nArchMeanShiftFreqTwoHundredLo}{-0.0007\xspace}

\newcommand{\nArchMeanShiftPtFiftyDiff}{-0.001\xspace}
\newcommand{\nArchMeanShiftPtFiftyDiffering}{994\xspace}
\newcommand{\nArchMeanShiftPtFiftyHi}{0.001\xspace}

\newcommand{\nArchMeanShiftPtFiftyLo}{-0.003\xspace}
\newcommand{\nArchMeanShiftPtFiftyMac}{0.061\xspace}

\newcommand{\nArchMeanShiftPtFiftyWorkers}{0.062\xspace}
\newcommand{\nArchMeanShiftPtFiveThousandDiff}{0.000\xspace}
\newcommand{\nArchMeanShiftPtFiveThousandDiffering}{1,000\xspace}
\newcommand{\nArchMeanShiftPtFiveThousandHi}{0.001\xspace}

\newcommand{\nArchMeanShiftPtFiveThousandLo}{-0.001\xspace}
\newcommand{\nArchMeanShiftPtFiveThousandMac}{0.048\xspace}

\newcommand{\nArchMeanShiftPtFiveThousandWorkers}{0.049\xspace}
\newcommand{\nArchMeanShiftPtTenDiff}{0.000\xspace}
\newcommand{\nArchMeanShiftPtTenDiffering}{993\xspace}
\newcommand{\nArchMeanShiftPtTenHi}{0.002\xspace}

\newcommand{\nArchMeanShiftPtTenLo}{-0.001\xspace}
\newcommand{\nArchMeanShiftPtTenMac}{0.108\xspace}

\newcommand{\nArchMeanShiftPtTenWorkers}{0.108\xspace}
\newcommand{\nArchMeanShiftPtThousandDiff}{0.000\xspace}
\newcommand{\nArchMeanShiftPtThousandDiffering}{1,000\xspace}
\newcommand{\nArchMeanShiftPtThousandHi}{0.002\xspace}

\newcommand{\nArchMeanShiftPtThousandLo}{-0.001\xspace}
\newcommand{\nArchMeanShiftPtThousandMac}{0.055\xspace}

\newcommand{\nArchMeanShiftPtThousandWorkers}{0.055\xspace}
\newcommand{\nArchMeanShiftPtTwentyFiveDiff}{0.000\xspace}
\newcommand{\nArchMeanShiftPtTwentyFiveDiffering}{993\xspace}
\newcommand{\nArchMeanShiftPtTwentyFiveHi}{0.001\xspace}

\newcommand{\nArchMeanShiftPtTwentyFiveLo}{-0.002\xspace}
\newcommand{\nArchMeanShiftPtTwentyFiveMac}{0.074\xspace}

\newcommand{\nArchMeanShiftPtTwentyFiveWorkers}{0.075\xspace}
\newcommand{\nArchMeanShiftPtTwoHundredDiff}{-0.001\xspace}
\newcommand{\nArchMeanShiftPtTwoHundredDiffering}{1,000\xspace}
\newcommand{\nArchMeanShiftPtTwoHundredHi}{0.000\xspace}

\newcommand{\nArchMeanShiftPtTwoHundredLo}{-0.003\xspace}
\newcommand{\nArchMeanShiftPtTwoHundredMac}{0.060\xspace}

\newcommand{\nArchMeanShiftPtTwoHundredWorkers}{0.061\xspace}

\newcommand{\nArchNoBreakFreqFiftyDiff}{-0.0002\xspace}
\newcommand{\nArchNoBreakFreqFiftyDiffering}{166\xspace}
\newcommand{\nArchNoBreakFreqFiftyHi}{0.0004\xspace}

\newcommand{\nArchNoBreakFreqFiftyLo}{-0.0007\xspace}

\newcommand{\nArchNoBreakFreqFiveThousandDiff}{0.0000\xspace}
\newcommand{\nArchNoBreakFreqFiveThousandDiffering}{923\xspace}
\newcommand{\nArchNoBreakFreqFiveThousandHi}{0.0001\xspace}

\newcommand{\nArchNoBreakFreqFiveThousandLo}{-0.0001\xspace}

\newcommand{\nArchNoBreakFreqTenDiff}{0.0013\xspace}
\newcommand{\nArchNoBreakFreqTenDiffering}{44\xspace}
\newcommand{\nArchNoBreakFreqTenHi}{0.0027\xspace}

\newcommand{\nArchNoBreakFreqTenLo}{0.0000\xspace}

\newcommand{\nArchNoBreakFreqThousandDiff}{0.0001\xspace}
\newcommand{\nArchNoBreakFreqThousandDiffering}{777\xspace}
\newcommand{\nArchNoBreakFreqThousandHi}{0.0002\xspace}

\newcommand{\nArchNoBreakFreqThousandLo}{-0.0001\xspace}

\newcommand{\nArchNoBreakFreqThousandPlaceMac}{5\xspace}
\newcommand{\nArchNoBreakFreqThousandPlaceWorkers}{4\xspace}

\newcommand{\nArchNoBreakFreqTwentyFiveDiff}{0.0000\xspace}
\newcommand{\nArchNoBreakFreqTwentyFiveDiffering}{61\xspace}
\newcommand{\nArchNoBreakFreqTwentyFiveHi}{0.0007\xspace}

\newcommand{\nArchNoBreakFreqTwentyFiveLo}{-0.0007\xspace}

\newcommand{\nArchNoBreakFreqTwoHundredDiff}{0.0004\xspace}
\newcommand{\nArchNoBreakFreqTwoHundredDiffering}{564\xspace}
\newcommand{\nArchNoBreakFreqTwoHundredHi}{0.0008\xspace}

\newcommand{\nArchNoBreakFreqTwoHundredLo}{0.0001\xspace}

\newcommand{\nArchNoBreakPtFiftyDiff}{0.001\xspace}
\newcommand{\nArchNoBreakPtFiftyDiffering}{996\xspace}
\newcommand{\nArchNoBreakPtFiftyHi}{0.001\xspace}

\newcommand{\nArchNoBreakPtFiftyLo}{0.000\xspace}
\newcommand{\nArchNoBreakPtFiftyMac}{0.053\xspace}

\newcommand{\nArchNoBreakPtFiftyWorkers}{0.052\xspace}
\newcommand{\nArchNoBreakPtFiveThousandDiff}{-0.001\xspace}
\newcommand{\nArchNoBreakPtFiveThousandDiffering}{1,000\xspace}
\newcommand{\nArchNoBreakPtFiveThousandHi}{0.000\xspace}

\newcommand{\nArchNoBreakPtFiveThousandLo}{-0.001\xspace}
\newcommand{\nArchNoBreakPtFiveThousandMac}{0.039\xspace}

\newcommand{\nArchNoBreakPtFiveThousandWorkers}{0.040\xspace}
\newcommand{\nArchNoBreakPtTenDiff}{0.000\xspace}
\newcommand{\nArchNoBreakPtTenDiffering}{996\xspace}
\newcommand{\nArchNoBreakPtTenHi}{0.001\xspace}

\newcommand{\nArchNoBreakPtTenLo}{0.000\xspace}
\newcommand{\nArchNoBreakPtTenMac}{0.058\xspace}

\newcommand{\nArchNoBreakPtTenWorkers}{0.058\xspace}
\newcommand{\nArchNoBreakPtThousandDiff}{0.000\xspace}
\newcommand{\nArchNoBreakPtThousandDiffering}{1,000\xspace}
\newcommand{\nArchNoBreakPtThousandHi}{0.000\xspace}

\newcommand{\nArchNoBreakPtThousandLo}{-0.001\xspace}
\newcommand{\nArchNoBreakPtThousandMac}{0.047\xspace}

\newcommand{\nArchNoBreakPtThousandWorkers}{0.047\xspace}
\newcommand{\nArchNoBreakPtTwentyFiveDiff}{0.000\xspace}
\newcommand{\nArchNoBreakPtTwentyFiveDiffering}{996\xspace}
\newcommand{\nArchNoBreakPtTwentyFiveHi}{0.001\xspace}

\newcommand{\nArchNoBreakPtTwentyFiveLo}{0.000\xspace}
\newcommand{\nArchNoBreakPtTwentyFiveMac}{0.053\xspace}

\newcommand{\nArchNoBreakPtTwentyFiveWorkers}{0.052\xspace}
\newcommand{\nArchNoBreakPtTwoHundredDiff}{-0.001\xspace}
\newcommand{\nArchNoBreakPtTwoHundredDiffering}{1,000\xspace}
\newcommand{\nArchNoBreakPtTwoHundredHi}{0.000\xspace}

\newcommand{\nArchNoBreakPtTwoHundredLo}{-0.002\xspace}
\newcommand{\nArchNoBreakPtTwoHundredMac}{0.056\xspace}

\newcommand{\nArchNoBreakPtTwoHundredWorkers}{0.057\xspace}

\newcommand{\nArchReps}{5,000\xspace}

\newcommand{\nArchVarShiftFreqFiftyDiff}{0.0005\xspace}
\newcommand{\nArchVarShiftFreqFiftyDiffering}{283\xspace}
\newcommand{\nArchVarShiftFreqFiftyHi}{0.0013\xspace}

\newcommand{\nArchVarShiftFreqFiftyLo}{-0.0003\xspace}

\newcommand{\nArchVarShiftFreqFiveThousandDiff}{0.0000\xspace}
\newcommand{\nArchVarShiftFreqFiveThousandDiffering}{932\xspace}
\newcommand{\nArchVarShiftFreqFiveThousandHi}{0.0001\xspace}

\newcommand{\nArchVarShiftFreqFiveThousandLo}{-0.0001\xspace}

\newcommand{\nArchVarShiftFreqTenDiff}{0.0010\xspace}
\newcommand{\nArchVarShiftFreqTenDiffering}{77\xspace}
\newcommand{\nArchVarShiftFreqTenHi}{0.0028\xspace}

\newcommand{\nArchVarShiftFreqTenLo}{-0.0007\xspace}

\newcommand{\nArchVarShiftFreqThousandDiff}{-0.0001\xspace}
\newcommand{\nArchVarShiftFreqThousandDiffering}{841\xspace}
\newcommand{\nArchVarShiftFreqThousandHi}{0.0000\xspace}

\newcommand{\nArchVarShiftFreqThousandLo}{-0.0003\xspace}

\newcommand{\nArchVarShiftFreqTwentyFiveDiff}{0.0005\xspace}
\newcommand{\nArchVarShiftFreqTwentyFiveDiffering}{166\xspace}
\newcommand{\nArchVarShiftFreqTwentyFiveHi}{0.0016\xspace}

\newcommand{\nArchVarShiftFreqTwentyFiveLo}{-0.0006\xspace}

\newcommand{\nArchVarShiftFreqTwoHundredDiff}{0.0000\xspace}
\newcommand{\nArchVarShiftFreqTwoHundredDiffering}{613\xspace}
\newcommand{\nArchVarShiftFreqTwoHundredHi}{0.0004\xspace}

\newcommand{\nArchVarShiftFreqTwoHundredLo}{-0.0004\xspace}

\newcommand{\nArchVarShiftPtFiftyDiff}{0.000\xspace}
\newcommand{\nArchVarShiftPtFiftyDiffering}{996\xspace}
\newcommand{\nArchVarShiftPtFiftyHi}{0.001\xspace}

\newcommand{\nArchVarShiftPtFiftyLo}{-0.001\xspace}
\newcommand{\nArchVarShiftPtFiftyMac}{0.054\xspace}

\newcommand{\nArchVarShiftPtFiftyWorkers}{0.054\xspace}
\newcommand{\nArchVarShiftPtFiveThousandDiff}{0.000\xspace}
\newcommand{\nArchVarShiftPtFiveThousandDiffering}{1,000\xspace}
\newcommand{\nArchVarShiftPtFiveThousandHi}{0.001\xspace}

\newcommand{\nArchVarShiftPtFiveThousandLo}{-0.001\xspace}
\newcommand{\nArchVarShiftPtFiveThousandMac}{0.040\xspace}

\newcommand{\nArchVarShiftPtFiveThousandWorkers}{0.040\xspace}
\newcommand{\nArchVarShiftPtTenDiff}{0.000\xspace}
\newcommand{\nArchVarShiftPtTenDiffering}{996\xspace}
\newcommand{\nArchVarShiftPtTenHi}{0.001\xspace}

\newcommand{\nArchVarShiftPtTenLo}{-0.001\xspace}
\newcommand{\nArchVarShiftPtTenMac}{0.095\xspace}

\newcommand{\nArchVarShiftPtTenWorkers}{0.095\xspace}
\newcommand{\nArchVarShiftPtThousandDiff}{0.000\xspace}
\newcommand{\nArchVarShiftPtThousandDiffering}{1,000\xspace}
\newcommand{\nArchVarShiftPtThousandHi}{0.001\xspace}

\newcommand{\nArchVarShiftPtThousandLo}{0.000\xspace}
\newcommand{\nArchVarShiftPtThousandMac}{0.047\xspace}

\newcommand{\nArchVarShiftPtThousandWorkers}{0.047\xspace}
\newcommand{\nArchVarShiftPtTwentyFiveDiff}{0.000\xspace}
\newcommand{\nArchVarShiftPtTwentyFiveDiffering}{996\xspace}
\newcommand{\nArchVarShiftPtTwentyFiveHi}{0.001\xspace}

\newcommand{\nArchVarShiftPtTwentyFiveLo}{-0.001\xspace}
\newcommand{\nArchVarShiftPtTwentyFiveMac}{0.070\xspace}

\newcommand{\nArchVarShiftPtTwentyFiveWorkers}{0.070\xspace}
\newcommand{\nArchVarShiftPtTwoHundredDiff}{-0.001\xspace}
\newcommand{\nArchVarShiftPtTwoHundredDiffering}{999\xspace}
\newcommand{\nArchVarShiftPtTwoHundredHi}{0.000\xspace}

\newcommand{\nArchVarShiftPtTwoHundredLo}{-0.001\xspace}
\newcommand{\nArchVarShiftPtTwoHundredMac}{0.048\xspace}

\newcommand{\nArchVarShiftPtTwoHundredWorkers}{0.048\xspace}

\newcommand{\nArchWorkersAvxFiveTwelve}{55\xspace}
\newcommand{\nArchWorkersAvxTwoOnly}{145\xspace}
\newcommand{\nArchWorkersChunks}{200\xspace}

\newcommand{\nArchWorkersMachine}{x86_64\xspace}
\newcommand{\nArchWorkersNumpy}{2.1.3\xspace}
\newcommand{\nArchWorkersPython}{3.13.0\xspace}

\newcommand{\nArmBoostFreqBelowEvery}{4\xspace}
\newcommand{\nArmBoostFreqBelowFifty}{4\xspace}

\newcommand{\nArmBothShiftAbaciFreqFiftyDiff}{-0.0365\xspace}
\newcommand{\nArmBothShiftAbaciFreqFiftyHi}{-0.0354\xspace}
\newcommand{\nArmBothShiftAbaciFreqFiftyLo}{-0.0376\xspace}

\newcommand{\nArmBothShiftAbaciFreqFiveThousandDiff}{-0.0004\xspace}
\newcommand{\nArmBothShiftAbaciFreqFiveThousandHi}{-0.0004\xspace}
\newcommand{\nArmBothShiftAbaciFreqFiveThousandLo}{-0.0004\xspace}

\newcommand{\nArmBothShiftAbaciFreqTenDiff}{-0.1024\xspace}
\newcommand{\nArmBothShiftAbaciFreqTenHi}{-0.0977\xspace}
\newcommand{\nArmBothShiftAbaciFreqTenLo}{-0.1072\xspace}

\newcommand{\nArmBothShiftAbaciFreqThousandDiff}{-0.0018\xspace}
\newcommand{\nArmBothShiftAbaciFreqThousandHi}{-0.0017\xspace}
\newcommand{\nArmBothShiftAbaciFreqThousandLo}{-0.0018\xspace}

\newcommand{\nArmBothShiftAbaciFreqTwentyFiveDiff}{-0.0654\xspace}
\newcommand{\nArmBothShiftAbaciFreqTwentyFiveHi}{-0.0634\xspace}
\newcommand{\nArmBothShiftAbaciFreqTwentyFiveLo}{-0.0673\xspace}

\newcommand{\nArmBothShiftAbaciFreqTwoHundredDiff}{-0.0089\xspace}
\newcommand{\nArmBothShiftAbaciFreqTwoHundredHi}{-0.0086\xspace}
\newcommand{\nArmBothShiftAbaciFreqTwoHundredLo}{-0.0092\xspace}

\newcommand{\nArmBothShiftBoostFreqFiftyDiff}{-0.0371\xspace}
\newcommand{\nArmBothShiftBoostFreqFiftyHi}{-0.0360\xspace}
\newcommand{\nArmBothShiftBoostFreqFiftyLo}{-0.0381\xspace}

\newcommand{\nArmBothShiftBoostFreqFiveThousandDiff}{-0.0004\xspace}
\newcommand{\nArmBothShiftBoostFreqFiveThousandHi}{-0.0004\xspace}
\newcommand{\nArmBothShiftBoostFreqFiveThousandLo}{-0.0004\xspace}

\newcommand{\nArmBothShiftBoostFreqTenDiff}{-0.1088\xspace}
\newcommand{\nArmBothShiftBoostFreqTenHi}{-0.1046\xspace}
\newcommand{\nArmBothShiftBoostFreqTenLo}{-0.1131\xspace}

\newcommand{\nArmBothShiftBoostFreqThousandDiff}{-0.0019\xspace}
\newcommand{\nArmBothShiftBoostFreqThousandHi}{-0.0018\xspace}
\newcommand{\nArmBothShiftBoostFreqThousandLo}{-0.0019\xspace}

\newcommand{\nArmBothShiftBoostFreqTwentyFiveDiff}{-0.0669\xspace}
\newcommand{\nArmBothShiftBoostFreqTwentyFiveHi}{-0.0650\xspace}
\newcommand{\nArmBothShiftBoostFreqTwentyFiveLo}{-0.0688\xspace}

\newcommand{\nArmBothShiftBoostFreqTwoHundredDiff}{-0.0092\xspace}
\newcommand{\nArmBothShiftBoostFreqTwoHundredHi}{-0.0090\xspace}
\newcommand{\nArmBothShiftBoostFreqTwoHundredLo}{-0.0095\xspace}

\newcommand{\nArmBothShiftMoveFreqFiftyDiff}{0.0039\xspace}
\newcommand{\nArmBothShiftMoveFreqFiftyHi}{0.0046\xspace}
\newcommand{\nArmBothShiftMoveFreqFiftyLo}{0.0031\xspace}

\newcommand{\nArmBothShiftMoveFreqFiveThousandFineDiff}{0.00004\xspace}
\newcommand{\nArmBothShiftMoveFreqFiveThousandFineHi}{0.00005\xspace}
\newcommand{\nArmBothShiftMoveFreqFiveThousandFineLo}{0.00004\xspace}

\newcommand{\nArmBothShiftMoveFreqTenDiff}{0.0156\xspace}
\newcommand{\nArmBothShiftMoveFreqTenHi}{0.0189\xspace}
\newcommand{\nArmBothShiftMoveFreqTenLo}{0.0123\xspace}

\newcommand{\nArmBothShiftMoveFreqThousandDiff}{0.0002\xspace}
\newcommand{\nArmBothShiftMoveFreqThousandHi}{0.0002\xspace}
\newcommand{\nArmBothShiftMoveFreqThousandLo}{0.0002\xspace}

\newcommand{\nArmBothShiftMoveFreqTwentyFiveDiff}{0.0072\xspace}
\newcommand{\nArmBothShiftMoveFreqTwentyFiveHi}{0.0086\xspace}
\newcommand{\nArmBothShiftMoveFreqTwentyFiveLo}{0.0057\xspace}

\newcommand{\nArmBothShiftMoveFreqTwoHundredDiff}{0.0011\xspace}
\newcommand{\nArmBothShiftMoveFreqTwoHundredHi}{0.0013\xspace}
\newcommand{\nArmBothShiftMoveFreqTwoHundredLo}{0.0009\xspace}

\newcommand{\nArmLargeMeanAbaciFreqFiftyDiff}{-0.0681\xspace}
\newcommand{\nArmLargeMeanAbaciFreqFiftyHi}{-0.0672\xspace}
\newcommand{\nArmLargeMeanAbaciFreqFiftyLo}{-0.0690\xspace}

\newcommand{\nArmLargeMeanAbaciFreqFiveThousandDiff}{-0.0007\xspace}
\newcommand{\nArmLargeMeanAbaciFreqFiveThousandHi}{-0.0007\xspace}
\newcommand{\nArmLargeMeanAbaciFreqFiveThousandLo}{-0.0007\xspace}

\newcommand{\nArmLargeMeanAbaciFreqTenDiff}{-0.2429\xspace}
\newcommand{\nArmLargeMeanAbaciFreqTenHi}{-0.2386\xspace}
\newcommand{\nArmLargeMeanAbaciFreqTenLo}{-0.2473\xspace}

\newcommand{\nArmLargeMeanAbaciFreqThousandDiff}{-0.0034\xspace}
\newcommand{\nArmLargeMeanAbaciFreqThousandHi}{-0.0034\xspace}
\newcommand{\nArmLargeMeanAbaciFreqThousandLo}{-0.0035\xspace}

\newcommand{\nArmLargeMeanAbaciFreqTwentyFiveDiff}{-0.1328\xspace}
\newcommand{\nArmLargeMeanAbaciFreqTwentyFiveHi}{-0.1310\xspace}
\newcommand{\nArmLargeMeanAbaciFreqTwentyFiveLo}{-0.1345\xspace}

\newcommand{\nArmLargeMeanAbaciFreqTwoHundredDiff}{-0.0171\xspace}
\newcommand{\nArmLargeMeanAbaciFreqTwoHundredHi}{-0.0168\xspace}
\newcommand{\nArmLargeMeanAbaciFreqTwoHundredLo}{-0.0173\xspace}

\newcommand{\nArmLargeMeanBoostFreqFiftyDiff}{-0.0685\xspace}
\newcommand{\nArmLargeMeanBoostFreqFiftyHi}{-0.0676\xspace}
\newcommand{\nArmLargeMeanBoostFreqFiftyLo}{-0.0694\xspace}

\newcommand{\nArmLargeMeanBoostFreqFiveThousandDiff}{-0.0007\xspace}
\newcommand{\nArmLargeMeanBoostFreqFiveThousandHi}{-0.0007\xspace}
\newcommand{\nArmLargeMeanBoostFreqFiveThousandLo}{-0.0007\xspace}

\newcommand{\nArmLargeMeanBoostFreqTenDiff}{-0.2540\xspace}
\newcommand{\nArmLargeMeanBoostFreqTenHi}{-0.2505\xspace}
\newcommand{\nArmLargeMeanBoostFreqTenLo}{-0.2576\xspace}

\newcommand{\nArmLargeMeanBoostFreqThousandDiff}{-0.0034\xspace}
\newcommand{\nArmLargeMeanBoostFreqThousandHi}{-0.0034\xspace}
\newcommand{\nArmLargeMeanBoostFreqThousandLo}{-0.0035\xspace}

\newcommand{\nArmLargeMeanBoostFreqTwentyFiveDiff}{-0.1325\xspace}
\newcommand{\nArmLargeMeanBoostFreqTwentyFiveHi}{-0.1307\xspace}
\newcommand{\nArmLargeMeanBoostFreqTwentyFiveLo}{-0.1342\xspace}

\newcommand{\nArmLargeMeanBoostFreqTwoHundredDiff}{-0.0172\xspace}
\newcommand{\nArmLargeMeanBoostFreqTwoHundredHi}{-0.0170\xspace}
\newcommand{\nArmLargeMeanBoostFreqTwoHundredLo}{-0.0174\xspace}

\newcommand{\nArmLargeMeanMoveFreqFiftyDiff}{0.0044\xspace}
\newcommand{\nArmLargeMeanMoveFreqFiftyHi}{0.0051\xspace}
\newcommand{\nArmLargeMeanMoveFreqFiftyLo}{0.0037\xspace}

\newcommand{\nArmLargeMeanMoveFreqFiveThousandFineDiff}{0.00004\xspace}
\newcommand{\nArmLargeMeanMoveFreqFiveThousandFineHi}{0.00005\xspace}
\newcommand{\nArmLargeMeanMoveFreqFiveThousandFineLo}{0.00004\xspace}

\newcommand{\nArmLargeMeanMoveFreqTenDiff}{0.0199\xspace}
\newcommand{\nArmLargeMeanMoveFreqTenHi}{0.0235\xspace}
\newcommand{\nArmLargeMeanMoveFreqTenLo}{0.0163\xspace}

\newcommand{\nArmLargeMeanMoveFreqThousandDiff}{0.0002\xspace}
\newcommand{\nArmLargeMeanMoveFreqThousandHi}{0.0002\xspace}
\newcommand{\nArmLargeMeanMoveFreqThousandLo}{0.0002\xspace}

\newcommand{\nArmLargeMeanMoveFreqTwentyFiveDiff}{0.0083\xspace}
\newcommand{\nArmLargeMeanMoveFreqTwentyFiveHi}{0.0097\xspace}
\newcommand{\nArmLargeMeanMoveFreqTwentyFiveLo}{0.0068\xspace}

\newcommand{\nArmLargeMeanMoveFreqTwoHundredDiff}{0.0011\xspace}
\newcommand{\nArmLargeMeanMoveFreqTwoHundredHi}{0.0013\xspace}
\newcommand{\nArmLargeMeanMoveFreqTwoHundredLo}{0.0009\xspace}

\newcommand{\nArmMeanShiftAbaciFreqFiftyDiff}{-0.0241\xspace}
\newcommand{\nArmMeanShiftAbaciFreqFiftyHi}{-0.0230\xspace}
\newcommand{\nArmMeanShiftAbaciFreqFiftyLo}{-0.0251\xspace}

\newcommand{\nArmMeanShiftAbaciFreqFiveThousandDiff}{-0.0002\xspace}
\newcommand{\nArmMeanShiftAbaciFreqFiveThousandHi}{-0.0002\xspace}
\newcommand{\nArmMeanShiftAbaciFreqFiveThousandLo}{-0.0003\xspace}

\newcommand{\nArmMeanShiftAbaciFreqTenDiff}{-0.0597\xspace}
\newcommand{\nArmMeanShiftAbaciFreqTenHi}{-0.0554\xspace}
\newcommand{\nArmMeanShiftAbaciFreqTenLo}{-0.0639\xspace}

\newcommand{\nArmMeanShiftAbaciFreqThousandDiff}{-0.0012\xspace}
\newcommand{\nArmMeanShiftAbaciFreqThousandHi}{-0.0012\xspace}
\newcommand{\nArmMeanShiftAbaciFreqThousandLo}{-0.0013\xspace}

\newcommand{\nArmMeanShiftAbaciFreqTwentyFiveDiff}{-0.0402\xspace}
\newcommand{\nArmMeanShiftAbaciFreqTwentyFiveHi}{-0.0385\xspace}
\newcommand{\nArmMeanShiftAbaciFreqTwentyFiveLo}{-0.0420\xspace}

\newcommand{\nArmMeanShiftAbaciFreqTwoHundredDiff}{-0.0058\xspace}
\newcommand{\nArmMeanShiftAbaciFreqTwoHundredHi}{-0.0056\xspace}
\newcommand{\nArmMeanShiftAbaciFreqTwoHundredLo}{-0.0061\xspace}

\newcommand{\nArmMeanShiftBoostFreqFiftyDiff}{-0.0220\xspace}
\newcommand{\nArmMeanShiftBoostFreqFiftyHi}{-0.0209\xspace}
\newcommand{\nArmMeanShiftBoostFreqFiftyLo}{-0.0231\xspace}

\newcommand{\nArmMeanShiftBoostFreqFiveThousandDiff}{-0.0002\xspace}
\newcommand{\nArmMeanShiftBoostFreqFiveThousandHi}{-0.0002\xspace}
\newcommand{\nArmMeanShiftBoostFreqFiveThousandLo}{-0.0002\xspace}

\newcommand{\nArmMeanShiftBoostFreqTenDiff}{-0.0717\xspace}
\newcommand{\nArmMeanShiftBoostFreqTenHi}{-0.0682\xspace}
\newcommand{\nArmMeanShiftBoostFreqTenLo}{-0.0752\xspace}

\newcommand{\nArmMeanShiftBoostFreqThousandDiff}{-0.0011\xspace}
\newcommand{\nArmMeanShiftBoostFreqThousandHi}{-0.0011\xspace}
\newcommand{\nArmMeanShiftBoostFreqThousandLo}{-0.0012\xspace}

\newcommand{\nArmMeanShiftBoostFreqTwentyFiveDiff}{-0.0394\xspace}
\newcommand{\nArmMeanShiftBoostFreqTwentyFiveHi}{-0.0377\xspace}
\newcommand{\nArmMeanShiftBoostFreqTwentyFiveLo}{-0.0411\xspace}

\newcommand{\nArmMeanShiftBoostFreqTwoHundredDiff}{-0.0054\xspace}
\newcommand{\nArmMeanShiftBoostFreqTwoHundredHi}{-0.0051\xspace}
\newcommand{\nArmMeanShiftBoostFreqTwoHundredLo}{-0.0057\xspace}

\newcommand{\nArmMeanShiftMoveFreqFiftyDiff}{0.0044\xspace}
\newcommand{\nArmMeanShiftMoveFreqFiftyHi}{0.0052\xspace}
\newcommand{\nArmMeanShiftMoveFreqFiftyLo}{0.0037\xspace}

\newcommand{\nArmMeanShiftMoveFreqFiveThousandFineDiff}{0.00004\xspace}
\newcommand{\nArmMeanShiftMoveFreqFiveThousandFineHi}{0.00005\xspace}
\newcommand{\nArmMeanShiftMoveFreqFiveThousandFineLo}{0.00004\xspace}

\newcommand{\nArmMeanShiftMoveFreqTenDiff}{0.0191\xspace}
\newcommand{\nArmMeanShiftMoveFreqTenHi}{0.0227\xspace}
\newcommand{\nArmMeanShiftMoveFreqTenLo}{0.0156\xspace}

\newcommand{\nArmMeanShiftMoveFreqThousandDiff}{0.0002\xspace}
\newcommand{\nArmMeanShiftMoveFreqThousandHi}{0.0002\xspace}
\newcommand{\nArmMeanShiftMoveFreqThousandLo}{0.0002\xspace}

\newcommand{\nArmMeanShiftMoveFreqTwentyFiveDiff}{0.0076\xspace}
\newcommand{\nArmMeanShiftMoveFreqTwentyFiveHi}{0.0090\xspace}
\newcommand{\nArmMeanShiftMoveFreqTwentyFiveLo}{0.0063\xspace}

\newcommand{\nArmMeanShiftMoveFreqTwoHundredDiff}{0.0010\xspace}
\newcommand{\nArmMeanShiftMoveFreqTwoHundredHi}{0.0012\xspace}
\newcommand{\nArmMeanShiftMoveFreqTwoHundredLo}{0.0008\xspace}

\newcommand{\nArmMoveFreqAboveEvery}{4\xspace}
\newcommand{\nArmMoveFreqAboveFifty}{4\xspace}

\newcommand{\nArmNoBreakAbaciFreqFiftyDiff}{-0.0002\xspace}
\newcommand{\nArmNoBreakAbaciFreqFiftyHi}{0.0007\xspace}
\newcommand{\nArmNoBreakAbaciFreqFiftyLo}{-0.0011\xspace}

\newcommand{\nArmNoBreakAbaciFreqFiveThousandDiff}{0.0000\xspace}
\newcommand{\nArmNoBreakAbaciFreqFiveThousandHi}{0.0000\xspace}
\newcommand{\nArmNoBreakAbaciFreqFiveThousandLo}{0.0000\xspace}

\newcommand{\nArmNoBreakAbaciFreqTenDiff}{-0.0336\xspace}
\newcommand{\nArmNoBreakAbaciFreqTenHi}{-0.0304\xspace}
\newcommand{\nArmNoBreakAbaciFreqTenLo}{-0.0369\xspace}

\newcommand{\nArmNoBreakAbaciFreqThousandDiff}{0.0000\xspace}
\newcommand{\nArmNoBreakAbaciFreqThousandHi}{0.0001\xspace}
\newcommand{\nArmNoBreakAbaciFreqThousandLo}{0.0000\xspace}

\newcommand{\nArmNoBreakAbaciFreqTwentyFiveDiff}{-0.0045\xspace}
\newcommand{\nArmNoBreakAbaciFreqTwentyFiveHi}{-0.0036\xspace}
\newcommand{\nArmNoBreakAbaciFreqTwentyFiveLo}{-0.0054\xspace}

\newcommand{\nArmNoBreakAbaciFreqTwoHundredDiff}{0.0000\xspace}
\newcommand{\nArmNoBreakAbaciFreqTwoHundredHi}{0.0002\xspace}
\newcommand{\nArmNoBreakAbaciFreqTwoHundredLo}{-0.0002\xspace}

\newcommand{\nArmNoBreakBoostFreqFiftyDiff}{0.0006\xspace}
\newcommand{\nArmNoBreakBoostFreqFiftyHi}{0.0013\xspace}
\newcommand{\nArmNoBreakBoostFreqFiftyLo}{-0.0002\xspace}

\newcommand{\nArmNoBreakBoostFreqFiveThousandDiff}{0.0000\xspace}
\newcommand{\nArmNoBreakBoostFreqFiveThousandHi}{0.0000\xspace}
\newcommand{\nArmNoBreakBoostFreqFiveThousandLo}{0.0000\xspace}

\newcommand{\nArmNoBreakBoostFreqTenDiff}{-0.0322\xspace}
\newcommand{\nArmNoBreakBoostFreqTenHi}{-0.0293\xspace}
\newcommand{\nArmNoBreakBoostFreqTenLo}{-0.0352\xspace}

\newcommand{\nArmNoBreakBoostFreqThousandDiff}{0.0001\xspace}
\newcommand{\nArmNoBreakBoostFreqThousandHi}{0.0001\xspace}
\newcommand{\nArmNoBreakBoostFreqThousandLo}{0.0000\xspace}

\newcommand{\nArmNoBreakBoostFreqTwentyFiveDiff}{-0.0032\xspace}
\newcommand{\nArmNoBreakBoostFreqTwentyFiveHi}{-0.0024\xspace}
\newcommand{\nArmNoBreakBoostFreqTwentyFiveLo}{-0.0040\xspace}

\newcommand{\nArmNoBreakBoostFreqTwoHundredDiff}{0.0003\xspace}
\newcommand{\nArmNoBreakBoostFreqTwoHundredHi}{0.0005\xspace}
\newcommand{\nArmNoBreakBoostFreqTwoHundredLo}{0.0001\xspace}

\newcommand{\nArmNoBreakMoveFreqFiftyDiff}{-0.0036\xspace}
\newcommand{\nArmNoBreakMoveFreqFiftyHi}{-0.0030\xspace}
\newcommand{\nArmNoBreakMoveFreqFiftyLo}{-0.0043\xspace}

\newcommand{\nArmNoBreakMoveFreqFiveThousandFineDiff}{-0.00003\xspace}
\newcommand{\nArmNoBreakMoveFreqFiveThousandFineHi}{-0.00003\xspace}
\newcommand{\nArmNoBreakMoveFreqFiveThousandFineLo}{-0.00004\xspace}

\newcommand{\nArmNoBreakMoveFreqTenDiff}{-0.0121\xspace}
\newcommand{\nArmNoBreakMoveFreqTenHi}{-0.0088\xspace}
\newcommand{\nArmNoBreakMoveFreqTenLo}{-0.0155\xspace}

\newcommand{\nArmNoBreakMoveFreqThousandDiff}{-0.0001\xspace}
\newcommand{\nArmNoBreakMoveFreqThousandHi}{-0.0001\xspace}
\newcommand{\nArmNoBreakMoveFreqThousandLo}{-0.0002\xspace}

\newcommand{\nArmNoBreakMoveFreqTwentyFiveDiff}{-0.0041\xspace}
\newcommand{\nArmNoBreakMoveFreqTwentyFiveHi}{-0.0033\xspace}
\newcommand{\nArmNoBreakMoveFreqTwentyFiveLo}{-0.0050\xspace}

\newcommand{\nArmNoBreakMoveFreqTwoHundredDiff}{-0.0009\xspace}
\newcommand{\nArmNoBreakMoveFreqTwoHundredHi}{-0.0007\xspace}
\newcommand{\nArmNoBreakMoveFreqTwoHundredLo}{-0.0011\xspace}

\newcommand{\nArmVarShiftAbaciFreqFiftyDiff}{-0.0197\xspace}
\newcommand{\nArmVarShiftAbaciFreqFiftyHi}{-0.0185\xspace}
\newcommand{\nArmVarShiftAbaciFreqFiftyLo}{-0.0209\xspace}

\newcommand{\nArmVarShiftAbaciFreqFiveThousandDiff}{-0.0002\xspace}
\newcommand{\nArmVarShiftAbaciFreqFiveThousandHi}{-0.0002\xspace}
\newcommand{\nArmVarShiftAbaciFreqFiveThousandLo}{-0.0002\xspace}

\newcommand{\nArmVarShiftAbaciFreqTenDiff}{-0.0509\xspace}
\newcommand{\nArmVarShiftAbaciFreqTenHi}{-0.0466\xspace}
\newcommand{\nArmVarShiftAbaciFreqTenLo}{-0.0553\xspace}

\newcommand{\nArmVarShiftAbaciFreqThousandDiff}{-0.0009\xspace}
\newcommand{\nArmVarShiftAbaciFreqThousandHi}{-0.0009\xspace}
\newcommand{\nArmVarShiftAbaciFreqThousandLo}{-0.0010\xspace}

\newcommand{\nArmVarShiftAbaciFreqTwentyFiveDiff}{-0.0350\xspace}
\newcommand{\nArmVarShiftAbaciFreqTwentyFiveHi}{-0.0332\xspace}
\newcommand{\nArmVarShiftAbaciFreqTwentyFiveLo}{-0.0370\xspace}

\newcommand{\nArmVarShiftAbaciFreqTwoHundredDiff}{-0.0050\xspace}
\newcommand{\nArmVarShiftAbaciFreqTwoHundredHi}{-0.0047\xspace}
\newcommand{\nArmVarShiftAbaciFreqTwoHundredLo}{-0.0053\xspace}

\newcommand{\nArmVarShiftBoostFreqFiftyDiff}{-0.0188\xspace}
\newcommand{\nArmVarShiftBoostFreqFiftyHi}{-0.0175\xspace}
\newcommand{\nArmVarShiftBoostFreqFiftyLo}{-0.0200\xspace}

\newcommand{\nArmVarShiftBoostFreqFiveThousandDiff}{-0.0002\xspace}
\newcommand{\nArmVarShiftBoostFreqFiveThousandHi}{-0.0002\xspace}
\newcommand{\nArmVarShiftBoostFreqFiveThousandLo}{-0.0002\xspace}

\newcommand{\nArmVarShiftBoostFreqTenDiff}{-0.0592\xspace}
\newcommand{\nArmVarShiftBoostFreqTenHi}{-0.0554\xspace}
\newcommand{\nArmVarShiftBoostFreqTenLo}{-0.0630\xspace}

\newcommand{\nArmVarShiftBoostFreqThousandDiff}{-0.0009\xspace}
\newcommand{\nArmVarShiftBoostFreqThousandHi}{-0.0009\xspace}
\newcommand{\nArmVarShiftBoostFreqThousandLo}{-0.0010\xspace}

\newcommand{\nArmVarShiftBoostFreqTwentyFiveDiff}{-0.0352\xspace}
\newcommand{\nArmVarShiftBoostFreqTwentyFiveHi}{-0.0334\xspace}
\newcommand{\nArmVarShiftBoostFreqTwentyFiveLo}{-0.0370\xspace}

\newcommand{\nArmVarShiftBoostFreqTwoHundredDiff}{-0.0049\xspace}
\newcommand{\nArmVarShiftBoostFreqTwoHundredHi}{-0.0046\xspace}
\newcommand{\nArmVarShiftBoostFreqTwoHundredLo}{-0.0052\xspace}

\newcommand{\nArmVarShiftMoveFreqFiftyDiff}{0.0038\xspace}
\newcommand{\nArmVarShiftMoveFreqFiftyHi}{0.0046\xspace}
\newcommand{\nArmVarShiftMoveFreqFiftyLo}{0.0031\xspace}

\newcommand{\nArmVarShiftMoveFreqFiveThousandFineDiff}{0.00004\xspace}
\newcommand{\nArmVarShiftMoveFreqFiveThousandFineHi}{0.00005\xspace}
\newcommand{\nArmVarShiftMoveFreqFiveThousandFineLo}{0.00004\xspace}

\newcommand{\nArmVarShiftMoveFreqTenDiff}{0.0142\xspace}
\newcommand{\nArmVarShiftMoveFreqTenHi}{0.0175\xspace}
\newcommand{\nArmVarShiftMoveFreqTenLo}{0.0109\xspace}

\newcommand{\nArmVarShiftMoveFreqThousandDiff}{0.0002\xspace}
\newcommand{\nArmVarShiftMoveFreqThousandHi}{0.0002\xspace}
\newcommand{\nArmVarShiftMoveFreqThousandLo}{0.0002\xspace}

\newcommand{\nArmVarShiftMoveFreqTwentyFiveDiff}{0.0058\xspace}
\newcommand{\nArmVarShiftMoveFreqTwentyFiveHi}{0.0073\xspace}
\newcommand{\nArmVarShiftMoveFreqTwentyFiveLo}{0.0044\xspace}

\newcommand{\nArmVarShiftMoveFreqTwoHundredDiff}{0.0011\xspace}
\newcommand{\nArmVarShiftMoveFreqTwoHundredHi}{0.0012\xspace}
\newcommand{\nArmVarShiftMoveFreqTwoHundredLo}{0.0009\xspace}

\newcommand{\nAssetPjmPageDate}{September 28, 2026\xspace}

\newcommand{\nBandHiPct}{97.5th\xspace}
\newcommand{\nBandLoPct}{2.5th\xspace}

\newcommand{\nBoost}{3.0\xspace}

\newcommand{\nBsAbCovMinOffset}{7\xspace}

\newcommand{\nBsAbCovMinusSeven}{0.8948\xspace}

\newcommand{\nBsAbCovPlusSeven}{0.8740\xspace}

\newcommand{\nBsAbWidthPlusSeven}{12.77\xspace}

\newcommand{\nBsAbWidthZero}{12.93\xspace}

\newcommand{\nBsCovRangePp}{2.08\xspace}

\newcommand{\nBsMpost}{200\xspace}
\newcommand{\nBsOffsetMax}{7\xspace}

\newcommand{\nBsReps}{500\xspace}
\newcommand{\nBsSalt}{110\xspace}
\newcommand{\nBurnin}{10\xspace}

\newcommand{\nCandAdmissionRows}{11\xspace}
\newcommand{\nCandAdmitted}{10\xspace}
\newcommand{\nCandDocuments}{29\xspace}

\newcommand{\nCandEarlier}{5\xspace}

\newcommand{\nCandReturned}{25\xspace}
\newcommand{\nCandReturnedDistinct}{24\xspace}
\newcommand{\nCandRun}{9\xspace}
\newcommand{\nCandUnreachable}{4\xspace}

\newcommand{\nCfErcotAbaciFifty}{0.091\xspace}

\newcommand{\nCfErcotAciFifty}{0.076\xspace}

\newcommand{\nCfErcotAgaciSrcFifty}{0.076\xspace}

\newcommand{\nCfErcotCellAci}{0.046\xspace}

\newcommand{\nCfErcotCellAllAci}{0.048\xspace}

\newcommand{\nCfErcotCellAllSeg}{0.067\xspace}

\newcommand{\nCfErcotCellForecasts}{98\xspace}
\newcommand{\nCfErcotCellLength}{117\xspace}

\newcommand{\nCfErcotCellSeg}{0.069\xspace}

\newcommand{\nCfErcotDtaciSrcFifty}{0.091\xspace}

\newcommand{\nCfErcotMinReference}{20\xspace}

\newcommand{\nCfErcotPairAbaciFiftyHi}{0.000\xspace}
\newcommand{\nCfErcotPairAbaciFiftyLo}{0.000\xspace}

\newcommand{\nCfErcotPairAciFiftyHi}{0.044\xspace}
\newcommand{\nCfErcotPairAciFiftyLo}{-0.029\xspace}

\newcommand{\nCfErcotPairAgaciSrcFiftyHi}{0.044\xspace}
\newcommand{\nCfErcotPairAgaciSrcFiftyLo}{-0.029\xspace}

\newcommand{\nCfErcotPairDtaciSrcFiftyHi}{0.000\xspace}
\newcommand{\nCfErcotPairDtaciSrcFiftyLo}{0.000\xspace}

\newcommand{\nCfErcotPairPidSrcFiftyHi}{-0.021\xspace}
\newcommand{\nCfErcotPairPidSrcFiftyLo}{-0.162\xspace}

\newcommand{\nCfErcotPairSaocpSrcFiftyHi}{-0.132\xspace}
\newcommand{\nCfErcotPairSaocpSrcFiftyLo}{-0.382\xspace}

\newcommand{\nCfErcotPidSrcFifty}{0.194\xspace}

\newcommand{\nCfErcotRefFifty}{68\xspace}

\newcommand{\nCfErcotSaocpSrcFifty}{0.341\xspace}

\newcommand{\nCfErcotSegFifty}{0.091\xspace}

\newcommand{\nCfErcotSideBelowCountWordFifty}{two\xspace}

\newcommand{\nCfErcotSideBelowWhoFifty}{conformal PID and SAOCP\xspace}

\newcommand{\nCfErcotSideIncludeWhoFifty}{ACI and AgACI\xspace}

\newcommand{\nCfErcotSidePointWhoFifty}{DtACI\xspace}

\newcommand{\nCfFomcAbaciTen}{0.085\xspace}

\newcommand{\nCfFomcAciTen}{0.088\xspace}

\newcommand{\nCfFomcAgaciSrcTen}{0.072\xspace}

\newcommand{\nCfFomcBestTen}{0.068\xspace}
\newcommand{\nCfFomcBestTenWho}{DtACI\xspace}

\newcommand{\nCfFomcDtaciSrcTen}{0.068\xspace}

\newcommand{\nCfFomcEndsDtaciSrcTwentyFive}{$[0.003,\, 0.075]$\xspace}
\newcommand{\nCfFomcEndsPidSrcTwentyFive}{$[0.001,\, 0.039]$\xspace}
\newcommand{\nCfFomcEndsSaocpSrcTwentyFive}{$[0.0002,\, 0.028]$\xspace}

\newcommand{\nCfFomcPairAbaciTenHi}{0.072\xspace}
\newcommand{\nCfFomcPairAbaciTenLo}{0.016\xspace}

\newcommand{\nCfFomcPairAciTenHi}{0.072\xspace}
\newcommand{\nCfFomcPairAciTenLo}{0.011\xspace}

\newcommand{\nCfFomcPairAgaciSrcTenHi}{0.088\xspace}
\newcommand{\nCfFomcPairAgaciSrcTenLo}{0.027\xspace}

\newcommand{\nCfFomcPairDtaciSrcTenHi}{0.091\xspace}
\newcommand{\nCfFomcPairDtaciSrcTenLo}{0.032\xspace}

\newcommand{\nCfFomcPairPidSrcTenHi}{0.080\xspace}
\newcommand{\nCfFomcPairPidSrcTenLo}{0.025\xspace}

\newcommand{\nCfFomcPairSaocpSrcTenHi}{0.084\xspace}
\newcommand{\nCfFomcPairSaocpSrcTenLo}{0.030\xspace}

\newcommand{\nCfFomcPidSrcTen}{0.077\xspace}

\newcommand{\nCfFomcSaocpSrcTen}{0.072\xspace}

\newcommand{\nCfFomcSegMultiTen}{0.128\xspace}
\newcommand{\nCfFomcSegMultiTwentyFive}{0.141\xspace}

\newcommand{\nCfFomcSideIncludeEndsTwentyFive}{$[-0.009,\, 0.031]$ and $[-0.004,\, 0.054]$\xspace}
\newcommand{\nCfFomcSideIncludeWhoTwentyFive}{ACI and AgACI\xspace}

\newcommand{\nCfSeed}{1234\xspace}

\newcommand{\nCompArchJobs}{200\xspace}
\newcommand{\nCompArchPostedUsd}{0.61462\xspace}
\newcommand{\nCompArchUsd}{0.54984\xspace}
\newcommand{\nCompArchWall}{31419.6\xspace}
\newcommand{\nCompBillingReadDate}{September 28, 2026\xspace}

\newcommand{\nCompCsixJobs}{10\xspace}
\newcommand{\nCompCsixUsd}{0.00101\xspace}
\newcommand{\nCompCsixWall}{57.7\xspace}

\newcommand{\nCompDoseJobs}{55\xspace}
\newcommand{\nCompDoseUsd}{0.00413\xspace}
\newcommand{\nCompDoseWall}{235.8\xspace}

\newcommand{\nCompErcotJobs}{5\xspace}
\newcommand{\nCompErcotUsd}{0.00253\xspace}
\newcommand{\nCompErcotWall}{144.7\xspace}

\newcommand{\nCompFomcJobs}{3\xspace}
\newcommand{\nCompFomcUsd}{0.00275\xspace}
\newcommand{\nCompFomcWall}{157.1\xspace}
\newcommand{\nCompJobs}{784\xspace}

\newcommand{\nCompLibJobs}{39\xspace}
\newcommand{\nCompLibUsd}{0.03218\xspace}
\newcommand{\nCompLibWall}{1839.0\xspace}
\newcommand{\nCompLocalDeposits}{18\xspace}
\newcommand{\nCompLocalSeconds}{41,677.8\xspace}

\newcommand{\nCompMixJobs}{41\xspace}
\newcommand{\nCompMixUsd}{0.00050\xspace}
\newcommand{\nCompMixWall}{28.3\xspace}
\newcommand{\nCompPostedDate}{September 26, 2026\xspace}
\newcommand{\nCompPostedTotalUsd}{4.57387\xspace}

\newcommand{\nCompPostedUsdFull}{1.35821\xspace}

\newcommand{\nCompProofJobs}{2\xspace}
\newcommand{\nCompProofUsd}{0.00210\xspace}
\newcommand{\nCompProofWall}{120.0\xspace}
\newcommand{\nCompRatePerSecond}{0.0000175\xspace}

\newcommand{\nCompRefitJobs}{27\xspace}
\newcommand{\nCompRefitPostedUsd}{0.06940\xspace}
\newcommand{\nCompRefitUsd}{0.01849\xspace}
\newcommand{\nCompRefitWall}{1056.4\xspace}

\newcommand{\nCompRerunPostedUsd}{0.02704\xspace}

\newcommand{\nCompSourcesPostedUsd}{2.50460\xspace}

\newcommand{\nCompSynJobs}{200\xspace}

\newcommand{\nCompSynSrcJobs}{202\xspace}
\newcommand{\nCompSynSrcUsd}{2.35659\xspace}
\newcommand{\nCompSynSrcWall}{134662.2\xspace}
\newcommand{\nCompSynUsd}{0.56584\xspace}
\newcommand{\nCompSynWall}{32333.7\xspace}
\newcommand{\nCompUsd}{3.53595\xspace}
\newcommand{\nCompWallSeconds}{202054.6\xspace}
\newcommand{\nContrastBothShiftFreqFiftyDiff}{0.0006\xspace}
\newcommand{\nContrastBothShiftFreqFiftyHi}{0.0013\xspace}
\newcommand{\nContrastBothShiftFreqFiftyLo}{-0.0001\xspace}
\newcommand{\nContrastBothShiftFreqFiveThousandDiff}{0.0000\xspace}
\newcommand{\nContrastBothShiftFreqFiveThousandHi}{0.0000\xspace}
\newcommand{\nContrastBothShiftFreqFiveThousandLo}{0.0000\xspace}
\newcommand{\nContrastBothShiftFreqTenDiff}{0.0064\xspace}
\newcommand{\nContrastBothShiftFreqTenHi}{0.0084\xspace}
\newcommand{\nContrastBothShiftFreqTenLo}{0.0044\xspace}
\newcommand{\nContrastBothShiftFreqThousandDiff}{0.0001\xspace}
\newcommand{\nContrastBothShiftFreqThousandHi}{0.0001\xspace}
\newcommand{\nContrastBothShiftFreqThousandLo}{0.0000\xspace}
\newcommand{\nContrastBothShiftFreqTwentyFiveDiff}{0.0016\xspace}
\newcommand{\nContrastBothShiftFreqTwentyFiveHi}{0.0025\xspace}
\newcommand{\nContrastBothShiftFreqTwentyFiveLo}{0.0006\xspace}
\newcommand{\nContrastBothShiftFreqTwoHundredDiff}{0.0003\xspace}
\newcommand{\nContrastBothShiftFreqTwoHundredHi}{0.0005\xspace}
\newcommand{\nContrastBothShiftFreqTwoHundredLo}{0.0001\xspace}

\newcommand{\nContrastFreqAboveTen}{4\xspace}

\newcommand{\nContrastFreqBelowFiftyWord}{two\xspace}

\newcommand{\nContrastFreqReachFiftyWord}{two\xspace}

\newcommand{\nContrastLargeMeanFreqFiftyDiff}{0.0004\xspace}
\newcommand{\nContrastLargeMeanFreqFiftyHi}{0.0010\xspace}
\newcommand{\nContrastLargeMeanFreqFiftyLo}{-0.0003\xspace}
\newcommand{\nContrastLargeMeanFreqFiveThousandDiff}{0.0000\xspace}
\newcommand{\nContrastLargeMeanFreqFiveThousandHi}{0.0000\xspace}
\newcommand{\nContrastLargeMeanFreqFiveThousandLo}{0.0000\xspace}
\newcommand{\nContrastLargeMeanFreqTenDiff}{0.0111\xspace}
\newcommand{\nContrastLargeMeanFreqTenHi}{0.0135\xspace}
\newcommand{\nContrastLargeMeanFreqTenLo}{0.0088\xspace}
\newcommand{\nContrastLargeMeanFreqThousandDiff}{0.0000\xspace}
\newcommand{\nContrastLargeMeanFreqThousandHi}{0.0000\xspace}
\newcommand{\nContrastLargeMeanFreqThousandLo}{0.0000\xspace}
\newcommand{\nContrastLargeMeanFreqTwentyFiveDiff}{-0.0003\xspace}
\newcommand{\nContrastLargeMeanFreqTwentyFiveHi}{0.0010\xspace}
\newcommand{\nContrastLargeMeanFreqTwentyFiveLo}{-0.0015\xspace}
\newcommand{\nContrastLargeMeanFreqTwoHundredDiff}{0.0002\xspace}
\newcommand{\nContrastLargeMeanFreqTwoHundredHi}{0.0003\xspace}
\newcommand{\nContrastLargeMeanFreqTwoHundredLo}{0.0000\xspace}

\newcommand{\nContrastMeanShiftFreqFiftyDiff}{-0.0021\xspace}
\newcommand{\nContrastMeanShiftFreqFiftyHi}{-0.0014\xspace}
\newcommand{\nContrastMeanShiftFreqFiftyLo}{-0.0027\xspace}
\newcommand{\nContrastMeanShiftFreqFiveThousandDiff}{0.0000\xspace}
\newcommand{\nContrastMeanShiftFreqFiveThousandHi}{0.0000\xspace}
\newcommand{\nContrastMeanShiftFreqFiveThousandLo}{0.0000\xspace}
\newcommand{\nContrastMeanShiftFreqTenDiff}{0.0120\xspace}
\newcommand{\nContrastMeanShiftFreqTenHi}{0.0144\xspace}
\newcommand{\nContrastMeanShiftFreqTenLo}{0.0096\xspace}
\newcommand{\nContrastMeanShiftFreqThousandDiff}{-0.0001\xspace}
\newcommand{\nContrastMeanShiftFreqThousandHi}{-0.0001\xspace}
\newcommand{\nContrastMeanShiftFreqThousandLo}{-0.0001\xspace}
\newcommand{\nContrastMeanShiftFreqTwentyFiveDiff}{-0.0008\xspace}
\newcommand{\nContrastMeanShiftFreqTwentyFiveHi}{-0.0001\xspace}
\newcommand{\nContrastMeanShiftFreqTwentyFiveLo}{-0.0016\xspace}
\newcommand{\nContrastMeanShiftFreqTwoHundredDiff}{-0.0005\xspace}
\newcommand{\nContrastMeanShiftFreqTwoHundredHi}{-0.0003\xspace}
\newcommand{\nContrastMeanShiftFreqTwoHundredLo}{-0.0006\xspace}
\newcommand{\nContrastMeanShiftPtFiftyDiff}{-0.0021\xspace}
\newcommand{\nContrastMeanShiftPtFiftyHi}{0.0013\xspace}
\newcommand{\nContrastMeanShiftPtFiftyLo}{-0.0055\xspace}

\newcommand{\nContrastMeanShiftPtTenDiff}{0.0294\xspace}
\newcommand{\nContrastMeanShiftPtTenHi}{0.0362\xspace}
\newcommand{\nContrastMeanShiftPtTenLo}{0.0227\xspace}

\newcommand{\nContrastNoBreakFreqFiftyDiff}{-0.0008\xspace}
\newcommand{\nContrastNoBreakFreqFiftyHi}{-0.0001\xspace}
\newcommand{\nContrastNoBreakFreqFiftyLo}{-0.0014\xspace}
\newcommand{\nContrastNoBreakFreqFiveThousandDiff}{0.0000\xspace}
\newcommand{\nContrastNoBreakFreqFiveThousandHi}{0.0000\xspace}
\newcommand{\nContrastNoBreakFreqFiveThousandLo}{0.0000\xspace}
\newcommand{\nContrastNoBreakFreqTenDiff}{-0.0014\xspace}
\newcommand{\nContrastNoBreakFreqTenHi}{0.0001\xspace}
\newcommand{\nContrastNoBreakFreqTenLo}{-0.0029\xspace}
\newcommand{\nContrastNoBreakFreqThousandDiff}{0.0000\xspace}
\newcommand{\nContrastNoBreakFreqThousandHi}{0.0000\xspace}
\newcommand{\nContrastNoBreakFreqThousandLo}{-0.0001\xspace}
\newcommand{\nContrastNoBreakFreqTwentyFiveDiff}{-0.0013\xspace}
\newcommand{\nContrastNoBreakFreqTwentyFiveHi}{-0.0008\xspace}
\newcommand{\nContrastNoBreakFreqTwentyFiveLo}{-0.0019\xspace}
\newcommand{\nContrastNoBreakFreqTwoHundredDiff}{-0.0003\xspace}
\newcommand{\nContrastNoBreakFreqTwoHundredHi}{-0.0001\xspace}
\newcommand{\nContrastNoBreakFreqTwoHundredLo}{-0.0005\xspace}

\newcommand{\nContrastVarShiftFreqFiftyDiff}{-0.0009\xspace}
\newcommand{\nContrastVarShiftFreqFiftyHi}{-0.0002\xspace}
\newcommand{\nContrastVarShiftFreqFiftyLo}{-0.0016\xspace}
\newcommand{\nContrastVarShiftFreqFiveThousandDiff}{0.0000\xspace}
\newcommand{\nContrastVarShiftFreqFiveThousandHi}{0.0000\xspace}
\newcommand{\nContrastVarShiftFreqFiveThousandLo}{0.0000\xspace}
\newcommand{\nContrastVarShiftFreqTenDiff}{0.0083\xspace}
\newcommand{\nContrastVarShiftFreqTenHi}{0.0105\xspace}
\newcommand{\nContrastVarShiftFreqTenLo}{0.0061\xspace}
\newcommand{\nContrastVarShiftFreqThousandDiff}{0.0000\xspace}
\newcommand{\nContrastVarShiftFreqThousandHi}{0.0000\xspace}
\newcommand{\nContrastVarShiftFreqThousandLo}{0.0000\xspace}
\newcommand{\nContrastVarShiftFreqTwentyFiveDiff}{0.0002\xspace}
\newcommand{\nContrastVarShiftFreqTwentyFiveHi}{0.0010\xspace}
\newcommand{\nContrastVarShiftFreqTwentyFiveLo}{-0.0006\xspace}
\newcommand{\nContrastVarShiftFreqTwoHundredDiff}{0.0000\xspace}
\newcommand{\nContrastVarShiftFreqTwoHundredHi}{0.0001\xspace}
\newcommand{\nContrastVarShiftFreqTwoHundredLo}{-0.0002\xspace}

\newcommand{\nCoreRequest}{1.0\xspace}

\newcommand{\nCountFifty}{50\xspace}
\newcommand{\nCountFiveThousand}{5,000\xspace}
\newcommand{\nCountTen}{10\xspace}
\newcommand{\nCountThousand}{1,000\xspace}
\newcommand{\nCountTwentyFive}{25\xspace}
\newcommand{\nCountTwoHundred}{200\xspace}
\newcommand{\nCountsDeclared}{6\xspace}
\newcommand{\nDecConstant}{1.0\xspace}
\newcommand{\nDecExponent}{0.5\xspace}

\newcommand{\nDoseFirstGainQ}{1.89\xspace}

\newcommand{\nDoseGridMaxDelta}{10\xspace}
\newcommand{\nDoseGridPoints}{11\xspace}
\newcommand{\nDoseLastNoGainQ}{1.06\xspace}

\newcommand{\nDosePeakDelta}{2.0\xspace}

\newcommand{\nDosePeakQ}{2.72\xspace}
\newcommand{\nDosePeakRedAbaci}{74.4\xspace}
\newcommand{\nDosePeakRedSe}{1.55\xspace}

\newcommand{\nDoseRedPtSegOne}{55.0\xspace}

\newcommand{\nDoseRedPtSegTwoHalf}{34.8\xspace}

\newcommand{\nDoseReps}{500\xspace}
\newcommand{\nDoseSalt}{105\xspace}
\newcommand{\nDoseTtotal}{600\xspace}
\newcommand{\nDtaciEta}{0.5\xspace}
\newcommand{\nDtaciSigmaDen}{200\xspace}

\newcommand{\nErcotBlock}{7\xspace}
\newcommand{\nErcotBoost}{3.0\xspace}
\newcommand{\nErcotBoostExcessAbaci}{0.0069\xspace}
\newcommand{\nErcotBoostExcessAbaciText}{0.0127\xspace}
\newcommand{\nErcotBoundAbaci}{0.0392\xspace}
\newcommand{\nErcotBoundAbaciText}{0.0250\xspace}
\newcommand{\nErcotBoundRatioAbaci}{3.52\xspace}
\newcommand{\nErcotBoundRatioAbaciText}{9.76\xspace}
\newcommand{\nErcotBreakDate}{December 5, 2025\xspace}

\newcommand{\nErcotBurnin}{10\xspace}
\newcommand{\nErcotCalib}{50\xspace}
\newcommand{\nErcotCalibEndIdx}{99\xspace}
\newcommand{\nErcotCalibStartIdx}{50\xspace}

\newcommand{\nErcotCiHiAbaci}{0.000\xspace}

\newcommand{\nErcotCiHiAgaciSrc}{0.094\xspace}

\newcommand{\nErcotCiHiDtaciSrc}{0.085\xspace}

\newcommand{\nErcotCiHiNaive}{-0.017\xspace}

\newcommand{\nErcotCiHiPidSrc}{0.077\xspace}

\newcommand{\nErcotCiHiSaocpSrc}{0.077\xspace}
\newcommand{\nErcotCiHiSeg}{0.068\xspace}
\newcommand{\nErcotCiHiSegWarm}{0.068\xspace}

\newcommand{\nErcotCiLoAbaci}{-0.026\xspace}

\newcommand{\nErcotCiLoAgaciSrc}{0.000\xspace}

\newcommand{\nErcotCiLoDtaciSrc}{-0.017\xspace}

\newcommand{\nErcotCiLoNaive}{-0.111\xspace}

\newcommand{\nErcotCiLoPidSrc}{-0.043\xspace}

\newcommand{\nErcotCiLoSaocpSrc}{-0.085\xspace}
\newcommand{\nErcotCiLoSeg}{-0.043\xspace}
\newcommand{\nErcotCiLoSegWarm}{-0.043\xspace}

\newcommand{\nErcotCovAbaci}{0.889\xspace}

\newcommand{\nErcotCovAci}{0.897\xspace}
\newcommand{\nErcotCovAgaci}{0.915\xspace}
\newcommand{\nErcotCovAgaciSrc}{0.932\xspace}

\newcommand{\nErcotCovDtaci}{0.863\xspace}
\newcommand{\nErcotCovDtaciSrc}{0.923\xspace}

\newcommand{\nErcotCovNaive}{0.838\xspace}
\newcommand{\nErcotCovPid}{0.761\xspace}
\newcommand{\nErcotCovPidSrc}{0.906\xspace}
\newcommand{\nErcotCovSaocp}{0.812\xspace}
\newcommand{\nErcotCovSaocpFiled}{0.778\xspace}
\newcommand{\nErcotCovSaocpSrc}{0.889\xspace}
\newcommand{\nErcotCovSeg}{0.897\xspace}
\newcommand{\nErcotCovSegWarm}{0.897\xspace}

\newcommand{\nErcotDiffAbaci}{-0.009\xspace}

\newcommand{\nErcotDiffAgaciSrc}{0.034\xspace}

\newcommand{\nErcotDiffDtaciSrc}{0.026\xspace}

\newcommand{\nErcotDiffNaive}{-0.060\xspace}

\newcommand{\nErcotDiffPidSrc}{0.009\xspace}

\newcommand{\nErcotDiffSaocpSrc}{-0.009\xspace}
\newcommand{\nErcotDiffSeg}{0.000\xspace}
\newcommand{\nErcotDiffSegWarm}{0.000\xspace}

\newcommand{\nErcotDose}{0.46\xspace}

\newcommand{\nErcotDoseUsd}{4.89\xspace}
\newcommand{\nErcotDraws}{10,000\xspace}
\newcommand{\nErcotEval}{204\xspace}
\newcommand{\nErcotEvalStart}{100\xspace}

\newcommand{\nErcotFiledCovAbaci}{0.889\xspace}

\newcommand{\nErcotFiledCovAci}{0.906\xspace}
\newcommand{\nErcotFiledCovAgaci}{0.897\xspace}
\newcommand{\nErcotFiledCovDtaci}{0.872\xspace}

\newcommand{\nErcotFiledCovNaive}{0.855\xspace}
\newcommand{\nErcotFiledCovPid}{0.692\xspace}
\newcommand{\nErcotFiledCovSaocp}{0.744\xspace}
\newcommand{\nErcotFiledCovSaocpFiled}{0.709\xspace}
\newcommand{\nErcotFiledCovSeg}{0.889\xspace}

\newcommand{\nErcotFirstDate}{June 1, 2025\xspace}

\newcommand{\nErcotFloorDraws}{1,000\xspace}

\newcommand{\nErcotFreqErrAbaci}{0.0111\xspace}
\newcommand{\nErcotFreqErrAbaciText}{0.0026\xspace}
\newcommand{\nErcotGamma}{2.0\xspace}
\newcommand{\nErcotHubSdPost}{35.8\xspace}
\newcommand{\nErcotHubSdPre}{13.6\xspace}
\newcommand{\nErcotInfAbaci}{0\xspace}

\newcommand{\nErcotInfAci}{0\xspace}

\newcommand{\nErcotInfAgaciSrc}{0\xspace}

\newcommand{\nErcotInfDtaciSrc}{0\xspace}

\newcommand{\nErcotInfNaive}{0\xspace}

\newcommand{\nErcotInfPidSrc}{0\xspace}

\newcommand{\nErcotInfSaocpSrc}{0\xspace}
\newcommand{\nErcotInfSeg}{2\xspace}
\newcommand{\nErcotInfSegWarm}{0\xspace}
\newcommand{\nErcotInfSegWord}{two\xspace}

\newcommand{\nErcotJumpAbaci}{4.94\xspace}
\newcommand{\nErcotJumpAbaciText}{4.72\xspace}
\newcommand{\nErcotLastDate}{March 31, 2026\xspace}

\newcommand{\nErcotLastIdx}{303\xspace}

\newcommand{\nErcotMaskedCovAbaci}{0.863\xspace}

\newcommand{\nErcotMaskedCovAci}{0.880\xspace}
\newcommand{\nErcotMaskedCovAgaci}{0.872\xspace}
\newcommand{\nErcotMaskedCovDtaci}{0.855\xspace}

\newcommand{\nErcotMaskedCovNaive}{0.838\xspace}
\newcommand{\nErcotMaskedCovPid}{0.675\xspace}
\newcommand{\nErcotMaskedCovSaocp}{0.726\xspace}
\newcommand{\nErcotMaskedCovSaocpFiled}{0.692\xspace}
\newcommand{\nErcotMaskedCovSeg}{0.863\xspace}

\newcommand{\nErcotMaskedDays}{3\xspace}

\newcommand{\nErcotMaskedFirstDateIso}{2026-03-29\xspace}

\newcommand{\nErcotPAbaci}{0.68\xspace}

\newcommand{\nErcotPAgaciSrc}{0.14\xspace}

\newcommand{\nErcotPDtaciSrc}{0.40\xspace}

\newcommand{\nErcotPNaive}{0.0062\xspace}

\newcommand{\nErcotPPidSrc}{0.93\xspace}

\newcommand{\nErcotPSaocpSrc}{0.88\xspace}
\newcommand{\nErcotPSeg}{1.00\xspace}
\newcommand{\nErcotPSegWarm}{1.00\xspace}

\newcommand{\nErcotPiSumAbaci}{0.0\xspace}
\newcommand{\nErcotPiSumAbaciText}{0.0\xspace}
\newcommand{\nErcotPost}{117\xspace}
\newcommand{\nErcotPre}{87\xspace}

\newcommand{\nErcotProjShareAbaci}{0.000\xspace}
\newcommand{\nErcotProjShareAbaciText}{0.000\xspace}

\newcommand{\nErcotRealignedCovAbaci}{0.880\xspace}

\newcommand{\nErcotRealignedCovAci}{0.897\xspace}
\newcommand{\nErcotRealignedCovAgaci}{0.906\xspace}
\newcommand{\nErcotRealignedCovDtaci}{0.872\xspace}

\newcommand{\nErcotRealignedCovNaive}{0.855\xspace}
\newcommand{\nErcotRealignedCovPid}{0.675\xspace}
\newcommand{\nErcotRealignedCovSaocp}{0.726\xspace}
\newcommand{\nErcotRealignedCovSaocpFiled}{0.684\xspace}
\newcommand{\nErcotRealignedCovSeg}{0.889\xspace}

\newcommand{\nErcotRows}{304\xspace}

\newcommand{\nErcotStreamRows}{371\xspace}

\newcommand{\nErcotTrainDays}{50\xspace}
\newcommand{\nErcotTrainEndIdx}{49\xspace}
\newcommand{\nErcotTwinAgaciSrcDiff}{0.017\xspace}

\newcommand{\nErcotTwinAgaciSrcHi}{0.043\xspace}
\newcommand{\nErcotTwinAgaciSrcLo}{0.000\xspace}
\newcommand{\nErcotTwinAgaciSrcP}{0.22\xspace}
\newcommand{\nErcotTwinDtaciSrcDiff}{0.060\xspace}

\newcommand{\nErcotTwinDtaciSrcHi}{0.137\xspace}
\newcommand{\nErcotTwinDtaciSrcLo}{0.009\xspace}
\newcommand{\nErcotTwinDtaciSrcP}{0.02\xspace}
\newcommand{\nErcotTwinPidSrcDiff}{0.145\xspace}

\newcommand{\nErcotTwinPidSrcHi}{0.231\xspace}
\newcommand{\nErcotTwinPidSrcLo}{0.060\xspace}
\newcommand{\nErcotTwinPidSrcP}{0.0002\xspace}
\newcommand{\nErcotTwinSaocpSrcDiff}{0.077\xspace}

\newcommand{\nErcotTwinSaocpSrcHi}{0.128\xspace}
\newcommand{\nErcotTwinSaocpSrcLo}{0.026\xspace}
\newcommand{\nErcotTwinSaocpSrcP}{0.0060\xspace}
\newcommand{\nErcotWidthAbaci}{55.92\xspace}

\newcommand{\nErcotWidthAci}{60.43\xspace}

\newcommand{\nErcotWidthAgaciSrc}{102.16\xspace}

\newcommand{\nErcotWidthDtaciSrc}{66.92\xspace}

\newcommand{\nErcotWidthNaive}{46.49\xspace}

\newcommand{\nErcotWidthPidSrc}{81.65\xspace}

\newcommand{\nErcotWidthSaocpSrc}{64.77\xspace}
\newcommand{\nErcotWidthSeg}{65.89\xspace}
\newcommand{\nErcotWidthSegWarm}{65.44\xspace}

\newcommand{\nExtAciGamma}{0.5\xspace}
\newcommand{\nExtCalib}{100\xspace}

\newcommand{\nExtHeavyAbaciFiledPost}{0.891\xspace}

\newcommand{\nExtHeavyAbaciPost}{0.892\xspace}
\newcommand{\nExtHeavyAbaciSe}{0.0007\xspace}

\newcommand{\nExtHeavyAciPost}{0.888\xspace}
\newcommand{\nExtHeavyAciSe}{0.0004\xspace}

\newcommand{\nExtHeavyDtaciPost}{0.893\xspace}
\newcommand{\nExtHeavyDtaciSe}{0.0006\xspace}

\newcommand{\nExtHeavyNaivePost}{0.771\xspace}
\newcommand{\nExtHeavyNaiveSe}{0.0020\xspace}

\newcommand{\nExtHeavyPidPost}{0.897\xspace}
\newcommand{\nExtHeavyPidSe}{0.0012\xspace}

\newcommand{\nExtHeavySaocpPost}{0.701\xspace}
\newcommand{\nExtHeavySaocpSe}{0.0048\xspace}

\newcommand{\nExtHeavySegCpPost}{0.889\xspace}
\newcommand{\nExtHeavySegCpSe}{0.0017\xspace}
\newcommand{\nExtHeavySegMultiPost}{0.889\xspace}
\newcommand{\nExtHeavySegMultiSe}{0.0017\xspace}
\newcommand{\nExtPostMean}{2.0\xspace}
\newcommand{\nExtPostScale}{1.5\xspace}
\newcommand{\nExtRangeMax}{0.918\xspace}
\newcommand{\nExtRangeMin}{0.888\xspace}
\newcommand{\nExtReps}{100\xspace}
\newcommand{\nExtSaocpLifetimeLog}{8\xspace}
\newcommand{\nExtSteps}{800\xspace}
\newcommand{\nExtTailDof}{3\xspace}
\newcommand{\nExtTau}{400\xspace}

\newcommand{\nExtTwoAbaciFiledPost}{0.896\xspace}

\newcommand{\nExtTwoAbaciPost}{0.896\xspace}
\newcommand{\nExtTwoAbaciSe}{0.0002\xspace}

\newcommand{\nExtTwoAciPost}{0.898\xspace}
\newcommand{\nExtTwoAciSe}{0.0002\xspace}

\newcommand{\nExtTwoDtaciPost}{0.903\xspace}
\newcommand{\nExtTwoDtaciSe}{0.0006\xspace}
\newcommand{\nExtTwoMeanOne}{1.5\xspace}
\newcommand{\nExtTwoMeanTwo}{-1.0\xspace}

\newcommand{\nExtTwoNaivePost}{0.841\xspace}
\newcommand{\nExtTwoNaiveSe}{0.0014\xspace}

\newcommand{\nExtTwoPidPost}{0.898\xspace}
\newcommand{\nExtTwoPidSe}{0.0005\xspace}

\newcommand{\nExtTwoSaocpPost}{0.754\xspace}
\newcommand{\nExtTwoSaocpSe}{0.0035\xspace}

\newcommand{\nExtTwoSegCpPost}{0.918\xspace}
\newcommand{\nExtTwoSegCpSe}{0.0012\xspace}
\newcommand{\nExtTwoSegMultiPost}{0.889\xspace}
\newcommand{\nExtTwoSegMultiSe}{0.0013\xspace}
\newcommand{\nExtTwoSteps}{900\xspace}
\newcommand{\nExtTwoTauOne}{300\xspace}
\newcommand{\nExtTwoTauTwo}{600\xspace}

\newcommand{\nFalsLmResetGap}{0.035\xspace}

\newcommand{\nFcFiledStreams}{9\xspace}

\newcommand{\nFigSkipOrd}{25th\xspace}

\newcommand{\nFiledErcotCovAbaci}{0.863\xspace}
\newcommand{\nFiledErcotCovAci}{0.880\xspace}
\newcommand{\nFiledErcotCovAgaci}{0.872\xspace}
\newcommand{\nFiledErcotCovDtaci}{0.855\xspace}
\newcommand{\nFiledErcotCovNaive}{0.838\xspace}
\newcommand{\nFiledErcotCovPid}{0.675\xspace}
\newcommand{\nFiledErcotCovSaocp}{0.692\xspace}
\newcommand{\nFiledErcotCovSeg}{0.863\xspace}

\newcommand{\nFiledFomcCiHi}{0.023\xspace}
\newcommand{\nFiledFomcCiLo}{-0.040\xspace}

\newcommand{\nFitFiledConstant}{31.15\xspace}

\newcommand{\nFitFiledConstantRelDiff}{$6.5\times 10^{-9}$\xspace}
\newcommand{\nFitFiledMinChild}{20\xspace}
\newcommand{\nFitFiledPairs}{36\xspace}

\newcommand{\nFitRealignedConstant}{31.49\xspace}

\newcommand{\nFitRealignedPairs}{35\xspace}

\newcommand{\nFitRowsSplitNeeds}{40\xspace}
\newcommand{\nFloorCritPct}{95th\xspace}

\newcommand{\nFloorErcotAbaci}{0.025\xspace}

\newcommand{\nFloorErcotAgaciSrc}{0.067\xspace}

\newcommand{\nFloorErcotDtaciSrc}{0.077\xspace}

\newcommand{\nFloorErcotNaive}{0.066\xspace}

\newcommand{\nFloorErcotPidSrc}{0.089\xspace}
\newcommand{\nFloorErcotPrimary}{0.025\xspace}

\newcommand{\nFloorErcotSaocpSrc}{none\xspace}
\newcommand{\nFloorErcotSeg}{0.085\xspace}
\newcommand{\nFloorErcotSegWarm}{0.085\xspace}

\newcommand{\nFloorFomcAbaci}{0.015\xspace}

\newcommand{\nFloorFomcAgaciSrc}{0.023\xspace}

\newcommand{\nFloorFomcDtaciSrc}{0.025\xspace}

\newcommand{\nFloorFomcNaive}{0.041\xspace}

\newcommand{\nFloorFomcPidSrc}{0.018\xspace}
\newcommand{\nFloorFomcPrimary}{0.015\xspace}

\newcommand{\nFloorFomcSaocpSrc}{0.016\xspace}
\newcommand{\nFloorFomcSegCp}{0.040\xspace}
\newcommand{\nFloorFomcSegMulti}{0.026\xspace}
\newcommand{\nFloorFomcSegMultiWarm}{0.025\xspace}

\newcommand{\nFloorFormC}{0.001\xspace}
\newcommand{\nFloorFormsAB}{0.01\xspace}
\newcommand{\nFloorGridList}{0, 0.005, 0.01, 0.02, 0.03, 0.05, 0.075 and 0.1\xspace}

\newcommand{\nFloorGridPoints}{8\xspace}
\newcommand{\nFloorLevelPct}{5\xspace}

\newcommand{\nFloorPowerPct}{80\xspace}
\newcommand{\nFomcArOrder}{5\xspace}
\newcommand{\nFomcBlock}{20\xspace}
\newcommand{\nFomcBoostMean}{0.0000\xspace}

\newcommand{\nFomcBoundMean}{0.0363\xspace}
\newcommand{\nFomcBoundRatio}{1.75\xspace}
\newcommand{\nFomcBreaks}{38\xspace}

\newcommand{\nFomcCalibEnd}{299\xspace}
\newcommand{\nFomcCalibStart}{250\xspace}
\newcommand{\nFomcCellEightDays}{34\xspace}
\newcommand{\nFomcCellEightFirst}{2022-03-16\xspace}
\newcommand{\nFomcCellEighteenDays}{28\xspace}
\newcommand{\nFomcCellEighteenFirst}{2023-06-14\xspace}
\newcommand{\nFomcCellElevenDays}{39\xspace}
\newcommand{\nFomcCellElevenFirst}{2022-07-27\xspace}
\newcommand{\nFomcCellFifteenDays}{34\xspace}
\newcommand{\nFomcCellFifteenFirst}{2023-02-01\xspace}
\newcommand{\nFomcCellFiveDays}{28\xspace}
\newcommand{\nFomcCellFiveFirst}{2021-11-03\xspace}
\newcommand{\nFomcCellFourDays}{29\xspace}
\newcommand{\nFomcCellFourFirst}{2021-09-22\xspace}
\newcommand{\nFomcCellFourteenDays}{32\xspace}
\newcommand{\nFomcCellFourteenFirst}{2022-12-14\xspace}
\newcommand{\nFomcCellMax}{39\xspace}
\newcommand{\nFomcCellMean}{30.8\xspace}
\newcommand{\nFomcCellMeanInt}{31\xspace}
\newcommand{\nFomcCellMin}{15\xspace}
\newcommand{\nFomcCellNineDays}{29\xspace}
\newcommand{\nFomcCellNineFirst}{2022-05-04\xspace}
\newcommand{\nFomcCellNineteenDays}{39\xspace}
\newcommand{\nFomcCellNineteenFirst}{2023-07-26\xspace}
\newcommand{\nFomcCellOneDays}{34\xspace}
\newcommand{\nFomcCellOneFirst}{2021-04-28\xspace}
\newcommand{\nFomcCellSevenDays}{34\xspace}
\newcommand{\nFomcCellSevenFirst}{2022-01-26\xspace}
\newcommand{\nFomcCellSeventeenDays}{29\xspace}
\newcommand{\nFomcCellSeventeenFirst}{2023-05-03\xspace}
\newcommand{\nFomcCellSixDays}{28\xspace}
\newcommand{\nFomcCellSixFirst}{2021-12-15\xspace}
\newcommand{\nFomcCellSixteenDays}{30\xspace}
\newcommand{\nFomcCellSixteenFirst}{2023-03-22\xspace}
\newcommand{\nFomcCellTenDays}{28\xspace}
\newcommand{\nFomcCellTenFirst}{2022-06-15\xspace}
\newcommand{\nFomcCellThirteenDays}{28\xspace}
\newcommand{\nFomcCellThirteenFirst}{2022-11-02\xspace}
\newcommand{\nFomcCellThirtyDays}{27\xspace}
\newcommand{\nFomcCellThirtyEightDays}{15\xspace}
\newcommand{\nFomcCellThirtyEightFirst}{2025-12-10\xspace}
\newcommand{\nFomcCellThirtyFirst}{2024-12-18\xspace}
\newcommand{\nFomcCellThirtyFiveDays}{34\xspace}
\newcommand{\nFomcCellThirtyFiveFirst}{2025-07-30\xspace}
\newcommand{\nFomcCellThirtyFourDays}{28\xspace}
\newcommand{\nFomcCellThirtyFourFirst}{2025-06-18\xspace}
\newcommand{\nFomcCellThirtyOneDays}{34\xspace}
\newcommand{\nFomcCellThirtyOneFirst}{2025-01-29\xspace}
\newcommand{\nFomcCellThirtySevenDays}{28\xspace}
\newcommand{\nFomcCellThirtySevenFirst}{2025-10-29\xspace}
\newcommand{\nFomcCellThirtySixDays}{29\xspace}
\newcommand{\nFomcCellThirtySixFirst}{2025-09-17\xspace}
\newcommand{\nFomcCellThirtyThreeDays}{29\xspace}
\newcommand{\nFomcCellThirtyThreeFirst}{2025-05-07\xspace}
\newcommand{\nFomcCellThirtyTwoDays}{34\xspace}
\newcommand{\nFomcCellThirtyTwoFirst}{2025-03-19\xspace}
\newcommand{\nFomcCellThreeDays}{39\xspace}
\newcommand{\nFomcCellThreeFirst}{2021-07-28\xspace}
\newcommand{\nFomcCellTwelveDays}{29\xspace}
\newcommand{\nFomcCellTwelveFirst}{2022-09-21\xspace}
\newcommand{\nFomcCellTwentyDays}{29\xspace}
\newcommand{\nFomcCellTwentyEightDays}{35\xspace}
\newcommand{\nFomcCellTwentyEightFirst}{2024-09-18\xspace}
\newcommand{\nFomcCellTwentyFirst}{2023-09-20\xspace}
\newcommand{\nFomcCellTwentyFiveDays}{29\xspace}
\newcommand{\nFomcCellTwentyFiveFirst}{2024-05-01\xspace}
\newcommand{\nFomcCellTwentyFourDays}{29\xspace}
\newcommand{\nFomcCellTwentyFourFirst}{2024-03-20\xspace}
\newcommand{\nFomcCellTwentyNineDays}{27\xspace}
\newcommand{\nFomcCellTwentyNineFirst}{2024-11-07\xspace}
\newcommand{\nFomcCellTwentyOneDays}{29\xspace}
\newcommand{\nFomcCellTwentyOneFirst}{2023-11-01\xspace}
\newcommand{\nFomcCellTwentySevenDays}{34\xspace}
\newcommand{\nFomcCellTwentySevenFirst}{2024-07-31\xspace}
\newcommand{\nFomcCellTwentySixDays}{33\xspace}
\newcommand{\nFomcCellTwentySixFirst}{2024-06-12\xspace}
\newcommand{\nFomcCellTwentyThreeDays}{34\xspace}
\newcommand{\nFomcCellTwentyThreeFirst}{2024-01-31\xspace}
\newcommand{\nFomcCellTwentyTwoDays}{32\xspace}
\newcommand{\nFomcCellTwentyTwoFirst}{2023-12-13\xspace}
\newcommand{\nFomcCellTwoDays}{29\xspace}
\newcommand{\nFomcCellTwoFirst}{2021-06-16\xspace}

\newcommand{\nFomcCellsTotal}{38\xspace}

\newcommand{\nFomcCertHolds}{38\xspace}

\newcommand{\nFomcCiHiAbaci}{-0.001\xspace}

\newcommand{\nFomcCiHiAgaciSrc}{0.016\xspace}

\newcommand{\nFomcCiHiDtaciSrc}{0.023\xspace}

\newcommand{\nFomcCiHiNaive}{-0.005\xspace}

\newcommand{\nFomcCiHiPidSrc}{0.010\xspace}

\newcommand{\nFomcCiHiSaocpSrc}{0.014\xspace}
\newcommand{\nFomcCiHiSegCp}{0.000\xspace}
\newcommand{\nFomcCiHiSegMulti}{-0.022\xspace}
\newcommand{\nFomcCiHiSegMultiWarm}{-0.033\xspace}

\newcommand{\nFomcCiLoAbaci}{-0.018\xspace}

\newcommand{\nFomcCiLoAgaciSrc}{-0.014\xspace}

\newcommand{\nFomcCiLoDtaciSrc}{-0.007\xspace}

\newcommand{\nFomcCiLoNaive}{-0.058\xspace}

\newcommand{\nFomcCiLoPidSrc}{-0.013\xspace}

\newcommand{\nFomcCiLoSaocpSrc}{-0.006\xspace}
\newcommand{\nFomcCiLoSegCp}{-0.049\xspace}
\newcommand{\nFomcCiLoSegMulti}{-0.056\xspace}
\newcommand{\nFomcCiLoSegMultiWarm}{-0.064\xspace}

\newcommand{\nFomcCovAbaci}{0.891\xspace}

\newcommand{\nFomcCovAci}{0.900\xspace}

\newcommand{\nFomcCovAgaciSrc}{0.902\xspace}

\newcommand{\nFomcCovDtaciSrc}{0.910\xspace}

\newcommand{\nFomcCovNaive}{0.870\xspace}

\newcommand{\nFomcCovPidSrc}{0.900\xspace}

\newcommand{\nFomcCovSaocpSrc}{0.903\xspace}
\newcommand{\nFomcCovSegCp}{0.877\xspace}
\newcommand{\nFomcCovSegMulti}{0.862\xspace}
\newcommand{\nFomcCovSegMultiWarm}{0.852\xspace}
\newcommand{\nFomcDates}{49\xspace}
\newcommand{\nFomcDatesBefore}{11\xspace}
\newcommand{\nFomcDayCellsTen}{38\xspace}
\newcommand{\nFomcDayCellsTwentyFive}{37\xspace}
\newcommand{\nFomcDayRefMedianTen}{20\xspace}
\newcommand{\nFomcDayRefMedianTwentyFive}{5\xspace}
\newcommand{\nFomcDayRefSmallestTen}{6\xspace}

\newcommand{\nFomcDiffAbaci}{-0.009\xspace}

\newcommand{\nFomcDiffAgaciSrc}{0.002\xspace}

\newcommand{\nFomcDiffDtaciSrc}{0.010\xspace}

\newcommand{\nFomcDiffNaive}{-0.030\xspace}

\newcommand{\nFomcDiffPidSrc}{0.000\xspace}

\newcommand{\nFomcDiffSaocpSrc}{0.003\xspace}
\newcommand{\nFomcDiffSegCp}{-0.023\xspace}
\newcommand{\nFomcDiffSegMulti}{-0.038\xspace}
\newcommand{\nFomcDiffSegMultiWarm}{-0.048\xspace}

\newcommand{\nFomcDifferences}{1,499\xspace}
\newcommand{\nFomcDose}{0.47\xspace}

\newcommand{\nFomcDraws}{10,000\xspace}

\newcommand{\nFomcEval}{1,199\xspace}
\newcommand{\nFomcEvalStart}{300\xspace}
\newcommand{\nFomcFiledScriptBlock}{20\xspace}
\newcommand{\nFomcFiledScriptDraws}{5,000\xspace}
\newcommand{\nFomcFiledScriptSeed}{42\xspace}
\newcommand{\nFomcFirstYear}{2020\xspace}
\newcommand{\nFomcFreqErrMean}{0.0207\xspace}

\newcommand{\nFomcGamma}{0.05\xspace}
\newcommand{\nFomcInfAbaci}{0\xspace}
\newcommand{\nFomcInfAci}{0\xspace}

\newcommand{\nFomcInfAgaciSrc}{0\xspace}

\newcommand{\nFomcInfDtaciSrc}{0\xspace}

\newcommand{\nFomcInfNaive}{0\xspace}

\newcommand{\nFomcInfPidSrc}{0\xspace}

\newcommand{\nFomcInfSaocpSrc}{0\xspace}
\newcommand{\nFomcInfSegCp}{2\xspace}
\newcommand{\nFomcInfSegMulti}{76\xspace}
\newcommand{\nFomcInfSegMultiWarm}{0\xspace}

\newcommand{\nFomcJumpMean}{0.0220\xspace}

\newcommand{\nFomcLastYear}{2025\xspace}

\newcommand{\nFomcMinReference}{20\xspace}
\newcommand{\nFomcMissingDays}{65\xspace}

\newcommand{\nFomcPiSumMean}{0.0\xspace}
\newcommand{\nFomcPreCellDays}{30\xspace}
\newcommand{\nFomcPreCellFirst}{2021-03-17\xspace}

\newcommand{\nFomcProjShareMean}{0.000\xspace}

\newcommand{\nFomcRawRows}{1,565\xspace}

\newcommand{\nFomcRerunAgree}{3,676\xspace}

\newcommand{\nFomcRerunP}{0.04\xspace}

\newcommand{\nFomcTrain}{200\xspace}
\newcommand{\nFomcTwinAgaciSrcDiff}{-0.018\xspace}

\newcommand{\nFomcTwinAgaciSrcHi}{-0.003\xspace}
\newcommand{\nFomcTwinAgaciSrcLo}{-0.035\xspace}
\newcommand{\nFomcTwinAgaciSrcP}{0.03\xspace}
\newcommand{\nFomcTwinDtaciSrcDiff}{-0.071\xspace}

\newcommand{\nFomcTwinDtaciSrcHi}{-0.057\xspace}
\newcommand{\nFomcTwinDtaciSrcLo}{-0.090\xspace}
\newcommand{\nFomcTwinDtaciSrcP}{0.0002\xspace}
\newcommand{\nFomcTwinPidSrcDiff}{0.000\xspace}

\newcommand{\nFomcTwinPidSrcHi}{0.010\xspace}
\newcommand{\nFomcTwinPidSrcLo}{-0.014\xspace}
\newcommand{\nFomcTwinPidSrcP}{0.83\xspace}
\newcommand{\nFomcTwinSaocpSrcDiff}{0.082\xspace}

\newcommand{\nFomcTwinSaocpSrcHi}{0.104\xspace}
\newcommand{\nFomcTwinSaocpSrcLo}{0.064\xspace}
\newcommand{\nFomcTwinSaocpSrcP}{0.0002\xspace}
\newcommand{\nFomcVintRerunBlock}{20\xspace}
\newcommand{\nFomcVintRerunDiff}{-0.0092\xspace}
\newcommand{\nFomcVintRerunDraws}{10,000\xspace}
\newcommand{\nFomcVintRerunHi}{-0.0008\xspace}
\newcommand{\nFomcVintRerunLo}{-0.0175\xspace}
\newcommand{\nFomcVintRerunP}{0.04\xspace}
\newcommand{\nFomcVintRerunSeed}{1234\xspace}

\newcommand{\nFomcVintWaveBlock}{20\xspace}
\newcommand{\nFomcVintWaveDiff}{-0.0083\xspace}
\newcommand{\nFomcVintWaveDraws}{10,000\xspace}
\newcommand{\nFomcVintWaveHi}{0.0000\xspace}
\newcommand{\nFomcVintWaveLo}{-0.0175\xspace}
\newcommand{\nFomcVintWaveP}{0.07\xspace}
\newcommand{\nFomcVintWaveSeed}{1234\xspace}

\newcommand{\nFomcWidthAbaci}{0.218\xspace}

\newcommand{\nFomcWidthAci}{0.225\xspace}

\newcommand{\nFomcWidthAgaciSrc}{0.216\xspace}

\newcommand{\nFomcWidthDtaciSrc}{0.212\xspace}

\newcommand{\nFomcWidthNaive}{0.198\xspace}

\newcommand{\nFomcWidthPidSrc}{0.209\xspace}

\newcommand{\nFomcWidthSaocpSrc}{0.219\xspace}
\newcommand{\nFomcWidthSegCp}{none\xspace}
\newcommand{\nFomcWidthSegMulti}{none\xspace}
\newcommand{\nFomcWidthSegMultiWarm}{0.193\xspace}

\newcommand{\nFreqFiftyAbaciLeads}{4\xspace}
\newcommand{\nFreqFiftyMethods}{17\xspace}

\newcommand{\nFreqLargeMeanAbaciTextFifty}{0.0876\xspace}

\newcommand{\nFreqLargeMeanAciFifty}{0.0922\xspace}

\newcommand{\nFreqLargeMeanSegFifty}{0.0472\xspace}

\newcommand{\nFreqMeanShiftAbaciFifty}{0.0081\xspace}

\newcommand{\nFreqMeanShiftAciFifty}{0.0322\xspace}

\newcommand{\nFreqMeanShiftDtaciSrcFifty}{0.2196\xspace}

\newcommand{\nFreqMeanShiftSaocpSrcFifty}{0.0566\xspace}

\newcommand{\nFreqMeanShiftSegFifty}{0.0489\xspace}

\newcommand{\nFrontErcotWidthAci}{60.43\xspace}

\newcommand{\nFrontErcotWidthAgaciSrc}{102.16\xspace}

\newcommand{\nFrontErcotWidthSaocpSrc}{64.77\xspace}

\newcommand{\nGenCovCoef}{0.5\xspace}
\newcommand{\nGenPeriod}{24\xspace}

\newcommand{\nGridAgaciFomcList}{0.005, 0.01, 0.05, 0.1 and 0.2\xspace}

\newcommand{\nGridAgaciList}{0.1, 0.5, 1.0, 2.0, 5.0 and 10.0\xspace}

\newcommand{\nGridDtaciList}{0.001, 0.005, 0.01, 0.05, 0.1, 0.5, 1.0 and 2.0\xspace}

\newcommand{\nGridSaocpList}{0.005, 0.02, 0.05, 0.2, 0.5 and 1.0\xspace}

\newcommand{\nInfGap}{0.1\xspace}

\newcommand{\nInteractBothShiftFreqFiftyDiff}{-0.0032\xspace}
\newcommand{\nInteractBothShiftFreqFiftyHi}{-0.0024\xspace}
\newcommand{\nInteractBothShiftFreqFiftyLo}{-0.0041\xspace}
\newcommand{\nInteractBothShiftFreqFiveThousandDiff}{0.0000\xspace}
\newcommand{\nInteractBothShiftFreqFiveThousandHi}{0.0000\xspace}
\newcommand{\nInteractBothShiftFreqFiveThousandLo}{0.0000\xspace}
\newcommand{\nInteractBothShiftFreqTenDiff}{-0.0092\xspace}
\newcommand{\nInteractBothShiftFreqTenHi}{-0.0060\xspace}
\newcommand{\nInteractBothShiftFreqTenLo}{-0.0124\xspace}
\newcommand{\nInteractBothShiftFreqThousandDiff}{-0.0001\xspace}
\newcommand{\nInteractBothShiftFreqThousandHi}{-0.0001\xspace}
\newcommand{\nInteractBothShiftFreqThousandLo}{-0.0002\xspace}
\newcommand{\nInteractBothShiftFreqTwentyFiveDiff}{-0.0056\xspace}
\newcommand{\nInteractBothShiftFreqTwentyFiveHi}{-0.0041\xspace}
\newcommand{\nInteractBothShiftFreqTwentyFiveLo}{-0.0071\xspace}
\newcommand{\nInteractBothShiftFreqTwoHundredDiff}{-0.0008\xspace}
\newcommand{\nInteractBothShiftFreqTwoHundredHi}{-0.0005\xspace}
\newcommand{\nInteractBothShiftFreqTwoHundredLo}{-0.0010\xspace}

\newcommand{\nInteractFreqBelowFifty}{4\xspace}

\newcommand{\nInteractLargeMeanFreqFiftyDiff}{-0.0040\xspace}
\newcommand{\nInteractLargeMeanFreqFiftyHi}{-0.0031\xspace}
\newcommand{\nInteractLargeMeanFreqFiftyLo}{-0.0049\xspace}
\newcommand{\nInteractLargeMeanFreqFiveThousandDiff}{0.0000\xspace}
\newcommand{\nInteractLargeMeanFreqFiveThousandHi}{0.0000\xspace}
\newcommand{\nInteractLargeMeanFreqFiveThousandLo}{0.0000\xspace}
\newcommand{\nInteractLargeMeanFreqTenDiff}{-0.0088\xspace}
\newcommand{\nInteractLargeMeanFreqTenHi}{-0.0055\xspace}
\newcommand{\nInteractLargeMeanFreqTenLo}{-0.0121\xspace}
\newcommand{\nInteractLargeMeanFreqThousandDiff}{-0.0002\xspace}
\newcommand{\nInteractLargeMeanFreqThousandHi}{-0.0001\xspace}
\newcommand{\nInteractLargeMeanFreqThousandLo}{-0.0002\xspace}
\newcommand{\nInteractLargeMeanFreqTwentyFiveDiff}{-0.0086\xspace}
\newcommand{\nInteractLargeMeanFreqTwentyFiveHi}{-0.0068\xspace}
\newcommand{\nInteractLargeMeanFreqTwentyFiveLo}{-0.0103\xspace}
\newcommand{\nInteractLargeMeanFreqTwoHundredDiff}{-0.0009\xspace}
\newcommand{\nInteractLargeMeanFreqTwoHundredHi}{-0.0007\xspace}
\newcommand{\nInteractLargeMeanFreqTwoHundredLo}{-0.0012\xspace}

\newcommand{\nInteractMeanShiftFreqFiftyDiff}{-0.0065\xspace}
\newcommand{\nInteractMeanShiftFreqFiftyHi}{-0.0055\xspace}
\newcommand{\nInteractMeanShiftFreqFiftyLo}{-0.0076\xspace}
\newcommand{\nInteractMeanShiftFreqFiveThousandDiff}{-0.0001\xspace}
\newcommand{\nInteractMeanShiftFreqFiveThousandHi}{0.0000\xspace}
\newcommand{\nInteractMeanShiftFreqFiveThousandLo}{-0.0001\xspace}
\newcommand{\nInteractMeanShiftFreqTenDiff}{-0.0071\xspace}
\newcommand{\nInteractMeanShiftFreqTenHi}{-0.0038\xspace}
\newcommand{\nInteractMeanShiftFreqTenLo}{-0.0103\xspace}
\newcommand{\nInteractMeanShiftFreqThousandDiff}{-0.0003\xspace}
\newcommand{\nInteractMeanShiftFreqThousandHi}{-0.0003\xspace}
\newcommand{\nInteractMeanShiftFreqThousandLo}{-0.0004\xspace}
\newcommand{\nInteractMeanShiftFreqTwentyFiveDiff}{-0.0085\xspace}
\newcommand{\nInteractMeanShiftFreqTwentyFiveHi}{-0.0070\xspace}
\newcommand{\nInteractMeanShiftFreqTwentyFiveLo}{-0.0100\xspace}
\newcommand{\nInteractMeanShiftFreqTwoHundredDiff}{-0.0014\xspace}
\newcommand{\nInteractMeanShiftFreqTwoHundredHi}{-0.0012\xspace}
\newcommand{\nInteractMeanShiftFreqTwoHundredLo}{-0.0017\xspace}

\newcommand{\nInteractMeanShiftPtFiftyHi}{-0.0009\xspace}
\newcommand{\nInteractMeanShiftPtFiftyLo}{-0.0096\xspace}

\newcommand{\nInteractNoBreakFreqFiftyDiff}{0.0029\xspace}
\newcommand{\nInteractNoBreakFreqFiftyHi}{0.0037\xspace}
\newcommand{\nInteractNoBreakFreqFiftyLo}{0.0021\xspace}
\newcommand{\nInteractNoBreakFreqFiveThousandDiff}{0.0000\xspace}
\newcommand{\nInteractNoBreakFreqFiveThousandHi}{0.0000\xspace}
\newcommand{\nInteractNoBreakFreqFiveThousandLo}{0.0000\xspace}
\newcommand{\nInteractNoBreakFreqTenDiff}{0.0107\xspace}
\newcommand{\nInteractNoBreakFreqTenHi}{0.0140\xspace}
\newcommand{\nInteractNoBreakFreqTenLo}{0.0075\xspace}
\newcommand{\nInteractNoBreakFreqThousandDiff}{0.0001\xspace}
\newcommand{\nInteractNoBreakFreqThousandHi}{0.0001\xspace}
\newcommand{\nInteractNoBreakFreqThousandLo}{0.0001\xspace}
\newcommand{\nInteractNoBreakFreqTwentyFiveDiff}{0.0028\xspace}
\newcommand{\nInteractNoBreakFreqTwentyFiveHi}{0.0036\xspace}
\newcommand{\nInteractNoBreakFreqTwentyFiveLo}{0.0020\xspace}
\newcommand{\nInteractNoBreakFreqTwoHundredDiff}{0.0006\xspace}
\newcommand{\nInteractNoBreakFreqTwoHundredHi}{0.0008\xspace}
\newcommand{\nInteractNoBreakFreqTwoHundredLo}{0.0004\xspace}

\newcommand{\nInteractVarShiftFreqFiftyDiff}{-0.0048\xspace}
\newcommand{\nInteractVarShiftFreqFiftyHi}{-0.0037\xspace}
\newcommand{\nInteractVarShiftFreqFiftyLo}{-0.0058\xspace}
\newcommand{\nInteractVarShiftFreqFiveThousandDiff}{-0.0001\xspace}
\newcommand{\nInteractVarShiftFreqFiveThousandHi}{0.0000\xspace}
\newcommand{\nInteractVarShiftFreqFiveThousandLo}{-0.0001\xspace}
\newcommand{\nInteractVarShiftFreqTenDiff}{-0.0059\xspace}
\newcommand{\nInteractVarShiftFreqTenHi}{-0.0027\xspace}
\newcommand{\nInteractVarShiftFreqTenLo}{-0.0091\xspace}
\newcommand{\nInteractVarShiftFreqThousandDiff}{-0.0002\xspace}
\newcommand{\nInteractVarShiftFreqThousandHi}{-0.0002\xspace}
\newcommand{\nInteractVarShiftFreqThousandLo}{-0.0003\xspace}
\newcommand{\nInteractVarShiftFreqTwentyFiveDiff}{-0.0057\xspace}
\newcommand{\nInteractVarShiftFreqTwentyFiveHi}{-0.0042\xspace}
\newcommand{\nInteractVarShiftFreqTwentyFiveLo}{-0.0072\xspace}
\newcommand{\nInteractVarShiftFreqTwoHundredDiff}{-0.0011\xspace}
\newcommand{\nInteractVarShiftFreqTwoHundredHi}{-0.0009\xspace}
\newcommand{\nInteractVarShiftFreqTwoHundredLo}{-0.0014\xspace}

\newcommand{\nIsFomcEndsAgaciSrcTwentyFive}{$[-0.021,\, 0.006]$\xspace}
\newcommand{\nIsFomcEndsDtaciSrcTwentyFive}{$[-0.024,\, 0.006]$\xspace}
\newcommand{\nIsFomcEndsPidSrcTwentyFive}{$[-0.021,\, 0.005]$\xspace}

\newcommand{\nIsFomcSideBelowEndsTwentyFive}{$[-0.052,\, -0.019]$ and $[-0.036,\, -0.009]$\xspace}
\newcommand{\nIsFomcSideBelowWhoTwentyFive}{ACI and SAOCP\xspace}

\newcommand{\nJumpShareFiftyMax}{0.89\xspace}
\newcommand{\nJumpShareFiftyMaxWho}{the mean shift\xspace}
\newcommand{\nJumpShareFiftyMin}{0.30\xspace}
\newcommand{\nJumpShareFiftyMinWho}{the large mean shift\xspace}

\newcommand{\nLgbEstimators}{200\xspace}

\newcommand{\nLgbLagFour}{14\xspace}
\newcommand{\nLgbLagOne}{1\xspace}
\newcommand{\nLgbLagThree}{7\xspace}
\newcommand{\nLgbLagTwo}{2\xspace}

\newcommand{\nLgbRolling}{7\xspace}

\newcommand{\nLibBlockFomc}{20\xspace}
\newcommand{\nLibBlockLevel}{7\xspace}
\newcommand{\nLibCases}{9\xspace}
\newcommand{\nLibDraws}{10,000\xspace}
\newcommand{\nLibEntered}{1\xspace}
\newcommand{\nLibEnteredWord}{one\xspace}

\newcommand{\nLibErcotCapBstale}{0.015\xspace}

\newcommand{\nLibErcotCapCfCellAbaci}{0.026\xspace}
\newcommand{\nLibErcotCapCfCellAci}{0.036\xspace}

\newcommand{\nLibErcotCapCfCellSeg}{0.053\xspace}

\newcommand{\nLibErcotCapDateIso}{2022-01-01\xspace}
\newcommand{\nLibErcotCapDose}{-0.09\xspace}
\newcommand{\nLibErcotCapDoseEq}{1.80\xspace}
\newcommand{\nLibErcotCapDoseEqHi}{1.86\xspace}
\newcommand{\nLibErcotCapDoseEqLo}{1.74\xspace}

\newcommand{\nLibErcotCapFloor}{0.009\xspace}

\newcommand{\nLibErcotCapGain}{-0.0005\xspace}

\newcommand{\nLibErcotCapPost}{400\xspace}

\newcommand{\nLibErcotCapPred}{reset neutral\xspace}

\newcommand{\nLibErcotCapRatio}{0.30\xspace}
\newcommand{\nLibErcotCapRule}{stays under\xspace}

\newcommand{\nLibErcotCapStaleBelow}{354\xspace}
\newcommand{\nLibErcotCapStaleScores}{400\xspace}

\newcommand{\nLibErcotCapWinnerCf}{AB-ACI\xspace}
\newcommand{\nLibErcotCapWinnerCfCov}{conformal PID\xspace}
\newcommand{\nLibErcotCapWinnerCfCovErr}{0.000\xspace}

\newcommand{\nLibErcotCapWinnerCfValue}{0.026\xspace}

\newcommand{\nLibErcotRtcbBstale}{0.131\xspace}

\newcommand{\nLibErcotRtcbCfCellAbaci}{0.065\xspace}
\newcommand{\nLibErcotRtcbCfCellAci}{0.046\xspace}

\newcommand{\nLibErcotRtcbCfCellSeg}{0.069\xspace}

\newcommand{\nLibErcotRtcbDateIso}{2025-12-05\xspace}
\newcommand{\nLibErcotRtcbDose}{0.46\xspace}
\newcommand{\nLibErcotRtcbDoseEq}{2.87\xspace}
\newcommand{\nLibErcotRtcbDoseEqHi}{2.97\xspace}
\newcommand{\nLibErcotRtcbDoseEqLo}{2.78\xspace}

\newcommand{\nLibErcotRtcbFloor}{0.025\xspace}

\newcommand{\nLibErcotRtcbGain}{-0.0007\xspace}

\newcommand{\nLibErcotRtcbPost}{117\xspace}

\newcommand{\nLibErcotRtcbPred}{reset helps\xspace}

\newcommand{\nLibErcotRtcbRatio}{1.41\xspace}
\newcommand{\nLibErcotRtcbRule}{stays under\xspace}

\newcommand{\nLibErcotRtcbStaleBelow}{90\xspace}
\newcommand{\nLibErcotRtcbStaleScores}{117\xspace}

\newcommand{\nLibErcotRtcbWinnerCf}{ACI\xspace}
\newcommand{\nLibErcotRtcbWinnerCfCov}{reset rule, warm start, Mondrian predictor and ACI\xspace}
\newcommand{\nLibErcotRtcbWinnerCfCovErr}{0.003\xspace}

\newcommand{\nLibErcotRtcbWinnerCfValue}{0.046\xspace}

\newcommand{\nLibFloorDraws}{1,000\xspace}
\newcommand{\nLibFloorGridPoints}{8\xspace}

\newcommand{\nLibFomcSingleBstale}{0.240\xspace}

\newcommand{\nLibFomcSingleCfCellAbaci}{0.081\xspace}
\newcommand{\nLibFomcSingleCfCellAci}{0.075\xspace}

\newcommand{\nLibFomcSingleCfCellSeg}{0.036\xspace}

\newcommand{\nLibFomcSingleDateIso}{2022-03-16\xspace}
\newcommand{\nLibFomcSingleDose}{0.60\xspace}
\newcommand{\nLibFomcSingleDoseEq}{2.46\xspace}
\newcommand{\nLibFomcSingleDoseEqHi}{2.51\xspace}
\newcommand{\nLibFomcSingleDoseEqLo}{2.40\xspace}

\newcommand{\nLibFomcSingleFloor}{0.019\xspace}

\newcommand{\nLibFomcSingleGain}{-0.0006\xspace}

\newcommand{\nLibFomcSinglePost}{400\xspace}

\newcommand{\nLibFomcSinglePred}{reset helps\xspace}

\newcommand{\nLibFomcSingleRatio}{4.80\xspace}
\newcommand{\nLibFomcSingleRule}{stays under\xspace}

\newcommand{\nLibFomcSingleStaleBelow}{264\xspace}
\newcommand{\nLibFomcSingleStaleScores}{400\xspace}

\newcommand{\nLibFomcSingleWinnerCf}{reset rule and Mondrian predictor\xspace}
\newcommand{\nLibFomcSingleWinnerCfCov}{AB-ACI, ACI and conformal PID\xspace}
\newcommand{\nLibFomcSingleWinnerCfCovErr}{0.002\xspace}

\newcommand{\nLibFomcSingleWinnerCfValue}{0.036\xspace}

\newcommand{\nLibGammaFomc}{0.05\xspace}
\newcommand{\nLibGammaLevel}{2.0\xspace}
\newcommand{\nLibMinCalib}{50\xspace}
\newcommand{\nLibMinPost}{100\xspace}
\newcommand{\nLibMinReference}{20\xspace}

\newcommand{\nLibNemApcBstale}{0.090\xspace}

\newcommand{\nLibNemApcCfCellAbaci}{0.095\xspace}
\newcommand{\nLibNemApcCfCellAci}{0.092\xspace}

\newcommand{\nLibNemApcCfCellSeg}{0.105\xspace}

\newcommand{\nLibNemApcDateIso}{2022-12-01\xspace}
\newcommand{\nLibNemApcDose}{-1.67\xspace}
\newcommand{\nLibNemApcDoseEq}{1.75\xspace}
\newcommand{\nLibNemApcDoseEqHi}{1.81\xspace}
\newcommand{\nLibNemApcDoseEqLo}{1.69\xspace}

\newcommand{\nLibNemApcFloor}{0.009\xspace}

\newcommand{\nLibNemApcGain}{-0.0005\xspace}

\newcommand{\nLibNemApcPost}{400\xspace}

\newcommand{\nLibNemApcPred}{reset helps\xspace}

\newcommand{\nLibNemApcRatio}{1.80\xspace}
\newcommand{\nLibNemApcRule}{stays under\xspace}

\newcommand{\nLibNemApcStaleBelow}{396\xspace}
\newcommand{\nLibNemApcStaleScores}{400\xspace}

\newcommand{\nLibNemApcWinnerCf}{AgACI\xspace}
\newcommand{\nLibNemApcWinnerCfCov}{DtACI\xspace}
\newcommand{\nLibNemApcWinnerCfCovErr}{0.000\xspace}

\newcommand{\nLibNemApcWinnerCfValue}{0.075\xspace}

\newcommand{\nLibNemFiveBstale}{0.385\xspace}

\newcommand{\nLibNemFiveCfCellAbaci}{0.416\xspace}
\newcommand{\nLibNemFiveCfCellAci}{0.433\xspace}

\newcommand{\nLibNemFiveCfCellSeg}{0.491\xspace}

\newcommand{\nLibNemFiveCfRef}{351\xspace}

\newcommand{\nLibNemFiveDate}{October 1, 2021\xspace}
\newcommand{\nLibNemFiveDateIso}{2021-10-01\xspace}
\newcommand{\nLibNemFiveDose}{2.52\xspace}
\newcommand{\nLibNemFiveDoseEq}{1.81\xspace}
\newcommand{\nLibNemFiveDoseEqHi}{1.87\xspace}
\newcommand{\nLibNemFiveDoseEqLo}{1.76\xspace}

\newcommand{\nLibNemFiveFloor}{0.010\xspace}

\newcommand{\nLibNemFiveGain}{0.0203\xspace}

\newcommand{\nLibNemFivePost}{400\xspace}

\newcommand{\nLibNemFivePred}{reset helps\xspace}

\newcommand{\nLibNemFiveRatio}{7.70\xspace}
\newcommand{\nLibNemFiveRule}{clears\xspace}

\newcommand{\nLibNemFiveStaleBelow}{206\xspace}
\newcommand{\nLibNemFiveStaleScores}{400\xspace}

\newcommand{\nLibNemFiveWinnerCf}{AgACI\xspace}
\newcommand{\nLibNemFiveWinnerCfCov}{AgACI\xspace}
\newcommand{\nLibNemFiveWinnerCfCovErr}{0.012\xspace}

\newcommand{\nLibNemFiveWinnerCfValue}{0.358\xspace}

\newcommand{\nLibNemMpcBstale}{0.005\xspace}

\newcommand{\nLibNemMpcCfCellAbaci}{0.162\xspace}
\newcommand{\nLibNemMpcCfCellAci}{0.156\xspace}

\newcommand{\nLibNemMpcCfCellSeg}{0.352\xspace}

\newcommand{\nLibNemMpcDateIso}{2025-07-01\xspace}
\newcommand{\nLibNemMpcDose}{0.02\xspace}
\newcommand{\nLibNemMpcDoseEq}{2.63\xspace}
\newcommand{\nLibNemMpcDoseEqHi}{2.68\xspace}
\newcommand{\nLibNemMpcDoseEqLo}{2.58\xspace}

\newcommand{\nLibNemMpcFloor}{0.022\xspace}

\newcommand{\nLibNemMpcGain}{-0.0005\xspace}

\newcommand{\nLibNemMpcPost}{275\xspace}

\newcommand{\nLibNemMpcPred}{reset neutral\xspace}

\newcommand{\nLibNemMpcRatio}{0.09\xspace}
\newcommand{\nLibNemMpcRule}{stays under\xspace}

\newcommand{\nLibNemMpcStaleBelow}{246\xspace}
\newcommand{\nLibNemMpcStaleScores}{275\xspace}

\newcommand{\nLibNemMpcWinnerCf}{stale rule\xspace}
\newcommand{\nLibNemMpcWinnerCfCov}{DtACI and conformal PID\xspace}
\newcommand{\nLibNemMpcWinnerCfCovErr}{0.002\xspace}

\newcommand{\nLibNemMpcWinnerCfValue}{0.079\xspace}

\newcommand{\nLibNewCases}{5\xspace}

\newcommand{\nLibOrdcOneBstale}{0.000\xspace}

\newcommand{\nLibOrdcOneCfCellAbaci}{0.045\xspace}
\newcommand{\nLibOrdcOneCfCellAci}{0.044\xspace}

\newcommand{\nLibOrdcOneCfCellSeg}{0.043\xspace}

\newcommand{\nLibOrdcOneDateIso}{2019-03-01\xspace}
\newcommand{\nLibOrdcOneDose}{-0.13\xspace}
\newcommand{\nLibOrdcOneDoseEq}{2.27\xspace}
\newcommand{\nLibOrdcOneDoseEqHi}{2.32\xspace}
\newcommand{\nLibOrdcOneDoseEqLo}{2.22\xspace}

\newcommand{\nLibOrdcOneFloor}{0.017\xspace}

\newcommand{\nLibOrdcOneGain}{-0.0005\xspace}

\newcommand{\nLibOrdcOnePost}{400\xspace}

\newcommand{\nLibOrdcOnePred}{reset neutral\xspace}

\newcommand{\nLibOrdcOneRatio}{0.00\xspace}
\newcommand{\nLibOrdcOneRule}{stays under\xspace}

\newcommand{\nLibOrdcOneStaleBelow}{360\xspace}
\newcommand{\nLibOrdcOneStaleScores}{400\xspace}

\newcommand{\nLibOrdcOneWinnerCf}{stale rule\xspace}
\newcommand{\nLibOrdcOneWinnerCfCov}{ACI, DtACI, SAOCP and stale rule\xspace}
\newcommand{\nLibOrdcOneWinnerCfCovErr}{0.000\xspace}

\newcommand{\nLibOrdcOneWinnerCfValue}{0.027\xspace}

\newcommand{\nLibOrdcTwoBstale}{0.008\xspace}

\newcommand{\nLibOrdcTwoCfCellAbaci}{0.064\xspace}
\newcommand{\nLibOrdcTwoCfCellAci}{0.062\xspace}

\newcommand{\nLibOrdcTwoCfCellSeg}{0.055\xspace}

\newcommand{\nLibOrdcTwoDateIso}{2020-03-01\xspace}
\newcommand{\nLibOrdcTwoDose}{-0.01\xspace}
\newcommand{\nLibOrdcTwoDoseEq}{2.37\xspace}
\newcommand{\nLibOrdcTwoDoseEqHi}{2.42\xspace}
\newcommand{\nLibOrdcTwoDoseEqLo}{2.31\xspace}

\newcommand{\nLibOrdcTwoFloor}{0.018\xspace}

\newcommand{\nLibOrdcTwoGain}{-0.0005\xspace}

\newcommand{\nLibOrdcTwoPost}{400\xspace}

\newcommand{\nLibOrdcTwoPred}{reset neutral\xspace}

\newcommand{\nLibOrdcTwoRatio}{0.15\xspace}
\newcommand{\nLibOrdcTwoRule}{stays under\xspace}

\newcommand{\nLibOrdcTwoStaleBelow}{357\xspace}
\newcommand{\nLibOrdcTwoStaleScores}{400\xspace}

\newcommand{\nLibOrdcTwoWinnerCf}{stale rule\xspace}
\newcommand{\nLibOrdcTwoWinnerCfCov}{AgACI, DtACI, naive rule and stale rule\xspace}
\newcommand{\nLibOrdcTwoWinnerCfCovErr}{0.008\xspace}

\newcommand{\nLibOrdcTwoWinnerCfValue}{0.047\xspace}

\newcommand{\nLibPjmRpfBstale}{0.080\xspace}

\newcommand{\nLibPjmRpfCfCellAbaci}{0.063\xspace}
\newcommand{\nLibPjmRpfCfCellAci}{0.072\xspace}

\newcommand{\nLibPjmRpfCfCellSeg}{0.110\xspace}

\newcommand{\nLibPjmRpfDateIso}{2022-10-01\xspace}
\newcommand{\nLibPjmRpfDose}{-1.29\xspace}
\newcommand{\nLibPjmRpfDoseEq}{2.45\xspace}
\newcommand{\nLibPjmRpfDoseEqHi}{2.50\xspace}
\newcommand{\nLibPjmRpfDoseEqLo}{2.39\xspace}

\newcommand{\nLibPjmRpfFloor}{0.019\xspace}

\newcommand{\nLibPjmRpfGain}{-0.0005\xspace}

\newcommand{\nLibPjmRpfPost}{400\xspace}

\newcommand{\nLibPjmRpfPred}{reset helps\xspace}

\newcommand{\nLibPjmRpfRatio}{1.60\xspace}
\newcommand{\nLibPjmRpfRule}{stays under\xspace}

\newcommand{\nLibPjmRpfStaleBelow}{392\xspace}
\newcommand{\nLibPjmRpfStaleScores}{400\xspace}

\newcommand{\nLibPjmRpfWinnerCf}{AB-ACI\xspace}
\newcommand{\nLibPjmRpfWinnerCfCov}{DtACI and SAOCP\xspace}
\newcommand{\nLibPjmRpfWinnerCfCovErr}{0.012\xspace}

\newcommand{\nLibPjmRpfWinnerCfValue}{0.063\xspace}

\newcommand{\nLibPredHelps}{5\xspace}
\newcommand{\nLibPredHelpsFailed}{4\xspace}

\newcommand{\nLibPredHelpsFailedFiftyOthersPrCf}{5\xspace}

\newcommand{\nLibPredHelpsFailedFiftyReadersPrCf}{4\xspace}

\newcommand{\nLibPredHelpsFailedPrCf}{3\xspace}

\newcommand{\nLibSeed}{1234\xspace}

\newcommand{\nMapErcotFirstPost}{187\xspace}
\newcommand{\nMapErcotFirstPostDate}{December 5, 2025\xspace}
\newcommand{\nMapErcotLastPre}{186\xspace}
\newcommand{\nMapErcotLastPreDate}{December 4, 2025\xspace}
\newcommand{\nMapErcotReadingDate}{January 23, 2026\xspace}
\newcommand{\nMapErcotReadingIdx}{236\xspace}
\newcommand{\nMapErcotReadingRef}{68\xspace}
\newcommand{\nMapErcotReadingScores}{49\xspace}
\newcommand{\nMapFomcCells}{38\xspace}

\newcommand{\nMapFomcEvalStartDate}{March 17, 2021\xspace}
\newcommand{\nMapFomcFirstFirstPost}{330\xspace}
\newcommand{\nMapFomcFirstFirstPostDate}{April 28, 2021\xspace}
\newcommand{\nMapFomcFirstLastPre}{329\xspace}
\newcommand{\nMapFomcFirstLastPreDate}{April 27, 2021\xspace}
\newcommand{\nMapFomcPreCell}{30\xspace}
\newcommand{\nMapFomcReadingScores}{9\xspace}
\newcommand{\nMapSynFirstPost}{500\xspace}
\newcommand{\nMapSynLastPre}{499\xspace}
\newcommand{\nMapSynReadingIdx}{549\xspace}
\newcommand{\nMapSynReadingScores}{49\xspace}
\newcommand{\nMbBound}{0.7425\xspace}

\newcommand{\nMbCutoff}{20\xspace}

\newcommand{\nMbFirst}{20\xspace}

\newcommand{\nMbLast}{200\xspace}
\newcommand{\nMbLevel}{0.95\xspace}
\newcommand{\nMbMmin}{4\xspace}
\newcommand{\nMbScores}{3\xspace}

\newcommand{\nMemMib}{2,048\xspace}

\newcommand{\nNemCfAllBestGap}{0.453\xspace}

\newcommand{\nNobreakBandDraws}{10,000\xspace}
\newcommand{\nNobreakBandSeed}{1234\xspace}
\newcommand{\nNobreakBurnin}{10\xspace}
\newcommand{\nNobreakCLessAciDiff}{0.00006\xspace}
\newcommand{\nNobreakCLessAciHi}{0.00022\xspace}
\newcommand{\nNobreakCLessAciLo}{-0.00010\xspace}
\newcommand{\nNobreakCovAbaci}{0.9001\xspace}

\newcommand{\nNobreakFloor}{$1.0\times 10^{-8}$\xspace}
\newcommand{\nNobreakGamma}{1.0\xspace}
\newcommand{\nNobreakLastPre}{499\xspace}
\newcommand{\nNobreakReps}{200\xspace}
\newcommand{\nNobreakSeedBase}{19407\xspace}
\newcommand{\nNobreakSeedStep}{1009\xspace}
\newcommand{\nNobreakT}{1,000\xspace}
\newcommand{\nNobreakTau}{500\xspace}
\newcommand{\nNominalCoverage}{0.9\xspace}

\newcommand{\nOrdFifty}{50th\xspace}
\newcommand{\nOrdFiveThousand}{5,000th\xspace}

\newcommand{\nOrdLevelsLow}{7\xspace}

\newcommand{\nOrdTen}{10th\xspace}

\newcommand{\nOrdTwentyFive}{25th\xspace}

\newcommand{\nOrderFirstFinite}{9\xspace}

\newcommand{\nOrderLargest}{2,000\xspace}

\newcommand{\nOrderSmallest}{2\xspace}
\newcommand{\nPFiveAEightTenths}{0.8\xspace}
\newcommand{\nPFiveAHalf}{0.5\xspace}
\newcommand{\nPFiveAOne}{1.0\xspace}
\newcommand{\nPFiveAOneHalf}{1.5\xspace}

\newcommand{\nPFiveDepLength}{200,000\xspace}
\newcommand{\nPFiveDepPredEightTenths}{-0.8\xspace}
\newcommand{\nPFiveDepPredHalf}{-0.5\xspace}
\newcommand{\nPFiveDepPredOne}{-1.0\xspace}
\newcommand{\nPFiveDepPredOneHalf}{-1.5\xspace}

\newcommand{\nPFiveDepSlopeEightTenths}{-0.734\xspace}
\newcommand{\nPFiveDepSlopeHalf}{-0.473\xspace}
\newcommand{\nPFiveDepSlopeOne}{-0.691\xspace}
\newcommand{\nPFiveDepSlopeOneHalf}{-0.915\xspace}
\newcommand{\nPFiveDepSlopeSeEightTenths}{0.111\xspace}
\newcommand{\nPFiveDepSlopeSeHalf}{0.025\xspace}
\newcommand{\nPFiveDepSlopeSeOne}{0.215\xspace}
\newcommand{\nPFiveDepSlopeSeOneHalf}{0.194\xspace}
\newcommand{\nPFiveDepStreams}{20\xspace}

\newcommand{\nPFiveIndexEightTenths}{-0.40\xspace}
\newcommand{\nPFiveIndexHalf}{-0.25\xspace}
\newcommand{\nPFiveIndexOne}{-0.50\xspace}
\newcommand{\nPFiveIndexOneHalf}{-0.50\xspace}
\newcommand{\nPFiveIndices}{4\xspace}
\newcommand{\nPFiveLagMax}{1,000\xspace}
\newcommand{\nPFiveLags}{10\xspace}
\newcommand{\nPFiveMargin}{0.1\xspace}
\newcommand{\nPFiveMeanHigh}{1.0\xspace}

\newcommand{\nPFivePost}{5,000\xspace}
\newcommand{\nPFiveReps}{1,000\xspace}
\newcommand{\nPFiveRuleHalf}{-0.35\xspace}
\newcommand{\nPFiveSaltFirst}{106\xspace}
\newcommand{\nPFiveSaltLast}{109\xspace}
\newcommand{\nPFiveSlopeFourEightTenths}{-0.504\xspace}
\newcommand{\nPFiveSlopeFourHalf}{-0.412\xspace}
\newcommand{\nPFiveSlopeFourOne}{-0.505\xspace}
\newcommand{\nPFiveSlopeFourOneHalf}{-0.516\xspace}
\newcommand{\nPFiveSlopeFourSeEightTenths}{0.008\xspace}
\newcommand{\nPFiveSlopeFourSeHalf}{0.012\xspace}
\newcommand{\nPFiveSlopeFourSeOne}{0.002\xspace}
\newcommand{\nPFiveSlopeFourSeOneHalf}{0.005\xspace}

\newcommand{\nPFiveSlopeSixEightTenths}{-0.517\xspace}
\newcommand{\nPFiveSlopeSixHalf}{-0.440\xspace}
\newcommand{\nPFiveSlopeSixOne}{-0.522\xspace}
\newcommand{\nPFiveSlopeSixOneHalf}{-0.531\xspace}
\newcommand{\nPFiveSlopeSixSeEightTenths}{0.008\xspace}
\newcommand{\nPFiveSlopeSixSeHalf}{0.016\xspace}
\newcommand{\nPFiveSlopeSixSeOne}{0.012\xspace}
\newcommand{\nPFiveSlopeSixSeOneHalf}{0.011\xspace}
\newcommand{\nPFiveTailEightTenths}{1.8\xspace}
\newcommand{\nPFiveTailHalf}{1.5\xspace}
\newcommand{\nPFiveTailOne}{2.0\xspace}
\newcommand{\nPFiveTailOneHalf}{2.5\xspace}

\newcommand{\nPSevenStreams}{16\xspace}
\newcommand{\nPSmallest}{0.0002\xspace}
\newcommand{\nPTwoDecimalsFrom}{0.01\xspace}
\newcommand{\nPermAemoCaptureDate}{September 23, 2026\xspace}

\newcommand{\nPiSumVarShiftNormFifty}{$2.5\times 10^{-6}$\xspace}
\newcommand{\nPiSumVarShiftRawFifty}{0.00013\xspace}
\newcommand{\nPidErcotCsat}{0.522\xspace}
\newcommand{\nPidErcotKi}{32.40\xspace}
\newcommand{\nPidErcotT}{254\xspace}
\newcommand{\nPidFloor}{0.01\xspace}
\newcommand{\nPidFomcCsat}{0.547\xspace}
\newcommand{\nPidFomcKi}{0.167\xspace}
\newcommand{\nPidFomcT}{1,249\xspace}
\newcommand{\nPidKi}{0.001\xspace}
\newcommand{\nPidKp}{0.05\xspace}
\newcommand{\nPidRealWindow}{50\xspace}
\newcommand{\nPidSynCsat}{0.563\xspace}
\newcommand{\nPidSynKiMax}{4.96\xspace}
\newcommand{\nPidSynKiMin}{1.73\xspace}
\newcommand{\nPidSynT}{5,510\xspace}
\newcommand{\nPidSynWindow}{100\xspace}

\newcommand{\nProfAboveAll}{10\xspace}
\newcommand{\nProfAboveAllDays}{3 to 7 and 9 to 13\xspace}

\newcommand{\nProfBelowAnyWord}{none\xspace}

\newcommand{\nProfReadable}{34\xspace}

\newcommand{\nProjThresholdPct}{1\xspace}

\newcommand{\nProvErcotCapCalibFirst}{2021-04-20\xspace}
\newcommand{\nProvErcotCapCalibLast}{2021-06-08\xspace}
\newcommand{\nProvErcotCapEvalFirst}{2021-06-09\xspace}
\newcommand{\nProvErcotCapEvalLast}{2023-02-04\xspace}
\newcommand{\nProvErcotCapFirstPost}{2022-01-01\xspace}
\newcommand{\nProvErcotCapTrainFirst}{2021-03-01\xspace}
\newcommand{\nProvErcotCapTrainLast}{2021-04-19\xspace}
\newcommand{\nProvErcotRtcbCalibFirst}{2025-07-21\xspace}
\newcommand{\nProvErcotRtcbCalibLast}{2025-09-08\xspace}
\newcommand{\nProvErcotRtcbEvalFirst}{2025-09-09\xspace}
\newcommand{\nProvErcotRtcbEvalLast}{2026-03-31\xspace}
\newcommand{\nProvErcotRtcbFirstPost}{2025-12-05\xspace}
\newcommand{\nProvErcotRtcbTrainFirst}{2025-06-01\xspace}
\newcommand{\nProvErcotRtcbTrainLast}{2025-07-20\xspace}
\newcommand{\nProvFomcSingleCalibFirst}{2021-05-27\xspace}
\newcommand{\nProvFomcSingleCalibLast}{2021-10-19\xspace}
\newcommand{\nProvFomcSingleEvalFirst}{2021-10-20\xspace}
\newcommand{\nProvFomcSingleEvalLast}{2023-10-19\xspace}
\newcommand{\nProvFomcSingleFirstPost}{2022-03-16\xspace}
\newcommand{\nProvFomcSingleTrainFirst}{2020-08-10\xspace}
\newcommand{\nProvFomcSingleTrainLast}{2021-03-17\xspace}
\newcommand{\nProvFomcSingleUnreadFirst}{2021-03-18\xspace}
\newcommand{\nProvFomcSingleUnreadLast}{2021-05-26\xspace}
\newcommand{\nProvNemApcCalibFirst}{2021-12-16\xspace}
\newcommand{\nProvNemApcCalibLast}{2022-02-03\xspace}
\newcommand{\nProvNemApcEvalFirst}{2022-02-04\xspace}
\newcommand{\nProvNemApcEvalLast}{2024-01-04\xspace}
\newcommand{\nProvNemApcFirstPost}{2022-12-01\xspace}
\newcommand{\nProvNemApcTrainFirst}{2021-10-27\xspace}
\newcommand{\nProvNemApcTrainLast}{2021-12-15\xspace}
\newcommand{\nProvNemFiveCalibFirst}{2020-10-16\xspace}
\newcommand{\nProvNemFiveCalibLast}{2020-12-04\xspace}
\newcommand{\nProvNemFiveEvalFirst}{2020-12-05\xspace}
\newcommand{\nProvNemFiveEvalLast}{2022-11-04\xspace}
\newcommand{\nProvNemFiveFirstPost}{2021-10-01\xspace}
\newcommand{\nProvNemFiveTrainFirst}{2020-08-27\xspace}
\newcommand{\nProvNemFiveTrainLast}{2020-10-15\xspace}
\newcommand{\nProvNemMpcCalibFirst}{2024-07-16\xspace}
\newcommand{\nProvNemMpcCalibLast}{2024-09-03\xspace}
\newcommand{\nProvNemMpcEvalFirst}{2024-09-04\xspace}
\newcommand{\nProvNemMpcEvalLast}{2026-04-01\xspace}
\newcommand{\nProvNemMpcFirstPost}{2025-07-01\xspace}
\newcommand{\nProvNemMpcTrainFirst}{2024-05-27\xspace}
\newcommand{\nProvNemMpcTrainLast}{2024-07-15\xspace}
\newcommand{\nProvOrdcOneCalibFirst}{2018-03-16\xspace}
\newcommand{\nProvOrdcOneCalibLast}{2018-05-04\xspace}
\newcommand{\nProvOrdcOneEvalFirst}{2018-05-05\xspace}
\newcommand{\nProvOrdcOneEvalLast}{2020-04-03\xspace}
\newcommand{\nProvOrdcOneFirstPost}{2019-03-01\xspace}
\newcommand{\nProvOrdcOneTrainFirst}{2018-01-25\xspace}
\newcommand{\nProvOrdcOneTrainLast}{2018-03-15\xspace}
\newcommand{\nProvOrdcTwoCalibFirst}{2019-03-17\xspace}
\newcommand{\nProvOrdcTwoCalibLast}{2019-05-05\xspace}
\newcommand{\nProvOrdcTwoEvalFirst}{2019-05-06\xspace}
\newcommand{\nProvOrdcTwoEvalLast}{2021-04-04\xspace}
\newcommand{\nProvOrdcTwoFirstPost}{2020-03-01\xspace}
\newcommand{\nProvOrdcTwoTrainFirst}{2019-01-26\xspace}
\newcommand{\nProvOrdcTwoTrainLast}{2019-03-16\xspace}
\newcommand{\nProvPjmRpfCalibFirst}{2021-10-16\xspace}
\newcommand{\nProvPjmRpfCalibLast}{2021-12-04\xspace}
\newcommand{\nProvPjmRpfEvalFirst}{2021-12-05\xspace}
\newcommand{\nProvPjmRpfEvalLast}{2023-11-04\xspace}
\newcommand{\nProvPjmRpfFirstPost}{2022-10-01\xspace}
\newcommand{\nProvPjmRpfTrainFirst}{2021-08-27\xspace}
\newcommand{\nProvPjmRpfTrainLast}{2021-10-15\xspace}

\newcommand{\nPtBandBothShiftAciTenHi}{-0.001\xspace}
\newcommand{\nPtBandBothShiftAciTenLo}{-0.010\xspace}

\newcommand{\nPtBandLargeMeanAbaciFiftyHi}{-0.027\xspace}
\newcommand{\nPtBandLargeMeanAbaciFiftyLo}{-0.035\xspace}

\newcommand{\nPtBandLargeMeanAciFiftyHi}{-0.043\xspace}
\newcommand{\nPtBandLargeMeanAciFiftyLo}{-0.049\xspace}

\newcommand{\nPtBandLargeMeanAciTenHi}{-0.022\xspace}
\newcommand{\nPtBandLargeMeanAciTenLo}{-0.034\xspace}

\newcommand{\nPtBandLargeMeanAgaciSrcFiftyHi}{-0.014\xspace}
\newcommand{\nPtBandLargeMeanAgaciSrcFiftyLo}{-0.019\xspace}

\newcommand{\nPtBandLargeMeanDtaciSrcFiftyHi}{-0.043\xspace}
\newcommand{\nPtBandLargeMeanDtaciSrcFiftyLo}{-0.050\xspace}

\newcommand{\nPtBandLargeMeanPidSrcFiftyHi}{-0.025\xspace}
\newcommand{\nPtBandLargeMeanPidSrcFiftyLo}{-0.032\xspace}

\newcommand{\nPtBandLargeMeanSaocpSrcFiftyHi}{-0.039\xspace}
\newcommand{\nPtBandLargeMeanSaocpSrcFiftyLo}{-0.052\xspace}

\newcommand{\nPtBandMeanShiftAciTenHi}{0.006\xspace}
\newcommand{\nPtBandMeanShiftAciTenLo}{-0.003\xspace}

\newcommand{\nPtBestLargeMeanFifty}{0.052\xspace}

\newcommand{\nPtBestNoBreakFifty}{0.017\xspace}
\newcommand{\nPtBestNoBreakFiftyWho}{DtACI\xspace}

\newcommand{\nPtBothShiftAbaciDecFifty}{0.062\xspace}

\newcommand{\nPtBothShiftAbaciDecTen}{0.187\xspace}

\newcommand{\nPtBothShiftAbaciFifty}{0.048\xspace}

\newcommand{\nPtBothShiftAbaciTen}{0.067\xspace}
\newcommand{\nPtBothShiftAbaciTextFifty}{0.051\xspace}

\newcommand{\nPtBothShiftAbaciTextTen}{0.104\xspace}

\newcommand{\nPtBothShiftAciFifty}{0.048\xspace}

\newcommand{\nPtBothShiftAciTen}{0.104\xspace}

\newcommand{\nPtBothShiftAgaciSrcFifty}{0.047\xspace}

\newcommand{\nPtBothShiftAgaciSrcTen}{0.324\xspace}

\newcommand{\nPtBothShiftDtaciSrcFifty}{0.093\xspace}

\newcommand{\nPtBothShiftDtaciSrcTen}{0.417\xspace}

\newcommand{\nPtBothShiftNaiveFifty}{0.445\xspace}

\newcommand{\nPtBothShiftNaiveTen}{0.495\xspace}

\newcommand{\nPtBothShiftOracleFifty}{0.000\xspace}

\newcommand{\nPtBothShiftOracleTen}{0.000\xspace}

\newcommand{\nPtBothShiftPidSrcFifty}{0.046\xspace}

\newcommand{\nPtBothShiftPidSrcTen}{0.217\xspace}

\newcommand{\nPtBothShiftSaocpSrcFifty}{0.072\xspace}

\newcommand{\nPtBothShiftSaocpSrcTen}{0.121\xspace}

\newcommand{\nPtBothShiftSegFifty}{0.037\xspace}

\newcommand{\nPtBothShiftSegTen}{0.099\xspace}

\newcommand{\nPtBothShiftSegWarmFifty}{0.037\xspace}

\newcommand{\nPtBothShiftSegWarmTen}{0.099\xspace}

\newcommand{\nPtErcotCellAci}{0.012\xspace}

\newcommand{\nPtErcotCellSeg}{0.034\xspace}

\newcommand{\nPtFomcPooledAciTen}{0.075\xspace}
\newcommand{\nPtFomcPooledAciTwentyFive}{0.080\xspace}

\newcommand{\nPtFomcPooledSegMultiTen}{0.129\xspace}
\newcommand{\nPtFomcPooledSegMultiTwentyFive}{0.086\xspace}

\newcommand{\nPtLargeMeanAbaciDecFifty}{0.046\xspace}

\newcommand{\nPtLargeMeanAbaciDecTen}{0.350\xspace}

\newcommand{\nPtLargeMeanAbaciFifty}{0.065\xspace}

\newcommand{\nPtLargeMeanAbaciTen}{0.183\xspace}
\newcommand{\nPtLargeMeanAbaciTextFifty}{0.070\xspace}

\newcommand{\nPtLargeMeanAbaciTextTen}{0.123\xspace}

\newcommand{\nPtLargeMeanAciFifty}{0.081\xspace}

\newcommand{\nPtLargeMeanAciTen}{0.123\xspace}

\newcommand{\nPtLargeMeanAgaciSrcFifty}{0.052\xspace}

\newcommand{\nPtLargeMeanAgaciSrcTen}{0.275\xspace}

\newcommand{\nPtLargeMeanDtaciSrcFifty}{0.081\xspace}

\newcommand{\nPtLargeMeanDtaciSrcTen}{0.802\xspace}

\newcommand{\nPtLargeMeanNaiveFifty}{0.894\xspace}

\newcommand{\nPtLargeMeanNaiveTen}{0.899\xspace}

\newcommand{\nPtLargeMeanOracleFifty}{0.000\xspace}

\newcommand{\nPtLargeMeanOracleTen}{0.000\xspace}

\newcommand{\nPtLargeMeanPidSrcFifty}{0.063\xspace}

\newcommand{\nPtLargeMeanPidSrcTen}{0.242\xspace}

\newcommand{\nPtLargeMeanSaocpSrcFifty}{0.080\xspace}

\newcommand{\nPtLargeMeanSaocpSrcTen}{0.105\xspace}

\newcommand{\nPtLargeMeanSegFifty}{0.035\xspace}

\newcommand{\nPtLargeMeanSegTen}{0.095\xspace}

\newcommand{\nPtLargeMeanSegWarmFifty}{0.035\xspace}

\newcommand{\nPtLargeMeanSegWarmTen}{0.095\xspace}

\newcommand{\nPtMeanShiftAbaciDecFifty}{0.033\xspace}

\newcommand{\nPtMeanShiftAbaciDecTen}{0.108\xspace}

\newcommand{\nPtMeanShiftAbaciFifty}{0.064\xspace}
\newcommand{\nPtMeanShiftAbaciFiveThousand}{0.065\xspace}
\newcommand{\nPtMeanShiftAbaciTen}{0.178\xspace}
\newcommand{\nPtMeanShiftAbaciTextFifty}{0.068\xspace}

\newcommand{\nPtMeanShiftAbaciTextTen}{0.093\xspace}

\newcommand{\nPtMeanShiftAciFifty}{0.080\xspace}
\newcommand{\nPtMeanShiftAciFiveThousand}{0.078\xspace}
\newcommand{\nPtMeanShiftAciTen}{0.093\xspace}

\newcommand{\nPtMeanShiftAgaciSrcFifty}{0.047\xspace}

\newcommand{\nPtMeanShiftAgaciSrcTen}{0.299\xspace}

\newcommand{\nPtMeanShiftDtaciSrcFifty}{0.079\xspace}

\newcommand{\nPtMeanShiftDtaciSrcTen}{0.392\xspace}

\newcommand{\nPtMeanShiftNaiveFifty}{0.442\xspace}

\newcommand{\nPtMeanShiftNaiveTen}{0.523\xspace}

\newcommand{\nPtMeanShiftOracleFifty}{0.000\xspace}

\newcommand{\nPtMeanShiftOracleTen}{0.000\xspace}

\newcommand{\nPtMeanShiftPidSrcFifty}{0.050\xspace}

\newcommand{\nPtMeanShiftPidSrcTen}{0.156\xspace}

\newcommand{\nPtMeanShiftSaocpSrcFifty}{0.062\xspace}

\newcommand{\nPtMeanShiftSaocpSrcTen}{0.108\xspace}

\newcommand{\nPtMeanShiftSegFifty}{0.035\xspace}

\newcommand{\nPtMeanShiftSegTen}{0.095\xspace}

\newcommand{\nPtMeanShiftSegWarmFifty}{0.035\xspace}

\newcommand{\nPtMeanShiftSegWarmTen}{0.095\xspace}

\newcommand{\nPtNoBreakAbaciDecFifty}{0.034\xspace}

\newcommand{\nPtNoBreakAbaciDecTen}{0.059\xspace}

\newcommand{\nPtNoBreakAbaciFifty}{0.063\xspace}

\newcommand{\nPtNoBreakAbaciTen}{0.049\xspace}
\newcommand{\nPtNoBreakAbaciTextFifty}{0.072\xspace}

\newcommand{\nPtNoBreakAbaciTextTen}{0.065\xspace}

\newcommand{\nPtNoBreakAciFifty}{0.061\xspace}

\newcommand{\nPtNoBreakAciTen}{0.065\xspace}

\newcommand{\nPtNoBreakAgaciSrcFifty}{0.026\xspace}

\newcommand{\nPtNoBreakAgaciSrcTen}{0.026\xspace}

\newcommand{\nPtNoBreakDtaciSrcFifty}{0.017\xspace}

\newcommand{\nPtNoBreakDtaciSrcTen}{0.018\xspace}

\newcommand{\nPtNoBreakNaiveFifty}{0.010\xspace}

\newcommand{\nPtNoBreakNaiveTen}{0.011\xspace}

\newcommand{\nPtNoBreakOracleFifty}{0.000\xspace}

\newcommand{\nPtNoBreakOracleTen}{0.000\xspace}

\newcommand{\nPtNoBreakPidSrcFifty}{0.045\xspace}

\newcommand{\nPtNoBreakPidSrcTen}{0.049\xspace}

\newcommand{\nPtNoBreakSaocpSrcFifty}{0.052\xspace}

\newcommand{\nPtNoBreakSaocpSrcTen}{0.058\xspace}

\newcommand{\nPtNoBreakSegFifty}{0.036\xspace}

\newcommand{\nPtNoBreakSegTen}{0.097\xspace}

\newcommand{\nPtNoBreakSegWarmFifty}{0.036\xspace}

\newcommand{\nPtNoBreakSegWarmTen}{0.097\xspace}

\newcommand{\nPtVarShiftAbaciDecFifty}{0.037\xspace}

\newcommand{\nPtVarShiftAbaciDecTen}{0.107\xspace}

\newcommand{\nPtVarShiftAbaciFifty}{0.050\xspace}

\newcommand{\nPtVarShiftAbaciTen}{0.138\xspace}
\newcommand{\nPtVarShiftAbaciTextFifty}{0.052\xspace}

\newcommand{\nPtVarShiftAbaciTextTen}{0.081\xspace}

\newcommand{\nPtVarShiftAciFifty}{0.058\xspace}

\newcommand{\nPtVarShiftAciTen}{0.081\xspace}

\newcommand{\nPtVarShiftAgaciSrcFifty}{0.057\xspace}

\newcommand{\nPtVarShiftAgaciSrcTen}{0.228\xspace}

\newcommand{\nPtVarShiftDtaciSrcFifty}{0.099\xspace}

\newcommand{\nPtVarShiftDtaciSrcTen}{0.252\xspace}

\newcommand{\nPtVarShiftNaiveFifty}{0.273\xspace}

\newcommand{\nPtVarShiftNaiveTen}{0.305\xspace}

\newcommand{\nPtVarShiftOracleFifty}{0.000\xspace}

\newcommand{\nPtVarShiftOracleTen}{0.000\xspace}

\newcommand{\nPtVarShiftPidSrcFifty}{0.043\xspace}

\newcommand{\nPtVarShiftPidSrcTen}{0.139\xspace}

\newcommand{\nPtVarShiftSaocpSrcFifty}{0.054\xspace}

\newcommand{\nPtVarShiftSaocpSrcTen}{0.095\xspace}

\newcommand{\nPtVarShiftSegFifty}{0.036\xspace}

\newcommand{\nPtVarShiftSegTen}{0.094\xspace}

\newcommand{\nPtVarShiftSegWarmFifty}{0.036\xspace}

\newcommand{\nPtVarShiftSegWarmTen}{0.094\xspace}

\newcommand{\nPythonHashSeed}{0\xspace}
\newcommand{\nReadBandHiPct}{97.5th\xspace}
\newcommand{\nReadBandLoPct}{2.5th\xspace}

\newcommand{\nRedAbaciBothShift}{63.9\xspace}

\newcommand{\nRedAbaciLargeMean}{73.8\xspace}
\newcommand{\nRedAbaciMeanShift}{74.9\xspace}
\newcommand{\nRedAbaciNoBreak}{2.2\xspace}
\newcommand{\nRedAbaciNonzeroMax}{74.9\xspace}
\newcommand{\nRedAbaciNonzeroMin}{60.3\xspace}
\newcommand{\nRedAbaciVarShift}{60.3\xspace}

\newcommand{\nRefitColsample}{0.8\xspace}

\newcommand{\nRefitDepth}{5\xspace}
\newcommand{\nRefitDistinctEval}{182\xspace}

\newcommand{\nRefitEstimators}{100\xspace}
\newcommand{\nRefitEvalDays}{204\xspace}
\newcommand{\nRefitEvalMax}{44.06\xspace}

\newcommand{\nRefitEvalMin}{21.11\xspace}

\newcommand{\nRefitFallback}{15\xspace}

\newcommand{\nRefitLeaves}{4\xspace}

\newcommand{\nRefitLevelDistinctMin}{182\xspace}

\newcommand{\nRefitLr}{0.05\xspace}
\newcommand{\nRefitMinChild}{5\xspace}
\newcommand{\nRefitPairs}{35\xspace}

\newcommand{\nRefitSplits}{300\xspace}

\newcommand{\nRefitTrainFirst}{14\xspace}
\newcommand{\nRefitTrainLast}{48\xspace}
\newcommand{\nRefitTrees}{100\xspace}

\newcommand{\nRefreshTen}{0.913517\xspace}
\newcommand{\nRefreshThousand}{0.999001\xspace}
\newcommand{\nRemovedFirstDate}{January 1, 2025\xspace}
\newcommand{\nReplayAgree}{110\xspace}
\newcommand{\nReplayPairs}{1,000\xspace}

\newcommand{\nResetEvents}{5,048\xspace}
\newcommand{\nResetGroups}{16\xspace}

\newcommand{\nRollingMinPeriods}{5\xspace}
\newcommand{\nRollingWindow}{20\xspace}
\newcommand{\nSaocpArchErcotSteps}{204\xspace}

\newcommand{\nSaocpEta}{0.5\xspace}
\newcommand{\nSaocpFloor}{0.001\xspace}
\newcommand{\nSaocpLifetimeLog}{9\xspace}
\newcommand{\nSaocpLifetimeLogCase}{8\xspace}
\newcommand{\nSaocpLifetimeLogErcot}{8\xspace}
\newcommand{\nSaocpLifetimeLogFomc}{10\xspace}

\newcommand{\nSaocpSynScaleMax}{8.58\xspace}
\newcommand{\nSaocpSynScaleMin}{2.99\xspace}

\newcommand{\nSeedMult}{1,000\xspace}

\newcommand{\nSlopeFreqAbaciDecMeanShiftFour}{-0.923\xspace}
\newcommand{\nSlopeFreqAbaciDecMeanShiftFourSe}{0.009\xspace}

\newcommand{\nSlopeFreqAbaciMax}{-0.993\xspace}
\newcommand{\nSlopeFreqAbaciMeanShiftFour}{-1.004\xspace}
\newcommand{\nSlopeFreqAbaciMeanShiftFourSe}{0.013\xspace}

\newcommand{\nSlopeFreqAbaciMin}{-1.004\xspace}

\newcommand{\nSlopeFreqAciMax}{-0.994\xspace}
\newcommand{\nSlopeFreqAciMeanShiftFour}{-0.994\xspace}
\newcommand{\nSlopeFreqAciMeanShiftFourSe}{0.004\xspace}

\newcommand{\nSlopeFreqAciMin}{-1.008\xspace}

\newcommand{\nSlopeFreqNaiveMax}{-0.438\xspace}
\newcommand{\nSlopeFreqNaiveMeanShiftFour}{-0.522\xspace}
\newcommand{\nSlopeFreqNaiveMeanShiftFourSe}{0.077\xspace}

\newcommand{\nSlopeFreqNaiveMin}{-0.614\xspace}

\newcommand{\nSlopeFreqSegMeanShiftFour}{-0.574\xspace}
\newcommand{\nSlopeFreqSegMeanShiftFourSe}{0.006\xspace}

\newcommand{\nSlopePtAbaciDecMeanShiftFour}{-0.288\xspace}
\newcommand{\nSlopePtAbaciDecMeanShiftFourSe}{0.017\xspace}

\newcommand{\nSlopePtAbaciMeanShiftFour}{0.001\xspace}

\newcommand{\nSlopePtSegMeanShiftFour}{-0.499\xspace}
\newcommand{\nSlopePtSegMeanShiftFourSe}{0.004\xspace}

\newcommand{\nSpendUsd}{0.58\xspace}
\newcommand{\nSrcAgaciMultiple}{100\xspace}

\newcommand{\nSrcFreqBelowFifty}{16\xspace}

\newcommand{\nSrcGridAgaciCount}{30\xspace}

\newcommand{\nSrcGridAgaciList}{0.0, 0.000005, 0.00005, 0.0001, 0.0002, 0.0003, 0.0004, 0.0005, 0.0006, 0.0007, 0.0008, 0.0009, 0.001, 0.002, 0.003, 0.004, 0.005, 0.006, 0.007, 0.008, 0.009, 0.01, 0.02, 0.03, 0.04, 0.05, 0.06, 0.07, 0.08 and 0.09\xspace}
\newcommand{\nSrcGridDtaciCount}{8\xspace}

\newcommand{\nSrcGridDtaciList}{0.001, 0.002, 0.004, 0.008, 0.016, 0.032, 0.064 and 0.128\xspace}
\newcommand{\nSrcPidLr}{0.1\xspace}

\newcommand{\nSrcPtAboveNoBreakTen}{4\xspace}

\newcommand{\nSrcPtBelowLargeMeanTen}{4\xspace}

\newcommand{\nSrcPtReachVarShiftTenWho}{SAOCP\xspace}

\newcommand{\nSrcSynDraws}{10,000\xspace}
\newcommand{\nSrcSynReps}{1,000\xspace}

\newcommand{\nSynCalib}{100\xspace}

\newcommand{\nSynDeltaBothShift}{2.0\xspace}
\newcommand{\nSynDeltaLargeMean}{5.0\xspace}
\newcommand{\nSynDeltaMeanShift}{2.0\xspace}
\newcommand{\nSynDeltaNoBreak}{0.0\xspace}
\newcommand{\nSynDeltaVarShift}{0.0\xspace}
\newcommand{\nSynGamma}{1.0\xspace}

\newcommand{\nSynReps}{1,000\xspace}

\newcommand{\nSynSaltLargeMean}{104\xspace}

\newcommand{\nSynSaltNoBreak}{100\xspace}

\newcommand{\nSynSigmaBothShift}{2.0\xspace}
\newcommand{\nSynSigmaLargeMean}{1.0\xspace}
\newcommand{\nSynSigmaMeanShift}{1.0\xspace}
\newcommand{\nSynSigmaNoBreak}{1.0\xspace}
\newcommand{\nSynSigmaVarShift}{2.0\xspace}
\newcommand{\nSynTpre}{500\xspace}
\newcommand{\nSynTtotal}{5,510\xspace}
\newcommand{\nTdAbaciBothShiftFiftyBoostGm}{0.04484\xspace}
\newcommand{\nTdAbaciBothShiftFiftyCert}{1.000\xspace}

\newcommand{\nTdAbaciBothShiftFiftyCoversPropNoBoost}{1.000\xspace}

\newcommand{\nTdAbaciBothShiftFiftyErr}{0.02064\xspace}

\newcommand{\nTdAbaciBothShiftFiftyJumpGm}{0.00714\xspace}
\newcommand{\nTdAbaciBothShiftFiftyJumpScore}{0.357\xspace}
\newcommand{\nTdAbaciBothShiftFiftyProjAny}{0.000\xspace}
\newcommand{\nTdAbaciBothShiftFiftyProjGm}{0.000000\xspace}
\newcommand{\nTdAbaciBothShiftFiftyProjOverOne}{0.000\xspace}
\newcommand{\nTdAbaciBothShiftFiftyProjShare}{0.000000\xspace}
\newcommand{\nTdAbaciBothShiftFiftyProp}{0.15281\xspace}
\newcommand{\nTdAbaciBothShiftFiftyPropNoBoost}{0.10797\xspace}
\newcommand{\nTdAbaciBothShiftFiftyRangeGm}{0.10083\xspace}

\newcommand{\nTdAbaciBothShiftFiftyRatioPropNoBoost}{5.23\xspace}
\newcommand{\nTdAbaciBothShiftFiveThousandBoostGm}{0.00045\xspace}
\newcommand{\nTdAbaciBothShiftFiveThousandCert}{1.000\xspace}

\newcommand{\nTdAbaciBothShiftFiveThousandCoversPropNoBoost}{1.000\xspace}

\newcommand{\nTdAbaciBothShiftFiveThousandErr}{0.00021\xspace}

\newcommand{\nTdAbaciBothShiftFiveThousandJumpGm}{0.00007\xspace}
\newcommand{\nTdAbaciBothShiftFiveThousandJumpScore}{0.357\xspace}
\newcommand{\nTdAbaciBothShiftFiveThousandProjAny}{0.000\xspace}
\newcommand{\nTdAbaciBothShiftFiveThousandProjGm}{0.000000\xspace}
\newcommand{\nTdAbaciBothShiftFiveThousandProjOverOne}{0.000\xspace}
\newcommand{\nTdAbaciBothShiftFiveThousandProjShare}{0.000000\xspace}
\newcommand{\nTdAbaciBothShiftFiveThousandProp}{0.00171\xspace}
\newcommand{\nTdAbaciBothShiftFiveThousandPropNoBoost}{0.00126\xspace}
\newcommand{\nTdAbaciBothShiftFiveThousandRangeGm}{0.00119\xspace}

\newcommand{\nTdAbaciBothShiftFiveThousandRatioPropNoBoost}{5.89\xspace}
\newcommand{\nTdAbaciBothShiftThousandBoostGm}{0.00224\xspace}
\newcommand{\nTdAbaciBothShiftThousandCert}{1.000\xspace}

\newcommand{\nTdAbaciBothShiftThousandCoversPropNoBoost}{1.000\xspace}

\newcommand{\nTdAbaciBothShiftThousandErr}{0.00110\xspace}

\newcommand{\nTdAbaciBothShiftThousandJumpGm}{0.00036\xspace}
\newcommand{\nTdAbaciBothShiftThousandJumpScore}{0.357\xspace}
\newcommand{\nTdAbaciBothShiftThousandProjAny}{0.000\xspace}
\newcommand{\nTdAbaciBothShiftThousandProjGm}{0.000000\xspace}
\newcommand{\nTdAbaciBothShiftThousandProjOverOne}{0.000\xspace}
\newcommand{\nTdAbaciBothShiftThousandProjShare}{0.000000\xspace}
\newcommand{\nTdAbaciBothShiftThousandProp}{0.00824\xspace}
\newcommand{\nTdAbaciBothShiftThousandPropNoBoost}{0.00600\xspace}
\newcommand{\nTdAbaciBothShiftThousandRangeGm}{0.00564\xspace}

\newcommand{\nTdAbaciBothShiftThousandRatioPropNoBoost}{5.47\xspace}
\newcommand{\nTdAbaciBothShiftTwoHundredBoostGm}{0.01121\xspace}
\newcommand{\nTdAbaciBothShiftTwoHundredCert}{1.000\xspace}

\newcommand{\nTdAbaciBothShiftTwoHundredCoversPropNoBoost}{1.000\xspace}

\newcommand{\nTdAbaciBothShiftTwoHundredErr}{0.00557\xspace}

\newcommand{\nTdAbaciBothShiftTwoHundredJumpGm}{0.00179\xspace}
\newcommand{\nTdAbaciBothShiftTwoHundredJumpScore}{0.357\xspace}
\newcommand{\nTdAbaciBothShiftTwoHundredProjAny}{0.000\xspace}
\newcommand{\nTdAbaciBothShiftTwoHundredProjGm}{0.000000\xspace}
\newcommand{\nTdAbaciBothShiftTwoHundredProjOverOne}{0.000\xspace}
\newcommand{\nTdAbaciBothShiftTwoHundredProjShare}{0.000000\xspace}
\newcommand{\nTdAbaciBothShiftTwoHundredProp}{0.03959\xspace}
\newcommand{\nTdAbaciBothShiftTwoHundredPropNoBoost}{0.02838\xspace}
\newcommand{\nTdAbaciBothShiftTwoHundredRangeGm}{0.02659\xspace}

\newcommand{\nTdAbaciBothShiftTwoHundredRatioPropNoBoost}{5.09\xspace}
\newcommand{\nTdAbaciLargeMeanFiftyBoostGm}{0.07258\xspace}
\newcommand{\nTdAbaciLargeMeanFiftyCert}{1.000\xspace}

\newcommand{\nTdAbaciLargeMeanFiftyCoversPropNoBoost}{1.000\xspace}

\newcommand{\nTdAbaciLargeMeanFiftyErr}{0.02412\xspace}

\newcommand{\nTdAbaciLargeMeanFiftyJumpGm}{0.00722\xspace}
\newcommand{\nTdAbaciLargeMeanFiftyJumpScore}{0.361\xspace}
\newcommand{\nTdAbaciLargeMeanFiftyProjAny}{0.000\xspace}
\newcommand{\nTdAbaciLargeMeanFiftyProjGm}{0.000000\xspace}
\newcommand{\nTdAbaciLargeMeanFiftyProjOverOne}{0.000\xspace}
\newcommand{\nTdAbaciLargeMeanFiftyProjShare}{0.000000\xspace}
\newcommand{\nTdAbaciLargeMeanFiftyProp}{0.21284\xspace}
\newcommand{\nTdAbaciLargeMeanFiftyPropNoBoost}{0.14026\xspace}
\newcommand{\nTdAbaciLargeMeanFiftyRangeGm}{0.13304\xspace}

\newcommand{\nTdAbaciLargeMeanFiftyRatioPropNoBoost}{5.82\xspace}
\newcommand{\nTdAbaciLargeMeanFiveThousandBoostGm}{0.00073\xspace}
\newcommand{\nTdAbaciLargeMeanFiveThousandCert}{1.000\xspace}

\newcommand{\nTdAbaciLargeMeanFiveThousandCoversPropNoBoost}{1.000\xspace}

\newcommand{\nTdAbaciLargeMeanFiveThousandErr}{0.00024\xspace}

\newcommand{\nTdAbaciLargeMeanFiveThousandJumpGm}{0.00007\xspace}
\newcommand{\nTdAbaciLargeMeanFiveThousandJumpScore}{0.361\xspace}
\newcommand{\nTdAbaciLargeMeanFiveThousandProjAny}{0.000\xspace}
\newcommand{\nTdAbaciLargeMeanFiveThousandProjGm}{0.000000\xspace}
\newcommand{\nTdAbaciLargeMeanFiveThousandProjOverOne}{0.000\xspace}
\newcommand{\nTdAbaciLargeMeanFiveThousandProjShare}{0.000000\xspace}
\newcommand{\nTdAbaciLargeMeanFiveThousandProp}{0.00219\xspace}
\newcommand{\nTdAbaciLargeMeanFiveThousandPropNoBoost}{0.00146\xspace}
\newcommand{\nTdAbaciLargeMeanFiveThousandRangeGm}{0.00139\xspace}

\newcommand{\nTdAbaciLargeMeanFiveThousandRatioPropNoBoost}{6.02\xspace}
\newcommand{\nTdAbaciLargeMeanThousandBoostGm}{0.00363\xspace}
\newcommand{\nTdAbaciLargeMeanThousandCert}{1.000\xspace}

\newcommand{\nTdAbaciLargeMeanThousandCoversPropNoBoost}{1.000\xspace}

\newcommand{\nTdAbaciLargeMeanThousandErr}{0.00119\xspace}

\newcommand{\nTdAbaciLargeMeanThousandJumpGm}{0.00036\xspace}
\newcommand{\nTdAbaciLargeMeanThousandJumpScore}{0.361\xspace}
\newcommand{\nTdAbaciLargeMeanThousandProjAny}{0.000\xspace}
\newcommand{\nTdAbaciLargeMeanThousandProjGm}{0.000000\xspace}
\newcommand{\nTdAbaciLargeMeanThousandProjOverOne}{0.000\xspace}
\newcommand{\nTdAbaciLargeMeanThousandProjShare}{0.000000\xspace}
\newcommand{\nTdAbaciLargeMeanThousandProp}{0.01082\xspace}
\newcommand{\nTdAbaciLargeMeanThousandPropNoBoost}{0.00719\xspace}
\newcommand{\nTdAbaciLargeMeanThousandRangeGm}{0.00683\xspace}

\newcommand{\nTdAbaciLargeMeanThousandRatioPropNoBoost}{6.04\xspace}
\newcommand{\nTdAbaciLargeMeanTwoHundredBoostGm}{0.01814\xspace}
\newcommand{\nTdAbaciLargeMeanTwoHundredCert}{1.000\xspace}

\newcommand{\nTdAbaciLargeMeanTwoHundredCoversPropNoBoost}{1.000\xspace}

\newcommand{\nTdAbaciLargeMeanTwoHundredErr}{0.00608\xspace}

\newcommand{\nTdAbaciLargeMeanTwoHundredJumpGm}{0.00181\xspace}
\newcommand{\nTdAbaciLargeMeanTwoHundredJumpScore}{0.361\xspace}
\newcommand{\nTdAbaciLargeMeanTwoHundredProjAny}{0.000\xspace}
\newcommand{\nTdAbaciLargeMeanTwoHundredProjGm}{0.000000\xspace}
\newcommand{\nTdAbaciLargeMeanTwoHundredProjOverOne}{0.000\xspace}
\newcommand{\nTdAbaciLargeMeanTwoHundredProjShare}{0.000000\xspace}
\newcommand{\nTdAbaciLargeMeanTwoHundredProp}{0.05358\xspace}
\newcommand{\nTdAbaciLargeMeanTwoHundredPropNoBoost}{0.03543\xspace}
\newcommand{\nTdAbaciLargeMeanTwoHundredRangeGm}{0.03363\xspace}

\newcommand{\nTdAbaciLargeMeanTwoHundredRatioPropNoBoost}{5.82\xspace}
\newcommand{\nTdAbaciMeanShiftFiftyBoostGm}{0.03228\xspace}
\newcommand{\nTdAbaciMeanShiftFiftyCert}{1.000\xspace}

\newcommand{\nTdAbaciMeanShiftFiftyCoversPropNoBoost}{1.000\xspace}

\newcommand{\nTdAbaciMeanShiftFiftyErr}{0.00808\xspace}

\newcommand{\nTdAbaciMeanShiftFiftyJumpGm}{0.00720\xspace}
\newcommand{\nTdAbaciMeanShiftFiftyJumpScore}{0.360\xspace}
\newcommand{\nTdAbaciMeanShiftFiftyProjAny}{0.000\xspace}
\newcommand{\nTdAbaciMeanShiftFiftyProjGm}{0.000000\xspace}
\newcommand{\nTdAbaciMeanShiftFiftyProjOverOne}{0.000\xspace}
\newcommand{\nTdAbaciMeanShiftFiftyProjShare}{0.000000\xspace}
\newcommand{\nTdAbaciMeanShiftFiftyProp}{0.11590\xspace}
\newcommand{\nTdAbaciMeanShiftFiftyPropNoBoost}{0.08362\xspace}
\newcommand{\nTdAbaciMeanShiftFiftyRangeGm}{0.07642\xspace}

\newcommand{\nTdAbaciMeanShiftFiftyRatioPropNoBoost}{10.35\xspace}
\newcommand{\nTdAbaciMeanShiftFiveThousandBoostGm}{0.00032\xspace}
\newcommand{\nTdAbaciMeanShiftFiveThousandCert}{1.000\xspace}

\newcommand{\nTdAbaciMeanShiftFiveThousandCoversPropNoBoost}{1.000\xspace}

\newcommand{\nTdAbaciMeanShiftFiveThousandErr}{0.00008\xspace}

\newcommand{\nTdAbaciMeanShiftFiveThousandJumpGm}{0.00007\xspace}
\newcommand{\nTdAbaciMeanShiftFiveThousandJumpScore}{0.360\xspace}
\newcommand{\nTdAbaciMeanShiftFiveThousandProjAny}{0.000\xspace}
\newcommand{\nTdAbaciMeanShiftFiveThousandProjGm}{0.000000\xspace}
\newcommand{\nTdAbaciMeanShiftFiveThousandProjOverOne}{0.000\xspace}
\newcommand{\nTdAbaciMeanShiftFiveThousandProjShare}{0.000000\xspace}
\newcommand{\nTdAbaciMeanShiftFiveThousandProp}{0.00122\xspace}
\newcommand{\nTdAbaciMeanShiftFiveThousandPropNoBoost}{0.00090\xspace}
\newcommand{\nTdAbaciMeanShiftFiveThousandRangeGm}{0.00083\xspace}

\newcommand{\nTdAbaciMeanShiftFiveThousandRatioPropNoBoost}{11.05\xspace}
\newcommand{\nTdAbaciMeanShiftThousandBoostGm}{0.00161\xspace}
\newcommand{\nTdAbaciMeanShiftThousandCert}{1.000\xspace}

\newcommand{\nTdAbaciMeanShiftThousandCoversPropNoBoost}{1.000\xspace}

\newcommand{\nTdAbaciMeanShiftThousandErr}{0.00038\xspace}

\newcommand{\nTdAbaciMeanShiftThousandJumpGm}{0.00036\xspace}
\newcommand{\nTdAbaciMeanShiftThousandJumpScore}{0.360\xspace}
\newcommand{\nTdAbaciMeanShiftThousandProjAny}{0.000\xspace}
\newcommand{\nTdAbaciMeanShiftThousandProjGm}{0.000000\xspace}
\newcommand{\nTdAbaciMeanShiftThousandProjOverOne}{0.000\xspace}
\newcommand{\nTdAbaciMeanShiftThousandProjShare}{0.000000\xspace}
\newcommand{\nTdAbaciMeanShiftThousandProp}{0.00597\xspace}
\newcommand{\nTdAbaciMeanShiftThousandPropNoBoost}{0.00436\xspace}
\newcommand{\nTdAbaciMeanShiftThousandRangeGm}{0.00400\xspace}

\newcommand{\nTdAbaciMeanShiftThousandRatioPropNoBoost}{11.56\xspace}
\newcommand{\nTdAbaciMeanShiftTwoHundredBoostGm}{0.00807\xspace}
\newcommand{\nTdAbaciMeanShiftTwoHundredCert}{1.000\xspace}

\newcommand{\nTdAbaciMeanShiftTwoHundredCoversPropNoBoost}{1.000\xspace}

\newcommand{\nTdAbaciMeanShiftTwoHundredErr}{0.00205\xspace}

\newcommand{\nTdAbaciMeanShiftTwoHundredJumpGm}{0.00180\xspace}
\newcommand{\nTdAbaciMeanShiftTwoHundredJumpScore}{0.360\xspace}
\newcommand{\nTdAbaciMeanShiftTwoHundredProjAny}{0.000\xspace}
\newcommand{\nTdAbaciMeanShiftTwoHundredProjGm}{0.000000\xspace}
\newcommand{\nTdAbaciMeanShiftTwoHundredProjOverOne}{0.000\xspace}
\newcommand{\nTdAbaciMeanShiftTwoHundredProjShare}{0.000000\xspace}
\newcommand{\nTdAbaciMeanShiftTwoHundredProp}{0.02924\xspace}
\newcommand{\nTdAbaciMeanShiftTwoHundredPropNoBoost}{0.02117\xspace}
\newcommand{\nTdAbaciMeanShiftTwoHundredRangeGm}{0.01937\xspace}

\newcommand{\nTdAbaciMeanShiftTwoHundredRatioPropNoBoost}{10.33\xspace}
\newcommand{\nTdAbaciNoBreakFiftyBoostGm}{0.00748\xspace}
\newcommand{\nTdAbaciNoBreakFiftyCert}{1.000\xspace}

\newcommand{\nTdAbaciNoBreakFiftyCoversPropNoBoost}{1.000\xspace}

\newcommand{\nTdAbaciNoBreakFiftyErr}{0.00884\xspace}

\newcommand{\nTdAbaciNoBreakFiftyJumpGm}{0.00707\xspace}
\newcommand{\nTdAbaciNoBreakFiftyJumpScore}{0.353\xspace}
\newcommand{\nTdAbaciNoBreakFiftyProjAny}{0.009\xspace}
\newcommand{\nTdAbaciNoBreakFiftyProjGm}{0.000022\xspace}
\newcommand{\nTdAbaciNoBreakFiftyProjOverOne}{0.009\xspace}
\newcommand{\nTdAbaciNoBreakFiftyProjShare}{0.000180\xspace}
\newcommand{\nTdAbaciNoBreakFiftyProp}{0.07459\xspace}
\newcommand{\nTdAbaciNoBreakFiftyPropNoBoost}{0.06710\xspace}
\newcommand{\nTdAbaciNoBreakFiftyRangeGm}{0.06001\xspace}

\newcommand{\nTdAbaciNoBreakFiftyRatioPropNoBoost}{7.59\xspace}
\newcommand{\nTdAbaciNoBreakFiveThousandBoostGm}{0.00007\xspace}
\newcommand{\nTdAbaciNoBreakFiveThousandCert}{1.000\xspace}

\newcommand{\nTdAbaciNoBreakFiveThousandCoversPropNoBoost}{1.000\xspace}

\newcommand{\nTdAbaciNoBreakFiveThousandErr}{0.00009\xspace}

\newcommand{\nTdAbaciNoBreakFiveThousandJumpGm}{0.00007\xspace}
\newcommand{\nTdAbaciNoBreakFiveThousandJumpScore}{0.353\xspace}
\newcommand{\nTdAbaciNoBreakFiveThousandProjAny}{0.010\xspace}
\newcommand{\nTdAbaciNoBreakFiveThousandProjGm}{0.000000\xspace}
\newcommand{\nTdAbaciNoBreakFiveThousandProjOverOne}{0.000\xspace}
\newcommand{\nTdAbaciNoBreakFiveThousandProjShare}{0.000002\xspace}
\newcommand{\nTdAbaciNoBreakFiveThousandProp}{0.00086\xspace}
\newcommand{\nTdAbaciNoBreakFiveThousandPropNoBoost}{0.00079\xspace}
\newcommand{\nTdAbaciNoBreakFiveThousandRangeGm}{0.00072\xspace}

\newcommand{\nTdAbaciNoBreakFiveThousandRatioPropNoBoost}{8.69\xspace}
\newcommand{\nTdAbaciNoBreakThousandBoostGm}{0.00037\xspace}
\newcommand{\nTdAbaciNoBreakThousandCert}{1.000\xspace}

\newcommand{\nTdAbaciNoBreakThousandCoversPropNoBoost}{1.000\xspace}

\newcommand{\nTdAbaciNoBreakThousandErr}{0.00045\xspace}

\newcommand{\nTdAbaciNoBreakThousandJumpGm}{0.00035\xspace}
\newcommand{\nTdAbaciNoBreakThousandJumpScore}{0.353\xspace}
\newcommand{\nTdAbaciNoBreakThousandProjAny}{0.009\xspace}
\newcommand{\nTdAbaciNoBreakThousandProjGm}{0.000001\xspace}
\newcommand{\nTdAbaciNoBreakThousandProjOverOne}{0.000\xspace}
\newcommand{\nTdAbaciNoBreakThousandProjShare}{0.000009\xspace}
\newcommand{\nTdAbaciNoBreakThousandProp}{0.00409\xspace}
\newcommand{\nTdAbaciNoBreakThousandPropNoBoost}{0.00372\xspace}
\newcommand{\nTdAbaciNoBreakThousandRangeGm}{0.00336\xspace}

\newcommand{\nTdAbaciNoBreakThousandRatioPropNoBoost}{8.21\xspace}
\newcommand{\nTdAbaciNoBreakTwoHundredBoostGm}{0.00187\xspace}
\newcommand{\nTdAbaciNoBreakTwoHundredCert}{1.000\xspace}

\newcommand{\nTdAbaciNoBreakTwoHundredCoversPropNoBoost}{1.000\xspace}

\newcommand{\nTdAbaciNoBreakTwoHundredErr}{0.00224\xspace}

\newcommand{\nTdAbaciNoBreakTwoHundredJumpGm}{0.00177\xspace}
\newcommand{\nTdAbaciNoBreakTwoHundredJumpScore}{0.353\xspace}
\newcommand{\nTdAbaciNoBreakTwoHundredProjAny}{0.009\xspace}
\newcommand{\nTdAbaciNoBreakTwoHundredProjGm}{0.000006\xspace}
\newcommand{\nTdAbaciNoBreakTwoHundredProjOverOne}{0.000\xspace}
\newcommand{\nTdAbaciNoBreakTwoHundredProjShare}{0.000045\xspace}
\newcommand{\nTdAbaciNoBreakTwoHundredProp}{0.01945\xspace}
\newcommand{\nTdAbaciNoBreakTwoHundredPropNoBoost}{0.01758\xspace}
\newcommand{\nTdAbaciNoBreakTwoHundredRangeGm}{0.01581\xspace}

\newcommand{\nTdAbaciNoBreakTwoHundredRatioPropNoBoost}{7.83\xspace}

\newcommand{\nTdAbaciVarShiftFiftyBoostGm}{0.03104\xspace}
\newcommand{\nTdAbaciVarShiftFiftyCert}{1.000\xspace}

\newcommand{\nTdAbaciVarShiftFiftyCoversPropNoBoost}{1.000\xspace}

\newcommand{\nTdAbaciVarShiftFiftyErr}{0.01298\xspace}

\newcommand{\nTdAbaciVarShiftFiftyJumpGm}{0.00708\xspace}
\newcommand{\nTdAbaciVarShiftFiftyJumpScore}{0.354\xspace}
\newcommand{\nTdAbaciVarShiftFiftyProjAny}{0.001\xspace}
\newcommand{\nTdAbaciVarShiftFiftyProjGm}{0.000003\xspace}
\newcommand{\nTdAbaciVarShiftFiftyProjOverOne}{0.001\xspace}
\newcommand{\nTdAbaciVarShiftFiftyProjShare}{0.000020\xspace}
\newcommand{\nTdAbaciVarShiftFiftyProp}{0.11900\xspace}
\newcommand{\nTdAbaciVarShiftFiftyPropNoBoost}{0.08796\xspace}
\newcommand{\nTdAbaciVarShiftFiftyRangeGm}{0.08087\xspace}

\newcommand{\nTdAbaciVarShiftFiftyRatioPropNoBoost}{6.78\xspace}
\newcommand{\nTdAbaciVarShiftFiveThousandBoostGm}{0.00031\xspace}
\newcommand{\nTdAbaciVarShiftFiveThousandCert}{1.000\xspace}

\newcommand{\nTdAbaciVarShiftFiveThousandCoversPropNoBoost}{1.000\xspace}

\newcommand{\nTdAbaciVarShiftFiveThousandErr}{0.00014\xspace}

\newcommand{\nTdAbaciVarShiftFiveThousandJumpGm}{0.00007\xspace}
\newcommand{\nTdAbaciVarShiftFiveThousandJumpScore}{0.354\xspace}
\newcommand{\nTdAbaciVarShiftFiveThousandProjAny}{0.001\xspace}
\newcommand{\nTdAbaciVarShiftFiveThousandProjGm}{0.000000\xspace}
\newcommand{\nTdAbaciVarShiftFiveThousandProjOverOne}{0.000\xspace}
\newcommand{\nTdAbaciVarShiftFiveThousandProjShare}{0.000000\xspace}
\newcommand{\nTdAbaciVarShiftFiveThousandProp}{0.00136\xspace}
\newcommand{\nTdAbaciVarShiftFiveThousandPropNoBoost}{0.00105\xspace}
\newcommand{\nTdAbaciVarShiftFiveThousandRangeGm}{0.00098\xspace}

\newcommand{\nTdAbaciVarShiftFiveThousandRatioPropNoBoost}{7.69\xspace}
\newcommand{\nTdAbaciVarShiftThousandBoostGm}{0.00155\xspace}
\newcommand{\nTdAbaciVarShiftThousandCert}{1.000\xspace}

\newcommand{\nTdAbaciVarShiftThousandCoversPropNoBoost}{1.000\xspace}

\newcommand{\nTdAbaciVarShiftThousandErr}{0.00067\xspace}

\newcommand{\nTdAbaciVarShiftThousandJumpGm}{0.00035\xspace}
\newcommand{\nTdAbaciVarShiftThousandJumpScore}{0.354\xspace}
\newcommand{\nTdAbaciVarShiftThousandProjAny}{0.001\xspace}
\newcommand{\nTdAbaciVarShiftThousandProjGm}{0.000000\xspace}
\newcommand{\nTdAbaciVarShiftThousandProjOverOne}{0.000\xspace}
\newcommand{\nTdAbaciVarShiftThousandProjShare}{0.000001\xspace}
\newcommand{\nTdAbaciVarShiftThousandProp}{0.00644\xspace}
\newcommand{\nTdAbaciVarShiftThousandPropNoBoost}{0.00489\xspace}
\newcommand{\nTdAbaciVarShiftThousandRangeGm}{0.00454\xspace}

\newcommand{\nTdAbaciVarShiftThousandRatioPropNoBoost}{7.30\xspace}
\newcommand{\nTdAbaciVarShiftTwoHundredBoostGm}{0.00776\xspace}
\newcommand{\nTdAbaciVarShiftTwoHundredCert}{1.000\xspace}

\newcommand{\nTdAbaciVarShiftTwoHundredCoversPropNoBoost}{1.000\xspace}

\newcommand{\nTdAbaciVarShiftTwoHundredErr}{0.00348\xspace}

\newcommand{\nTdAbaciVarShiftTwoHundredJumpGm}{0.00177\xspace}
\newcommand{\nTdAbaciVarShiftTwoHundredJumpScore}{0.354\xspace}
\newcommand{\nTdAbaciVarShiftTwoHundredProjAny}{0.001\xspace}
\newcommand{\nTdAbaciVarShiftTwoHundredProjGm}{0.000001\xspace}
\newcommand{\nTdAbaciVarShiftTwoHundredProjOverOne}{0.000\xspace}
\newcommand{\nTdAbaciVarShiftTwoHundredProjShare}{0.000005\xspace}
\newcommand{\nTdAbaciVarShiftTwoHundredProp}{0.03080\xspace}
\newcommand{\nTdAbaciVarShiftTwoHundredPropNoBoost}{0.02304\xspace}
\newcommand{\nTdAbaciVarShiftTwoHundredRangeGm}{0.02127\xspace}

\newcommand{\nTdAbaciVarShiftTwoHundredRatioPropNoBoost}{6.61\xspace}

\newcommand{\nTdNonzeroFiftyRatioBoostMax}{4.0\xspace}
\newcommand{\nTdNonzeroFiftyRatioBoostMin}{2.2\xspace}
\newcommand{\nTdNonzeroFiftyRatioPropNoBoostMax}{10.35\xspace}
\newcommand{\nTdNonzeroFiftyRatioPropNoBoostMin}{5.23\xspace}

\newcommand{\nThmChecksSeed}{20260927\xspace}
\newcommand{\nThmCondCount}{1,000\xspace}
\newcommand{\nThmCondHalfThousand}{0.0912\xspace}
\newcommand{\nThmCountsAboveAlpha}{8\xspace}
\newcommand{\nThmCountsAboveAlphaLast}{9\xspace}
\newcommand{\nThmCountsChecked}{50\xspace}
\newcommand{\nThmDkwConst}{0.8883\xspace}
\newcommand{\nThmEndpointCellOne}{20\xspace}
\newcommand{\nThmEndpointCellThree}{200\xspace}
\newcommand{\nThmEndpointCellTwo}{50\xspace}
\newcommand{\nThmEndpointLower}{0.0051\xspace}
\newcommand{\nThmEndpointMax}{0.0621\xspace}
\newcommand{\nThmEndpointUpper}{0.1134\xspace}

\newcommand{\nThmGridPoints}{42\xspace}
\newcommand{\nThmLargestGap}{0.2733\xspace}

\newcommand{\nThmLawCountMax}{1,000\xspace}
\newcommand{\nThmLawCounts}{9\xspace}
\newcommand{\nThmLaws}{4\xspace}
\newcommand{\nThmLowerConst}{0.0464\xspace}
\newcommand{\nThmPerTimeHalfThousand}{0.0131\xspace}
\newcommand{\nThmRefreshCountTen}{10\xspace}
\newcommand{\nThmRefreshCountThousand}{1,000\xspace}
\newcommand{\nThmRefreshFastTen}{0.3986\xspace}
\newcommand{\nThmRefreshFastThousand}{0.4095\xspace}

\newcommand{\nThmSimReps}{200,000\xspace}
\newcommand{\nThmSmallRangeConst}{0.0541\xspace}
\newcommand{\nThmStateHolds}{9\xspace}

\newcommand{\nThmStateStepOne}{0.1\xspace}
\newcommand{\nThmStateStepThree}{1.0\xspace}
\newcommand{\nThmStateStepTwo}{0.5\xspace}
\newcommand{\nThmStateUpdates}{100,000\xspace}
\newcommand{\nThmTwoScoreCov}{0.6333\xspace}

\newcommand{\nThmVarFirst}{0.1064\xspace}
\newcommand{\nThmVarFirstCount}{10\xspace}
\newcommand{\nThmVarLast}{0.0916\xspace}
\newcommand{\nThmVarLastCount}{1,000\xspace}

\newcommand{\nTwinLargeMeanAgaciSrcFiftyPtDiff}{-0.044\xspace}
\newcommand{\nTwinLargeMeanAgaciSrcFiftyPtHi}{-0.041\xspace}
\newcommand{\nTwinLargeMeanAgaciSrcFiftyPtLo}{-0.046\xspace}

\newcommand{\nTwinLargeMeanAgaciSrcTenPtDiff}{0.171\xspace}
\newcommand{\nTwinLargeMeanAgaciSrcTenPtHi}{0.186\xspace}
\newcommand{\nTwinLargeMeanAgaciSrcTenPtLo}{0.157\xspace}

\newcommand{\nTwinLargeMeanDtaciSrcFiftyPtDiff}{-0.006\xspace}
\newcommand{\nTwinLargeMeanDtaciSrcFiftyPtHi}{-0.003\xspace}
\newcommand{\nTwinLargeMeanDtaciSrcFiftyPtLo}{-0.009\xspace}

\newcommand{\nTwinLargeMeanDtaciSrcTenPtDiff}{0.693\xspace}
\newcommand{\nTwinLargeMeanDtaciSrcTenPtHi}{0.704\xspace}
\newcommand{\nTwinLargeMeanDtaciSrcTenPtLo}{0.682\xspace}

\newcommand{\nTwinLargeMeanPidSrcFiftyPtDiff}{-0.443\xspace}
\newcommand{\nTwinLargeMeanPidSrcFiftyPtHi}{-0.437\xspace}
\newcommand{\nTwinLargeMeanPidSrcFiftyPtLo}{-0.448\xspace}

\newcommand{\nTwinLargeMeanPidSrcTenPtDiff}{-0.656\xspace}
\newcommand{\nTwinLargeMeanPidSrcTenPtHi}{-0.646\xspace}
\newcommand{\nTwinLargeMeanPidSrcTenPtLo}{-0.665\xspace}

\newcommand{\nTwinLargeMeanSaocpSrcFiftyPtDiff}{-0.657\xspace}
\newcommand{\nTwinLargeMeanSaocpSrcFiftyPtHi}{-0.649\xspace}
\newcommand{\nTwinLargeMeanSaocpSrcFiftyPtLo}{-0.664\xspace}

\newcommand{\nTwinLargeMeanSaocpSrcTenPtDiff}{-0.696\xspace}
\newcommand{\nTwinLargeMeanSaocpSrcTenPtHi}{-0.689\xspace}
\newcommand{\nTwinLargeMeanSaocpSrcTenPtLo}{-0.703\xspace}

\newcommand{\nTwinMeanShiftAgaciSrcFiftyFreqDiff}{0.1044\xspace}
\newcommand{\nTwinMeanShiftAgaciSrcFiftyFreqHi}{0.1073\xspace}
\newcommand{\nTwinMeanShiftAgaciSrcFiftyFreqLo}{0.1015\xspace}

\newcommand{\nTwinMeanShiftDtaciSrcFiftyFreqDiff}{0.1521\xspace}
\newcommand{\nTwinMeanShiftDtaciSrcFiftyFreqHi}{0.1549\xspace}
\newcommand{\nTwinMeanShiftDtaciSrcFiftyFreqLo}{0.1494\xspace}

\newcommand{\nTwinMeanShiftPidSrcFiftyFreqDiff}{-0.2018\xspace}
\newcommand{\nTwinMeanShiftPidSrcFiftyFreqHi}{-0.1984\xspace}
\newcommand{\nTwinMeanShiftPidSrcFiftyFreqLo}{-0.2053\xspace}

\newcommand{\nTwinMeanShiftSaocpSrcFiftyFreqDiff}{-0.2400\xspace}
\newcommand{\nTwinMeanShiftSaocpSrcFiftyFreqHi}{-0.2359\xspace}
\newcommand{\nTwinMeanShiftSaocpSrcFiftyFreqLo}{-0.2441\xspace}

\newcommand{\nTwoScoreExp}{0.6491\xspace}

\newcommand{\nVerLightgbm}{4.6.0\xspace}
\newcommand{\nVerNumpy}{2.1.3\xspace}
\newcommand{\nVerOpera}{1.2.0\xspace}
\newcommand{\nVerPandas}{2.3.2\xspace}

\newcommand{\nVerPython}{3.13.3\xspace}
\newcommand{\nVerPythonLater}{3.13.0\xspace}
\newcommand{\nVerR}{4.5.3\xspace}
\newcommand{\nVerScipy}{1.18.0\xspace}
\newcommand{\nVerSklearn}{1.6.1\xspace}
\newcommand{\nVintReplayBlock}{20\xspace}
\newcommand{\nVintReplayDiff}{-0.0083\xspace}
\newcommand{\nVintReplayDraws}{5,000\xspace}
\newcommand{\nVintReplayHi}{0.0000\xspace}
\newcommand{\nVintReplayLo}{-0.0175\xspace}
\newcommand{\nVintReplayP}{0.07\xspace}
\newcommand{\nVintReplaySeed}{42\xspace}

\newcommand{\nAbstractNominalPct}{90\xspace}
\newcommand{\nAbstractResetGapPoints}{3.5\xspace}
\newcommand{\nAbstractBestGapPoints}{5.2\xspace}
\newcommand{\nAbstractReductionWord}{a third\xspace}
\newcommand{\nAbstractMissTargetPct}{10\xspace}

\newcommand{\nAbstractNemFrozenCoveragePct}{51.5\xspace}

\newcommand{\nAbstractBreaksWord}{nine\xspace}

\renewcommand{\nLibErcotCapPred}{restart neutral\xspace}
\renewcommand{\nLibErcotCapWinnerCf}{restarted ACI\xspace}
\renewcommand{\nLibErcotRtcbPred}{restart helps\xspace}
\renewcommand{\nLibErcotRtcbWinnerCfCov}{restarted quantile, warm start, Mondrian predictor and ACI\xspace}
\renewcommand{\nLibFomcSinglePred}{restart helps\xspace}
\renewcommand{\nLibFomcSingleWinnerCf}{restarted quantile and Mondrian predictor\xspace}
\renewcommand{\nLibFomcSingleWinnerCfCov}{restarted ACI, ACI and conformal PID\xspace}
\renewcommand{\nLibNemApcPred}{restart helps\xspace}
\renewcommand{\nLibNemFivePred}{restart helps\xspace}
\renewcommand{\nLibNemMpcPred}{restart neutral\xspace}
\renewcommand{\nLibNemMpcWinnerCf}{frozen quantile\xspace}
\renewcommand{\nLibOrdcOnePred}{restart neutral\xspace}
\renewcommand{\nLibOrdcOneWinnerCf}{frozen quantile\xspace}
\renewcommand{\nLibOrdcOneWinnerCfCov}{ACI, DtACI, SAOCP and frozen quantile\xspace}
\renewcommand{\nLibOrdcTwoPred}{restart neutral\xspace}
\renewcommand{\nLibOrdcTwoWinnerCf}{frozen quantile\xspace}
\renewcommand{\nLibOrdcTwoWinnerCfCov}{AgACI, DtACI, naive rule and frozen quantile\xspace}
\renewcommand{\nLibPjmRpfPred}{restart helps\xspace}
\renewcommand{\nLibPjmRpfWinnerCf}{restarted ACI\xspace}

\newlength{\assetcolA}
\newlength{\assetcolB}
\newlength{\assetcolC}
\newlength{\winnercol}
\newlength{\vintcolA}
\newlength{\vintcolB}
\newlength{\compcolA}
\newlength{\twincolA}
\newlength{\mapcolA}
\newlength{\mapcolB}
\newlength{\mapcolC}
\newlength{\mapcolD}
\newlength{\provcolA}
\newlength{\provcolB}
\newlength{\provcolC}
\newlength{\doccolA}
\newlength{\doccolB}
\newlength{\anncolA}
\newlength{\anncolB}
\newlength{\anncolC}
\newlength{\anncolD}
\newlength{\libcolA}
\newlength{\archcolA}
\newlength{\termcolA}
\newlength{\termcolB}
\newlength{\schcolA}
\newlength{\schcolB}
\makeatletter
\g@addto@macro{\UrlBreaks}{\do\-\do\%}
\makeatother
\newcommand{\authoremail}{kl3685@columbia.edu}
\newcommand{\codehost}{\url{https://doi.org/10.5281/zenodo.24203546}\xspace}
\newcommand{\spread}[1]{\unskip\textsubscript{\fontsize{6.2}{7}\selectfont #1}}
\newcommand{\spreadmathsizes}{\DeclareMathSizes{6.2}{6.2}{5}{5}}\spreadmathsizes

\newtheorem{theorem}{Theorem}
\newtheorem{corollary}{Corollary}
\newtheorem{proposition}{Proposition}
\theoremstyle{definition}
\newtheorem{definition}{Definition}
\newtheorem{remark}{Remark}

\title{Announced Breaks Separate Conformal Reliability from Frequency Calibration}

\author{%
  Karl Li \\
  Columbia University \\
  {\normalfont\small\texttt{\authoremail}}
}

\begin{document}
\hypersetup{pdfauthor={Karl Li}}
\raggedbottom

\maketitle

\begin{abstract}
A prediction interval should contain the outcome at a stated rate. Conformal prediction sets its width from past forecast errors. When the process behind a time series changes on an announced date, intervals set from errors before that break can stop covering. After Australia moved to five minute electricity settlement, a \nAbstractNominalPct{}\% interval fixed beforehand covered \nAbstractNemFrozenCoveragePct{}\% of the next \nLibNemFivePost days. One forecaster needs the next interval to cover, another a finished period's share of misses at \nAbstractMissTargetPct{}\%. We show that each need calls for its own restart at the date. We recompute the next interval from the $n$ errors since the break. If these errors are independent, its coverage comes within $O(n^{-1/2})$ of the stated rate, the best order attainable. For a finished period, we restart adaptive conformal inference (ACI). On bounded errors, if a floor under its width adds a bounded total, it brings the share of misses within $O(n^{-1})$ of target but promises nothing about the next interval. After a simulated mean shift of \nSynDeltaLargeMean noise standard deviations, recomputed intervals cover within \nAbstractResetGapPoints percentage points of \nAbstractNominalPct{}\% on average at the \nOrdFifty forecast, \nAbstractReductionWord closer than the \nAbstractBestGapPoints of the best of five baselines. Of \nAbstractBreaksWord announced breaks in power markets and at a central bank, restarting ACI gives a detectable gain only at the \nLibEnteredWord large break.
\end{abstract}

\section{Introduction}
\label{sec:intro}

\begin{figure}[t]
\centering
\includegraphics{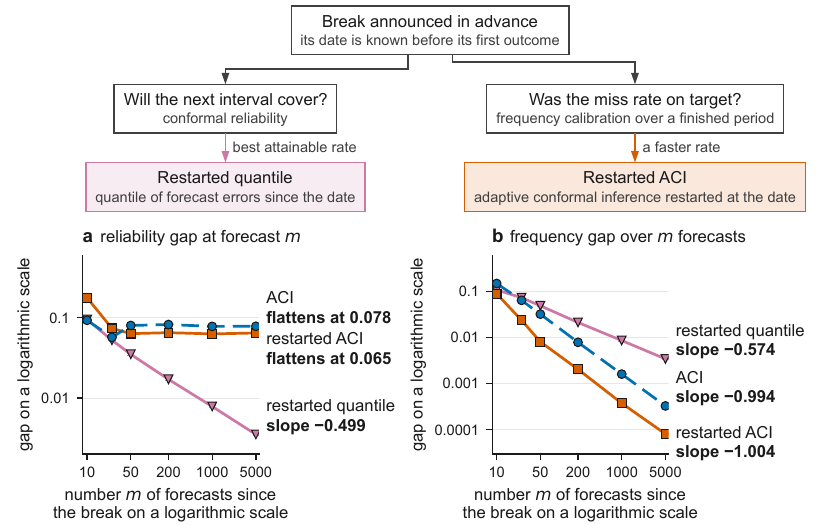}
\caption{An announced break admits two restarts, and each attains the rate of one criterion and misses the other's. The second row of boxes asks what the forecaster needs, coverage of the next interval or a finished period's miss rate on target. Against the number $m$ of forecasts since a simulated mean shift, panel a plots the \emph{reliability gap}, the expected distance of the next interval's coverage from nominal coverage, and panel b the \emph{frequency gap}, the miss rate's distance from its target.}
\label{fig:overview}
\end{figure}

A forecaster of a time series often reports a prediction interval, a range around the forecast that should contain the outcome at a stated rate such as \nAbstractNominalPct{}\%, called its nominal coverage. Conformal prediction sets the interval's half width from the sizes of past forecast errors. A dispatcher bidding on the next delivery period, or a risk manager setting tomorrow's limit, needs the next interval to contain the next outcome at nominal coverage. A period's miss rate is the share of its outcomes that fell outside their intervals. An auditor reviewing a finished quarter needs something else, that quarter's miss rate on its target of \nAbstractMissTargetPct{}\%. We call the need for the next interval to cover at nominal coverage conformal reliability, and the need for a finished period's miss rate to meet its target frequency calibration, the two columns of Figure~\ref{fig:overview}.

The process behind a series can change on a single date, a structural break, after which errors from before the date no longer describe the errors to come. Many breaks are announced in advance. When a market operator announces a rule change, or a central bank publishes its meeting calendar, every user knows the date of the break before its first outcome arrives. Australia's electricity market, for example, moved to five minute settlement on \nLibNemFiveDate, and a \nAbstractNominalPct{}\% interval set from the errors before the switch, then held fixed, covered \nAbstractNemFrozenCoveragePct{}\% of the \nLibNemFivePost days that followed.

Adaptive conformal inference (ACI) widens the interval after each miss and narrows it after each covered outcome \citep{gibbs2021aci}. Under arbitrary change this keeps the miss rate on target in the long run, so ACI serves the auditor. ACI and its variants learn of a break only from their own errors \citep{zaffran2022agaci,gibbs2024dtaci,bhatnagar2023saocp,angelopoulos2023pid}. None of their guarantees uses the date. Mondrian and group-weighted conformal prediction do take a known partition of the data, and they bound coverage within each part as an average over random draws of that part's data \citep{vovk2003mondrian,bhattacharyya2024groupweighted}. We ask what a restart that starts a conformal method afresh at the announced date guarantees a known number of forecasts later, and which of the two needs it serves.

We study two restarts at the announced date, and Figure~\ref{fig:overview} sets them side by side. Conformal methods call the absolute forecast errors scores. Split conformal prediction sets the half width at a quantile of scores held apart from the forecaster's training. The restarted quantile is split conformal prediction that discards its scores at each announced date. Restarted ACI runs ACI's recursion from the quantile of the scores at that date. Each restart serves one criterion and misses the other, the restarted quantile serving conformal reliability and restarted ACI frequency calibration.

\begin{enumerate}
\item \textbf{Two criteria.} In Section~\ref{sec:setup} we cut the stream into cells at the announced dates and measure each criterion by its gap from its target.
\item \textbf{The best attainable rate for the next interval.} In Section~\ref{sec:theory} we prove that, for independent scores, the restarted quantile's next interval approaches nominal coverage as one over the square root of the number of scores since the break. No method told the dates does better in the worst case over score distributions whose density near the quantile lies within fixed bounds.
\item \textbf{A faster rate for a finished period.} We prove there that, on bounded scores, restarted ACI's miss rate over a finished period approaches its target as one over the number of forecasts, as long as the total by which a floor lifts its width stays bounded. A miss rate on target promises nothing about the next interval.
\item \textbf{Evidence across break sizes.} In Sections~\ref{sec:pertime} to~\ref{sec:dose} we test both rates on simulated breaks and on \nLibCases real breaks, each dated by the document that announced it. The restarted quantile's next interval lies closest to nominal coverage after a large simulated break, and most real breaks are too small for restarted ACI to give a detectable gain.
\end{enumerate}

\section{Problem setup}
\label{sec:setup}

We cut the stream into cells at the announced dates, one cell from each break to the next, and we measure each criterion by a gap from its target. Each observation pairs a covariate vector $X_t$ with a real outcome $Y_t$. A forecaster $\hat f$ is trained once on data from before the stream and then frozen. We measure each forecast's miss by its score, the absolute forecast error $s_t=|Y_t-\hat f(X_t)|$. At time $t$ a method issues $C_t=[\hat f(X_t)-\hat q_t,\,\hat f(X_t)+\hat q_t]$, with the threshold $\hat q_t$ fixed from the past before $Y_t$ arrives, so the threshold is the interval's half width. The miscoverage level $\alpha$ is the target miss rate, $\alpha=\nAlpha$ on every stream, so nominal coverage is \nNominalCoverage.

Informally, the class below holds every stream whose scores keep one law between announced dates, with dependence that fades over time so that a cell's scores can estimate its quantile. Mixing is the field's measure of that fading, and it means that scores far apart in time are close to independent.

\begin{definition}[Announced break class]
\label{def:ssi}
A structural partition $\pi\colon[T]\to[K]$ cuts the stream into contiguous cells at boundaries $\tau_1<\cdots<\tau_{K-1}$, each known before the first outcome of the cell it opens. Within cell $k$ the scores share a marginal distribution function $F_k$, and their $\beta$-mixing coefficients satisfy $\beta_k(j)\le C_\beta\rho^j$. The method receives $\pi$ but not the laws. For fixed $\alpha\in(0,1)$, $\rho\in[0,1)$, $C_\beta>0$ and density bounds $0<f_{\min}\le f_{\max}$, the class $\mathcal{H}$ contains every partition whose $F_k$ are continuous with a density between $f_{\min}$ and $f_{\max}$ on a neighbourhood of the $(1-\alpha)$-quantile, and whose $\beta_k$ meet that bound.
\end{definition}

Intuitively, the two gaps measure the two needs of Section~\ref{sec:intro} on one cell. The frequency gap looks back over the cell's first forecasts and asks how far their miss rate fell from the miscoverage level. The reliability gap looks at a single forecast and asks how far the chance that its interval covers lies from nominal coverage, on average over the scores its threshold was built from. For example, a threshold that sits too high for part of a period and too low for the rest can leave the period's miss rate on target while no single interval covers at nominal coverage.

\begin{definition}[Two coverage metrics]
\label{def:metrics}
Fix a cell $k$ of length at least $M$ whose last outcome before the boundary is $\tau$, so the cell runs from $\tau+1$, and average over its first $M$ forecasts. Its frequency gap is
\[
G^{\mathrm{freq}}_M
:=
\left|
\bar{\alpha}_M - \alpha
\right|,
\qquad
\bar{\alpha}_M
=
\frac{1}{M}\sum_{t=\tau+1}^{\tau+M}
\mathbf{1}\{Y_t \notin C_t\}.
\]
Its reliability gap at time $t$ is the $L^1$ distance
\[
\Delta_t
:=
\mathbb{E}\bigl[\,|F_k(\hat q_t) - (1-\alpha)|\,\bigr],
\]
where the expectation runs over everything the threshold uses, the cell's scores before $t$ and, for a method that keeps it, the history before the boundary. The frequency gap is our measure of frequency calibration, and the reliability gap is our measure of conformal reliability.
\end{definition}

\paragraph{Counting.} We index a boundary by the last outcome before it, so a break effective at index $d$ has its boundary at $d-1$. The $m$th forecast of a cell is at $\tau+m$ and uses the cell's $m-1$ completed scores, so the first uses none. A reliability gap at $m$ is the gap of the $m$th forecast, and a frequency gap at $m$ averages the first $m$ forecasts.

\paragraph{Real streams.} On a real stream the cell's law is unknown, so we measure each forecast's threshold against the cell's scores from that forecast on. We call these unseen scores the holdout reference. A threshold's coverage is the share of reference scores at or below it.

Every guarantee below is valid for any frozen forecaster, correct or misspecified, because each concerns the law of the score within a cell and that law already includes the forecaster's error.

\subsection{The two restarts}
\label{sec:abaci}

At each announced break the restarted quantile starts afresh from an empty score set, while restarted ACI moves its threshold to a fresh quantile and then takes larger steps for a while. Both methods learn of a break once the outcome before it has arrived, so the restarted quantile's first forecast in a cell, by Algorithm~\ref{alg:segsplit}, uses no score. We write $\mathrm{err}_t := \mathbf{1}\{|Y_t - \hat f(X_t)| > \hat q_t\}$ for the miss indicator. Restarted ACI, Algorithm~\ref{alg:abaci}, runs ACI's recursion on the score scale, moving the threshold after each outcome by its step times $\mathrm{err}_t-\alpha$ and holding it above a floor. At a boundary it first takes this update on the boundary's score and then jumps to the quantile of every score evaluated so far.

\begin{algorithm}[H]
\caption{\textbf{The restarted quantile.} This is split conformal prediction restarted at each boundary. It takes the plug-in quantile of the current cell's $n$ scores, interpolating linearly between the order statistics whose ranks bracket $1+(n-1)(1-\alpha)$, and it empties its score set after each boundary's outcome.}
\label{alg:segsplit}
\begin{algorithmic}[1]
\Require Forecaster $\hat f$, coverage $1{-}\alpha$, boundaries $\{\tau_j\}$, calibration scores. \quad $\mathcal{S}\gets$ calibration.
\For{$t = 1, 2, \ldots$}
  \State $\hat q_t \gets \mathrm{Quantile}_{1{-}\alpha}(\mathcal{S})$ if $|\mathcal{S}|\ge\lceil 1/(1{-}\alpha)\rceil$, else $\hat q_t \gets +\infty$
  \State Output $C_t = [\hat f(X_t)\pm\hat q_t]$ and observe $Y_t$
  \If{$t\in\{\tau_j\}$}
    \State $\mathcal{S}\gets\emptyset$
  \Else
    \State Append $|Y_t-\hat f(X_t)|$ to $\mathcal{S}$
  \EndIf
\EndFor
\end{algorithmic}
\end{algorithm}

Restarted ACI has four forms, which Appendix~\ref{app:synthetic} states. The boosted form, Algorithm~\ref{alg:abaci}, is our default, and the burn-in form restarts at every boundary with a fixed step. With $\kappa=1$ and no jump, and with its floor never reached, Algorithm~\ref{alg:abaci} is the pure recursion of Theorem~\ref{thm:freq}.

\begin{algorithm}[H]
\caption{\textbf{Restarted adaptive conformal inference (ACI) in its boosted form.} After the outcome at boundary $\tau$ the threshold jumps by $J_b$ to the linearly interpolated quantile of every evaluated score, and the new threshold enters the forecast at $\tau+1$. The step then rises to $\kappa\gamma$ for $b$ updates and decays back to $\gamma$.}
\label{alg:abaci}
\begin{algorithmic}[1]
\Require Forecaster $\hat f$, coverage $1{-}\alpha$, calibration scores, one boundary $\tau$, step $\gamma$, boost $\kappa\ge 1$, boost length $b$ and floor $\hat q_{\min}>0$. \quad $\hat q\gets\mathrm{Quantile}_{1{-}\alpha}(\text{calibration})$ and $\mathcal{S}\gets\emptyset$.
\For{$t = 1, 2, \ldots$}
  \State Output $C_t=[\hat f(X_t)\pm\hat q]$ and observe $Y_t$
  \State $\mathrm{err}_t \gets \mathbf{1}\{|Y_t-\hat f(X_t)|>\hat q\}$ and append $|Y_t-\hat f(X_t)|$ to $\mathcal{S}$
  \State $\gamma_t\gets\gamma$ if $t\le\tau$, else $\gamma_t\gets\gamma\max\{1,\kappa/\sqrt{m_t}\}$ with $m_t=\max\{1,t-\tau-b+1\}$
  \State $\hat q\gets\max\{\hat q+\gamma_t(\mathrm{err}_t-\alpha),\,\hat q_{\min}\}$
  \If{$t=\tau$}
    \State $J_b\gets\mathrm{Quantile}_{1{-}\alpha}(\mathcal{S})-\hat q$ and $\hat q\gets\hat q+J_b$
  \EndIf
\EndFor
\end{algorithmic}
\end{algorithm}

\section{The two rates}
\label{sec:theory}

This section proves what each restart guarantees on its own criterion, which tells a forecaster which restart to use. The restarted quantile's reliability gap falls as one over the square root of the number of scores since the break, and no method told the dates does better. Restarted ACI's frequency gap falls as one over the number of forecasts, a faster rate that says nothing about the next interval.

Intuitively, the restarted quantile is a sample quantile of the scores since the break, and the share of a cell's law below a sample quantile of $n$ independent scores wanders about one over the square root of $n$ around its target, as the empirical distribution function does.

\begin{theorem}[Reliability gap of the restarted quantile]
\label{thm:upper}
Fix $\alpha\in(0,1)$ and a cell $k$ of $\mathcal{H}$ opened by an announced break, and run Algorithm~\ref{alg:segsplit} with cutoff $c=\lceil 1/(1-\alpha)\rceil$. Take a forecast at time $t$ on $n_t$ completed scores since the restart, with threshold $\hat q_t$, and let $D_{n_t}=\sup_x|\widehat F_{n_t}(x)-F_k(x)|$ for the empirical distribution function $\widehat F_{n_t}$ of those scores. If $n_t<c$, the threshold is infinite and $\Delta_t=\alpha$. If $n_t\ge c$, then on every realization and under any dependence within the cell,
\begin{equation}
\bigl|F_k(\hat q_t)-(1-\alpha)\bigr|\le D_{n_t}+\frac{\max(\alpha,1-\alpha)}{n_t}.
\label{eq:sandwich}
\end{equation}
For independent scores within the cell and every $n_t\ge c$,
\begin{equation}
\Delta_t\le\frac{\nThmDkwConst}{\sqrt{n_t}}+\frac{\max(\alpha,1-\alpha)}{n_t}
\quad\text{and}\quad
\Delta_t\le\Bigl(\frac{1}{2(n_t+2)}+\frac{8\max(\alpha,1-\alpha)^2}{(n_t+1)^2}\Bigr)^{1/2}.
\label{eq:upperbound}
\end{equation}
Under the class's geometric $\beta$-mixing, $\Delta_t\le C\log n_t/\sqrt{n_t}$ for every $n_t\ge c$, with $C$ depending only on $C_\beta$ and $\rho$.
\end{theorem}

For independent scores the restarted quantile's coverage from the cutoff on is thus within a known distance of nominal coverage whatever the forecaster and the cell's law, and that distance halves each time the number of scores quadruples. Appendix~\ref{app:mondrian} compares it with a Mondrian predictor.

Intuitively, each miss raises the threshold and each cover lowers it, so a threshold confined to a bounded range forces the period's miss rate back toward the miscoverage level as the period grows.

\begin{theorem}[Frequency gap of the restarted recursion]
\label{thm:freq}
Run the pure recursion $\hat q_{t+1}=\hat q_t+\gamma(\mathrm{err}_t-\alpha)$ from the quantile of the scores before the boundary $\tau$ of an announced break, with no jump and no floor, on scores bounded almost surely by $B$. Its threshold stays in $[-\alpha\gamma,\,B+(1-\alpha)\gamma]$ at every $t$, and for every $M\ge1$ its frequency gap satisfies $G^{\mathrm{freq}}_M\le(B/\gamma+1)/M$ almost surely.
\end{theorem}

This is the telescoping identity of \citet[Proposition~4.1]{gibbs2021aci} started at the boundary, so the restart changes the constant and leaves the exponent.

Intuitively, Algorithm~\ref{alg:abaci} departs from the pure recursion by its jump at the break and afterwards by its floor and its boosted step, and each departure adds its own term to the telescoping sum.

\begin{proposition}[The terms Algorithm~\ref{alg:abaci} adds]
\label{prop:practical-corrections}
Run Algorithm~\ref{alg:abaci} on scores in $[0,B]$ almost surely, with floor $\hat q_{\min}\in(0,B]$ and with its initial threshold and its threshold after the restart in $[0,B]$, over its first $M$ forecasts after the boundary. Let $\gamma_t$ be its step at update $t$ and $J_b$ its jump at the boundary. Write $\Pi_t:=\bigl[\hat q_{\min}-(\hat q_t+\gamma_t(\mathrm{err}_t-\alpha))\bigr]_+$ for the floor correction and $B_M:=\bigl|\sum_{t=\tau+1}^{\tau+M}(\gamma_t-\gamma)(\mathrm{err}_t-\alpha)\bigr|$ for the boost term. Every threshold lies in $[0,\,B+(1-\alpha)\kappa\gamma]$, so the range $R_M$ of the thresholds $\hat q_{\tau+1},\ldots,\hat q_{\tau+M+1}$ is at most $B+(1-\alpha)\kappa\gamma\le B+\kappa\gamma$. Then
\begin{equation}
G^{\mathrm{freq}}_M\le\frac{R_M+|J_b|+\sum_{t=\tau+1}^{\tau+M}\Pi_t+B_M}{\gamma M},
\label{eq:practical-correction-main}
\end{equation}
with $B_M\le\max\{\alpha,1-\alpha\}(\kappa-1)(b+\kappa^2)\gamma$ for every $M$. Algorithm~\ref{alg:abaci} therefore keeps the $O(1/M)$ rate on every cell where $|J_b|+\sum_t\Pi_t=O(1)$.
\end{proposition}

A user can compute every term of \eqref{eq:practical-correction-main} online, and Appendix~\ref{app:synthetic} checks the bound on every run.

Intuitively, ACI without a restart moves its threshold by at most one step per outcome, so after a break it needs as many outcomes as the steps that separate the old quantile from the new one.

\begin{proposition}[Recovery time of ACI at a fixed step]
\label{thm:detection}
Run ACI without a restart at the step $\gamma=c_\gamma/\sqrt{T}$ that gives the rate of the frequency gap of \citet{gibbs2021aci} in the long run, on i.i.d.\ scores from $F_{\mathrm{pre}}$ followed by i.i.d.\ scores from $F_{\mathrm{post}}$. Let their $(1-\alpha)$-quantiles be $q^*_{\mathrm{pre}}$ and $q^*_{\mathrm{post}}$, with $\xi=|q^*_{\mathrm{post}}-q^*_{\mathrm{pre}}|>0$. If $|\hat q_{\tau+1}-q^*_{\mathrm{pre}}|\le\xi/4$ and the density after the break is at least $f_{\min}$ between the two quantiles, then $\Delta_{\tau+M}\le\epsilon$ with $\epsilon<\xi f_{\min}/2$ requires $M\ge c_5\sqrt{T}$, where $c_5=\xi/(4c_\gamma)$ up to the indexing convention of one step. On the same sequence Algorithm~\ref{alg:segsplit} reaches $\Delta_t\le\epsilon$ after $n_t=O(1/\epsilon^2)$ scores since the break, whatever $T$ and $\gamma$.
\end{proposition}

ACI at that step thus needs a number of forecasts that grows with the square root of the stream's length before its threshold recovers, while the restarted quantile's recovery does not depend on it.

Intuitively, two laws of the scores that differ only near the quantile are hard to tell apart from a few scores, and a method that cannot tell them apart must miss the quantile of one of them.

\begin{theorem}[Le Cam lower bound for the reliability gap]
\label{thm:lower}
Let $h=\min(\alpha,1-\alpha)$, and let the class's density bounds admit $[1/2,3/2]$. Fix a cell $k$ and the forecast $t_M=\tau_k+M+1$ after $M\ge1$ completed scores since the restart. Over every conformal method $P$ given the announced partition, $\inf_P\sup_{\mathcal{H}}\Delta_{t_M}\ge c_Ah/\sqrt M$, and the bound rises to $c_A\sqrt{h/M}$ once $M\ge 3/(2h)$, where $c_A=\sqrt{3/8}\,e^{-1/2}/8=\nThmLowerConst$.
\end{theorem}

No method told the boundaries can therefore beat the restarted quantile's order on the worst law of the class, so that order is the minimax rate. We prove the bound in Appendix~\ref{app:proof4} by Le~Cam's two-point method. The same $n^{-1/2}$ rate is minimax for the coverage of each sample over an independent calibration sample of size $n$ \citep{areces2024two,duchi2025few}, with the number of scores since the restart in place of $n$.

\begin{corollary}[Minimax rate of the reliability gap at fixed parameters]
\label{cor:rate}
At fixed $\alpha$, $C_\beta$, $\rho$ and density bounds admitting $[1/2,3/2]$, the minimax reliability gap over the cells of independent scores of $\mathcal{H}$ after $M$ completed scores since the restart is $\Theta(M^{-1/2})$, and Algorithm~\ref{alg:segsplit} attains it. Under geometric $\beta$-mixing the restarted quantile attains $O(\log M/\sqrt M)$. Theorem~\ref{thm:freq}'s $G^{\mathrm{freq}}_M=O(M^{-1})$ concerns a different quantity, so a small realized frequency gap bounds nothing about the reliability gap of the threshold in force.
\end{corollary}

The rate needs fixed parameters, as Appendix~\ref{app:corollary} shows.

Intuitively, a restart at a date where nothing changes throws away scores that still describe the stream, so its quantile is only as precise as the scores since the date allow.

\begin{proposition}[Expected cost of a restart when no break comes]
\label{prop:safety}
Suppose the announced break does not come, so the scores on both sides of the boundary form one stream of the class with distribution function $F$ and $(1-\alpha)$-quantile $q^\star$, and run Algorithm~\ref{alg:segsplit} after the boundary. Its reliability gap then satisfies the bounds of Theorem~\ref{thm:upper}. Let $t_M=\tau+M+1$, let $\hat q^{\mathrm{seg}}_{t_M}$ be the restarted quantile of the $M$ scores since the restart, and let $\hat q^{\mathrm{NR}}_{t_M}$ be the quantile of all $T_M=\tau+M$ scores before the same forecast. If $\mathbb{E}|\hat q^{\mathrm{seg}}_{t_M}-q^\star|\le C_q/\sqrt{M}$ and $\mathbb{E}|\hat q^{\mathrm{NR}}_{t_M}-q^\star|\ge c_q/\sqrt{T_M}$, then the widths $W=2\hat q$ satisfy $\mathbb{E}|W^{\mathrm{seg}}_{t_M}-2q^\star|\le(C_q/c_q)\sqrt{T_M/M}\,\mathbb{E}|W^{\mathrm{NR}}_{t_M}-2q^\star|$.
\end{proposition}

A restart at a date where nothing changes thus costs at most a factor in the width's error that grows with the square root of the history it discards. We check restarted ACI in the same case in Appendix~\ref{app:nobreak}.

Intuitively, the reliability gap averages coverage over the scores the threshold was built from, and only dependence on those scores separates it from coverage given them.

\begin{proposition}[Conditional coverage]
\label{prop:conditional}
In a cell $k$ of $\mathcal{H}$, let $G_n(x)$ be the conditional probability that the next score is at most $x$ given the cell's first $n$ scores, and let $\hat q$ be the restarted quantile's threshold after those scores. Then $\bigl|\mathbb{E}|G_n(\hat q)-(1-\alpha)|-\Delta\bigr|\le\beta_k(1)$. For independent scores within the cell the two are equal, and $\Delta$ is the expected error of the next interval's conditional coverage.
\end{proposition}

\section{Reliability after a break}
\label{sec:pertime}

After a large synthetic break the restarted quantile has the smallest reliability gap in Table~\ref{tab:pertime} at the \nOrdFifty forecast, \nFalsLmResetGap against \nPtBestLargeMeanFifty for the best of five baselines. On two real streams, whose breaks are smaller, it beats \nCfErcotSideBelowCountWordFifty baselines but not ACI after a market reform. It trails all five after policy decisions. The experiment tests whether the restarted quantile's advantage on the next interval, which Section~\ref{sec:theory} proves in order, shows at the numbers of forecasts a user meets after a break.

\begin{table}[!t]
\caption{The restarted quantile has the smallest reliability gap after the large synthetic break and beats \nCfErcotSideBelowCountWordFifty baselines on the ERCOT window, but trails all five on the FOMC stream. Each subscript is a band, from the \nReadBandLoPct to the \nReadBandHiPct percentile of the restarted quantile's mean gap less the method's over \nSrcSynDraws resamples.}
\label{tab:pertime}
\centering
\footnotesize
\setlength{\tabcolsep}{6pt}
\begin{tabular}{llll}
\toprule
method & \begin{tabular}[c]{@{}c@{}}large mean shift\\at the \nOrdFifty forecast\end{tabular} & \begin{tabular}[c]{@{}c@{}}ERCOT window\\at the \nOrdFifty day\end{tabular} & \begin{tabular}[c]{@{}c@{}}FOMC stream\\at the \nOrdTen day\end{tabular} \\
\midrule
restarted quantile & \nPtLargeMeanSegFifty & \nCfErcotSegFifty & \nCfFomcSegMultiTen \\
restarted ACI & \nPtLargeMeanAbaciFifty\spread{$[\nPtBandLargeMeanAbaciFiftyLo,\, \nPtBandLargeMeanAbaciFiftyHi]$} & \nCfErcotAbaciFifty\spread{$[\nCfErcotPairAbaciFiftyLo,\, \nCfErcotPairAbaciFiftyHi]$} & \nCfFomcAbaciTen\spread{$[\nCfFomcPairAbaciTenLo,\, \nCfFomcPairAbaciTenHi]$} \\
ACI & \nPtLargeMeanAciFifty\spread{$[\nPtBandLargeMeanAciFiftyLo,\, \nPtBandLargeMeanAciFiftyHi]$} & \nCfErcotAciFifty\spread{$[\nCfErcotPairAciFiftyLo,\, \nCfErcotPairAciFiftyHi]$} & \nCfFomcAciTen\spread{$[\nCfFomcPairAciTenLo,\, \nCfFomcPairAciTenHi]$} \\
DtACI & \nPtLargeMeanDtaciSrcFifty\spread{$[\nPtBandLargeMeanDtaciSrcFiftyLo,\, \nPtBandLargeMeanDtaciSrcFiftyHi]$} & \nCfErcotDtaciSrcFifty\spread{$[\nCfErcotPairDtaciSrcFiftyLo,\, \nCfErcotPairDtaciSrcFiftyHi]$} & \nCfFomcDtaciSrcTen\spread{$[\nCfFomcPairDtaciSrcTenLo,\, \nCfFomcPairDtaciSrcTenHi]$} \\
SAOCP & \nPtLargeMeanSaocpSrcFifty\spread{$[\nPtBandLargeMeanSaocpSrcFiftyLo,\, \nPtBandLargeMeanSaocpSrcFiftyHi]$} & \nCfErcotSaocpSrcFifty\spread{$[\nCfErcotPairSaocpSrcFiftyLo,\, \nCfErcotPairSaocpSrcFiftyHi]$} & \nCfFomcSaocpSrcTen\spread{$[\nCfFomcPairSaocpSrcTenLo,\, \nCfFomcPairSaocpSrcTenHi]$} \\
conformal PID & \nPtLargeMeanPidSrcFifty\spread{$[\nPtBandLargeMeanPidSrcFiftyLo,\, \nPtBandLargeMeanPidSrcFiftyHi]$} & \nCfErcotPidSrcFifty\spread{$[\nCfErcotPairPidSrcFiftyLo,\, \nCfErcotPairPidSrcFiftyHi]$} & \nCfFomcPidSrcTen\spread{$[\nCfFomcPairPidSrcTenLo,\, \nCfFomcPairPidSrcTenHi]$} \\
AgACI & \nPtLargeMeanAgaciSrcFifty\spread{$[\nPtBandLargeMeanAgaciSrcFiftyLo,\, \nPtBandLargeMeanAgaciSrcFiftyHi]$} & \nCfErcotAgaciSrcFifty\spread{$[\nCfErcotPairAgaciSrcFiftyLo,\, \nCfErcotPairAgaciSrcFiftyHi]$} & \nCfFomcAgaciSrcTen\spread{$[\nCfFomcPairAgaciSrcTenLo,\, \nCfFomcPairAgaciSrcTenHi]$} \\
\bottomrule
\end{tabular}
\end{table}

\paragraph{Baselines.} Five baselines, ACI and four of its variants, learn of a break only from their errors, so they show what a forecaster gets without the date. Dynamically-tuned ACI (DtACI) tunes ACI's step online \citep{gibbs2024dtaci}. Two variants aggregate experts, strongly adaptive online conformal prediction (SAOCP) over experts started at different times \citep{bhatnagar2023saocp} and aggregated ACI (AgACI) over ACI experts run at different steps \citep{zaffran2022agaci}. Conformal PID steers the miss rate toward its target with a proportional-integral-derivative (PID) controller \citep{angelopoulos2023pid}. The variants run from their official implementations. One rule, which Appendix~\ref{app:synthetic} states, sets every constant they leave open, and no tuning budget is matched across methods.

\paragraph{Streams.} The synthetic family gives breaks of known size and law, and its five configurations run \nSynReps replications each. The main comparison uses its large mean shift of \nSynDeltaLargeMean noise standard deviations. We call the days of a real series around one announced break its window. The first real stream is the ERCOT window, the daily real-time price at a Texas hub around \nErcotBreakDate, when the Electric Reliability Council of Texas (ERCOT) launched real-time co-optimization plus batteries \citep{li2026battery}, with \nErcotPost evaluated days after the reform. The second is the daily change in the ten-year Treasury yield across \nFomcBreaks decisions of the Federal Open Market Committee (FOMC), each of which opens a cell. We fit one forecaster per stream once and freeze it, and Appendix~\ref{app:ercot} describes them.

\paragraph{Statistics.} We measure the reliability gap against the known law after a synthetic break and against the holdout reference on a real stream. The holdout reference is finite, and it stands in for a cell's law whose constancy within the cell we did not test. Each comparison reports the restarted quantile's gap less another method's with a paired band, the \nReadBandLoPct to \nReadBandHiPct percentile over bootstrap resamples, and a band wholly below zero favours the restarted quantile. We fixed the rule for constants before reading any evaluated outcome, and the rule of Section~\ref{sec:dose} for a gain a window can detect before reading any price. We chose the \nOrdTen day after an FOMC decision, having seen results at the \nOrdTen and the \nOrdTwentyFive days, as the latest day at which every cell contributes and the median reference holds at least \nFomcMinReference scores. We also chose, after seeing the counts, to group the methods that do best on each announced break by whether they use the announcement, as Appendix~\ref{app:library} states.

\paragraph{The restarted quantile leads after a large break.} At the \nOrdFifty forecast its gap in Table~\ref{tab:pertime} is \nFalsLmResetGap, and the five baselines range from \nPtLargeMeanAgaciSrcFifty for AgACI to \nPtLargeMeanDtaciSrcFifty for DtACI, each band below zero. The bands already lie below zero at the \nOrdTen forecast. Without a break the restart pays for the history it discards, with a gap of \nPtNoBreakSegFifty against \nPtBestNoBreakFifty for \nPtBestNoBreakFiftyWho. Appendix~\ref{app:synthetic} reports every configuration.

\paragraph{After a market reform it beats \nCfErcotSideBelowCountWordFifty baselines.} At the \nOrdFifty day after the reform the restarted quantile's gap is \nCfErcotSegFifty on a holdout reference of \nCfErcotRefFifty scores. Its bands lie below zero against \nCfErcotSideBelowWhoFifty and include zero against \nCfErcotSideIncludeWhoFifty. Against \nCfErcotSidePointWhoFifty both thresholds cover the same reference days, so the gaps agree on every resample. Averaged over the first \nCfErcotCellForecasts forecasts, whose references have at least \nCfErcotMinReference scores, ACI has the smallest gap of all seven methods, \nCfErcotCellAci.

\paragraph{After policy decisions it trails all five.} The restarted quantile starts afresh at every decision. At the \nOrdTen day after a decision its gap averages \nCfFomcSegMultiTen over the \nFomcDayCellsTen cells, whose references have a median of \nFomcDayRefMedianTen scores and at least \nFomcDayRefSmallestTen. Its band lies above zero against all five baselines, of which \nCfFomcBestTenWho does best at \nCfFomcBestTen.

\section{Calibration after a break}
\label{sec:frequency}

\begin{figure}[!t]
\centering
\includegraphics{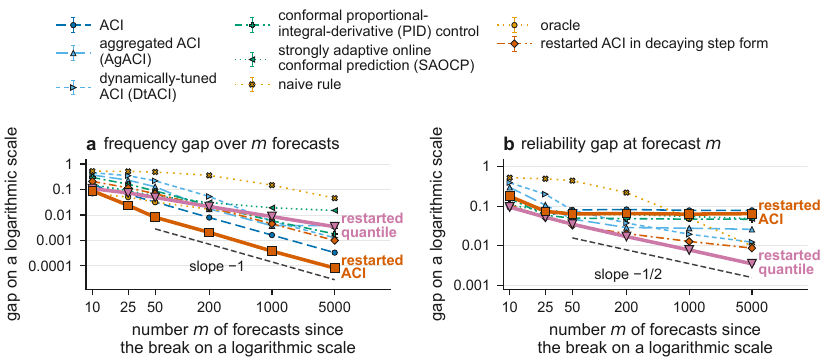}
\caption{The frequency gap of restarted adaptive conformal inference (ACI) falls at slope minus one and the restarted quantile's reliability gap at minus one half, as Theorems~\ref{thm:freq} and~\ref{thm:upper} give. Against the number $m$ of forecasts since a simulated mean shift, panels a and b plot the two gaps as means with one standard error each way. A dashed segment beside each restart's line has its theorem's slope. The naive rule is split conformal prediction on every score since calibration. The oracle uses the true quantile after the break, and its reliability gap of zero lies off panel b.}
\label{fig:rates}
\end{figure}

Restarted ACI has the smallest frequency gap over the first \nCountFifty forecasts after every synthetic break. Each restart's fitted slope in Figure~\ref{fig:rates} matches the rate Section~\ref{sec:theory} proves for it. The experiment tests whether the two rates show in the slopes of the two gaps against the number of forecasts, and whether restarted ACI's gain over ACI holds at every simulated break.

\paragraph{Each restart attains its own rate.} Figure~\ref{fig:rates} plots both gaps against the number of forecasts on the mean shift of \nSynDeltaMeanShift. Over the four periods from the first \nCountFifty to the first \nCountFiveThousand forecasts, the frequency gaps of restarted ACI and of ACI fall at slopes of $\nSlopeFreqAbaciMeanShiftFour$ and $\nSlopeFreqAciMeanShiftFour$, with standard errors of \nSlopeFreqAbaciMeanShiftFourSe and \nSlopeFreqAciMeanShiftFourSe, the rate of Theorem~\ref{thm:freq}, while the restarted quantile's falls at $\nSlopeFreqSegMeanShiftFour$ with a standard error of \nSlopeFreqSegMeanShiftFourSe. For the reliability gap the order reverses. The restarted quantile's falls at $\nSlopePtSegMeanShiftFour$ with a standard error of \nSlopePtSegMeanShiftFourSe, the rate of Theorem~\ref{thm:upper}, while restarted ACI and ACI flatten at \nPtMeanShiftAbaciFiveThousand and \nPtMeanShiftAciFiveThousand by the \nOrdFiveThousand forecast, the plateau their fixed step leaves.

\paragraph{Restarted ACI cuts ACI's frequency gap.} Over the first \nCountFifty forecasts it cuts that gap by \nRedAbaciNonzeroMin to \nRedAbaciNonzeroMax{}\% on every configuration with a break, from \nFreqMeanShiftAciFifty to \nFreqMeanShiftAbaciFifty on the mean shift. Its band lies below zero against every variant on every such configuration, and on all \nFreqFiftyAbaciLeads its gap is the smallest of the \nFreqFiftyMethods methods run there, Table~\ref{tab:pertime}'s seven among them. The restarted quantile, at \nFreqMeanShiftSegFifty on the mean shift, trails it. Appendix~\ref{app:synthetic} separates the jump from the boost and checks Proposition~\ref{prop:practical-corrections} on every run.

\section{When an announced date helps}
\label{sec:dose}

An announced date helps restarted ACI only after a large break, and of \nLibCases real announced breaks \nLibEnteredWord is large enough for its window to detect a gain. The quantile shift of a break is the $(1-\alpha)$-quantile of the scores after it less that of the scores before it, in standard deviations of the scores before it. A window's minimum detectable difference is the smallest coverage difference between two methods that the window detects with a \nFloorLevelPct{}\% test at \nFloorPowerPct{}\% power. The gain of restarted ACI is ACI's frequency gap less its own over the first \nCountFifty forecasts. The experiment sets the gain on a grid of simulated breaks against each real break's minimum detectable difference, which tells whether a real break is large enough for the date to show. On the grid restarted ACI gains nothing up to a quantile shift of \nDoseLastNoGainQ and gains from \nDoseFirstGainQ on.

\begin{figure}[!t]
\centering
\includegraphics{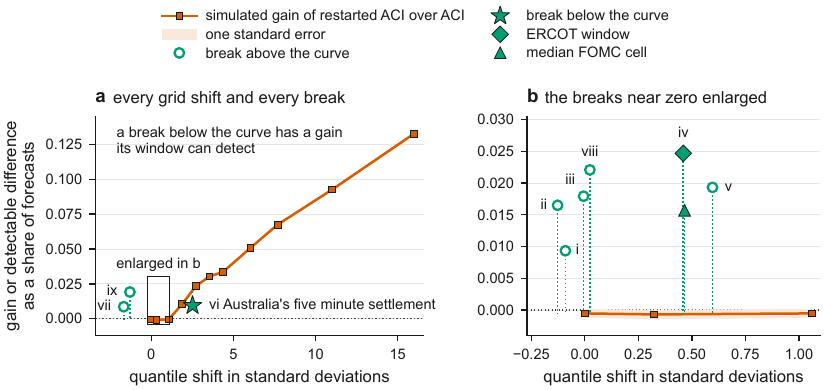}
\caption{Restarted adaptive conformal inference (ACI) gains over ACI only at quantile shifts above \nDoseLastNoGainQ, and \nLibEnteredWord of \nLibCases announced breaks has a gain its window can detect. The curve plots the simulated \emph{gain}, ACI's frequency gap less restarted ACI's, against the quantile shift. Each other marker is a real break at its window's minimum detectable difference, labelled as in Table~\ref{tab:library}, and its dotted segment runs to the curve's gain at its shift. The legend names the window of the Electric Reliability Council of Texas (ERCOT) and the median cell of the Federal Open Market Committee (FOMC).}
\label{fig:dose}
\end{figure}

\paragraph{The gain comes at large shifts.} Figure~\ref{fig:dose} plots this gain against the quantile shift. At a shift of \nDosePeakQ, the mean shift of \nDosePeakDelta, restarted ACI cuts ACI's frequency gap by \nDosePeakRedAbaci{}\%. The restarted quantile's reliability gap at the \nOrdFifty forecast lies below ACI's at every grid shift, by \nDoseRedPtSegTwoHalf to \nDoseRedPtSegOne{}\%.

\paragraph{Both real breaks are too small for a gain.} Each has a quantile shift below \nDoseLastNoGainQ, and each window's minimum detectable difference lies above the curve there. Across the ERCOT reform the standard deviation of the hub price rises from \nErcotHubSdPre to \nErcotHubSdPost USD per MWh, yet the score quantile moves by \nErcotDoseUsd USD per MWh, a quantile shift of \nErcotDose. Restarted ACI covers \nErcotCovAbaci of the days after the reform against ACI's \nErcotCovAci, a difference inside the \nFloorErcotPrimary the window can detect, so the result reflects the window's power and leaves the effect's size open. The median FOMC cell has a quantile shift of \nFomcDose, and there the burn-in form covers \nFomcCovAbaci against ACI's \nFomcCovAci, with a band below zero beside a minimum detectable difference of \nFloorFomcPrimary, so neither stream's band favours restarted ACI. Appendix~\ref{app:ercot} gives every method's coverage and band.

\paragraph{Only \nLibEnteredWord of \nLibCases breaks has a gain its window can detect.} We compare the curve's gain over the first \nCountFifty forecasts with a window's minimum detectable difference over all its days after the break, on one scale. A break has a gain its window can detect when its quantile shift exceeds the shift at which the gain meets the minimum detectable difference, and its window has at least \nLibMinPost days after the break and \nLibMinCalib calibration scores. The one such break is \nLibNemFiveDate, when the National Electricity Market (NEM) of Australia moved to five minute settlement. In its New South Wales region the quantile shift was \nLibNemFiveDose, above the \nLibNemFiveDoseEq at which the curve's gain meets that window's minimum detectable difference of \nLibNemFiveFloor. At the \nOrdFifty forecast, on \nLibNemFiveCfRef unseen scores, every method's reliability gap exceeds the miscoverage level, the smallest at \nNemCfAllBestGap. Appendix~\ref{app:library} tests a rule of thumb for when a restart gains, and Appendices~\ref{app:nobreak} and~\ref{app:sensitivity} show that a misdated or absent break costs little.

\section{Related work}
\label{sec:related}

\paragraph{Online methods that adapt from errors.} ACI guarantees coverage in the long run \citep{gibbs2021aci}, AgACI and DtACI aggregate or tune its step online \citep{zaffran2022agaci,gibbs2024dtaci}, SAOCP bounds regret on every time interval \citep{bhatnagar2023saocp}, and conformal PID keeps that coverage for any bounded score sequence \citep{angelopoulos2023pid}. A decaying step adds convergence under exchangeable scores, whose joint law ignores their order \citep{angelopoulos2024decaying}. \citet{liang2026minimax} show that split conformal prediction with drift detection and restarts attains the minimax training-conditional regret over change points of unknown number and place. \citet{vaze2026simultaneous} views ACI as projected gradient descent on the pinball loss and bounds the coverage violation that does not cancel by the path length of the oracle quantile, assuming smooth drift where we assume a known boundary. \citet{junohn2025retroadj} define the recovery time, the steps coverage takes to return to nominal after a change, and shorten it by retrospective adjustment. None of these guarantees uses a break's date, and we show what the date adds to each criterion.

\paragraph{Methods that use a known partition.} Weighted conformal prediction reweights calibration scores when the test law departs from the calibration law \citep{tibshirani2019weighted,podkopaev2021distribution,barber2023beyondexch,cauchois2023robust}. Mondrian conformal prediction calibrates within a category \citep{vovk2003mondrian,vovk2005book}, and group-weighted conformal prediction covers many groups with small counts \citep{bhattacharyya2024groupweighted}. \citet{bharti2026parameter} give parameter-free online coverage conditional on intersecting groups, a frequency in the long run within each group, which is frequency calibration with the cell as the group. We bound instead the gap of the threshold in force at a known number of scores within a cell.

\paragraph{Methods for time series.} These build intervals from ensembles and sequential residuals, from past errors weighted by learned similarity, by planning over a horizon or by switching score sets on a state variable \citep{xu2021enbpi,xu2023spci,auer2023hopcpt,yang2024bci,sunyu2025cptc}. Split conformal prediction stays near nominal coverage under temporal dependence \citep{barber2026timeseries}, and adaptive conformal inference for causal effects recovers the exchangeable rate under geometric mixing \citep{koukorinis2026doubly}. Conditional guarantees come from reweighting over a function class and from transductive bounds on adaptive scores \citep{gibbs2025conditional,gazin2024dkw}. Other work corrects the bias of spectrally weighted quantiles on a stream, bounds the power of anytime-valid shift detection, sweeps shift sizes under delayed feedback and benchmarks calibration by regime on economic series whose regimes have no announced date \citep{opoku2026drift,kuang2026anytime,halabi2026adaptive,oancea2026dynamic}. Conformal methods also localize a change point offline and adapt a classifier in service to a synthetic shift at a known step \citep{peng2026arc,leong2026online}. None of these gives a rate at a known count after an announced date, as we do.

\section{Limitations and broader impacts}
\label{sec:limitations}

On the real streams no band against ACI favours either restart, and the reliability results rest on finite holdout references. Under polynomial mixing of index $a<1$ we apply to the reliability gap the exponent $a/2$ that \citet[Section~3.6]{halkiewicz2026rolling} gives for the dependence term of a rolling calibration window's risk bound, without proving that it transfers. On a renewal family whose mixing coefficient we did not measure, the fitted slope misses the criterion we fixed before the run. On the one announced break with a gain its window can detect, every method's reliability gap at the \nOrdFifty forecast exceeds \nAlpha.

\paragraph{Broader impacts.}
\label{sec:impacts}
The intervals here serve a bid on the next delivery period and a limit set before the next outcome, and both depend on the reliability gap, as Proposition~\ref{prop:conditional} states. A user who checks a finished period relies on the rate of the frequency gap. A bid or limit set on a threshold that undercovers after a break exposes its user to the outcomes the interval misses, and a finished period on target taken as a promise about the next interval hides that failure. A restart at a date announced in error discards history the threshold needed, a cost Proposition~\ref{prop:safety} bounds. Each guarantee is valid only under its hypotheses, independent scores within a cell for the rate of the reliability gap and bounded scores with a bounded total floor correction for the rate of the frequency gap. The methods forecast no price, so they affect a market only through the intervals users set.

\section{Conclusion}
\label{sec:conclusion}

An announced date gives each criterion its own guarantee. The restarted quantile serves a forecaster who must act before the next outcome, and its reliability gap falls at the minimax rate when the scores within a cell are independent. Restarted ACI serves a user who checks a finished period, and on bounded scores its frequency gap falls as one over the number of forecasts when its total floor correction stays bounded. On the mean shift the fitted slopes match these rates, $\nSlopePtSegMeanShiftFour$ for the restarted quantile's reliability gap and $\nSlopeFreqAbaciMeanShiftFour$ for restarted ACI's frequency gap. Each restart misses the other's rate, with the restarted quantile's frequency gap at fitted slope $\nSlopeFreqSegMeanShiftFour$ and restarted ACI's reliability gap flat at slope $\nSlopePtAbaciMeanShiftFour$.

A report on a recursion's finished period should therefore give both gaps, the frequency gap the period attains and the reliability gap of the threshold in force at its end. The same two restarts apply to the next date an operator announces, such as the California operator's extended day-ahead market. Whether one online method can attain both rates is open.

\begin{ack}
I thank Roger Fouquet, John Hewitt, and Guillaume Roger, whose guidance on academic writing improved how I express ideas. I also thank Ioannis Karatzas, whose extensive expertise in mathematics and statistics formed the basis of my research. I am grateful to Tony Dear, William Hixon and Daniel Steingart for their steady support throughout my studies, and to Mike Lock, Dawsyn Martone, Jay Rubenstein, Ilya Rudkevich and Xander Stover for their time and encouragement. Names within each group are in alphabetical order of surname. This work received no external funding, and I declare no competing interests.
\end{ack}

\bibliographystyle{plainnat}
\bibliography{references_camera_ready}

\clearpage
\appendix
\raggedbottom

\section{Full proofs}
\label{app:proofs}

We prove each result of Section~\ref{sec:theory} here and report a numerical check of each proof. The checks run in double precision, by exact integration where the law allows it and otherwise by simulation at seed \nThmChecksSeed over \nThmSimReps replications, unless a check names its own count. Their code and output are in the archive of Appendix~\ref{app:compute}.

Fix $\alpha\in(0,1)$, write $\theta=1-\alpha$ and $h=\min(\alpha,1-\alpha)$, and let $c=\lceil 1/(1-\alpha)\rceil$ be the cutoff. The proofs index thresholds by completed scores since the restart, so $\hat q_n$ is the threshold the forecast after $n$ scores uses, the $\hat q_t$ of Section~\ref{sec:setup} with $n_t=n$. For $n<c$ it is infinite. For $n\ge c$ it is the linearly interpolated quantile at $\theta$ of those scores, lying the fraction $g=(n-1)\theta-j$ of the way from the $(j+1)$th to the $(j+2)$th order statistic, where $j=\lfloor (n-1)\theta\rfloor$. Let $D_n=\sup_x|\widehat F_n(x)-F_k(x)|$ be the uniform distance between the scores' empirical distribution function and the cell's law.

\subsection{Proof of Theorem~\ref{thm:upper}}
\label{app:proof1}

\paragraph{Inequality~\eqref{eq:sandwich}.}
Let $F_k$ be continuous and the dependence arbitrary. The threshold lies between the two order statistics it joins and $F_k$ is nondecreasing, so $F_k(\hat q_n)$ lies between $F_k(S_{(j+1)})$ and $F_k(S_{(j+2)})$. At most $j+1$ scores lie strictly below $S_{(j+2)}$, so continuity gives $F_k(S_{(j+2)})\le (j+1)/n+D_n$. At least $j+1$ scores lie at or below $S_{(j+1)}$, so $F_k(S_{(j+1)})\ge (j+1)/n-D_n$. Since $(j+1)/n-\theta=(\alpha-g)/n$ with $0\le g<1$, and $|\alpha-g|\le\max(\alpha,1-\alpha)$ for every such $g$, every realization with $n\ge c$ satisfies \eqref{eq:sandwich}. The rank $j+2$ exists wherever $g>0$, because $(n-1)\theta<n-1$ for every $\alpha>0$. Taking expectations gives $\Delta\le\mathbb{E}D_n+\max(\alpha,1-\alpha)/n$.

\paragraph{Independent scores.}
Massart's inequality \citep{massart1990tight} gives $\mathbb{P}(\sqrt n D_n>\lambda)\le 2e^{-2\lambda^2}$ for $\lambda\ge0$. The bound reaches one at $\lambda_0=\sqrt{\log 2/2}$, and integrating the tail beyond it gives $\mathbb{E}\sqrt n D_n\le\lambda_0+\sqrt{\pi/2}\,\mathrm{erfc}(\sqrt2\lambda_0)=\nThmDkwConst$. With \eqref{eq:sandwich} this is the first bound of \eqref{eq:upperbound}, for every continuous $F_k$ and with no condition on the density or the tail.

\paragraph{A sharper bound.}
For independent scores the $U_i=F_k(S_i)$ are independent uniforms, and $F_k$ maps the two order statistics the threshold joins to the uniform order statistics of ranks $j+1$ and $j+2$. The squared distance to $\theta$ is largest at an end of the interval between them, so $\mathbb{E}(F_k(\hat q_n)-\theta)^2$ is at most the sum over the two ranks $i$ of $\mathbb{E}(U_{(i)}-\theta)^2=i(n+1-i)/((n+1)^2(n+2))+(i/(n+1)-\theta)^2$. Each variance term is at most $1/(4(n+2))$ and each bias at most $2\max(\alpha,1-\alpha)/(n+1)$, and Jensen's inequality gives the second bound of \eqref{eq:upperbound}.

\paragraph{Below the cutoff.}
For $n<c$ the threshold is infinite, the interval covers every outcome, and the gap is exactly $1-\theta=\alpha$.

\paragraph{Geometric mixing.}
Inequality~\eqref{eq:sandwich} holds on every realization, so the rate is that of $\mathbb{E}D_n$. Split the $n$ scores into alternating blocks of length $v$, with $\mu$ blocks of each parity. Berbee's coupling lemma \citep{berbee1979coupling} replaces the blocks of one parity by independent blocks with the same laws, which differ from them with probability at most $\mu\beta(v)$. Let $L=\lceil\sqrt\mu\rceil$. For independent blocks, the distance at the points where $F_k$ takes the values $i/L$ is the largest of $2L$ centred signed deviations, each the average of $\mu$ independent centred variables of range one or its negative. Hoeffding's lemma bounds the expected largest deviation by $\sqrt{\log(2L)/(2\mu)}$, and the discretization adds $1/L$. So
\[
\mathbb{E}D_n\le\sqrt{\frac{\log(2\sqrt\mu+2)}{2\mu}}+\frac{1}{\sqrt\mu}+\mu\beta(v)+\frac{2v}{n}.
\]
The coupling term $\mu\beta(v)$ and the edge term $2v/n$ are all the dependence adds. Under $\beta(v)\le C_\beta\rho^v$, a block length of order $\log n/\log(1/\rho)$ keeps both at order $\log n/\sqrt n$ or below and makes $\mathbb{E}D_n$ of order $\log n/\sqrt{n\log(1/\rho)}$, with constants depending only on $C_\beta$ and $\rho$.

\paragraph{Checks.}
Integration over the two order statistics gives the exact gap on independent uniform scores at every count from one to \nThmCountsChecked, and every count lies within both bounds of \eqref{eq:upperbound}. A simulation measures the gap at \nThmLawCounts counts up to \nThmLawCountMax on \nThmLaws laws, the uniform, the exponential, a Pareto law of index one and a half, and the law with survival function $1/\log(e+s)$, and the bound holds at every point. That last law has a bounded density near its quantile while the mean of its quantile estimate is infinite at every count, and the proof needs neither a density nor a moment. Exact computation at \nAddLevels miscoverage levels, \nAddLevelTenth, \nAddLevelHalf, \nAddLevelSixTenths, \nAddLevelThreeQuarters, \nAddLevelNineTenths and \nAddLevelNineteenTwentieths, with cutoffs \nAddCutoffTenth, \nAddCutoffHalf, \nAddCutoffSixTenths, \nAddCutoffThreeQuarters, \nAddCutoffNineTenths and \nAddCutoffNineteenTwentieths, covers every count from one to \nAddCountMax. Below the cutoff the gap equals the miscoverage level. From the cutoff on it lies within both bounds of \eqref{eq:upperbound}, and in \nAddRealizations realizations at each count, \eqref{eq:sandwich} never fails.

\subsection{Proof of Theorem~\ref{thm:freq}}
\label{app:proof2}

\paragraph{The bounded threshold.}
Let the scores lie in $[0,B]$ almost surely and the recursion start in $[0,B]$, as a quantile of such scores does. A threshold above $B$ covers every score and falls by $\alpha\gamma$, a threshold below zero misses every score and rises by $(1-\alpha)\gamma$, and a threshold in $[0,B]$ moves by at most one of those steps. By induction the threshold stays in $[-\alpha\gamma,\,B+(1-\alpha)\gamma]$. A simulation on scores bounded by one, at steps \nThmStateStepOne, \nThmStateStepTwo and \nThmStateStepThree for \nThmStateUpdates updates each, under uniform scores and under two sequences that push the threshold up and down, keeps the threshold inside the bound in all \nThmStateHolds runs.

\paragraph{The rate.}
Summing the recursion over the period telescopes, $\gamma\sum_{t=\tau+1}^{\tau+M}(\mathrm{err}_t-\alpha)=\hat q_{\tau+M+1}-\hat q_{\tau+1}$. Dividing by $\gamma M$ and applying the bounded threshold gives, on every realization,
\[
G^{\mathrm{freq}}_M=\frac{|\hat q_{\tau+M+1}-\hat q_{\tau+1}|}{\gamma M}\le\frac{B+\gamma}{\gamma M}.
\]
This is the telescoping identity of \citet[Proposition~4.1]{gibbs2021aci} after a restart.

\subsection{Proof of Proposition~\ref{prop:practical-corrections}}
\label{app:practical-corrections}

Algorithm~\ref{alg:abaci} adds to the pure recursion a jump at the break and, afterwards, a floor and a boosted step, and each enters the telescoping sum as its own term. Let $E_t=\mathrm{err}_t-\alpha$. The unconstrained step is $\tilde q_{t+1}=\hat q_t+\gamma_t E_t$ and the projected step $\hat q_{t+1}=\max(\tilde q_{t+1},\hat q_{\min})$, so
\[
\Pi_t:=\hat q_{t+1}-\tilde q_{t+1}=\bigl[\hat q_{\min}-(\hat q_t+\gamma_t E_t)\bigr]_+ .
\]
Let $J$ be the jump inside the period, equal to $J_b$ when the restart falls in $(\tau,\tau+M]$ and zero otherwise, so $|J|\le|J_b|$. Summing over $t=\tau+1,\ldots,\tau+M$ gives
\[
\sum_{t=\tau+1}^{\tau+M}\gamma_t E_t=\hat q_{\tau+M+1}-\hat q_{\tau+1}-J-\sum_{t=\tau+1}^{\tau+M}\Pi_t .
\]
Writing $\gamma\sum_t E_t=\sum_t\gamma_t E_t-\sum_t(\gamma_t-\gamma)E_t$ and taking absolute values gives \eqref{eq:practical-correction-main}.

\paragraph{The bounded range.}
The initial threshold lies in $[0,B]$ by assumption, and so does the threshold after the restart, a quantile of scores in $[0,B]$. Take a threshold $q$ in $[0,\,B+(1-\alpha)\kappa\gamma]$ and a step of at most $\kappa\gamma$. A miss means a score above $q$, so $q<B$ and the unprojected step stays below $B+(1-\alpha)\kappa\gamma$, and the projection returns the larger of that value and a floor of at most $B$. A cover lowers the unprojected threshold, and the projection returns at most the larger of $q$ and $B$. Every projected threshold is at least the positive floor. The range $R_M$ is therefore at most $B+(1-\alpha)\kappa\gamma\le B+\kappa\gamma$ uniformly in $M$.

The floor must lie inside the bound, because a floor above $B$ lifts every projected threshold while the threshold after the restart stays at or below $B$. At a floor of \nAddObjElevenFloor on scores bounded by \nAddObjElevenB, at unit step and boost, the threshold starts at \nAddObjElevenInitial, jumps to \nAddObjElevenReset and climbs to \nAddObjElevenStateMax, a range of \nAddObjElevenRange against the bound of \nAddObjElevenBound. A simulation of \nAddStateRuns runs, each a grid point over the score bound, the floor in $(0,B]$, the step and the boost under one of two score sequences, keeps every threshold inside the interval and every period's range within its bound in all \nAddStateInside runs. One update from every point of a lattice of thresholds in the interval, under every point of a lattice of scores in $[0,B]$ and every step the recursion takes, never leaves the interval at any of \nAddInductionPoints grid points.

\paragraph{The boost term.}
The excess $\gamma_t-\gamma=\gamma(\max\{1,\kappa/\sqrt{m_t}\}-1)$ is $(\kappa-1)\gamma$ on the $b$ updates with $m_t=1$ and positive on fewer than $\kappa^2$ later updates, each below $(\kappa-1)\gamma$. With $|E_t|\le\max\{\alpha,1-\alpha\}$ this gives $B_M\le\max\{\alpha,1-\alpha\}(\kappa-1)(b+\kappa^2)\gamma$ for every $M$. So on scores bounded almost surely by $B$, with the floor in $(0,B]$ and the entering threshold and the threshold after the restart in $[0,B]$, Algorithm~\ref{alg:abaci} has the $O(1/M)$ rate whenever $|J_b|+\sum_t\Pi_t=O(1)$, and a total floor correction growing in proportion to $M$ keeps the bound away from zero.
\hfill$\square$

\subsection{Proof of Proposition~\ref{thm:detection}}
\label{app:proof3}

Let $q^*_{\mathrm{pre}}=F_{\mathrm{pre}}^{-1}(1-\alpha)$ and $q^*_{\mathrm{post}}=F_{\mathrm{post}}^{-1}(1-\alpha)$. At the first prediction after the break,
\[
|\hat q_{\tau+1}-q^*_{\mathrm{post}}|\ge|q^*_{\mathrm{post}}-q^*_{\mathrm{pre}}|-|\hat q_{\tau+1}-q^*_{\mathrm{pre}}|\ge\frac{3\xi}{4}.
\]
ACI moves the threshold by at most $\gamma$ per step, since $\mathrm{err}_t-\alpha\in[-\alpha,1-\alpha]$, so after $M$ updates since the break $|\hat q_{\tau+M}-q^*_{\mathrm{post}}|\ge 3\xi/4-M\gamma$. If $M<\xi/(4\gamma)$, this exceeds $\xi/2$, and the density bound between the two quantiles gives
\[
\Delta_{\tau+M}=\mathbb{E}\bigl|F_{\mathrm{post}}(\hat q_{\tau+M})-(1-\alpha)\bigr|\ge\frac{\xi f_{\min}}{2}.
\]
So $\Delta_{\tau+M}\le\epsilon<\xi f_{\min}/2$ requires $M\ge\xi/(4\gamma)=(\xi/(4c_\gamma))\sqrt T$. Algorithm~\ref{alg:segsplit} instead restarts at the boundary, and \eqref{eq:upperbound} gives $\Delta_t\le\nThmDkwConst/\sqrt{n_t}+\max(\alpha,1-\alpha)/n_t$ on independent scores after the break, so $n_t$ of order $1/\epsilon^2$ gives $\Delta_t\le\epsilon$ whatever $T$ and $\gamma$.
\hfill$\square$

\subsection{Proof of Theorem~\ref{thm:lower}}
\label{app:proof4}

\paragraph{The construction.}
The supremum over the class is at least the supremum over cells of independent scores drawn from one of two members, so the proof restricts to such cells and keeps every cell other than $k$ at a fixed law. Rescale the scores to $[0,1]$ and let $F_0$ be uniform, with quantile $\theta$ at $\theta$. For $\varepsilon\le h/2$ let $F_1$ have density $1+\varepsilon\phi$, with $\phi=1/h$ on $[\theta-h,\theta]$, $\phi=-1/h$ on $(\theta,\theta+h]$ and $\phi=0$ elsewhere. The interval lies in $[0,1]$ because $h\le\min(\theta,1-\theta)$, and the density lies in $[1-\varepsilon/h,\,1+\varepsilon/h]\subseteq[1/2,3/2]$, so both laws belong to the class Theorem~\ref{thm:lower} assumes.

\paragraph{The divergence.}
One draw has Kullback-Leibler (KL) divergence $\mathrm{KL}(F_0\|F_1)=-\int\log(1+\varepsilon\phi)=-h\log(1-(\varepsilon/h)^2)$, at most $(4/3)\varepsilon^2/h$ for $\varepsilon/h\le 1/2$, and $M$ independent draws have $M$ times that.

\paragraph{The separation.}
Here $F_1(\theta)=\theta+\varepsilon$, and the quantile of $F_1$ at $\theta$ lies below $\theta$ by $\Delta_\varepsilon=\varepsilon h/(h+\varepsilon)\ge\varepsilon/2$. Since $F_1-F_0$ is $\varepsilon$ times the tent that is one at $\theta$ and zero outside $[\theta-h,\theta+h]$, every threshold $x$ satisfies $|F_0(x)-\theta|+|F_1(x)-\theta|\ge\Delta_\varepsilon$.

\paragraph{The two-point bound.}
For any method, the larger of the two expected gaps after $M$ scores is at least half their sum. That sum is at least $\Delta_\varepsilon$ times the overlap $\int\min(dF_0^{\otimes M},dF_1^{\otimes M})$, which the Bretagnolle and Huber inequality bounds below by $\tfrac12\exp(-M\,\mathrm{KL}(F_0\|F_1))$ \citep{tsybakov2009nonparametric}. The larger gap is therefore at least $(\Delta_\varepsilon/4)\exp(-M\,\mathrm{KL}(F_0\|F_1))$.

\paragraph{The two ranges.}
For $M\ge 3/(2h)$, the choice $\varepsilon=\sqrt{3/8}\sqrt{h/M}$ keeps $\varepsilon/h\le1/2$ and the divergence at most one half, so the bound is at least $c_A\sqrt{h/M}$ with $c_A=\sqrt{3/8}\,e^{-1/2}/8=\nThmLowerConst$. For every $M\ge1$, the choice $\varepsilon=h/2$ gives $(h/12)(3/4)^{Mh}$, at least $\nThmSmallRangeConst\,h$ for $M<3/(2h)$. Since $c_A<\nThmSmallRangeConst$ and $h\le\sqrt h$, the bound $c_Ah/\sqrt M$ holds at every $M\ge1$. At \nThmGridPoints pairs of miscoverage level and count, a check in double precision confirms that the density of $F_1$ lies in $[1/2,3/2]$, that the divergence equals its exact form, that the separation holds and that the bound lies below $\alpha$.
\hfill$\square$

\subsection{Corollary~\ref{cor:rate} and the class's parameters}
\label{app:corollary}

For independent scores, \eqref{eq:upperbound} bounds the restarted quantile's gap after $M$ scores by a constant over $\sqrt M$ for every continuous law, and Theorem~\ref{thm:lower} bounds the minimax gap below by $c_Ah/\sqrt M$ over independent cells whose density bounds admit $[1/2,3/2]$. Together they give $\Theta(M^{-1/2})$ at fixed $\alpha$. Under geometric mixing at fixed $C_\beta$ and $\rho$, Theorem~\ref{thm:upper} gives order $\log M/\sqrt M$.

The fixed parameters matter. The retain or refresh process keeps its last score with probability $1-r$ and otherwise draws a fresh uniform score, a Markov chain whose $\beta$ coefficient at lag $k$ is $(1-r)^k$. With refresh probability one over the square of the count $n$, no refresh follows the first score with probability $(1-1/n^2)^{n-1}$, which is \nRefreshTen at \nThmRefreshCountTen scores and \nRefreshThousand at \nThmRefreshCountThousand, so the scores are most likely one repeated draw. The restarted quantile's simulated gap is then \nThmRefreshFastTen after \nThmRefreshCountTen scores and \nThmRefreshFastThousand after \nThmRefreshCountThousand. No rate holds over a class whose $\rho$ may approach one with the count, which is why Definition~\ref{def:ssi} fixes $\rho$.

\subsection{Proof of Proposition~\ref{prop:safety}}
\label{app:proof_safety}

Suppose the scores on both sides of the boundary $\tau$ share the distribution function $F$, so both quantiles equal $q^\star$ and $\xi=0$. The hypothesis makes the stream one $\beta$-mixing process of the class. The stretch after the restart is a segment of it, with the same distribution function and density bounds near $q^\star$, and a restriction of a $\beta$-mixing sequence keeps the coefficient bound, so Definition~\ref{def:ssi} holds on the cell after the restart. Algorithm~\ref{alg:segsplit} takes the empirical quantile of the scores since the restart, so its reliability gap satisfies Theorem~\ref{thm:upper} there.

For the widths, let $t_M=\tau+M+1$, so $n_{t_M}=M$, and let $T_M=\tau+M$. Under the stated moment conditions $\mathbb{E}|\hat q_{t_M}^{\mathrm{seg}}-q^\star|\le C_q/\sqrt M$ and $\mathbb{E}|\hat q_{t_M}^{\mathrm{NR}}-q^\star|\ge c_q/\sqrt{T_M}$. With $W_t=2\hat q_t$,
\[
\frac{\mathbb{E}|W_{t_M}^{\mathrm{seg}}-2q^\star|}{\mathbb{E}|W_{t_M}^{\mathrm{NR}}-2q^\star|}\le\frac{C_q}{c_q}\sqrt{\frac{T_M}{M}},
\]
and after $n$ completed scores since the restart the factor is $(C_q/c_q)\sqrt{T_n/n}$ with $T_n=\tau+n$.
\hfill$\square$

\subsection{Proof of Proposition~\ref{prop:conditional}}
\label{app:conditional}

The threshold $\hat q$ is a function of the cell's first $n$ scores, so the next interval's conditional coverage is $G_n(\hat q)$. The difference $|G_n(\hat q)-F_k(\hat q)|$ is at most the largest gap, over sets $C$, between the conditional and marginal probabilities that the next score lies in $C$. That largest gap has expectation equal to the $\beta$ coefficient between the cell's past and its next score, at most $\beta_k(1)$, and the triangle inequality gives Proposition~\ref{prop:conditional}. Under independence $G_n=F_k$ and the two gaps are equal. Simulation confirms the equality on every replication of the retain or refresh process at refresh probability one, where $\beta_k(1)$ is zero, and the bound at refresh probabilities one half and one tenth. At one half the conditional error is \nThmCondHalfThousand after \nThmCondCount scores, against a reliability gap of \nThmPerTimeHalfThousand.
\hfill$\square$

\subsection{The plug-in quantile and the rank corrected Mondrian set}
\label{app:mondrian}

Algorithm~\ref{alg:segsplit} differs from a Mondrian predictor in two ways. At every miscoverage level below one half it takes the plug-in quantile of its cell, where the rank corrected Mondrian set of \citet{vovk2003mondrian} takes a larger rank, as this appendix proves. And its guarantee bounds a two-sided gap at a known number of scores within the cell, where Mondrian and group-weighted guarantees \citep{bhattacharyya2024groupweighted} bound coverage from one side over pooled calibration draws.

The rank corrected Mondrian set of \citet{vovk2003mondrian} takes the score of rank $k_n=\lceil(n+1)\theta\rceil$ and is infinite where $k_n>n$. For exchangeable scores without ties, the next score falls at or below it with probability exactly $k_n/(n+1)\ge\theta$, and with ties with probability at least $k_n/(n+1)$. At $\alpha=\nAlpha$ the rank exceeds the count below \nOrderFirstFinite scores, so the set is infinite there while the plug-in threshold is finite from two scores on. At two scores the set covers with probability one, and the plug-in threshold covers with expected probability \nThmTwoScoreCov under the uniform law, by integration, and \nTwoScoreExp under the unit exponential. From \nOrderFirstFinite scores on, $k_n\ge(n+1)\theta=(n-1)\theta+2\theta$. Below $\alpha=1/2$, $2\theta>1$, so $k_n\ge j+2$, and the set's order statistic lies at or above the upper one the plug-in joins, which makes the set the wider of the two on scores without ties. A check at every count from \nOrderSmallest to \nOrderLargest compares the ranks and, from \nOrderFirstFinite scores on, the two thresholds on one draw of uniform scores, without a failure, and exact arithmetic at \nOrdLevelsLow levels below one half and every count up to \nOrderLargest finds no failure of the ordering. The plug-in Mondrian predictor in the experiments uses the plug-in quantile, and on all \nPSevenStreams streams, the five synthetic configurations, the two real streams and the \nLibCases announced breaks, its threshold agrees with the restarted quantile's to four decimals at every step.

\subsection{Scope of the propositions and the last forecast of each cell}
\label{app:prop-scope}

Proposition~\ref{thm:detection} covers ACI at the fixed step $c_\gamma/\sqrt{T}$. The four variants adapt their steps or aggregate experts online, so they fall outside it, and Section~\ref{sec:pertime} measures them directly.

At small counts the gap exceeds the miscoverage level. On independent uniform scores the exact gap exceeds $\alpha$ at \nThmCountsAboveAlpha counts from two scores to \nThmCountsAboveAlphaLast, the most, \nThmLargestGap, at two scores. At every count Theorem~\ref{thm:upper} gives the bound that holds, $\alpha$ below the cutoff and \eqref{eq:upperbound} from it on, and the integration check of Appendix~\ref{app:proof1} confirms it up to \nThmCountsChecked.

On every realization the coverage $F_k(\hat q_n)$ of the interpolated threshold lies between the coverages of the two order statistics it joins, which \eqref{eq:sandwich} uses in place of an exact law. A simulation confirms this on every replication of exponential scores, with $n$ times the coverage variance between \nThmVarLast and \nThmVarFirst over counts from \nThmVarFirstCount to \nThmVarLastCount, so the variance is of order $1/n$.

Let the boundaries $\tau_1<\cdots<\tau_{K-1}$ of the announced breaks cut a stream starting at time one into the $K$ cells of Definition~\ref{def:ssi}, and let cells $2$ to $K$ have lengths $M_2,\ldots,M_K$. The last forecast in cell $j$ is its $M_j$th, at $\tau_{j-1}+M_j$, on $M_j-1$ scores. Write $M_{\min}=\min_jM_j$ over those cells. With independence within cells and $M_{\min}-1\ge c$, so that every such forecast has at least the cutoff's count of scores, Theorems~\ref{thm:upper} and~\ref{thm:lower} applied cell by cell give
\begin{align*}
\frac{c_Ah}{\sqrt{M_{\min}-1}}
&\le\inf_P\sup_{\mathcal{H}}\max_{2\le j\le K}\Delta_{\tau_{j-1}+M_j}
\le\sup_{\mathcal{H}}\max_{2\le j\le K}\Delta^{\mathrm{reset}}_{\tau_{j-1}+M_j}\\
&\le\Bigl(\frac{1}{2(M_{\min}+1)}+\frac{8\max(\alpha,1-\alpha)^2}{M_{\min}^2}\Bigr)^{1/2},
\end{align*}
with the lower bound taken on the shortest cell and every other cell at a reference law.

The condition on the minimum count is needed. Below the cutoff a cell's last threshold is infinite and its gap equals the miscoverage level. At a miscoverage level of \nMbLevel the cutoff is \nMbCutoff scores, and a shortest cell of \nMbMmin forecasts ends on \nMbScores scores with a gap of \nMbLevel, against the display's bound of \nMbBound. Under the condition, integration gives the exact gap at that level at every count from \nMbFirst to \nMbLast scores, and it never exceeds the bound. Integration also gives the restarted quantile's gap at the last forecast of cells of \nThmEndpointCellOne, \nThmEndpointCellTwo and \nThmEndpointCellThree uniform scores. The largest is \nThmEndpointMax, in the cell of \nThmEndpointCellOne, below the exact second moment bound of \nThmEndpointUpper there, while the minimax lower bound on that cell, a bound on the class's worst law, is \nThmEndpointLower.

\section{Safety when no break comes}
\label{app:nobreak}

\begin{figure}[H]
\centering
\includegraphics{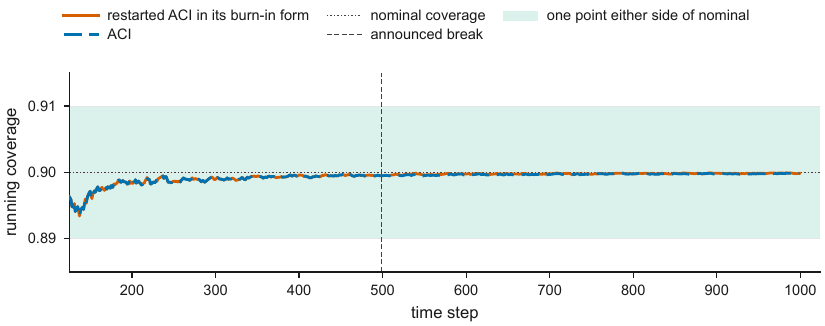}
\caption{Restarted adaptive conformal inference (ACI) and ACI both stay within one percentage point of nominal coverage when the announced break does not come. Lines plot the running coverage of restarted ACI's burn-in form and of ACI against the time step, as means over replications, and the dashed line marks the announced date.}
\label{fig:nobreak}
\end{figure}

When the announced break does not come, restarted ACI matches ACI. Over \nNobreakReps replications of \nNobreakT steps with the announced date at step \nNobreakTau, so the boundary at \nNobreakLastPre, both have mean coverage \nNobreakCovAbaci after the restart. Proposition~\ref{prop:safety} covers the restarted quantile, and this simulation covers restarted ACI. Each replication draws standard normal outcomes around a zero forecast from a generator seeded at \nNobreakSeedBase plus \nNobreakSeedStep times its index, and it calibrates on its first \nSynCalib scores. ACI steps at \nNobreakGamma. Restarted ACI empties its score set when told the boundary and steps at \nNobreakGamma with no boost. Once the score set has \nNobreakBurnin scores from after the boundary it moves to their quantile, which is the burn-in form's schedule at one boundary. Both methods take numpy's higher empirical quantile and keep the threshold at or above \nNobreakFloor. The burn-in form's mean coverage less ACI's is \nNobreakCLessAciDiff, with a paired band of $[\nNobreakCLessAciLo, \nNobreakCLessAciHi]$ over \nNobreakBandDraws resamples of the replications at seed \nNobreakBandSeed. Both running coverages are drawn in Figure~\ref{fig:nobreak} from the \nFigSkipOrd step after calibration.

\section{The synthetic family and the baselines}
\label{app:synthetic}

We give every setting of the synthetic family and of the baselines here and report every configuration. The baselines are ACI and the four variants of Section~\ref{sec:pertime}, DtACI, SAOCP, conformal PID and AgACI. Configurations are named by their mean shift and their noise scale after the break, $(\delta,\sigma_2)$, in noise standard deviations before the break. Unless a caption says otherwise, every synthetic band runs from the \nReadBandLoPct to the \nReadBandHiPct percentile of a mean difference over \nSrcSynDraws resamples of the \nSynReps shared replications.

\paragraph{The family.}
One generator per replication draws a standard normal covariate $x_t$ and standard normal noise $\eta_t$. The outcome is $\sin(2\pi t/\nGenPeriod)+\nGenCovCoef\,x_t+\eta_t$ before the boundary and $\sin(2\pi t/\nGenPeriod)+\nGenCovCoef\,x_t+\delta+\sigma_2\eta_t$ from it on. The forecast is $\sin(2\pi t/\nGenPeriod)+\nGenCovCoef\,x_t$ throughout, so the score is $|\eta_t|$ before the boundary and $|\delta+\sigma_2\eta_t|$ after it. The five configurations are $(\nSynDeltaNoBreak,\nSynSigmaNoBreak)$ without a break, $(\nSynDeltaMeanShift,\nSynSigmaMeanShift)$ for the mean shift, $(\nSynDeltaVarShift,\nSynSigmaVarShift)$ for the variance shift, $(\nSynDeltaBothShift,\nSynSigmaBothShift)$ for both shifts and $(\nSynDeltaLargeMean,\nSynSigmaLargeMean)$ for the large mean shift. Each draws \nSynTtotal steps with the break taking effect at index \nSynTpre, so the boundary is at \nMapSynLastPre, and calibrates every method on its first \nSynCalib scores. A replication's seed is its index times \nSeedMult plus the configuration's seed offset, from \nSynSaltNoBreak without a break to \nSynSaltLargeMean for the large mean shift in the order above, with the interpreter's hash seed at \nPythonHashSeed. Every method runs on the stream that seed draws.

\paragraph{Forms and reference rules.}
Restarted ACI has four forms. The boosted form, Algorithm~\ref{alg:abaci} with boost length $b$, is our default. The decaying step form jumps as the boosted form does at the boundary but takes the step $\nDecConstant/k^{\nDecExponent}$ at its $k$th later update in place of the fixed step and the boost. The two burn-in forms take the fixed step through a burn-in of $b$ scores after the boundary and then jump to their quantile. The boosted burn-in form then takes the boosted step, while the burn-in form does this at every boundary and keeps the fixed step with no boost. With $\kappa=1$ and no jump, and with its floor never reached, Algorithm~\ref{alg:abaci} is the pure recursion of Theorem~\ref{thm:freq}.

\begin{remark}[Warm start]
\label{rem:warm}
The restarted quantile's threshold is infinite until the cell has $\lceil 1/(1-\alpha)\rceil$ scores. We call this count the cutoff, and it is two at $\alpha=\nAlpha$. A warm start keeps the quantile in force at the boundary until then, and that quantile is finite on every stream. From the cutoff on it returns what Algorithm~\ref{alg:segsplit} returns, so its reliability gap equals the restarted quantile's. Below the cutoff its finite threshold has a gap of its own, where the restarted quantile's gap is $\alpha$. It thus replaces each infinite width with a finite one, as this appendix and Appendix~\ref{app:ercot} measure.
\end{remark}

The boosted form runs at step \nSynGamma with boost \nBoost over a boost length of \nBurnin. Its jump goes to the linearly interpolated quantile of every score evaluated through the boundary's outcome, without a floor, and later updates are floored at \nFloorFormsAB. Over all \nResetEvents restarts in \nResetGroups stream groups, the five configurations, the cells of the two real streams and the \nLibCases announced breaks, the threshold after the restart was above the floor. The two burn-in forms, with burn-in $b$, step $\gamma$ and floor $\hat q_{\min}$, start each score set with the first score after the boundary. While the score set has fewer than $b$ scores, each update sets the threshold to $\max\{\hat q+\gamma(\mathrm{err}_t-\alpha),\,\hat q_{\min}\}$ and appends the new score. The update that brings the score set to $b$ scores takes that step and then moves the threshold to the score set's linearly interpolated quantile at $1-\alpha$. The boosted burn-in form then takes the boosted form's boosted step, counted from that update, while the burn-in form keeps the fixed step. The burn-in form repeats this at every boundary with a floor of \nFloorFormC, and the boosted burn-in form's floor is the boosted form's. The frozen quantile is the quantile of the calibration scores and the scores before the boundary, kept unchanged from the boundary on. The plug-in Mondrian predictor files each score under its cell and uses the plug-in quantile of the next outcome's cell, and the oracle uses the true quantile after the break.

\paragraph{Variants.}
The four variants run from their official implementations, DtACI \citep{gibbs2024dtaci}, SAOCP \citep{bhatnagar2023saocp}, conformal PID \citep{angelopoulos2023pid} and AgACI \citep{zaffran2022agaci}. SAOCP's repository states the Apache licence, and those of conformal PID and AgACI the MIT licence. DtACI's states none, so the released code omits it and names the repository from which a reader can clone it. AgACI aggregates its experts with the R package opera, released under the GNU Lesser General Public License.

\paragraph{Constants.}
One rule sets every constant a variant leaves open. A default in the implementation stays. A constant the implementation leaves open takes the value the variant's paper uses in its own experiments. A constant neither fixes for a stream of this kind comes from the stream's calibration fold, from its length or its largest score, before any evaluated outcome is read. Under the second clause, DtACI takes its paper's \nSrcGridDtaciCount candidate steps, \nSrcGridDtaciList. Conformal PID takes a step of \nSrcPidLr times the range of the scores over its trailing window and its paper's saturation constant, two over $\pi$ times the ceiling of $\alpha\log T$ less one over $\log T$, with $T$ the sequence length. That constant is \nPidSynCsat on the synthetic family's \nPidSynT steps, \nPidErcotCsat on the ERCOT window's \nPidErcotT days and \nPidFomcCsat on the FOMC stream's \nPidFomcT days. AgACI's experts take the \nSrcGridAgaciCount steps its code lists, \nSrcGridAgaciList. SAOCP sets each expert's scale to the square root of three times the calibration fold's largest score, so each expert's base step is that score, the rule its paper states for time series. Under the third clause, conformal PID's burn-in, which is also the window its step uses, is the calibration fold's length, \nPidSynWindow steps on the synthetic family and \nPidRealWindow days on each real stream. Its integral constant is the fold's largest score, which runs from \nPidSynKiMin to \nPidSynKiMax over the mean shift's replications, where SAOCP's scale runs from \nSaocpSynScaleMin to \nSaocpSynScaleMax, and is \nPidErcotKi USD per MWh on the ERCOT window and \nPidFomcKi percentage points on the FOMC stream. AgACI's experts use a window as long as the calibration fold, and an infinite expert bound becomes the forecast minus or plus \nSrcAgaciMultiple times the fold's largest score.

\paragraph{Inputs.}
No baseline sees a boundary. Each takes the calibration fold as its history before the stream and issues its interval before every evaluated outcome. DtACI's interval is the quantile of every earlier score at the weighted mean of its experts' miscoverage levels, the version its paper runs as DtACI. AgACI's experts follow the original algorithm step for step with one change. The original refits a random forest at every step on a sliding window, trained on the window's earlier rows, and takes the scores of its later rows as calibration scores. Here the experts use the stream's frozen forecast and its most recent scores, as many as the calibration fold contains, so every method shares one forecaster. AgACI gives its interval as offsets $a$ below and $b$ above the forecast, and such an interval covers the reference days whose signed residual, outcome less forecast, lies in $[-a,b]$. Both gaps are measured with AgACI's two offsets against signed residuals, and code written apart from the analyses recomputes every reported AgACI value on a real stream from the runs' rows, agreeing at all \nAgaciSignedAgree comparisons.

\paragraph{Threshold forms.}
ACI moves the score threshold by its step times the miss indicator less $\alpha$ \citep{bhatnagar2023saocp,angelopoulos2023pid}, starting at the calibration quantile with a floor of \nAciFloor. It steps at \nSynGamma on the synthetic family, at \nLibGammaLevel on the ERCOT window and the price series of the announced breaks and at \nLibGammaFomc on the FOMC stream. We also wrote threshold forms of the four variants, called threshold DtACI, threshold SAOCP, threshold conformal PID and threshold AgACI. Threshold AgACI weighs ACI experts at steps \nGridAgaciList, or \nGridAgaciFomcList on the FOMC stream, by exponential weights at rate \nAgaciEta on each expert's miss plus \nAgaciWidthWeight times its width. Threshold DtACI weighs ACI experts at steps \nGridDtaciList, from uniform weights, by exponential weights at rate \nDtaciEta on the distance outside each interval, mixed with a uniform share of one over \nDtaciSigmaDen. Threshold SAOCP starts, at every update, one expert at each of its fixed steps, the multiples \nGridSaocpList of the calibration quantile, for every power of two that divides the update up to $2^{\nSaocpLifetimeLog}$ on the synthetic family, $2^{\nSaocpLifetimeLogErcot}$ on the ERCOT window, $2^{\nSaocpLifetimeLogFomc}$ on the FOMC stream and $2^{\nSaocpLifetimeLogCase}$ on each announced break. It weighs live experts by exponential weights at rate \nSaocpEta on each one's mean loss over its life, where a step's loss is $\alpha$ times the outcome's distance below the interval plus $1-\alpha$ times its distance above, and it floors each expert at \nSaocpFloor. Threshold conformal PID adds to its threshold the proportional gain \nPidKp times the miss indicator less $\alpha$ and the integral gain \nPidKi times the running sum of those errors, with no score forecast and a floor of \nPidFloor. We fixed these settings, ACI's step, the boosted form's boost and boost length and every other method's setting before reading any evaluated outcome.

\paragraph{How each threshold form departs from the original.}
Threshold DtACI's experts move the score threshold over a grid of steps on the score scale, where the original moves the miscoverage level over a grid of steps on that level \citep{gibbs2024dtaci}. Its weights charge misses alone in place of the pinball loss, at a fixed share and rate in place of values set from an interval, with the uniform share mixed in before renormalizing. Threshold SAOCP's experts are trackers at a fixed step in place of scale-free gradient learners, each starting at the calibration quantile with the largest weight, where the original starts it at the current aggregate with weight zero \citep{bhatnagar2023saocp}. Its weights are exponential in place of coin betting under a prior, its lifetimes have no multiplier, its loss charges misses alone in place of the quantile loss, and each expert is floored. Threshold conformal PID adds a scaled running sum of errors to the threshold's increment, a second integration, where the original adds a saturating function of that sum to the threshold \citep{angelopoulos2023pid}. Its step is fixed in score units, where the official implementation uses a fraction of the recent scores' range, and its threshold is floored. Threshold AgACI aggregates one threshold for both ends of the interval in place of two separate aggregations \citep{zaffran2022agaci}. Its exponential weights use the miss plus a width term in place of each end's pinball loss under Bernstein online aggregation, over a grid of steps on the score scale in place of the original's larger grid of steps on the miscoverage level. Appendix~\ref{app:corrections} compares each official implementation with its threshold form.

\paragraph{Further results.}
Across the five configurations, restarted ACI's frequency slope runs from $\nSlopeFreqAbaciMin$ to $\nSlopeFreqAbaciMax$ and ACI's from $\nSlopeFreqAciMin$ to $\nSlopeFreqAciMax$, while the naive rule's runs from $\nSlopeFreqNaiveMin$ to $\nSlopeFreqNaiveMax$. Without a break, restarted ACI's frequency gap over the first \nCountFifty forecasts is \nRedAbaciNoBreak{}\% below ACI's as a point estimate, a difference of $\nArmNoBreakAbaciFreqFiftyDiff$ whose band of $[\nArmNoBreakAbaciFreqFiftyLo, \nArmNoBreakAbaciFreqFiftyHi]$ includes zero. With a break the cut runs from \nRedAbaciVarShift{}\% on the variance shift to \nRedAbaciMeanShift{}\% on the mean shift, with \nRedAbaciBothShift{}\% on both shifts and \nRedAbaciLargeMean{}\% on the large mean shift. The boosted burn-in form changes both the scores its restart uses and the time it comes, and over the same period on the large mean shift its frequency gap is \nFreqLargeMeanAbaciTextFifty against ACI's \nFreqLargeMeanAciFifty. The decaying step form, at constant \nDecConstant and exponent \nDecExponent, fits a frequency slope of $\nSlopeFreqAbaciDecMeanShiftFour$ with standard error \nSlopeFreqAbaciDecMeanShiftFourSe and a reliability slope of $\nSlopePtAbaciDecMeanShiftFour$ with standard error \nSlopePtAbaciDecMeanShiftFourSe on the mean shift, against the $-1$ of Theorem~\ref{thm:freq} and the $-1/2$ of Theorem~\ref{thm:upper}. On the mean shift the naive rule's frequency gap falls at a slope of $\nSlopeFreqNaiveMeanShiftFour$, with a standard error of \nSlopeFreqNaiveMeanShiftFourSe, over the four periods of Section~\ref{sec:frequency}.

Over the first \nCountFifty forecasts on the mean shift, the frequency gaps of the four variants range from \nFreqMeanShiftSaocpSrcFifty for SAOCP to \nFreqMeanShiftDtaciSrcFifty for DtACI. Restarted ACI's band lies below zero against every variant on every configuration with a break, in all \nSrcFreqBelowFifty pairs. The restarted quantile's frequency gap over that period is \nFreqLargeMeanSegFifty on the large mean shift.

\paragraph{The boost helps and the jump alone hurts.} Restarted ACI pairs a jump at the boundary with a boosted step. Over the first \nCountFifty forecasts, the boost alone, which keeps ACI's threshold at the boundary, has a smaller frequency gap than ACI with a band below zero on all \nArmBoostFreqBelowFifty configurations with a break, at a difference of $\nArmMeanShiftBoostFreqFiftyDiff$ on the mean shift against restarted ACI's $\nArmMeanShiftAbaciFreqFiftyDiff$. The jump alone, which keeps ACI's step, has a larger gap than ACI on all \nArmMoveFreqAboveFifty, at \nArmMeanShiftMoveFreqFiftyDiff on the mean shift. The jump lowers the gap more with the boosted step than with ACI's step, an interaction whose band lies below zero on all \nInteractFreqBelowFifty.

Table~\ref{tab:arms} separates restarted ACI's jump from its boosted step with two ablations. The boost alone lies below ACI with a band below zero on all \nArmBoostFreqBelowEvery configurations with a break at each of the \nCountsDeclared counts from the first \nCountTen to the first \nCountFiveThousand forecasts, and the jump alone lies above ACI on all \nArmMoveFreqAboveEvery at every count.

Restarted ACI jumps to the quantile of the evaluated scores at the boundary and boosts its step. The jump alone's cells over the first \nCountFiveThousand forecasts print at five decimals to show their sign.

\begin{table}[H]
\caption{Frequency gap of restarted ACI and of its two parts, each less ACI's at step \nSynGamma, over the first $m$ forecasts at every prespecified count. The boost alone keeps ACI's threshold at the boundary, and the jump alone keeps ACI's step. A negative difference is closer to nominal than ACI.}
\label{tab:arms}
\centering
\scriptsize
\setlength{\tabcolsep}{3pt}
\begin{tabular}{llccc}
\toprule
configuration & first $m$ & restarted ACI & boost alone & jump alone \\
\midrule
no break & \nCountTen & $\nArmNoBreakAbaciFreqTenDiff$ $[\nArmNoBreakAbaciFreqTenLo,\, \nArmNoBreakAbaciFreqTenHi]$ & $\nArmNoBreakBoostFreqTenDiff$ $[\nArmNoBreakBoostFreqTenLo,\, \nArmNoBreakBoostFreqTenHi]$ & $\nArmNoBreakMoveFreqTenDiff$ $[\nArmNoBreakMoveFreqTenLo,\, \nArmNoBreakMoveFreqTenHi]$ \\
$(\nSynDeltaNoBreak,\, \nSynSigmaNoBreak)$ & \nCountTwentyFive & $\nArmNoBreakAbaciFreqTwentyFiveDiff$ $[\nArmNoBreakAbaciFreqTwentyFiveLo,\, \nArmNoBreakAbaciFreqTwentyFiveHi]$ & $\nArmNoBreakBoostFreqTwentyFiveDiff$ $[\nArmNoBreakBoostFreqTwentyFiveLo,\, \nArmNoBreakBoostFreqTwentyFiveHi]$ & $\nArmNoBreakMoveFreqTwentyFiveDiff$ $[\nArmNoBreakMoveFreqTwentyFiveLo,\, \nArmNoBreakMoveFreqTwentyFiveHi]$ \\
 & \nCountFifty & $\nArmNoBreakAbaciFreqFiftyDiff$ $[\nArmNoBreakAbaciFreqFiftyLo,\, \nArmNoBreakAbaciFreqFiftyHi]$ & \nArmNoBreakBoostFreqFiftyDiff $[\nArmNoBreakBoostFreqFiftyLo,\, \nArmNoBreakBoostFreqFiftyHi]$ & $\nArmNoBreakMoveFreqFiftyDiff$ $[\nArmNoBreakMoveFreqFiftyLo,\, \nArmNoBreakMoveFreqFiftyHi]$ \\
 & \nCountTwoHundred & \nArmNoBreakAbaciFreqTwoHundredDiff $[\nArmNoBreakAbaciFreqTwoHundredLo,\, \nArmNoBreakAbaciFreqTwoHundredHi]$ & \nArmNoBreakBoostFreqTwoHundredDiff $[\nArmNoBreakBoostFreqTwoHundredLo,\, \nArmNoBreakBoostFreqTwoHundredHi]$ & $\nArmNoBreakMoveFreqTwoHundredDiff$ $[\nArmNoBreakMoveFreqTwoHundredLo,\, \nArmNoBreakMoveFreqTwoHundredHi]$ \\
 & \nCountThousand & \nArmNoBreakAbaciFreqThousandDiff $[\nArmNoBreakAbaciFreqThousandLo,\, \nArmNoBreakAbaciFreqThousandHi]$ & \nArmNoBreakBoostFreqThousandDiff $[\nArmNoBreakBoostFreqThousandLo,\, \nArmNoBreakBoostFreqThousandHi]$ & $\nArmNoBreakMoveFreqThousandDiff$ $[\nArmNoBreakMoveFreqThousandLo,\, \nArmNoBreakMoveFreqThousandHi]$ \\
 & \nCountFiveThousand & \nArmNoBreakAbaciFreqFiveThousandDiff $[\nArmNoBreakAbaciFreqFiveThousandLo,\, \nArmNoBreakAbaciFreqFiveThousandHi]$ & \nArmNoBreakBoostFreqFiveThousandDiff $[\nArmNoBreakBoostFreqFiveThousandLo,\, \nArmNoBreakBoostFreqFiveThousandHi]$ & $\nArmNoBreakMoveFreqFiveThousandFineDiff$ $[\nArmNoBreakMoveFreqFiveThousandFineLo,\, \nArmNoBreakMoveFreqFiveThousandFineHi]$ \\
\midrule
mean shift & \nCountTen & $\nArmMeanShiftAbaciFreqTenDiff$ $[\nArmMeanShiftAbaciFreqTenLo,\, \nArmMeanShiftAbaciFreqTenHi]$ & $\nArmMeanShiftBoostFreqTenDiff$ $[\nArmMeanShiftBoostFreqTenLo,\, \nArmMeanShiftBoostFreqTenHi]$ & \nArmMeanShiftMoveFreqTenDiff $[\nArmMeanShiftMoveFreqTenLo,\, \nArmMeanShiftMoveFreqTenHi]$ \\
$(\nSynDeltaMeanShift,\, \nSynSigmaMeanShift)$ & \nCountTwentyFive & $\nArmMeanShiftAbaciFreqTwentyFiveDiff$ $[\nArmMeanShiftAbaciFreqTwentyFiveLo,\, \nArmMeanShiftAbaciFreqTwentyFiveHi]$ & $\nArmMeanShiftBoostFreqTwentyFiveDiff$ $[\nArmMeanShiftBoostFreqTwentyFiveLo,\, \nArmMeanShiftBoostFreqTwentyFiveHi]$ & \nArmMeanShiftMoveFreqTwentyFiveDiff $[\nArmMeanShiftMoveFreqTwentyFiveLo,\, \nArmMeanShiftMoveFreqTwentyFiveHi]$ \\
 & \nCountFifty & $\nArmMeanShiftAbaciFreqFiftyDiff$ $[\nArmMeanShiftAbaciFreqFiftyLo,\, \nArmMeanShiftAbaciFreqFiftyHi]$ & $\nArmMeanShiftBoostFreqFiftyDiff$ $[\nArmMeanShiftBoostFreqFiftyLo,\, \nArmMeanShiftBoostFreqFiftyHi]$ & \nArmMeanShiftMoveFreqFiftyDiff $[\nArmMeanShiftMoveFreqFiftyLo,\, \nArmMeanShiftMoveFreqFiftyHi]$ \\
 & \nCountTwoHundred & $\nArmMeanShiftAbaciFreqTwoHundredDiff$ $[\nArmMeanShiftAbaciFreqTwoHundredLo,\, \nArmMeanShiftAbaciFreqTwoHundredHi]$ & $\nArmMeanShiftBoostFreqTwoHundredDiff$ $[\nArmMeanShiftBoostFreqTwoHundredLo,\, \nArmMeanShiftBoostFreqTwoHundredHi]$ & \nArmMeanShiftMoveFreqTwoHundredDiff $[\nArmMeanShiftMoveFreqTwoHundredLo,\, \nArmMeanShiftMoveFreqTwoHundredHi]$ \\
 & \nCountThousand & $\nArmMeanShiftAbaciFreqThousandDiff$ $[\nArmMeanShiftAbaciFreqThousandLo,\, \nArmMeanShiftAbaciFreqThousandHi]$ & $\nArmMeanShiftBoostFreqThousandDiff$ $[\nArmMeanShiftBoostFreqThousandLo,\, \nArmMeanShiftBoostFreqThousandHi]$ & \nArmMeanShiftMoveFreqThousandDiff $[\nArmMeanShiftMoveFreqThousandLo,\, \nArmMeanShiftMoveFreqThousandHi]$ \\
 & \nCountFiveThousand & $\nArmMeanShiftAbaciFreqFiveThousandDiff$ $[\nArmMeanShiftAbaciFreqFiveThousandLo,\, \nArmMeanShiftAbaciFreqFiveThousandHi]$ & $\nArmMeanShiftBoostFreqFiveThousandDiff$ $[\nArmMeanShiftBoostFreqFiveThousandLo,\, \nArmMeanShiftBoostFreqFiveThousandHi]$ & \nArmMeanShiftMoveFreqFiveThousandFineDiff $[\nArmMeanShiftMoveFreqFiveThousandFineLo,\, \nArmMeanShiftMoveFreqFiveThousandFineHi]$ \\
\midrule
variance shift & \nCountTen & $\nArmVarShiftAbaciFreqTenDiff$ $[\nArmVarShiftAbaciFreqTenLo,\, \nArmVarShiftAbaciFreqTenHi]$ & $\nArmVarShiftBoostFreqTenDiff$ $[\nArmVarShiftBoostFreqTenLo,\, \nArmVarShiftBoostFreqTenHi]$ & \nArmVarShiftMoveFreqTenDiff $[\nArmVarShiftMoveFreqTenLo,\, \nArmVarShiftMoveFreqTenHi]$ \\
$(\nSynDeltaVarShift,\, \nSynSigmaVarShift)$ & \nCountTwentyFive & $\nArmVarShiftAbaciFreqTwentyFiveDiff$ $[\nArmVarShiftAbaciFreqTwentyFiveLo,\, \nArmVarShiftAbaciFreqTwentyFiveHi]$ & $\nArmVarShiftBoostFreqTwentyFiveDiff$ $[\nArmVarShiftBoostFreqTwentyFiveLo,\, \nArmVarShiftBoostFreqTwentyFiveHi]$ & \nArmVarShiftMoveFreqTwentyFiveDiff $[\nArmVarShiftMoveFreqTwentyFiveLo,\, \nArmVarShiftMoveFreqTwentyFiveHi]$ \\
 & \nCountFifty & $\nArmVarShiftAbaciFreqFiftyDiff$ $[\nArmVarShiftAbaciFreqFiftyLo,\, \nArmVarShiftAbaciFreqFiftyHi]$ & $\nArmVarShiftBoostFreqFiftyDiff$ $[\nArmVarShiftBoostFreqFiftyLo,\, \nArmVarShiftBoostFreqFiftyHi]$ & \nArmVarShiftMoveFreqFiftyDiff $[\nArmVarShiftMoveFreqFiftyLo,\, \nArmVarShiftMoveFreqFiftyHi]$ \\
 & \nCountTwoHundred & $\nArmVarShiftAbaciFreqTwoHundredDiff$ $[\nArmVarShiftAbaciFreqTwoHundredLo,\, \nArmVarShiftAbaciFreqTwoHundredHi]$ & $\nArmVarShiftBoostFreqTwoHundredDiff$ $[\nArmVarShiftBoostFreqTwoHundredLo,\, \nArmVarShiftBoostFreqTwoHundredHi]$ & \nArmVarShiftMoveFreqTwoHundredDiff $[\nArmVarShiftMoveFreqTwoHundredLo,\, \nArmVarShiftMoveFreqTwoHundredHi]$ \\
 & \nCountThousand & $\nArmVarShiftAbaciFreqThousandDiff$ $[\nArmVarShiftAbaciFreqThousandLo,\, \nArmVarShiftAbaciFreqThousandHi]$ & $\nArmVarShiftBoostFreqThousandDiff$ $[\nArmVarShiftBoostFreqThousandLo,\, \nArmVarShiftBoostFreqThousandHi]$ & \nArmVarShiftMoveFreqThousandDiff $[\nArmVarShiftMoveFreqThousandLo,\, \nArmVarShiftMoveFreqThousandHi]$ \\
 & \nCountFiveThousand & $\nArmVarShiftAbaciFreqFiveThousandDiff$ $[\nArmVarShiftAbaciFreqFiveThousandLo,\, \nArmVarShiftAbaciFreqFiveThousandHi]$ & $\nArmVarShiftBoostFreqFiveThousandDiff$ $[\nArmVarShiftBoostFreqFiveThousandLo,\, \nArmVarShiftBoostFreqFiveThousandHi]$ & \nArmVarShiftMoveFreqFiveThousandFineDiff $[\nArmVarShiftMoveFreqFiveThousandFineLo,\, \nArmVarShiftMoveFreqFiveThousandFineHi]$ \\
\midrule
both shifts & \nCountTen & $\nArmBothShiftAbaciFreqTenDiff$ $[\nArmBothShiftAbaciFreqTenLo,\, \nArmBothShiftAbaciFreqTenHi]$ & $\nArmBothShiftBoostFreqTenDiff$ $[\nArmBothShiftBoostFreqTenLo,\, \nArmBothShiftBoostFreqTenHi]$ & \nArmBothShiftMoveFreqTenDiff $[\nArmBothShiftMoveFreqTenLo,\, \nArmBothShiftMoveFreqTenHi]$ \\
$(\nSynDeltaBothShift,\, \nSynSigmaBothShift)$ & \nCountTwentyFive & $\nArmBothShiftAbaciFreqTwentyFiveDiff$ $[\nArmBothShiftAbaciFreqTwentyFiveLo,\, \nArmBothShiftAbaciFreqTwentyFiveHi]$ & $\nArmBothShiftBoostFreqTwentyFiveDiff$ $[\nArmBothShiftBoostFreqTwentyFiveLo,\, \nArmBothShiftBoostFreqTwentyFiveHi]$ & \nArmBothShiftMoveFreqTwentyFiveDiff $[\nArmBothShiftMoveFreqTwentyFiveLo,\, \nArmBothShiftMoveFreqTwentyFiveHi]$ \\
 & \nCountFifty & $\nArmBothShiftAbaciFreqFiftyDiff$ $[\nArmBothShiftAbaciFreqFiftyLo,\, \nArmBothShiftAbaciFreqFiftyHi]$ & $\nArmBothShiftBoostFreqFiftyDiff$ $[\nArmBothShiftBoostFreqFiftyLo,\, \nArmBothShiftBoostFreqFiftyHi]$ & \nArmBothShiftMoveFreqFiftyDiff $[\nArmBothShiftMoveFreqFiftyLo,\, \nArmBothShiftMoveFreqFiftyHi]$ \\
 & \nCountTwoHundred & $\nArmBothShiftAbaciFreqTwoHundredDiff$ $[\nArmBothShiftAbaciFreqTwoHundredLo,\, \nArmBothShiftAbaciFreqTwoHundredHi]$ & $\nArmBothShiftBoostFreqTwoHundredDiff$ $[\nArmBothShiftBoostFreqTwoHundredLo,\, \nArmBothShiftBoostFreqTwoHundredHi]$ & \nArmBothShiftMoveFreqTwoHundredDiff $[\nArmBothShiftMoveFreqTwoHundredLo,\, \nArmBothShiftMoveFreqTwoHundredHi]$ \\
 & \nCountThousand & $\nArmBothShiftAbaciFreqThousandDiff$ $[\nArmBothShiftAbaciFreqThousandLo,\, \nArmBothShiftAbaciFreqThousandHi]$ & $\nArmBothShiftBoostFreqThousandDiff$ $[\nArmBothShiftBoostFreqThousandLo,\, \nArmBothShiftBoostFreqThousandHi]$ & \nArmBothShiftMoveFreqThousandDiff $[\nArmBothShiftMoveFreqThousandLo,\, \nArmBothShiftMoveFreqThousandHi]$ \\
 & \nCountFiveThousand & $\nArmBothShiftAbaciFreqFiveThousandDiff$ $[\nArmBothShiftAbaciFreqFiveThousandLo,\, \nArmBothShiftAbaciFreqFiveThousandHi]$ & $\nArmBothShiftBoostFreqFiveThousandDiff$ $[\nArmBothShiftBoostFreqFiveThousandLo,\, \nArmBothShiftBoostFreqFiveThousandHi]$ & \nArmBothShiftMoveFreqFiveThousandFineDiff $[\nArmBothShiftMoveFreqFiveThousandFineLo,\, \nArmBothShiftMoveFreqFiveThousandFineHi]$ \\
\midrule
large mean shift & \nCountTen & $\nArmLargeMeanAbaciFreqTenDiff$ $[\nArmLargeMeanAbaciFreqTenLo,\, \nArmLargeMeanAbaciFreqTenHi]$ & $\nArmLargeMeanBoostFreqTenDiff$ $[\nArmLargeMeanBoostFreqTenLo,\, \nArmLargeMeanBoostFreqTenHi]$ & \nArmLargeMeanMoveFreqTenDiff $[\nArmLargeMeanMoveFreqTenLo,\, \nArmLargeMeanMoveFreqTenHi]$ \\
$(\nSynDeltaLargeMean,\, \nSynSigmaLargeMean)$ & \nCountTwentyFive & $\nArmLargeMeanAbaciFreqTwentyFiveDiff$ $[\nArmLargeMeanAbaciFreqTwentyFiveLo,\, \nArmLargeMeanAbaciFreqTwentyFiveHi]$ & $\nArmLargeMeanBoostFreqTwentyFiveDiff$ $[\nArmLargeMeanBoostFreqTwentyFiveLo,\, \nArmLargeMeanBoostFreqTwentyFiveHi]$ & \nArmLargeMeanMoveFreqTwentyFiveDiff $[\nArmLargeMeanMoveFreqTwentyFiveLo,\, \nArmLargeMeanMoveFreqTwentyFiveHi]$ \\
 & \nCountFifty & $\nArmLargeMeanAbaciFreqFiftyDiff$ $[\nArmLargeMeanAbaciFreqFiftyLo,\, \nArmLargeMeanAbaciFreqFiftyHi]$ & $\nArmLargeMeanBoostFreqFiftyDiff$ $[\nArmLargeMeanBoostFreqFiftyLo,\, \nArmLargeMeanBoostFreqFiftyHi]$ & \nArmLargeMeanMoveFreqFiftyDiff $[\nArmLargeMeanMoveFreqFiftyLo,\, \nArmLargeMeanMoveFreqFiftyHi]$ \\
 & \nCountTwoHundred & $\nArmLargeMeanAbaciFreqTwoHundredDiff$ $[\nArmLargeMeanAbaciFreqTwoHundredLo,\, \nArmLargeMeanAbaciFreqTwoHundredHi]$ & $\nArmLargeMeanBoostFreqTwoHundredDiff$ $[\nArmLargeMeanBoostFreqTwoHundredLo,\, \nArmLargeMeanBoostFreqTwoHundredHi]$ & \nArmLargeMeanMoveFreqTwoHundredDiff $[\nArmLargeMeanMoveFreqTwoHundredLo,\, \nArmLargeMeanMoveFreqTwoHundredHi]$ \\
 & \nCountThousand & $\nArmLargeMeanAbaciFreqThousandDiff$ $[\nArmLargeMeanAbaciFreqThousandLo,\, \nArmLargeMeanAbaciFreqThousandHi]$ & $\nArmLargeMeanBoostFreqThousandDiff$ $[\nArmLargeMeanBoostFreqThousandLo,\, \nArmLargeMeanBoostFreqThousandHi]$ & \nArmLargeMeanMoveFreqThousandDiff $[\nArmLargeMeanMoveFreqThousandLo,\, \nArmLargeMeanMoveFreqThousandHi]$ \\
 & \nCountFiveThousand & $\nArmLargeMeanAbaciFreqFiveThousandDiff$ $[\nArmLargeMeanAbaciFreqFiveThousandLo,\, \nArmLargeMeanAbaciFreqFiveThousandHi]$ & $\nArmLargeMeanBoostFreqFiveThousandDiff$ $[\nArmLargeMeanBoostFreqFiveThousandLo,\, \nArmLargeMeanBoostFreqFiveThousandHi]$ & \nArmLargeMeanMoveFreqFiveThousandFineDiff $[\nArmLargeMeanMoveFreqFiveThousandFineLo,\, \nArmLargeMeanMoveFreqFiveThousandFineHi]$ \\
\bottomrule
\end{tabular}
\par\raggedright ACI is adaptive conformal inference.
\end{table}

Table~\ref{tab:contrast} sets the boosted form against the boost alone, beside the interaction of jump and boost. Over the first \nCountFifty forecasts the boosted form less the boost alone is $\nContrastMeanShiftFreqFiftyDiff$ on the mean shift, with a band below zero on \nContrastFreqBelowFiftyWord configurations with a break and including zero on the other \nContrastFreqReachFiftyWord. Over the first \nCountTen forecasts the boosted form lies above the boost alone on all \nContrastFreqAboveTen configurations with a break. For the reliability gap the same contrast on the mean shift is $\nContrastMeanShiftPtFiftyDiff$ at the \nOrdFifty forecast, with a band of $[\nContrastMeanShiftPtFiftyLo, \nContrastMeanShiftPtFiftyHi]$ that includes zero, and $\nContrastMeanShiftPtTenDiff$ at the \nOrdTen, with a band of $[\nContrastMeanShiftPtTenLo, \nContrastMeanShiftPtTenHi]$ above zero. The interaction's reliability band at the \nOrdFifty forecast lies below zero, at $[\nInteractMeanShiftPtFiftyLo, \nInteractMeanShiftPtFiftyHi]$.

\begin{table}[H]
\caption{Restarted ACI's frequency gap less the boost alone's over the first $m$ forecasts, beside the interaction of jump and boost. The interaction is restarted ACI less the boost alone less the jump alone plus ACI, the jump's effect with the boosted step less its effect with ACI's step. A negative contrast favours restarted ACI, and a negative interaction means the jump lowers the gap more with the boosted step than without it.}
\label{tab:contrast}
\centering
\scriptsize
\setlength{\tabcolsep}{6pt}
\begin{tabular}{llcc}
\toprule
configuration & first $m$ & restarted ACI less the boost alone & interaction \\
\midrule
no break & \nCountTen & $\nContrastNoBreakFreqTenDiff$ $[\nContrastNoBreakFreqTenLo,\, \nContrastNoBreakFreqTenHi]$ & \nInteractNoBreakFreqTenDiff $[\nInteractNoBreakFreqTenLo,\, \nInteractNoBreakFreqTenHi]$ \\
$(\nSynDeltaNoBreak,\, \nSynSigmaNoBreak)$ & \nCountTwentyFive & $\nContrastNoBreakFreqTwentyFiveDiff$ $[\nContrastNoBreakFreqTwentyFiveLo,\, \nContrastNoBreakFreqTwentyFiveHi]$ & \nInteractNoBreakFreqTwentyFiveDiff $[\nInteractNoBreakFreqTwentyFiveLo,\, \nInteractNoBreakFreqTwentyFiveHi]$ \\
 & \nCountFifty & $\nContrastNoBreakFreqFiftyDiff$ $[\nContrastNoBreakFreqFiftyLo,\, \nContrastNoBreakFreqFiftyHi]$ & \nInteractNoBreakFreqFiftyDiff $[\nInteractNoBreakFreqFiftyLo,\, \nInteractNoBreakFreqFiftyHi]$ \\
 & \nCountTwoHundred & $\nContrastNoBreakFreqTwoHundredDiff$ $[\nContrastNoBreakFreqTwoHundredLo,\, \nContrastNoBreakFreqTwoHundredHi]$ & \nInteractNoBreakFreqTwoHundredDiff $[\nInteractNoBreakFreqTwoHundredLo,\, \nInteractNoBreakFreqTwoHundredHi]$ \\
 & \nCountThousand & \nContrastNoBreakFreqThousandDiff $[\nContrastNoBreakFreqThousandLo,\, \nContrastNoBreakFreqThousandHi]$ & \nInteractNoBreakFreqThousandDiff $[\nInteractNoBreakFreqThousandLo,\, \nInteractNoBreakFreqThousandHi]$ \\
 & \nCountFiveThousand & \nContrastNoBreakFreqFiveThousandDiff $[\nContrastNoBreakFreqFiveThousandLo,\, \nContrastNoBreakFreqFiveThousandHi]$ & \nInteractNoBreakFreqFiveThousandDiff $[\nInteractNoBreakFreqFiveThousandLo,\, \nInteractNoBreakFreqFiveThousandHi]$ \\
\midrule
mean shift & \nCountTen & \nContrastMeanShiftFreqTenDiff $[\nContrastMeanShiftFreqTenLo,\, \nContrastMeanShiftFreqTenHi]$ & $\nInteractMeanShiftFreqTenDiff$ $[\nInteractMeanShiftFreqTenLo,\, \nInteractMeanShiftFreqTenHi]$ \\
$(\nSynDeltaMeanShift,\, \nSynSigmaMeanShift)$ & \nCountTwentyFive & $\nContrastMeanShiftFreqTwentyFiveDiff$ $[\nContrastMeanShiftFreqTwentyFiveLo,\, \nContrastMeanShiftFreqTwentyFiveHi]$ & $\nInteractMeanShiftFreqTwentyFiveDiff$ $[\nInteractMeanShiftFreqTwentyFiveLo,\, \nInteractMeanShiftFreqTwentyFiveHi]$ \\
 & \nCountFifty & $\nContrastMeanShiftFreqFiftyDiff$ $[\nContrastMeanShiftFreqFiftyLo,\, \nContrastMeanShiftFreqFiftyHi]$ & $\nInteractMeanShiftFreqFiftyDiff$ $[\nInteractMeanShiftFreqFiftyLo,\, \nInteractMeanShiftFreqFiftyHi]$ \\
 & \nCountTwoHundred & $\nContrastMeanShiftFreqTwoHundredDiff$ $[\nContrastMeanShiftFreqTwoHundredLo,\, \nContrastMeanShiftFreqTwoHundredHi]$ & $\nInteractMeanShiftFreqTwoHundredDiff$ $[\nInteractMeanShiftFreqTwoHundredLo,\, \nInteractMeanShiftFreqTwoHundredHi]$ \\
 & \nCountThousand & $\nContrastMeanShiftFreqThousandDiff$ $[\nContrastMeanShiftFreqThousandLo,\, \nContrastMeanShiftFreqThousandHi]$ & $\nInteractMeanShiftFreqThousandDiff$ $[\nInteractMeanShiftFreqThousandLo,\, \nInteractMeanShiftFreqThousandHi]$ \\
 & \nCountFiveThousand & \nContrastMeanShiftFreqFiveThousandDiff $[\nContrastMeanShiftFreqFiveThousandLo,\, \nContrastMeanShiftFreqFiveThousandHi]$ & $\nInteractMeanShiftFreqFiveThousandDiff$ $[\nInteractMeanShiftFreqFiveThousandLo,\, \nInteractMeanShiftFreqFiveThousandHi]$ \\
\midrule
variance shift & \nCountTen & \nContrastVarShiftFreqTenDiff $[\nContrastVarShiftFreqTenLo,\, \nContrastVarShiftFreqTenHi]$ & $\nInteractVarShiftFreqTenDiff$ $[\nInteractVarShiftFreqTenLo,\, \nInteractVarShiftFreqTenHi]$ \\
$(\nSynDeltaVarShift,\, \nSynSigmaVarShift)$ & \nCountTwentyFive & \nContrastVarShiftFreqTwentyFiveDiff $[\nContrastVarShiftFreqTwentyFiveLo,\, \nContrastVarShiftFreqTwentyFiveHi]$ & $\nInteractVarShiftFreqTwentyFiveDiff$ $[\nInteractVarShiftFreqTwentyFiveLo,\, \nInteractVarShiftFreqTwentyFiveHi]$ \\
 & \nCountFifty & $\nContrastVarShiftFreqFiftyDiff$ $[\nContrastVarShiftFreqFiftyLo,\, \nContrastVarShiftFreqFiftyHi]$ & $\nInteractVarShiftFreqFiftyDiff$ $[\nInteractVarShiftFreqFiftyLo,\, \nInteractVarShiftFreqFiftyHi]$ \\
 & \nCountTwoHundred & \nContrastVarShiftFreqTwoHundredDiff $[\nContrastVarShiftFreqTwoHundredLo,\, \nContrastVarShiftFreqTwoHundredHi]$ & $\nInteractVarShiftFreqTwoHundredDiff$ $[\nInteractVarShiftFreqTwoHundredLo,\, \nInteractVarShiftFreqTwoHundredHi]$ \\
 & \nCountThousand & \nContrastVarShiftFreqThousandDiff $[\nContrastVarShiftFreqThousandLo,\, \nContrastVarShiftFreqThousandHi]$ & $\nInteractVarShiftFreqThousandDiff$ $[\nInteractVarShiftFreqThousandLo,\, \nInteractVarShiftFreqThousandHi]$ \\
 & \nCountFiveThousand & \nContrastVarShiftFreqFiveThousandDiff $[\nContrastVarShiftFreqFiveThousandLo,\, \nContrastVarShiftFreqFiveThousandHi]$ & $\nInteractVarShiftFreqFiveThousandDiff$ $[\nInteractVarShiftFreqFiveThousandLo,\, \nInteractVarShiftFreqFiveThousandHi]$ \\
\midrule
both shifts & \nCountTen & \nContrastBothShiftFreqTenDiff $[\nContrastBothShiftFreqTenLo,\, \nContrastBothShiftFreqTenHi]$ & $\nInteractBothShiftFreqTenDiff$ $[\nInteractBothShiftFreqTenLo,\, \nInteractBothShiftFreqTenHi]$ \\
$(\nSynDeltaBothShift,\, \nSynSigmaBothShift)$ & \nCountTwentyFive & \nContrastBothShiftFreqTwentyFiveDiff $[\nContrastBothShiftFreqTwentyFiveLo,\, \nContrastBothShiftFreqTwentyFiveHi]$ & $\nInteractBothShiftFreqTwentyFiveDiff$ $[\nInteractBothShiftFreqTwentyFiveLo,\, \nInteractBothShiftFreqTwentyFiveHi]$ \\
 & \nCountFifty & \nContrastBothShiftFreqFiftyDiff $[\nContrastBothShiftFreqFiftyLo,\, \nContrastBothShiftFreqFiftyHi]$ & $\nInteractBothShiftFreqFiftyDiff$ $[\nInteractBothShiftFreqFiftyLo,\, \nInteractBothShiftFreqFiftyHi]$ \\
 & \nCountTwoHundred & \nContrastBothShiftFreqTwoHundredDiff $[\nContrastBothShiftFreqTwoHundredLo,\, \nContrastBothShiftFreqTwoHundredHi]$ & $\nInteractBothShiftFreqTwoHundredDiff$ $[\nInteractBothShiftFreqTwoHundredLo,\, \nInteractBothShiftFreqTwoHundredHi]$ \\
 & \nCountThousand & \nContrastBothShiftFreqThousandDiff $[\nContrastBothShiftFreqThousandLo,\, \nContrastBothShiftFreqThousandHi]$ & $\nInteractBothShiftFreqThousandDiff$ $[\nInteractBothShiftFreqThousandLo,\, \nInteractBothShiftFreqThousandHi]$ \\
 & \nCountFiveThousand & \nContrastBothShiftFreqFiveThousandDiff $[\nContrastBothShiftFreqFiveThousandLo,\, \nContrastBothShiftFreqFiveThousandHi]$ & \nInteractBothShiftFreqFiveThousandDiff $[\nInteractBothShiftFreqFiveThousandLo,\, \nInteractBothShiftFreqFiveThousandHi]$ \\
\midrule
large mean shift & \nCountTen & \nContrastLargeMeanFreqTenDiff $[\nContrastLargeMeanFreqTenLo,\, \nContrastLargeMeanFreqTenHi]$ & $\nInteractLargeMeanFreqTenDiff$ $[\nInteractLargeMeanFreqTenLo,\, \nInteractLargeMeanFreqTenHi]$ \\
$(\nSynDeltaLargeMean,\, \nSynSigmaLargeMean)$ & \nCountTwentyFive & $\nContrastLargeMeanFreqTwentyFiveDiff$ $[\nContrastLargeMeanFreqTwentyFiveLo,\, \nContrastLargeMeanFreqTwentyFiveHi]$ & $\nInteractLargeMeanFreqTwentyFiveDiff$ $[\nInteractLargeMeanFreqTwentyFiveLo,\, \nInteractLargeMeanFreqTwentyFiveHi]$ \\
 & \nCountFifty & \nContrastLargeMeanFreqFiftyDiff $[\nContrastLargeMeanFreqFiftyLo,\, \nContrastLargeMeanFreqFiftyHi]$ & $\nInteractLargeMeanFreqFiftyDiff$ $[\nInteractLargeMeanFreqFiftyLo,\, \nInteractLargeMeanFreqFiftyHi]$ \\
 & \nCountTwoHundred & \nContrastLargeMeanFreqTwoHundredDiff $[\nContrastLargeMeanFreqTwoHundredLo,\, \nContrastLargeMeanFreqTwoHundredHi]$ & $\nInteractLargeMeanFreqTwoHundredDiff$ $[\nInteractLargeMeanFreqTwoHundredLo,\, \nInteractLargeMeanFreqTwoHundredHi]$ \\
 & \nCountThousand & \nContrastLargeMeanFreqThousandDiff $[\nContrastLargeMeanFreqThousandLo,\, \nContrastLargeMeanFreqThousandHi]$ & $\nInteractLargeMeanFreqThousandDiff$ $[\nInteractLargeMeanFreqThousandLo,\, \nInteractLargeMeanFreqThousandHi]$ \\
 & \nCountFiveThousand & \nContrastLargeMeanFreqFiveThousandDiff $[\nContrastLargeMeanFreqFiveThousandLo,\, \nContrastLargeMeanFreqFiveThousandHi]$ & \nInteractLargeMeanFreqFiveThousandDiff $[\nInteractLargeMeanFreqFiveThousandLo,\, \nInteractLargeMeanFreqFiveThousandHi]$ \\
\bottomrule
\end{tabular}
\par\raggedright ACI is adaptive conformal inference.
\end{table}

Every method's reliability gap at the \nOrdTen and \nOrdFifty forecasts is in Table~\ref{tab:pertime-full}. At the \nOrdTen, the restarted quantile's band lies below zero against all \nSrcPtBelowLargeMeanTen variants on every configuration whose break shifts the mean. On the variance shift it lies below zero against every variant except \nSrcPtReachVarShiftTenWho, where it includes zero, and without a break it lies above zero against all \nSrcPtAboveNoBreakTen. Against ACI it lies below zero on the large mean shift and on both shifts, at $[\nPtBandLargeMeanAciTenLo, \nPtBandLargeMeanAciTenHi]$ and $[\nPtBandBothShiftAciTenLo, \nPtBandBothShiftAciTenHi]$. It includes zero on the mean shift, at $[\nPtBandMeanShiftAciTenLo, \nPtBandMeanShiftAciTenHi]$, and lies above zero on the variance shift and without a break.

An infinite threshold counts as a gap of \nInfGap, though no forecast at either count has one. Restarted ACI appears in its boosted form, in its boosted burn-in form, which empties its score set when told the boundary and jumps after a burn-in, and in its decaying step form, whose step decays as \nDecConstant over the number of updates since the restart to the power \nDecExponent. The oracle's gap is zero by construction. The warm start returns the restarted quantile's threshold from the first finite one on, so their gaps are equal. The released data include every paired band.

\begin{table}[H]
\caption{Reliability gap of every method on the five synthetic configurations at the \nOrdTen and \nOrdFifty forecasts, as means over \nSynReps replications under the known law after the break.}
\label{tab:pertime-full}
\centering
\scriptsize
\setlength{\tabcolsep}{2.6pt}
\begin{tabular}{lcccccccccc}
\toprule
 & \multicolumn{2}{c}{\begin{tabular}[c]{@{}c@{}}no break\\{}$(\nSynDeltaNoBreak,\, \nSynSigmaNoBreak)$\end{tabular}} & \multicolumn{2}{c}{\begin{tabular}[c]{@{}c@{}}mean shift\\{}$(\nSynDeltaMeanShift,\, \nSynSigmaMeanShift)$\end{tabular}} & \multicolumn{2}{c}{\begin{tabular}[c]{@{}c@{}}variance shift\\{}$(\nSynDeltaVarShift,\, \nSynSigmaVarShift)$\end{tabular}} & \multicolumn{2}{c}{\begin{tabular}[c]{@{}c@{}}both shifts\\{}$(\nSynDeltaBothShift,\, \nSynSigmaBothShift)$\end{tabular}} & \multicolumn{2}{c}{\begin{tabular}[c]{@{}c@{}}large mean shift\\{}$(\nSynDeltaLargeMean,\, \nSynSigmaLargeMean)$\end{tabular}} \\
\cmidrule(lr){2-3} \cmidrule(lr){4-5} \cmidrule(lr){6-7} \cmidrule(lr){8-9} \cmidrule(lr){10-11}
method & \nOrdTen & \nOrdFifty & \nOrdTen & \nOrdFifty & \nOrdTen & \nOrdFifty & \nOrdTen & \nOrdFifty & \nOrdTen & \nOrdFifty \\
\midrule
restarted quantile & \nPtNoBreakSegTen & \nPtNoBreakSegFifty & \nPtMeanShiftSegTen & \nPtMeanShiftSegFifty & \nPtVarShiftSegTen & \nPtVarShiftSegFifty & \nPtBothShiftSegTen & \nPtBothShiftSegFifty & \nPtLargeMeanSegTen & \nPtLargeMeanSegFifty \\
warm start & \nPtNoBreakSegWarmTen & \nPtNoBreakSegWarmFifty & \nPtMeanShiftSegWarmTen & \nPtMeanShiftSegWarmFifty & \nPtVarShiftSegWarmTen & \nPtVarShiftSegWarmFifty & \nPtBothShiftSegWarmTen & \nPtBothShiftSegWarmFifty & \nPtLargeMeanSegWarmTen & \nPtLargeMeanSegWarmFifty \\
restarted ACI & \nPtNoBreakAbaciTen & \nPtNoBreakAbaciFifty & \nPtMeanShiftAbaciTen & \nPtMeanShiftAbaciFifty & \nPtVarShiftAbaciTen & \nPtVarShiftAbaciFifty & \nPtBothShiftAbaciTen & \nPtBothShiftAbaciFifty & \nPtLargeMeanAbaciTen & \nPtLargeMeanAbaciFifty \\
boosted burn-in form & \nPtNoBreakAbaciTextTen & \nPtNoBreakAbaciTextFifty & \nPtMeanShiftAbaciTextTen & \nPtMeanShiftAbaciTextFifty & \nPtVarShiftAbaciTextTen & \nPtVarShiftAbaciTextFifty & \nPtBothShiftAbaciTextTen & \nPtBothShiftAbaciTextFifty & \nPtLargeMeanAbaciTextTen & \nPtLargeMeanAbaciTextFifty \\
decaying step form & \nPtNoBreakAbaciDecTen & \nPtNoBreakAbaciDecFifty & \nPtMeanShiftAbaciDecTen & \nPtMeanShiftAbaciDecFifty & \nPtVarShiftAbaciDecTen & \nPtVarShiftAbaciDecFifty & \nPtBothShiftAbaciDecTen & \nPtBothShiftAbaciDecFifty & \nPtLargeMeanAbaciDecTen & \nPtLargeMeanAbaciDecFifty \\
ACI & \nPtNoBreakAciTen & \nPtNoBreakAciFifty & \nPtMeanShiftAciTen & \nPtMeanShiftAciFifty & \nPtVarShiftAciTen & \nPtVarShiftAciFifty & \nPtBothShiftAciTen & \nPtBothShiftAciFifty & \nPtLargeMeanAciTen & \nPtLargeMeanAciFifty \\
DtACI & \nPtNoBreakDtaciSrcTen & \nPtNoBreakDtaciSrcFifty & \nPtMeanShiftDtaciSrcTen & \nPtMeanShiftDtaciSrcFifty & \nPtVarShiftDtaciSrcTen & \nPtVarShiftDtaciSrcFifty & \nPtBothShiftDtaciSrcTen & \nPtBothShiftDtaciSrcFifty & \nPtLargeMeanDtaciSrcTen & \nPtLargeMeanDtaciSrcFifty \\
SAOCP & \nPtNoBreakSaocpSrcTen & \nPtNoBreakSaocpSrcFifty & \nPtMeanShiftSaocpSrcTen & \nPtMeanShiftSaocpSrcFifty & \nPtVarShiftSaocpSrcTen & \nPtVarShiftSaocpSrcFifty & \nPtBothShiftSaocpSrcTen & \nPtBothShiftSaocpSrcFifty & \nPtLargeMeanSaocpSrcTen & \nPtLargeMeanSaocpSrcFifty \\
conformal PID & \nPtNoBreakPidSrcTen & \nPtNoBreakPidSrcFifty & \nPtMeanShiftPidSrcTen & \nPtMeanShiftPidSrcFifty & \nPtVarShiftPidSrcTen & \nPtVarShiftPidSrcFifty & \nPtBothShiftPidSrcTen & \nPtBothShiftPidSrcFifty & \nPtLargeMeanPidSrcTen & \nPtLargeMeanPidSrcFifty \\
AgACI & \nPtNoBreakAgaciSrcTen & \nPtNoBreakAgaciSrcFifty & \nPtMeanShiftAgaciSrcTen & \nPtMeanShiftAgaciSrcFifty & \nPtVarShiftAgaciSrcTen & \nPtVarShiftAgaciSrcFifty & \nPtBothShiftAgaciSrcTen & \nPtBothShiftAgaciSrcFifty & \nPtLargeMeanAgaciSrcTen & \nPtLargeMeanAgaciSrcFifty \\
naive & \nPtNoBreakNaiveTen & \nPtNoBreakNaiveFifty & \nPtMeanShiftNaiveTen & \nPtMeanShiftNaiveFifty & \nPtVarShiftNaiveTen & \nPtVarShiftNaiveFifty & \nPtBothShiftNaiveTen & \nPtBothShiftNaiveFifty & \nPtLargeMeanNaiveTen & \nPtLargeMeanNaiveFifty \\
oracle & \nPtNoBreakOracleTen & \nPtNoBreakOracleFifty & \nPtMeanShiftOracleTen & \nPtMeanShiftOracleFifty & \nPtVarShiftOracleTen & \nPtVarShiftOracleFifty & \nPtBothShiftOracleTen & \nPtBothShiftOracleFifty & \nPtLargeMeanOracleTen & \nPtLargeMeanOracleFifty \\
\bottomrule
\end{tabular}
\par\raggedright ACI is adaptive conformal inference. DtACI is dynamically-tuned ACI. SAOCP is strongly adaptive online conformal prediction. Conformal PID is conformal proportional-integral-derivative control. AgACI is aggregated ACI.
\end{table}

\paragraph{The bound covers every frequency gap.} A user can compute every term of \eqref{eq:practical-correction-main} online. Our synthetic scores are absolute Gaussian residuals and so unbounded, and there the bound is a check rather than a guarantee. The complete bound lies at or above the frequency gap in all \nAcctWindows finished periods of the boosted form and the boosted burn-in form, across five configurations and six numbers of forecasts from the first \nCountTen to the first \nCountFiveThousand, though scores reach \nAcctLargestScore. We compute each term of Proposition~\ref{prop:practical-corrections} on every run and divide it by the step times the number of forecasts, the scale of the frequency gap. Over the first \nCountFifty forecasts of the four configurations with a break, even the bound without its boost term, which lies below the complete bound the proposition guarantees, is \nTdNonzeroFiftyRatioPropNoBoostMin to \nTdNonzeroFiftyRatioPropNoBoostMax times the frequency gap and covers it on every replication. On the ERCOT window that bound is \nErcotBoundAbaci against a frequency gap of \nErcotFreqErrAbaci, and Appendix~\ref{app:ercot} gives every term on the real streams.

Over the first \nCountFifty forecasts of the four configurations with a break, the jump is \nJumpShareFiftyMin of the frequency gap on \nJumpShareFiftyMinWho and \nJumpShareFiftyMax on \nJumpShareFiftyMaxWho, and the boost term is \nTdNonzeroFiftyRatioBoostMin to \nTdNonzeroFiftyRatioBoostMax times that gap. The total floor correction is zero except on the variance shift, where it averages \nPiSumVarShiftRawFifty on the score scale and \nPiSumVarShiftNormFifty after division. Tables~\ref{tab:correction} and~\ref{tab:projection} give the terms of Proposition~\ref{prop:practical-corrections} and the diagnostic of the bounded threshold over the first \nCountFifty, \nCountTwoHundred, \nCountThousand and \nCountFiveThousand forecasts.

In the table, the frequency gap is the distance of the period's miss rate from the miscoverage level. The synthetic scores are unbounded, so each value is a check rather than the proposition's guarantee. The jump, the sum of the absolute moves the restart makes, appears in score units and then, like the total floor correction and the boost term, divided by the step times $M$, the scale on which each enters the bound. Restarted ACI jumps at the boundary's own outcome and the period's telescoping sum starts after that jump, so the jump enters the bound as an allowance the period's identity does not need. The ratio is the mean range bound without its boost term over the mean frequency gap, and the last column is the share of replications on which that bound covers the replication's own frequency gap.

\begin{table}[H]
\caption{Terms of Proposition~\ref{prop:practical-corrections} for restarted ACI at step \nSynGamma, boost \nBoost and boost length \nBurnin, as means over \nSynReps replications. The range bound is the right side of \eqref{eq:practical-correction-main}, printed without and with its boost term, and only the complete bound is the guarantee.}
\label{tab:correction}
\centering
\scriptsize
\setlength{\tabcolsep}{2.6pt}
\begin{tabular}{llccccccccc}
\toprule
configuration & period & \begin{tabular}[c]{@{}c@{}}jump in\\score units\end{tabular} & jump & \begin{tabular}[c]{@{}c@{}}total\\floor\\correction\end{tabular} & \begin{tabular}[c]{@{}c@{}}boost\\term\end{tabular} & \begin{tabular}[c]{@{}c@{}}frequency\\gap\end{tabular} & \begin{tabular}[c]{@{}c@{}}range bound\\without boost\end{tabular} & \begin{tabular}[c]{@{}c@{}}range\\bound\end{tabular} & ratio & covered \\
\midrule
no break & \nCountFifty & \nTdAbaciNoBreakFiftyJumpScore & \nTdAbaciNoBreakFiftyJumpGm & \nTdAbaciNoBreakFiftyProjGm & \nTdAbaciNoBreakFiftyBoostGm & \nTdAbaciNoBreakFiftyErr & \nTdAbaciNoBreakFiftyPropNoBoost & \nTdAbaciNoBreakFiftyProp & \nTdAbaciNoBreakFiftyRatioPropNoBoost & \nTdAbaciNoBreakFiftyCoversPropNoBoost \\
$(\nSynDeltaNoBreak,\, \nSynSigmaNoBreak)$ & \nCountTwoHundred & \nTdAbaciNoBreakTwoHundredJumpScore & \nTdAbaciNoBreakTwoHundredJumpGm & \nTdAbaciNoBreakTwoHundredProjGm & \nTdAbaciNoBreakTwoHundredBoostGm & \nTdAbaciNoBreakTwoHundredErr & \nTdAbaciNoBreakTwoHundredPropNoBoost & \nTdAbaciNoBreakTwoHundredProp & \nTdAbaciNoBreakTwoHundredRatioPropNoBoost & \nTdAbaciNoBreakTwoHundredCoversPropNoBoost \\
 & \nCountThousand & \nTdAbaciNoBreakThousandJumpScore & \nTdAbaciNoBreakThousandJumpGm & \nTdAbaciNoBreakThousandProjGm & \nTdAbaciNoBreakThousandBoostGm & \nTdAbaciNoBreakThousandErr & \nTdAbaciNoBreakThousandPropNoBoost & \nTdAbaciNoBreakThousandProp & \nTdAbaciNoBreakThousandRatioPropNoBoost & \nTdAbaciNoBreakThousandCoversPropNoBoost \\
 & \nCountFiveThousand & \nTdAbaciNoBreakFiveThousandJumpScore & \nTdAbaciNoBreakFiveThousandJumpGm & \nTdAbaciNoBreakFiveThousandProjGm & \nTdAbaciNoBreakFiveThousandBoostGm & \nTdAbaciNoBreakFiveThousandErr & \nTdAbaciNoBreakFiveThousandPropNoBoost & \nTdAbaciNoBreakFiveThousandProp & \nTdAbaciNoBreakFiveThousandRatioPropNoBoost & \nTdAbaciNoBreakFiveThousandCoversPropNoBoost \\
\midrule
mean shift & \nCountFifty & \nTdAbaciMeanShiftFiftyJumpScore & \nTdAbaciMeanShiftFiftyJumpGm & \nTdAbaciMeanShiftFiftyProjGm & \nTdAbaciMeanShiftFiftyBoostGm & \nTdAbaciMeanShiftFiftyErr & \nTdAbaciMeanShiftFiftyPropNoBoost & \nTdAbaciMeanShiftFiftyProp & \nTdAbaciMeanShiftFiftyRatioPropNoBoost & \nTdAbaciMeanShiftFiftyCoversPropNoBoost \\
$(\nSynDeltaMeanShift,\, \nSynSigmaMeanShift)$ & \nCountTwoHundred & \nTdAbaciMeanShiftTwoHundredJumpScore & \nTdAbaciMeanShiftTwoHundredJumpGm & \nTdAbaciMeanShiftTwoHundredProjGm & \nTdAbaciMeanShiftTwoHundredBoostGm & \nTdAbaciMeanShiftTwoHundredErr & \nTdAbaciMeanShiftTwoHundredPropNoBoost & \nTdAbaciMeanShiftTwoHundredProp & \nTdAbaciMeanShiftTwoHundredRatioPropNoBoost & \nTdAbaciMeanShiftTwoHundredCoversPropNoBoost \\
 & \nCountThousand & \nTdAbaciMeanShiftThousandJumpScore & \nTdAbaciMeanShiftThousandJumpGm & \nTdAbaciMeanShiftThousandProjGm & \nTdAbaciMeanShiftThousandBoostGm & \nTdAbaciMeanShiftThousandErr & \nTdAbaciMeanShiftThousandPropNoBoost & \nTdAbaciMeanShiftThousandProp & \nTdAbaciMeanShiftThousandRatioPropNoBoost & \nTdAbaciMeanShiftThousandCoversPropNoBoost \\
 & \nCountFiveThousand & \nTdAbaciMeanShiftFiveThousandJumpScore & \nTdAbaciMeanShiftFiveThousandJumpGm & \nTdAbaciMeanShiftFiveThousandProjGm & \nTdAbaciMeanShiftFiveThousandBoostGm & \nTdAbaciMeanShiftFiveThousandErr & \nTdAbaciMeanShiftFiveThousandPropNoBoost & \nTdAbaciMeanShiftFiveThousandProp & \nTdAbaciMeanShiftFiveThousandRatioPropNoBoost & \nTdAbaciMeanShiftFiveThousandCoversPropNoBoost \\
\midrule
variance shift & \nCountFifty & \nTdAbaciVarShiftFiftyJumpScore & \nTdAbaciVarShiftFiftyJumpGm & \nTdAbaciVarShiftFiftyProjGm & \nTdAbaciVarShiftFiftyBoostGm & \nTdAbaciVarShiftFiftyErr & \nTdAbaciVarShiftFiftyPropNoBoost & \nTdAbaciVarShiftFiftyProp & \nTdAbaciVarShiftFiftyRatioPropNoBoost & \nTdAbaciVarShiftFiftyCoversPropNoBoost \\
$(\nSynDeltaVarShift,\, \nSynSigmaVarShift)$ & \nCountTwoHundred & \nTdAbaciVarShiftTwoHundredJumpScore & \nTdAbaciVarShiftTwoHundredJumpGm & \nTdAbaciVarShiftTwoHundredProjGm & \nTdAbaciVarShiftTwoHundredBoostGm & \nTdAbaciVarShiftTwoHundredErr & \nTdAbaciVarShiftTwoHundredPropNoBoost & \nTdAbaciVarShiftTwoHundredProp & \nTdAbaciVarShiftTwoHundredRatioPropNoBoost & \nTdAbaciVarShiftTwoHundredCoversPropNoBoost \\
 & \nCountThousand & \nTdAbaciVarShiftThousandJumpScore & \nTdAbaciVarShiftThousandJumpGm & \nTdAbaciVarShiftThousandProjGm & \nTdAbaciVarShiftThousandBoostGm & \nTdAbaciVarShiftThousandErr & \nTdAbaciVarShiftThousandPropNoBoost & \nTdAbaciVarShiftThousandProp & \nTdAbaciVarShiftThousandRatioPropNoBoost & \nTdAbaciVarShiftThousandCoversPropNoBoost \\
 & \nCountFiveThousand & \nTdAbaciVarShiftFiveThousandJumpScore & \nTdAbaciVarShiftFiveThousandJumpGm & \nTdAbaciVarShiftFiveThousandProjGm & \nTdAbaciVarShiftFiveThousandBoostGm & \nTdAbaciVarShiftFiveThousandErr & \nTdAbaciVarShiftFiveThousandPropNoBoost & \nTdAbaciVarShiftFiveThousandProp & \nTdAbaciVarShiftFiveThousandRatioPropNoBoost & \nTdAbaciVarShiftFiveThousandCoversPropNoBoost \\
\midrule
both shifts & \nCountFifty & \nTdAbaciBothShiftFiftyJumpScore & \nTdAbaciBothShiftFiftyJumpGm & \nTdAbaciBothShiftFiftyProjGm & \nTdAbaciBothShiftFiftyBoostGm & \nTdAbaciBothShiftFiftyErr & \nTdAbaciBothShiftFiftyPropNoBoost & \nTdAbaciBothShiftFiftyProp & \nTdAbaciBothShiftFiftyRatioPropNoBoost & \nTdAbaciBothShiftFiftyCoversPropNoBoost \\
$(\nSynDeltaBothShift,\, \nSynSigmaBothShift)$ & \nCountTwoHundred & \nTdAbaciBothShiftTwoHundredJumpScore & \nTdAbaciBothShiftTwoHundredJumpGm & \nTdAbaciBothShiftTwoHundredProjGm & \nTdAbaciBothShiftTwoHundredBoostGm & \nTdAbaciBothShiftTwoHundredErr & \nTdAbaciBothShiftTwoHundredPropNoBoost & \nTdAbaciBothShiftTwoHundredProp & \nTdAbaciBothShiftTwoHundredRatioPropNoBoost & \nTdAbaciBothShiftTwoHundredCoversPropNoBoost \\
 & \nCountThousand & \nTdAbaciBothShiftThousandJumpScore & \nTdAbaciBothShiftThousandJumpGm & \nTdAbaciBothShiftThousandProjGm & \nTdAbaciBothShiftThousandBoostGm & \nTdAbaciBothShiftThousandErr & \nTdAbaciBothShiftThousandPropNoBoost & \nTdAbaciBothShiftThousandProp & \nTdAbaciBothShiftThousandRatioPropNoBoost & \nTdAbaciBothShiftThousandCoversPropNoBoost \\
 & \nCountFiveThousand & \nTdAbaciBothShiftFiveThousandJumpScore & \nTdAbaciBothShiftFiveThousandJumpGm & \nTdAbaciBothShiftFiveThousandProjGm & \nTdAbaciBothShiftFiveThousandBoostGm & \nTdAbaciBothShiftFiveThousandErr & \nTdAbaciBothShiftFiveThousandPropNoBoost & \nTdAbaciBothShiftFiveThousandProp & \nTdAbaciBothShiftFiveThousandRatioPropNoBoost & \nTdAbaciBothShiftFiveThousandCoversPropNoBoost \\
\midrule
large mean shift & \nCountFifty & \nTdAbaciLargeMeanFiftyJumpScore & \nTdAbaciLargeMeanFiftyJumpGm & \nTdAbaciLargeMeanFiftyProjGm & \nTdAbaciLargeMeanFiftyBoostGm & \nTdAbaciLargeMeanFiftyErr & \nTdAbaciLargeMeanFiftyPropNoBoost & \nTdAbaciLargeMeanFiftyProp & \nTdAbaciLargeMeanFiftyRatioPropNoBoost & \nTdAbaciLargeMeanFiftyCoversPropNoBoost \\
$(\nSynDeltaLargeMean,\, \nSynSigmaLargeMean)$ & \nCountTwoHundred & \nTdAbaciLargeMeanTwoHundredJumpScore & \nTdAbaciLargeMeanTwoHundredJumpGm & \nTdAbaciLargeMeanTwoHundredProjGm & \nTdAbaciLargeMeanTwoHundredBoostGm & \nTdAbaciLargeMeanTwoHundredErr & \nTdAbaciLargeMeanTwoHundredPropNoBoost & \nTdAbaciLargeMeanTwoHundredProp & \nTdAbaciLargeMeanTwoHundredRatioPropNoBoost & \nTdAbaciLargeMeanTwoHundredCoversPropNoBoost \\
 & \nCountThousand & \nTdAbaciLargeMeanThousandJumpScore & \nTdAbaciLargeMeanThousandJumpGm & \nTdAbaciLargeMeanThousandProjGm & \nTdAbaciLargeMeanThousandBoostGm & \nTdAbaciLargeMeanThousandErr & \nTdAbaciLargeMeanThousandPropNoBoost & \nTdAbaciLargeMeanThousandProp & \nTdAbaciLargeMeanThousandRatioPropNoBoost & \nTdAbaciLargeMeanThousandCoversPropNoBoost \\
 & \nCountFiveThousand & \nTdAbaciLargeMeanFiveThousandJumpScore & \nTdAbaciLargeMeanFiveThousandJumpGm & \nTdAbaciLargeMeanFiveThousandProjGm & \nTdAbaciLargeMeanFiveThousandBoostGm & \nTdAbaciLargeMeanFiveThousandErr & \nTdAbaciLargeMeanFiveThousandPropNoBoost & \nTdAbaciLargeMeanFiveThousandProp & \nTdAbaciLargeMeanFiveThousandRatioPropNoBoost & \nTdAbaciLargeMeanFiveThousandCoversPropNoBoost \\
\bottomrule
\end{tabular}
\end{table}

The projection share is the mean share of the period's steps on which the floor acted, and the next two columns give the shares of replications on which the floor acted at all and on which it acted on more than the prespecified \nProjThresholdPct{}\% of steps. The last column gives the share of replications whose threshold stayed within the bound that the observed score bound and the largest step give, an accounting check for each period.

\begin{table}[H]
\caption{Diagnostic of the bounded threshold for restarted ACI over \nSynReps replications. The threshold range is the largest less the smallest threshold over the period, divided by the step times $M$, and the bound beside it adds the jump and the total floor correction on that scale, which is the bound of Proposition~\ref{prop:practical-corrections} without its boost term.}
\label{tab:projection}
\centering
\scriptsize
\setlength{\tabcolsep}{3.2pt}
\begin{tabular}{llcccccc}
\toprule
configuration & period & threshold range & \begin{tabular}[c]{@{}c@{}}bound without\\boost\end{tabular} & \begin{tabular}[c]{@{}c@{}}projection\\share of steps\end{tabular} & \begin{tabular}[c]{@{}c@{}}replications on\\which the floor acted\end{tabular} & \begin{tabular}[c]{@{}c@{}}replications over\\\nProjThresholdPct{}\%\end{tabular} & \begin{tabular}[c]{@{}c@{}}threshold\\within its bound\end{tabular} \\
\midrule
no break & \nCountFifty & \nTdAbaciNoBreakFiftyRangeGm & \nTdAbaciNoBreakFiftyPropNoBoost & \nTdAbaciNoBreakFiftyProjShare & \nTdAbaciNoBreakFiftyProjAny & \nTdAbaciNoBreakFiftyProjOverOne & \nTdAbaciNoBreakFiftyCert \\
$(\nSynDeltaNoBreak,\, \nSynSigmaNoBreak)$ & \nCountTwoHundred & \nTdAbaciNoBreakTwoHundredRangeGm & \nTdAbaciNoBreakTwoHundredPropNoBoost & \nTdAbaciNoBreakTwoHundredProjShare & \nTdAbaciNoBreakTwoHundredProjAny & \nTdAbaciNoBreakTwoHundredProjOverOne & \nTdAbaciNoBreakTwoHundredCert \\
 & \nCountThousand & \nTdAbaciNoBreakThousandRangeGm & \nTdAbaciNoBreakThousandPropNoBoost & \nTdAbaciNoBreakThousandProjShare & \nTdAbaciNoBreakThousandProjAny & \nTdAbaciNoBreakThousandProjOverOne & \nTdAbaciNoBreakThousandCert \\
 & \nCountFiveThousand & \nTdAbaciNoBreakFiveThousandRangeGm & \nTdAbaciNoBreakFiveThousandPropNoBoost & \nTdAbaciNoBreakFiveThousandProjShare & \nTdAbaciNoBreakFiveThousandProjAny & \nTdAbaciNoBreakFiveThousandProjOverOne & \nTdAbaciNoBreakFiveThousandCert \\
\midrule
mean shift & \nCountFifty & \nTdAbaciMeanShiftFiftyRangeGm & \nTdAbaciMeanShiftFiftyPropNoBoost & \nTdAbaciMeanShiftFiftyProjShare & \nTdAbaciMeanShiftFiftyProjAny & \nTdAbaciMeanShiftFiftyProjOverOne & \nTdAbaciMeanShiftFiftyCert \\
$(\nSynDeltaMeanShift,\, \nSynSigmaMeanShift)$ & \nCountTwoHundred & \nTdAbaciMeanShiftTwoHundredRangeGm & \nTdAbaciMeanShiftTwoHundredPropNoBoost & \nTdAbaciMeanShiftTwoHundredProjShare & \nTdAbaciMeanShiftTwoHundredProjAny & \nTdAbaciMeanShiftTwoHundredProjOverOne & \nTdAbaciMeanShiftTwoHundredCert \\
 & \nCountThousand & \nTdAbaciMeanShiftThousandRangeGm & \nTdAbaciMeanShiftThousandPropNoBoost & \nTdAbaciMeanShiftThousandProjShare & \nTdAbaciMeanShiftThousandProjAny & \nTdAbaciMeanShiftThousandProjOverOne & \nTdAbaciMeanShiftThousandCert \\
 & \nCountFiveThousand & \nTdAbaciMeanShiftFiveThousandRangeGm & \nTdAbaciMeanShiftFiveThousandPropNoBoost & \nTdAbaciMeanShiftFiveThousandProjShare & \nTdAbaciMeanShiftFiveThousandProjAny & \nTdAbaciMeanShiftFiveThousandProjOverOne & \nTdAbaciMeanShiftFiveThousandCert \\
\midrule
variance shift & \nCountFifty & \nTdAbaciVarShiftFiftyRangeGm & \nTdAbaciVarShiftFiftyPropNoBoost & \nTdAbaciVarShiftFiftyProjShare & \nTdAbaciVarShiftFiftyProjAny & \nTdAbaciVarShiftFiftyProjOverOne & \nTdAbaciVarShiftFiftyCert \\
$(\nSynDeltaVarShift,\, \nSynSigmaVarShift)$ & \nCountTwoHundred & \nTdAbaciVarShiftTwoHundredRangeGm & \nTdAbaciVarShiftTwoHundredPropNoBoost & \nTdAbaciVarShiftTwoHundredProjShare & \nTdAbaciVarShiftTwoHundredProjAny & \nTdAbaciVarShiftTwoHundredProjOverOne & \nTdAbaciVarShiftTwoHundredCert \\
 & \nCountThousand & \nTdAbaciVarShiftThousandRangeGm & \nTdAbaciVarShiftThousandPropNoBoost & \nTdAbaciVarShiftThousandProjShare & \nTdAbaciVarShiftThousandProjAny & \nTdAbaciVarShiftThousandProjOverOne & \nTdAbaciVarShiftThousandCert \\
 & \nCountFiveThousand & \nTdAbaciVarShiftFiveThousandRangeGm & \nTdAbaciVarShiftFiveThousandPropNoBoost & \nTdAbaciVarShiftFiveThousandProjShare & \nTdAbaciVarShiftFiveThousandProjAny & \nTdAbaciVarShiftFiveThousandProjOverOne & \nTdAbaciVarShiftFiveThousandCert \\
\midrule
both shifts & \nCountFifty & \nTdAbaciBothShiftFiftyRangeGm & \nTdAbaciBothShiftFiftyPropNoBoost & \nTdAbaciBothShiftFiftyProjShare & \nTdAbaciBothShiftFiftyProjAny & \nTdAbaciBothShiftFiftyProjOverOne & \nTdAbaciBothShiftFiftyCert \\
$(\nSynDeltaBothShift,\, \nSynSigmaBothShift)$ & \nCountTwoHundred & \nTdAbaciBothShiftTwoHundredRangeGm & \nTdAbaciBothShiftTwoHundredPropNoBoost & \nTdAbaciBothShiftTwoHundredProjShare & \nTdAbaciBothShiftTwoHundredProjAny & \nTdAbaciBothShiftTwoHundredProjOverOne & \nTdAbaciBothShiftTwoHundredCert \\
 & \nCountThousand & \nTdAbaciBothShiftThousandRangeGm & \nTdAbaciBothShiftThousandPropNoBoost & \nTdAbaciBothShiftThousandProjShare & \nTdAbaciBothShiftThousandProjAny & \nTdAbaciBothShiftThousandProjOverOne & \nTdAbaciBothShiftThousandCert \\
 & \nCountFiveThousand & \nTdAbaciBothShiftFiveThousandRangeGm & \nTdAbaciBothShiftFiveThousandPropNoBoost & \nTdAbaciBothShiftFiveThousandProjShare & \nTdAbaciBothShiftFiveThousandProjAny & \nTdAbaciBothShiftFiveThousandProjOverOne & \nTdAbaciBothShiftFiveThousandCert \\
\midrule
large mean shift & \nCountFifty & \nTdAbaciLargeMeanFiftyRangeGm & \nTdAbaciLargeMeanFiftyPropNoBoost & \nTdAbaciLargeMeanFiftyProjShare & \nTdAbaciLargeMeanFiftyProjAny & \nTdAbaciLargeMeanFiftyProjOverOne & \nTdAbaciLargeMeanFiftyCert \\
$(\nSynDeltaLargeMean,\, \nSynSigmaLargeMean)$ & \nCountTwoHundred & \nTdAbaciLargeMeanTwoHundredRangeGm & \nTdAbaciLargeMeanTwoHundredPropNoBoost & \nTdAbaciLargeMeanTwoHundredProjShare & \nTdAbaciLargeMeanTwoHundredProjAny & \nTdAbaciLargeMeanTwoHundredProjOverOne & \nTdAbaciLargeMeanTwoHundredCert \\
 & \nCountThousand & \nTdAbaciLargeMeanThousandRangeGm & \nTdAbaciLargeMeanThousandPropNoBoost & \nTdAbaciLargeMeanThousandProjShare & \nTdAbaciLargeMeanThousandProjAny & \nTdAbaciLargeMeanThousandProjOverOne & \nTdAbaciLargeMeanThousandCert \\
 & \nCountFiveThousand & \nTdAbaciLargeMeanFiveThousandRangeGm & \nTdAbaciLargeMeanFiveThousandPropNoBoost & \nTdAbaciLargeMeanFiveThousandProjShare & \nTdAbaciLargeMeanFiveThousandProjAny & \nTdAbaciLargeMeanFiveThousandProjOverOne & \nTdAbaciLargeMeanFiveThousandCert \\
\bottomrule
\end{tabular}
\par\raggedright ACI is adaptive conformal inference.
\end{table}

\section{Sensitivity checks and the mixing check}
\label{app:sensitivity}

We report four checks of restarted ACI and the restarted quantile here, of a misdated break, of the grid of shifts, of heavy tails with two breaks and of polynomial mixing.

\begin{figure}[H]
\centering
\includegraphics{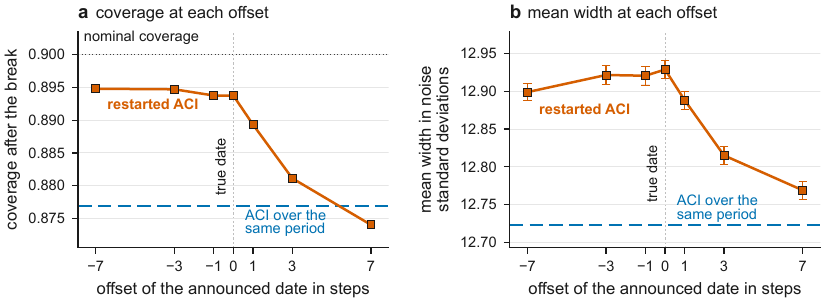}
\caption{A misdated break changes the coverage of restarted adaptive conformal inference (ACI) by \nBsCovRangePp percentage points. Against the offset of the announced date from the true date on the large mean shift, panel a plots coverage over the forecasts after the true break and panel b the mean width. The vertical line marks the true date.}
\label{fig:offsets}
\end{figure}

\paragraph{A misdated break.}
On the large mean shift, moving the boosted form's restart anywhere from \nBsOffsetMax steps early to \nBsOffsetMax steps late changes its coverage by \nBsCovRangePp percentage points. The check uses \nBsReps replications per offset at seeds from the scheme of Appendix~\ref{app:synthetic} with seed offset \nBsSalt, with ACI beside it, and measures the \nBsMpost forecasts after the true boundary. Early restarts keep coverage near \nBsAbCovMinusSeven, and a restart \nBsAbCovMinOffset steps late lowers it to \nBsAbCovPlusSeven. The mean width stays between \nBsAbWidthPlusSeven and \nBsAbWidthZero noise standard deviations, as Figure~\ref{fig:offsets} shows with one standard error each way.

\begin{figure}[H]
\centering
\includegraphics{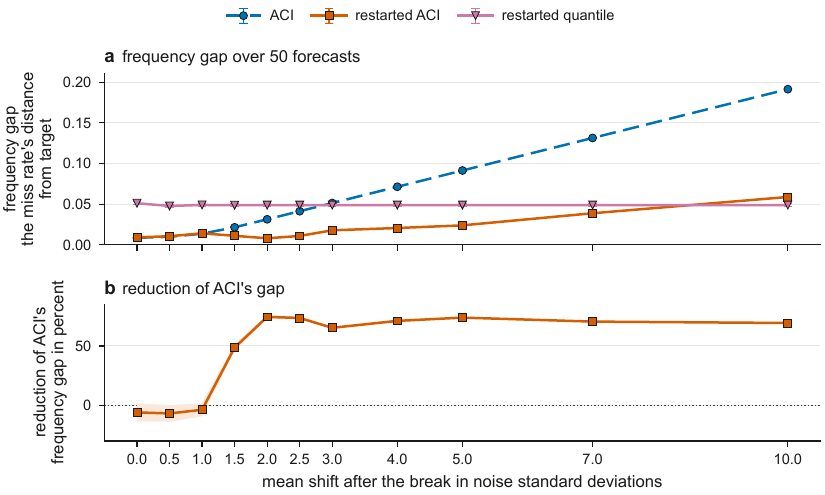}
\caption{Restarted adaptive conformal inference (ACI) cuts ACI's frequency gap most at a mean shift of \nDosePeakDelta. Against the mean shift, panel a plots the frequency gaps of both restarts and ACI over the first \nCountFifty forecasts, and panel b restarted ACI's relative reduction of ACI's gap, each with one standard error. Both vertical axes rest on the frequency gap.}
\label{fig:dosegrid}
\end{figure}

\paragraph{The grid of shifts.}
The grid runs the family at a noise scale of one after the break and at \nDoseGridPoints mean shifts from zero to \nDoseGridMaxDelta, over \nDoseTtotal steps and \nDoseReps replications at seed offset \nDoseSalt, and measures the frequency gaps over the first \nCountFifty forecasts. The relative reduction is one hundred times one less the ratio of restarted ACI's mean gap to ACI's, with a standard error from resampling the replications. It peaks at a mean shift of \nDosePeakDelta, at \nDosePeakRedAbaci{}\% with a standard error of \nDosePeakRedSe points. The gaps are drawn in Figure~\ref{fig:dosegrid}, and Section~\ref{sec:dose} sets them against the quantile shift.

\paragraph{Heavy tails and two breaks.}
Two further checks run \nExtReps replications each, one under Student noise with \nExtTailDof degrees of freedom and one with two announced breaks. On both, the restarted quantile and the recursions ACI and the burn-in form keep coverage after the break between \nExtRangeMin and \nExtRangeMax, while threshold SAOCP falls to \nExtHeavySaocpPost and \nExtTwoSaocpPost, as Table~\ref{tab:stress} reports. On the check with two breaks the restarted quantile covers \nExtTwoSegMultiPost when it restarts at both boundaries, as Algorithm~\ref{alg:segsplit} does, and \nExtTwoSegCpPost at the first alone.

The first check draws \nExtSteps steps of Student noise with \nExtTailDof degrees of freedom whose mean moves to \nExtPostMean and scale to \nExtPostScale at step \nExtTau. The second draws \nExtTwoSteps steps of unit Gaussian noise whose mean moves to \nExtTwoMeanOne at step \nExtTwoTauOne and to $\nExtTwoMeanTwo$ at step \nExtTwoTauTwo. Both calibrate on \nExtCalib scores and measure from the first boundary. ACI and the burn-in form step at \nExtAciGamma. The threshold rows are the threshold forms of Appendix~\ref{app:synthetic}, and the longest lifetime of threshold SAOCP is $2^{\nExtSaocpLifetimeLog}$. The restarted quantile takes two rows, one restarting at the first boundary alone and one at every boundary, which coincide on the check with one boundary.

\begin{table}[H]
\caption{Coverage after the break on two checks of \nExtReps replications, under Student noise and with two announced breaks. Each subscript is a standard error, the standard deviation over replications divided by the square root of their count.}
\label{tab:stress}
\centering
\footnotesize
\setlength{\tabcolsep}{8pt}
\begin{tabular}{lll}
\toprule
method & Student noise & two announced breaks \\
\midrule
naive & \nExtHeavyNaivePost\spread{\nExtHeavyNaiveSe} & \nExtTwoNaivePost\spread{\nExtTwoNaiveSe} \\
ACI & \nExtHeavyAciPost\spread{\nExtHeavyAciSe} & \nExtTwoAciPost\spread{\nExtTwoAciSe} \\
restarted ACI in its burn-in form & \nExtHeavyAbaciPost\spread{\nExtHeavyAbaciSe} & \nExtTwoAbaciPost\spread{\nExtTwoAbaciSe} \\
restarted quantile at the first boundary & \nExtHeavySegCpPost\spread{\nExtHeavySegCpSe} & \nExtTwoSegCpPost\spread{\nExtTwoSegCpSe} \\
restarted quantile at every boundary & \nExtHeavySegMultiPost\spread{\nExtHeavySegMultiSe} & \nExtTwoSegMultiPost\spread{\nExtTwoSegMultiSe} \\
threshold DtACI & \nExtHeavyDtaciPost\spread{\nExtHeavyDtaciSe} & \nExtTwoDtaciPost\spread{\nExtTwoDtaciSe} \\
threshold conformal PID & \nExtHeavyPidPost\spread{\nExtHeavyPidSe} & \nExtTwoPidPost\spread{\nExtTwoPidSe} \\
threshold SAOCP & \nExtHeavySaocpPost\spread{\nExtHeavySaocpSe} & \nExtTwoSaocpPost\spread{\nExtTwoSaocpSe} \\
\bottomrule
\end{tabular}
\end{table}

\paragraph{The mixing check.}
The mixing check runs the restarted quantile on an alternating renewal family with two states whose location switches between zero and \nPFiveMeanHigh. Each sojourn is a Pareto draw of scale one and index one plus the dependence index, rounded up to a whole step. The first state is zero or one with probability one half, and its first sojourn is an equilibrium residual, so the process is stationary from the start. Each score is the absolute value of the location plus a standard normal draw $Z$, and the reliability gap measures each threshold against the stationary score law, an equal mixture of the laws of $|Z|$ and $|Z+\nPFiveMeanHigh|$.

Each of \nPFiveIndices indices runs \nPFiveReps replications of \nPFivePost scores at seed offsets \nPFiveSaltFirst to \nPFiveSaltLast. Under the criterion we fixed before the run, the statement is not confirmed if the slope at index \nPFiveAHalf is steeper than $\nPFiveRuleHalf$ or the slope at index \nPFiveAOneHalf lies more than \nPFiveMargin from minus one half. The criterion compares each fitted slope with the borrowed exponent as if that upper rate were an equality. At index \nPFiveAHalf the slope over six counts is $\nPFiveSlopeSixHalf$ against the borrowed $\nPFiveIndexHalf$, so the criterion is not met. At index \nPFiveAOneHalf it is $\nPFiveSlopeSixOneHalf$, within \nPFiveMargin of minus one half. The state indicator's autocorrelation, measured on \nPFiveDepStreams streams of \nPFiveDepLength steps at \nPFiveLags lags up to \nPFiveLagMax, decays at $\nPFiveDepSlopeHalf$ at index \nPFiveAHalf against the targeted $\nPFiveDepPredHalf$, a measurement of correlation that bounds no mixing coefficient. Every index is in Table~\ref{tab:mixing}.

The exponent column gives the borrowed upper rate, minus the index over two below index one and minus one half from index one on, which the prespecified criterion treats as an equality. The first slope is over the six counts \nCountTen, \nCountTwentyFive, \nCountFifty, \nCountTwoHundred, \nCountThousand and \nCountFiveThousand and the second over the four from \nCountFifty on. The last two columns give the targeted decay of the state indicator's autocorrelation and the slope on logarithmic axes measured on \nPFiveDepStreams streams of \nPFiveDepLength steps.

\begin{table}[H]
\caption{The check under polynomial mixing on the family with two states, over \nPFiveReps replications of \nPFivePost scores since the restart per tail index. The two slopes are least squares fits of the log mean reliability gap on the log count of completed scores, each with its regression's standard error as a subscript.}
\label{tab:mixing}
\centering
\footnotesize
\setlength{\tabcolsep}{5pt}
\begin{tabular}{lllllll}
\toprule
index & \begin{tabular}[c]{@{}c@{}}sojourn\\tail index\end{tabular} & \begin{tabular}[c]{@{}c@{}}borrowed\\exponent\end{tabular} & \begin{tabular}[c]{@{}c@{}}slope over\\six counts\end{tabular} & \begin{tabular}[c]{@{}c@{}}slope over\\four counts\end{tabular} & \begin{tabular}[c]{@{}c@{}}targeted\\decay\end{tabular} & \begin{tabular}[c]{@{}c@{}}measured\\decay\end{tabular} \\
\midrule
\nPFiveAHalf & \nPFiveTailHalf & $\nPFiveIndexHalf$ & $\nPFiveSlopeSixHalf$\spread{\nPFiveSlopeSixSeHalf} & $\nPFiveSlopeFourHalf$\spread{\nPFiveSlopeFourSeHalf} & $\nPFiveDepPredHalf$ & $\nPFiveDepSlopeHalf$\spread{\nPFiveDepSlopeSeHalf} \\
\nPFiveAEightTenths & \nPFiveTailEightTenths & $\nPFiveIndexEightTenths$ & $\nPFiveSlopeSixEightTenths$\spread{\nPFiveSlopeSixSeEightTenths} & $\nPFiveSlopeFourEightTenths$\spread{\nPFiveSlopeFourSeEightTenths} & $\nPFiveDepPredEightTenths$ & $\nPFiveDepSlopeEightTenths$\spread{\nPFiveDepSlopeSeEightTenths} \\
\nPFiveAOne & \nPFiveTailOne & $\nPFiveIndexOne$ & $\nPFiveSlopeSixOne$\spread{\nPFiveSlopeSixSeOne} & $\nPFiveSlopeFourOne$\spread{\nPFiveSlopeFourSeOne} & $\nPFiveDepPredOne$ & $\nPFiveDepSlopeOne$\spread{\nPFiveDepSlopeSeOne} \\
\nPFiveAOneHalf & \nPFiveTailOneHalf & $\nPFiveIndexOneHalf$ & $\nPFiveSlopeSixOneHalf$\spread{\nPFiveSlopeSixSeOneHalf} & $\nPFiveSlopeFourOneHalf$\spread{\nPFiveSlopeFourSeOneHalf} & $\nPFiveDepPredOneHalf$ & $\nPFiveDepSlopeOneHalf$\spread{\nPFiveDepSlopeSeOneHalf} \\
\bottomrule
\end{tabular}
\end{table}

\section{The two real streams}
\label{app:ercot}

We describe the two real streams here and report every method on them. The first is the ERCOT window around its reform, and the second the stream of FOMC decisions. ACI is the comparator throughout. Bands on the real streams use the paired moving block bootstrap, which a paragraph below states.

\paragraph{Indices.}
The index of every boundary and reported result is in Table~\ref{tab:index-map}. On the ERCOT window, which counts days from \nErcotFirstDate, the reform's operating day of \nMapErcotFirstPostDate falls at index \nMapErcotFirstPost, so the boundary is at \nMapErcotLastPre and the \nOrdFifty forecast falls on \nMapErcotReadingDate. The FOMC stream's evaluation starts at index \nFomcEvalStart on \nMapFomcEvalStartDate, the day of a decision whose boundary is the last calibration index. That decision opens the partition's first cell, whose \nMapFomcPreCell evaluated days run to the first evaluated boundary, and the results after decisions use the \nFomcCellsTotal cells after it.

\begin{table}[H]
\caption{Index of every boundary and of every reported result, under the counting convention of Section~\ref{sec:setup}.}
\label{tab:index-map}
\centering
\scriptsize
\setlength{\tabcolsep}{3pt}
\begin{tabular}{>{\raggedright\arraybackslash}p{\mapcolA}>{\raggedright\arraybackslash}p{\mapcolB}>{\raggedright\arraybackslash}p{\mapcolB}>{\raggedright\arraybackslash}p{\mapcolC}c>{\raggedright\arraybackslash}p{\mapcolD}}
\toprule
stream & \begin{tabular}[c]{@{}c@{}}last outcome\\before the boundary\end{tabular} & \begin{tabular}[c]{@{}c@{}}first forecast\\after the boundary\end{tabular} & \begin{tabular}[c]{@{}c@{}}the reported\\forecast\end{tabular} & \begin{tabular}[c]{@{}c@{}}completed\\scores\end{tabular} & its reference \\
\midrule
synthetic family & stream index \nMapSynLastPre & stream index \nMapSynFirstPost & the \nOrdFifty, stream index \nMapSynReadingIdx & \nMapSynReadingScores & the known law after the break \\
ERCOT window & window index \nMapErcotLastPre, \nMapErcotLastPreDate & window index \nMapErcotFirstPost, \nMapErcotFirstPostDate & the \nOrdFifty, window index \nMapErcotReadingIdx, \nMapErcotReadingDate & \nMapErcotReadingScores & \nMapErcotReadingRef unseen scores \\
FOMC stream & the observation before each decision's first forecast, index \nMapFomcFirstLastPre, \nMapFomcFirstLastPreDate, at the first evaluated boundary & the first observation on or after each decision, index \nMapFomcFirstFirstPost, \nMapFomcFirstFirstPostDate, after the first evaluated boundary & the \nOrdTen of each of the \nMapFomcCells cells & \nMapFomcReadingScores & a median of \nFomcDayRefMedianTen unseen scores, the smallest \nFomcDayRefSmallestTen \\
\bottomrule
\end{tabular}
\end{table}

\paragraph{The ERCOT window.}
The series is the daily mean of the fifteen minute real-time settlement point price at ERCOT's bus average hub, in USD per MWh, by calendar date in Central time of each interval's start. Of the \nErcotStreamRows days in the daily summary, the window keeps the first \nErcotRows, from \nErcotFirstDate to \nErcotLastDate, all priced. The forecaster trains on the first \nErcotTrainDays days, indices zero to \nErcotTrainEndIdx, and calibrates on the next \nErcotCalib, indices \nErcotCalibStartIdx to \nErcotCalibEndIdx. Evaluation covers the remaining \nErcotEval days, from \nErcotEvalStart to \nErcotLastIdx, \nErcotPre days before the reform and \nErcotPost after it. The forecaster is a LightGBM regressor of \nRefitEstimators trees at learning rate \nRefitLr and column subsample \nRefitColsample, each tree with depth at most \nRefitDepth and at most \nRefitLeaves leaves of at least \nRefitMinChild rows. Six features describe recent prices, the prices \nLgbLagOne, \nLgbLagTwo, \nLgbLagThree and \nLgbLagFour days back and the mean and standard deviation of the \nLgbRolling prices ending on the row's day. Three describe the calendar, the weekday, the month and the day of the year. The model is fitted once on \nRefitPairs pairs, each matching the feature row of a training day from index \nRefitTrainFirst to \nRefitTrainLast with the next day's price, and a day's forecast is the model's output on the previous day's row. The first \nRefitFallback forecasts, inside the training fold, take the mean of earlier prices, with index zero taking its own price. A later day with a missing feature would take the mean of the last week's prices, though no evaluated day has one. The fit makes \nRefitSplits splits over its \nRefitTrees trees, and its \nRefitEvalDays evaluated forecasts take \nRefitDistinctEval distinct values from \nRefitEvalMin to \nRefitEvalMax USD per MWh. Every method's coverage after the reform and its difference from ACI are in Table~\ref{tab:ercot}. Appendix~\ref{app:corrections} gives each method's coverage under forecasters at LightGBM's default leaf size, which return one value, and on a shorter daily summary.

Restarted ACI runs its boosted form at step \nErcotGamma with a boost of \nErcotBoost over its first \nErcotBurnin updates after the reform. A negative difference covers less than ACI. The minimum detectable difference is the difference the window detects at \nFloorPowerPct{}\% power, with none marking one beyond the grid. The last two columns give the mean finite width in USD per MWh and the count of infinite widths.

\begin{table}[H]
\caption{Restarted ACI covers \nErcotCovAbaci of the \nErcotPost days after the ERCOT reform against ACI's \nErcotCovAci, a difference inside what the window can detect. Rows give every method's coverage under the forecaster of Appendix~\ref{app:ercot}, at a quantile shift of \nErcotDose. Each difference from ACI has its band as a subscript, over \nErcotDraws resamples in blocks of \nErcotBlock days, with its two-sided $p$ beside it.}
\label{tab:ercot}
\centering
\footnotesize
\setlength{\tabcolsep}{5pt}
\begin{tabular}{lllllll}
\toprule
method & coverage & \begin{tabular}[c]{@{}c@{}}difference\\from ACI\end{tabular} & $p$ & \begin{tabular}[c]{@{}c@{}}detectable\\difference\end{tabular} & \begin{tabular}[c]{@{}c@{}}finite\\width\end{tabular} & infinite \\
\midrule
ACI & \nErcotCovAci & comparator &  &  & \nErcotWidthAci & \nErcotInfAci \\
restarted ACI & \nErcotCovAbaci & $\nErcotDiffAbaci$\spread{$[\nErcotCiLoAbaci,\, \nErcotCiHiAbaci]$} & \nErcotPAbaci & \nFloorErcotAbaci & \nErcotWidthAbaci & \nErcotInfAbaci \\
restarted quantile & \nErcotCovSeg & \nErcotDiffSeg\spread{$[\nErcotCiLoSeg,\, \nErcotCiHiSeg]$} & \nErcotPSeg & \nFloorErcotSeg & \nErcotWidthSeg & \nErcotInfSeg \\
restarted quantile with warm start & \nErcotCovSegWarm & \nErcotDiffSegWarm\spread{$[\nErcotCiLoSegWarm,\, \nErcotCiHiSegWarm]$} & \nErcotPSegWarm & \nFloorErcotSegWarm & \nErcotWidthSegWarm & \nErcotInfSegWarm \\
naive & \nErcotCovNaive & $\nErcotDiffNaive$\spread{$[\nErcotCiLoNaive,\, \nErcotCiHiNaive]$} & \nErcotPNaive & \nFloorErcotNaive & \nErcotWidthNaive & \nErcotInfNaive \\
DtACI & \nErcotCovDtaciSrc & \nErcotDiffDtaciSrc\spread{$[\nErcotCiLoDtaciSrc,\, \nErcotCiHiDtaciSrc]$} & \nErcotPDtaciSrc & \nFloorErcotDtaciSrc & \nErcotWidthDtaciSrc & \nErcotInfDtaciSrc \\
SAOCP & \nErcotCovSaocpSrc & $\nErcotDiffSaocpSrc$\spread{$[\nErcotCiLoSaocpSrc,\, \nErcotCiHiSaocpSrc]$} & \nErcotPSaocpSrc & \nFloorErcotSaocpSrc & \nErcotWidthSaocpSrc & \nErcotInfSaocpSrc \\
conformal PID & \nErcotCovPidSrc & \nErcotDiffPidSrc\spread{$[\nErcotCiLoPidSrc,\, \nErcotCiHiPidSrc]$} & \nErcotPPidSrc & \nFloorErcotPidSrc & \nErcotWidthPidSrc & \nErcotInfPidSrc \\
AgACI & \nErcotCovAgaciSrc & \nErcotDiffAgaciSrc\spread{$[\nErcotCiLoAgaciSrc,\, \nErcotCiHiAgaciSrc]$} & \nErcotPAgaciSrc & \nFloorErcotAgaciSrc & \nErcotWidthAgaciSrc & \nErcotInfAgaciSrc \\
\bottomrule
\end{tabular}
\par\raggedright ACI is adaptive conformal inference. ERCOT is the Electric Reliability Council of Texas. DtACI is dynamically-tuned ACI. SAOCP is strongly adaptive online conformal prediction. Conformal PID is conformal proportional-integral-derivative control. AgACI is aggregated ACI.
\end{table}

\paragraph{The FOMC stream.}
The raw series has \nFomcRawRows daily rows of the ten-year constant maturity Treasury yield. Dropping the \nFomcMissingDays rows without a value leaves yields whose \nFomcDifferences first differences, in percentage points and dated at the later day, form the series. An autoregression of order \nFomcArOrder with an intercept is fitted once on the first \nFomcTrain differences and frozen, and its first forecasts, inside the training block, take the mean of the differences to date. Calibration takes indices \nFomcCalibStart to \nFomcCalibEnd, and evaluation covers the \nFomcEval days from index \nFomcEvalStart. A decision's first forecast falls on the first observation on or after its date, and its boundary is the observation before. Of the \nFomcDates decisions from \nFomcFirstYear to \nFomcLastYear, \nFomcDatesBefore have boundaries before evaluation and \nFomcBreaks inside it, which open the \nFomcCellsTotal cells of Table~\ref{tab:fomc-cells}, from \nFomcCellMin to \nFomcCellMax days long with a mean of \nFomcCellMean. Restarted ACI runs its burn-in form at step \nFomcGamma with a burn-in of \nBurnin scores, and Table~\ref{tab:fomc} gives every method's coverage and width. Appendix~\ref{app:corrections} gives the band under other resampling settings and score set conventions.

The partition's first cell, the \nFomcPreCellDays evaluated days from \nFomcPreCellFirst to the boundary at index \nMapFomcFirstLastPre, precedes the listed cells.

\begin{table}[H]
\caption{The \nFomcCellsTotal evaluated cells of the FOMC stream in order, read down each pair of columns. A cell runs from the first forecast after one boundary through the next boundary, so the first listed cell opens after the boundary at index \nMapFomcFirstLastPre, with its first forecast at index \nMapFomcFirstFirstPost.}
\label{tab:fomc-cells}
\centering
\scriptsize
\setlength{\tabcolsep}{8pt}
\begin{tabular}{cccccc}
\toprule
first day & days & first day & days & first day & days \\
\midrule
\nFomcCellOneFirst & \nFomcCellOneDays & \nFomcCellFourteenFirst & \nFomcCellFourteenDays & \nFomcCellTwentySevenFirst & \nFomcCellTwentySevenDays \\
\nFomcCellTwoFirst & \nFomcCellTwoDays & \nFomcCellFifteenFirst & \nFomcCellFifteenDays & \nFomcCellTwentyEightFirst & \nFomcCellTwentyEightDays \\
\nFomcCellThreeFirst & \nFomcCellThreeDays & \nFomcCellSixteenFirst & \nFomcCellSixteenDays & \nFomcCellTwentyNineFirst & \nFomcCellTwentyNineDays \\
\nFomcCellFourFirst & \nFomcCellFourDays & \nFomcCellSeventeenFirst & \nFomcCellSeventeenDays & \nFomcCellThirtyFirst & \nFomcCellThirtyDays \\
\nFomcCellFiveFirst & \nFomcCellFiveDays & \nFomcCellEighteenFirst & \nFomcCellEighteenDays & \nFomcCellThirtyOneFirst & \nFomcCellThirtyOneDays \\
\nFomcCellSixFirst & \nFomcCellSixDays & \nFomcCellNineteenFirst & \nFomcCellNineteenDays & \nFomcCellThirtyTwoFirst & \nFomcCellThirtyTwoDays \\
\nFomcCellSevenFirst & \nFomcCellSevenDays & \nFomcCellTwentyFirst & \nFomcCellTwentyDays & \nFomcCellThirtyThreeFirst & \nFomcCellThirtyThreeDays \\
\nFomcCellEightFirst & \nFomcCellEightDays & \nFomcCellTwentyOneFirst & \nFomcCellTwentyOneDays & \nFomcCellThirtyFourFirst & \nFomcCellThirtyFourDays \\
\nFomcCellNineFirst & \nFomcCellNineDays & \nFomcCellTwentyTwoFirst & \nFomcCellTwentyTwoDays & \nFomcCellThirtyFiveFirst & \nFomcCellThirtyFiveDays \\
\nFomcCellTenFirst & \nFomcCellTenDays & \nFomcCellTwentyThreeFirst & \nFomcCellTwentyThreeDays & \nFomcCellThirtySixFirst & \nFomcCellThirtySixDays \\
\nFomcCellElevenFirst & \nFomcCellElevenDays & \nFomcCellTwentyFourFirst & \nFomcCellTwentyFourDays & \nFomcCellThirtySevenFirst & \nFomcCellThirtySevenDays \\
\nFomcCellTwelveFirst & \nFomcCellTwelveDays & \nFomcCellTwentyFiveFirst & \nFomcCellTwentyFiveDays & \nFomcCellThirtyEightFirst & \nFomcCellThirtyEightDays \\
\nFomcCellThirteenFirst & \nFomcCellThirteenDays & \nFomcCellTwentySixFirst & \nFomcCellTwentySixDays &  &  \\
\bottomrule
\end{tabular}
\end{table}

A negative difference covers less than ACI, and the minimum detectable difference is the difference the stream detects at \nFloorPowerPct{}\% power. Mean widths are in percentage points of yield, with none where a width is infinite, and the last column counts infinite widths. The restarted quantile takes three rows, restarting at the first boundary alone, at every boundary, and at every boundary with the warm start of Remark~\ref{rem:warm}. Restarted ACI runs its burn-in form at step \nFomcGamma with a burn-in of \nBurnin scores, no boost and a floor of \nFloorFormC.

\begin{table}[H]
\caption{The burn-in form of restarted ACI covers \nFomcCovAbaci on the FOMC stream against ACI's \nFomcCovAci. Rows give every method's coverage over the stream's \nFomcEval evaluation days and \nFomcBreaks evaluated boundaries, at a median cell quantile shift of \nFomcDose. Each difference from ACI has its band as a subscript, over \nFomcDraws resamples in blocks of \nFomcBlock days.}
\label{tab:fomc}
\centering
\footnotesize
\setlength{\tabcolsep}{5pt}
\begin{tabular}{llllll}
\toprule
method & coverage & \begin{tabular}[c]{@{}c@{}}difference\\from ACI\end{tabular} & \begin{tabular}[c]{@{}c@{}}detectable\\difference\end{tabular} & mean width & infinite \\
\midrule
naive & \nFomcCovNaive & $\nFomcDiffNaive$\spread{$[\nFomcCiLoNaive,\, \nFomcCiHiNaive]$} & \nFloorFomcNaive & \nFomcWidthNaive & \nFomcInfNaive \\
ACI & \nFomcCovAci & comparator &  & \nFomcWidthAci & \nFomcInfAci \\
restarted ACI in its burn-in form & \nFomcCovAbaci & $\nFomcDiffAbaci$\spread{$[\nFomcCiLoAbaci,\, \nFomcCiHiAbaci]$} & \nFloorFomcAbaci & \nFomcWidthAbaci & \nFomcInfAbaci \\
DtACI & \nFomcCovDtaciSrc & \nFomcDiffDtaciSrc\spread{$[\nFomcCiLoDtaciSrc,\, \nFomcCiHiDtaciSrc]$} & \nFloorFomcDtaciSrc & \nFomcWidthDtaciSrc & \nFomcInfDtaciSrc \\
SAOCP & \nFomcCovSaocpSrc & \nFomcDiffSaocpSrc\spread{$[\nFomcCiLoSaocpSrc,\, \nFomcCiHiSaocpSrc]$} & \nFloorFomcSaocpSrc & \nFomcWidthSaocpSrc & \nFomcInfSaocpSrc \\
conformal PID & \nFomcCovPidSrc & \nFomcDiffPidSrc\spread{$[\nFomcCiLoPidSrc,\, \nFomcCiHiPidSrc]$} & \nFloorFomcPidSrc & \nFomcWidthPidSrc & \nFomcInfPidSrc \\
AgACI & \nFomcCovAgaciSrc & \nFomcDiffAgaciSrc\spread{$[\nFomcCiLoAgaciSrc,\, \nFomcCiHiAgaciSrc]$} & \nFloorFomcAgaciSrc & \nFomcWidthAgaciSrc & \nFomcInfAgaciSrc \\
restarted quantile at the first boundary & \nFomcCovSegCp & $\nFomcDiffSegCp$\spread{$[\nFomcCiLoSegCp,\, \nFomcCiHiSegCp]$} & \nFloorFomcSegCp & \nFomcWidthSegCp & \nFomcInfSegCp \\
restarted quantile at every boundary & \nFomcCovSegMulti & $\nFomcDiffSegMulti$\spread{$[\nFomcCiLoSegMulti,\, \nFomcCiHiSegMulti]$} & \nFloorFomcSegMulti & \nFomcWidthSegMulti & \nFomcInfSegMulti \\
warm start at every boundary & \nFomcCovSegMultiWarm & $\nFomcDiffSegMultiWarm$\spread{$[\nFomcCiLoSegMultiWarm,\, \nFomcCiHiSegMultiWarm]$} & \nFloorFomcSegMultiWarm & \nFomcWidthSegMultiWarm & \nFomcInfSegMultiWarm \\
\bottomrule
\end{tabular}
\end{table}

\paragraph{The bootstrap and the minimum detectable difference.}
For two methods on one stream, the paired moving block bootstrap resamples the day series of the difference of their coverage indicators. Each resample concatenates blocks of $\ell$ consecutive days until the series is covered and keeps the first $n$ days, with each block starting at a day drawn uniformly from those where a whole block fits. The block length is \nErcotBlock on the ERCOT window and \nFomcBlock on the FOMC stream, and each pair draws $N=\nErcotDraws$ resamples from its own generator at seed \nCfSeed. The band is the \nBandLoPct to \nBandHiPct percentile of the resampled mean differences. With $k$ the smaller of the counts of resampled means at or below zero and at or above it, the two-sided $p$ is the smaller of one and $2(1+k)/(N+1)$. It prints to two decimals from \nPTwoDecimalsFrom up and to four below, so the smallest printable $p$ is \nPSmallest.

For a window's minimum detectable difference we centre the difference series on its mean and draw two sets of \nErcotFloorDraws moving block resampled means. The critical value is the \nFloorCritPct percentile of the absolute values in the first set, a two-sided test at the \nFloorLevelPct{}\% level. The power at a shift is the share of the second set whose absolute value, after adding the shift, exceeds the critical value, on a grid of \nFloorGridPoints shifts, \nFloorGridList. The minimum detectable difference is the smallest shift with power of \nFloorPowerPct{}\%, interpolated linearly between the grid shifts that bracket it.

A reliability band on a stream with one boundary resamples the reference scores by moving blocks with every threshold held fixed, and on the FOMC stream it resamples the contributing cells with replacement. Either band is the spread of the gap over resamples of a finite reference, the uncertainty of the holdout reference. On the ERCOT window it shows how the gap moves with the draw of one reference, and on the FOMC stream how it moves across decisions.

\paragraph{Other references.}
The in-sample reference measures each forecast's threshold against every score of its cell after the break, including the scores the threshold was built from. On the ERCOT window it gives ACI a cell mean of \nPtErcotCellAci and the restarted quantile \nPtErcotCellSeg, where the holdout reference gives \nCfErcotCellAci and \nCfErcotCellSeg over the first \nCfErcotCellForecasts forecasts and \nCfErcotCellAllAci and \nCfErcotCellAllSeg over all \nCfErcotCellLength. On the FOMC stream, across the first \nProfReadable days after a decision, the pointwise band on the holdout reference lies above zero against all five baselines on \nProfAboveAll days, \nProfAboveAllDays, and below zero against any of them on \nProfBelowAnyWord.

At the \nOrdTwentyFive day \nFomcDayCellsTwentyFive cells contribute, and their references have a median of \nFomcDayRefMedianTwentyFive scores. The restarted quantile's gap on the holdout reference averages \nCfFomcSegMultiTwentyFive there. Its band lies above zero against DtACI and conformal PID, at \nCfFomcEndsDtaciSrcTwentyFive and \nCfFomcEndsPidSrcTwentyFive, and against SAOCP at \nCfFomcEndsSaocpSrcTwentyFive, and it includes zero against \nCfFomcSideIncludeWhoTwentyFive, at \nCfFomcSideIncludeEndsTwentyFive. The in-sample band at that day lies below zero against \nIsFomcSideBelowWhoTwentyFive, at \nIsFomcSideBelowEndsTwentyFive. It includes zero against AgACI and DtACI, at \nIsFomcEndsAgaciSrcTwentyFive and \nIsFomcEndsDtaciSrcTwentyFive, and against conformal PID at \nIsFomcEndsPidSrcTwentyFive. The advantage the in-sample reference shows on the decision calendar thus disappears under the holdout reference. The pooled reference, every score after a decision from the first boundary on, gives the restarted quantile \nPtFomcPooledSegMultiTen at the \nOrdTen day and \nPtFomcPooledSegMultiTwentyFive at the \nOrdTwentyFive, against \nPtFomcPooledAciTen and \nPtFomcPooledAciTwentyFive for ACI.

\paragraph{The terms of the bound.}
The terms of Proposition~\ref{prop:practical-corrections} on both streams are in Table~\ref{tab:terms-real}, as means over the cells on the FOMC stream, where the check of Lemma~1 of \citet{angelopoulos2024decaying}, that every threshold stayed within its bound, holds on all \nFomcCertHolds cells.

On the ERCOT window, under the forecaster of this appendix, restarted ACI runs at step \nErcotGamma in its boosted form and in its boosted burn-in form, whose jump comes after \nBurnin scores since the break, with jump and total floor correction in USD per MWh. On the FOMC stream it runs its burn-in form at step \nFomcGamma with no boost, and every value is a mean over the \nFomcCellsTotal cells, in percentage points of yield. The boost term is divided by the step times the count. On the FOMC stream the ratio divides the bound's mean over cells by the error's. The last column reports the check of Lemma~1 of \citet{angelopoulos2024decaying}, an accounting check for each period.

\begin{table}[H]
\caption{Terms of Proposition~\ref{prop:practical-corrections} on the two real streams. The displacement bound is the threshold's displacement over the period plus the jump and the total floor correction, over the step times the count, printed without its boost term, and the telescoping identity puts the complete bound at or above the frequency gap.}
\label{tab:terms-real}
\centering
\scriptsize
\setlength{\tabcolsep}{2.2pt}
\begin{tabular}{>{\raggedright\arraybackslash}p{\termcolA}c c c c c c c c>{\raggedright\arraybackslash}p{\termcolB}}
\toprule
run & steps & jump & \begin{tabular}[c]{@{}c@{}}total\\floor\\correction\end{tabular} & \begin{tabular}[c]{@{}c@{}}projection\\share\end{tabular} & boost & \begin{tabular}[c]{@{}c@{}}frequency\\gap\end{tabular} & \begin{tabular}[c]{@{}c@{}}displacement bound\\without boost\end{tabular} & \begin{tabular}[c]{@{}c@{}}bound over\\frequency gap\end{tabular} & \begin{tabular}[c]{@{}c@{}}threshold\\within\\its bound\end{tabular} \\
\midrule
the boosted form on ERCOT & \nErcotPost & \nErcotJumpAbaci & \nErcotPiSumAbaci & \nErcotProjShareAbaci & \nErcotBoostExcessAbaci & \nErcotFreqErrAbaci & \nErcotBoundAbaci & \nErcotBoundRatioAbaci & holds \\
the boosted burn-in form on ERCOT & \nErcotPost & \nErcotJumpAbaciText & \nErcotPiSumAbaciText & \nErcotProjShareAbaciText & \nErcotBoostExcessAbaciText & \nErcotFreqErrAbaciText & \nErcotBoundAbaciText & \nErcotBoundRatioAbaciText & holds \\
the burn-in form over the FOMC cells & \nFomcCellMeanInt & \nFomcJumpMean & \nFomcPiSumMean & \nFomcProjShareMean & \nFomcBoostMean & \nFomcFreqErrMean & \nFomcBoundMean & \nFomcBoundRatio & holds on all \nFomcCertHolds \\
\bottomrule
\end{tabular}
\par\raggedright ERCOT is the Electric Reliability Council of Texas. FOMC is the Federal Open Market Committee.
\end{table}

\begin{figure}[H]
\centering
\includegraphics{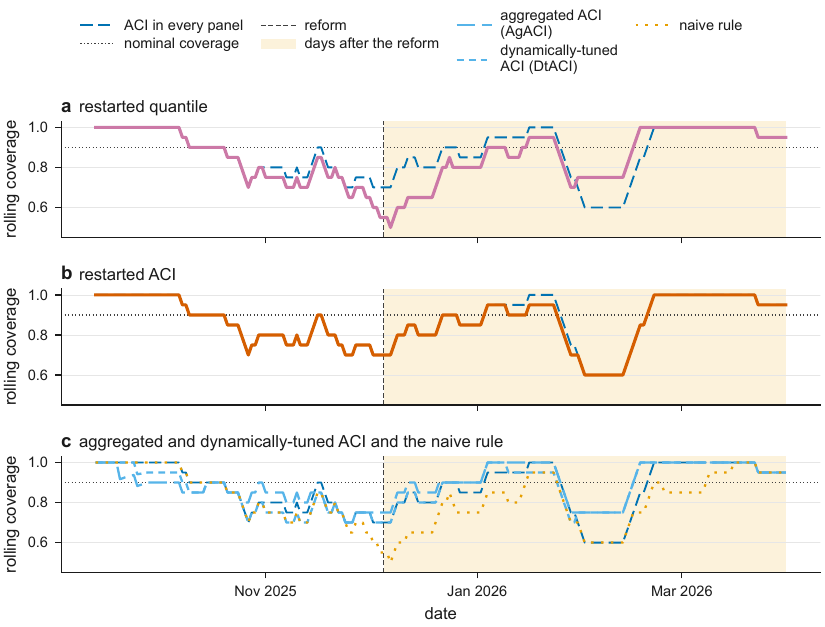}
\caption{Every method's rolling coverage falls below nominal coverage after the reform of the Electric Reliability Council of Texas (ERCOT) and ends at or above it. Lines plot coverage over a rolling span of days against the date, and the dashed line marks the reform. Each panel draws adaptive conformal inference (ACI) as the common reference.}
\label{fig:ercot_rolling}
\end{figure}

\begin{figure}[H]
\centering
\includegraphics{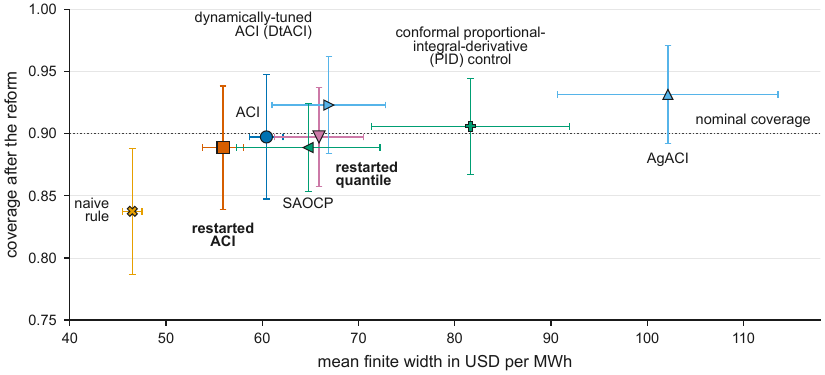}
\caption{The four variants of adaptive conformal inference (ACI) have wider mean intervals than ACI's \nFrontErcotWidthAci USD per MWh after the Texas market reform, from \nFrontErcotWidthSaocpSrc for strongly adaptive online conformal prediction (SAOCP) to \nFrontErcotWidthAgaciSrc for aggregated ACI (AgACI). Markers plot coverage against mean finite width, with one standard error each way, and each mark carries its method's name.}
\label{fig:ercot_frontier}
\end{figure}

Figures~\ref{fig:ercot_rolling} and~\ref{fig:ercot_frontier} show rolling coverage on the ERCOT window and coverage after the reform against mean finite width, under the forecaster of this appendix. The first computes coverage over a rolling span of \nRollingWindow days and draws a span once it holds \nRollingMinPeriods days. The second plots each method over the \nErcotPost days after the reform, and the restarted quantile's mean finite width there leaves out its \nErcotInfSegWord infinite widths. In the second, each error bar is a moving block standard error at block length \nErcotBlock over \nErcotDraws draws, computed for coverage from the day series of coverage indicators and for width from the finite widths in day order. The warm start covers as the restarted quantile does at the finite width Table~\ref{tab:ercot} prints, and its marker would coincide with the restarted quantile's at print size, so the figure omits it.

The same protocol applies to any break an operator announces before its first affected price, among them the California operator's extended day-ahead market, whose series joins the announced breaks once it spans a window after the break of the length Appendix~\ref{app:library} uses.

\section{The announced breaks}
\label{app:library}

We took every candidate break's effective date from the operator's or regulator's own document before reading any price. The breaks fall in the markets of ERCOT, Australia's NEM and PJM Interconnection, and at the FOMC, and ACI is the comparator throughout. The candidate search requested \nCandDocuments addresses, of which \nCandReturned returned a document and \nCandUnreachable none, giving \nCandReturnedDistinct distinct documents. The search checked \nCandAdmissionRows candidate breaks against the protocol's inclusion rules. The protocol included \nCandEarlier of them, the ERCOT offer cap, the ERCOT co-optimization, one FOMC decision, the NEM five minute settlement and the expiry of the gas transit contract between Russia and Ukraine. The search included \nLibNewCases more, whose documents dated their boundaries and whose series spanned the windows, the two shifts of ERCOT's reserve curve, the NEM administered price cap, the NEM market price cap and PJM Interconnection's reserve price formation. It dropped a California offer cap change whose price series no public service returned. The experiments run \nCandRun of the \nCandAdmitted included breaks. They leave out the gas transit expiry of \nRemovedFirstDate, because that series, the daily close of the Dutch gas futures, came from Yahoo Finance, whose terms forbid the automated collection that gathered it. Every exhibit and count in the paper uses the other \nLibCases breaks.

For each break in a power market we average the publisher's interval prices by day, in USD per MWh on ERCOT and on PJM Interconnection's market and in Australian dollars per MWh on the NEM, and train the forecaster of the ERCOT window on the window's first days. ACI and restarted ACI step at \nLibGammaLevel in each break's price unit. The break at the central bank uses the FOMC stream's differences around one decision with its own autoregression, at step \nLibGammaFomc. A break's quantile shift compares the $(1-\alpha)$-quantiles of its scores after the boundary and of its evaluated scores before it, over the standard deviation of the evaluated scores before it with one degree of freedom removed. Breaks in power markets resample in blocks of \nLibBlockLevel days and the break at the central bank in blocks of \nLibBlockFomc days, each with \nLibDraws draws at seed \nLibSeed and a minimum detectable difference from \nLibFloorDraws draws on a grid of \nLibFloorGridPoints shifts. Every break's series and windows are in Table~\ref{tab:provenance}.

The break at the central bank fits its autoregression on its first differences and leaves the next ones unread, then calibrates on the rest before its evaluation. PJM Interconnection operates the market of the reserve price formation.

\begin{table}[H]
\caption{Each announced break of Table~\ref{tab:library} with its series and day and its windows, each window running from its first day to its last. For each break in a power market, the forecaster is trained on the window's first days, and the methods calibrate on as many more. Evaluation starts on the day after.}
\label{tab:provenance}
\centering
\scriptsize
\setlength{\tabcolsep}{2.5pt}
\begin{tabular}{l>{\raggedright\arraybackslash}p{\provcolA}>{\raggedright\arraybackslash}p{\provcolB}>{\raggedright\arraybackslash}p{\provcolC}>{\raggedright\arraybackslash}p{\provcolC}>{\raggedright\arraybackslash}p{\provcolC}c}
\toprule
label & \begin{tabular}[c]{@{}c@{}}announced\\break\end{tabular} & series and day & training & calibration & evaluation & \begin{tabular}[c]{@{}c@{}}first day\\after the\\break\end{tabular} \\
\midrule
i & ERCOT offer cap & ERCOT's North hub real-time price in USD per MWh, every interval averaged over the calendar day of its start in the market's own time & \nProvErcotCapTrainFirst to \nProvErcotCapTrainLast & \nProvErcotCapCalibFirst to \nProvErcotCapCalibLast & \nProvErcotCapEvalFirst to \nProvErcotCapEvalLast & \nProvErcotCapFirstPost \\
ii & first ERCOT reserve curve shift & ERCOT's North hub real-time price in USD per MWh, every interval averaged over the calendar day of its start in the market's own time & \nProvOrdcOneTrainFirst to \nProvOrdcOneTrainLast & \nProvOrdcOneCalibFirst to \nProvOrdcOneCalibLast & \nProvOrdcOneEvalFirst to \nProvOrdcOneEvalLast & \nProvOrdcOneFirstPost \\
iii & second ERCOT reserve curve shift & ERCOT's North hub real-time price in USD per MWh, every interval averaged over the calendar day of its start in the market's own time & \nProvOrdcTwoTrainFirst to \nProvOrdcTwoTrainLast & \nProvOrdcTwoCalibFirst to \nProvOrdcTwoCalibLast & \nProvOrdcTwoEvalFirst to \nProvOrdcTwoEvalLast & \nProvOrdcTwoFirstPost \\
iv & ERCOT real-time co-optimization & the ERCOT window's own daily series of Appendix~\ref{app:ercot} & \nProvErcotRtcbTrainFirst to \nProvErcotRtcbTrainLast & \nProvErcotRtcbCalibFirst to \nProvErcotRtcbCalibLast & \nProvErcotRtcbEvalFirst to \nProvErcotRtcbEvalLast & \nProvErcotRtcbFirstPost \\
v & one FOMC decision & the ten-year Treasury yield's first differences, each dated at its later day & \nProvFomcSingleTrainFirst to \nProvFomcSingleTrainLast, then \nProvFomcSingleUnreadFirst to \nProvFomcSingleUnreadLast unread & \nProvFomcSingleCalibFirst to \nProvFomcSingleCalibLast & \nProvFomcSingleEvalFirst to \nProvFomcSingleEvalLast & \nProvFomcSingleFirstPost \\
vi & NEM five minute settlement & the New South Wales region's trading price in Australian dollars per MWh, every interval averaged over the calendar day of its settlement timestamp & \nProvNemFiveTrainFirst to \nProvNemFiveTrainLast & \nProvNemFiveCalibFirst to \nProvNemFiveCalibLast & \nProvNemFiveEvalFirst to \nProvNemFiveEvalLast & \nProvNemFiveFirstPost \\
vii & NEM administered price cap & the New South Wales region's trading price in Australian dollars per MWh, every interval averaged over the calendar day of its settlement timestamp & \nProvNemApcTrainFirst to \nProvNemApcTrainLast & \nProvNemApcCalibFirst to \nProvNemApcCalibLast & \nProvNemApcEvalFirst to \nProvNemApcEvalLast & \nProvNemApcFirstPost \\
viii & NEM market price cap & the New South Wales region's trading price in Australian dollars per MWh, every interval averaged over the calendar day of its settlement timestamp & \nProvNemMpcTrainFirst to \nProvNemMpcTrainLast & \nProvNemMpcCalibFirst to \nProvNemMpcCalibLast & \nProvNemMpcEvalFirst to \nProvNemMpcEvalLast & \nProvNemMpcFirstPost \\
ix & PJM reserve price formation & PJM's day-ahead price for its whole market in USD per MWh, every hour averaged over the calendar day of its start in the market's own time & \nProvPjmRpfTrainFirst to \nProvPjmRpfTrainLast & \nProvPjmRpfCalibFirst to \nProvPjmRpfCalibLast & \nProvPjmRpfEvalFirst to \nProvPjmRpfEvalLast & \nProvPjmRpfFirstPost \\
\bottomrule
\end{tabular}
\par\raggedright ERCOT is the Electric Reliability Council of Texas. FOMC is the Federal Open Market Committee. NEM is Australia's National Electricity Market. PJM is PJM Interconnection.
\end{table}

Each of the \nAnnLibAccepted announced breaks comes with the document that announced it, dated before its first affected day, and Table~\ref{tab:documents} prints each document's issuer, date, first affected day, announcing passage and address.

\begin{table}[H]
\caption{The document that announced each announced break of Table~\ref{tab:library}, with its issuer, title, own date, first affected day, announcing passage as the page prints it and address. Every page was read on \nAnnReadDate. The cells abbreviate the Australian Energy Market Commission (AEMC), the low and the high system-wide offer caps of ERCOT's rule (LCAP and HCAP) and the Texas Administrative Code (TAC), and \nAnnOrdcRevision is the ERCOT revision request that shifted the reserve curve.}
\label{tab:documents}
\centering
\scriptsize
\setlength{\tabcolsep}{2.5pt}
\begin{tabular}{l>{\raggedright\arraybackslash}p{\anncolA}>{\raggedright\arraybackslash}p{\anncolB}>{\raggedright\arraybackslash}p{\anncolC}c>{\raggedright\arraybackslash}p{\anncolD}}
\toprule
label & \begin{tabular}[c]{@{}c@{}}announced\\break\end{tabular} & the document that announced it & its own date & \begin{tabular}[c]{@{}c@{}}first day\\it reaches\end{tabular} & the passage that announces it \\
\midrule
i & ERCOT offer cap & \nAnnErcotCapIssuer, \emph{\nAnnErcotCapTitle}, \nAnnErcotCapAddress & \nAnnErcotCapOwnDate, \nAnnErcotCapOwnDateKind & \nAnnErcotCapFirstAffected & ``\nAnnErcotCapPassage'' \\
ii & first ERCOT reserve curve shift & \nAnnOrdcOneIssuer, \emph{\nAnnOrdcOneTitle}, \nAnnOrdcOneAddress & \nAnnOrdcOneOwnDate, \nAnnOrdcOneOwnDateKind & \nAnnOrdcOneFirstAffected & ``\nAnnOrdcOnePassage'' \\
iii & second ERCOT reserve curve shift & \nAnnOrdcTwoIssuer, \emph{\nAnnOrdcTwoTitle}, \nAnnOrdcTwoAddress & \nAnnOrdcTwoOwnDate, \nAnnOrdcTwoOwnDateKind & \nAnnOrdcTwoFirstAffected & ``\nAnnOrdcTwoPassage'' \\
iv & ERCOT real-time co-optimization & \nAnnErcotRtcbIssuer, \emph{\nAnnErcotRtcbTitle}, \nAnnErcotRtcbAddress & \nAnnErcotRtcbOwnDate, \nAnnErcotRtcbOwnDateKind & \nAnnErcotRtcbFirstAffected & ``\nAnnErcotRtcbPassage'' \\
v & one FOMC decision & \nAnnFomcSingleIssuer, \emph{\nAnnFomcSingleTitle}, \nAnnFomcSingleAddress & \nAnnFomcSingleOwnDate, \nAnnFomcSingleOwnDateKind & \nAnnFomcSingleFirstAffected & ``\nAnnFomcSinglePassage'' \\
vi & NEM five minute settlement & \nAnnNemFiveIssuer, \emph{\nAnnNemFiveTitle}, \nAnnNemFiveAddress & \nAnnNemFiveOwnDate, \nAnnNemFiveOwnDateKind & \nAnnNemFiveFirstAffected & ``\nAnnNemFivePassage'' \\
vii & NEM administered price cap & \nAnnNemApcIssuer, \emph{\nAnnNemApcTitle}, \nAnnNemApcAddress & \nAnnNemApcOwnDate, \nAnnNemApcOwnDateKind & \nAnnNemApcFirstAffected & ``\nAnnNemApcPassage'' \\
viii & NEM market price cap & \nAnnNemMpcIssuer, \emph{\nAnnNemMpcTitle}, \nAnnNemMpcAddress & \nAnnNemMpcOwnDate, \nAnnNemMpcOwnDateKind & \nAnnNemMpcFirstAffected & ``\nAnnNemMpcPassage'' \\
ix & PJM reserve price formation & \nAnnPjmRpfIssuer, \emph{\nAnnPjmRpfTitle}, \nAnnPjmRpfAddress & \nAnnPjmRpfOwnDate, \nAnnPjmRpfOwnDateKind & \nAnnPjmRpfFirstAffected & ``\nAnnPjmRpfPassage'' \\
\bottomrule
\end{tabular}
\par\raggedright ERCOT is the Electric Reliability Council of Texas. FOMC is the Federal Open Market Committee. NEM is Australia's National Electricity Market. PJM is PJM Interconnection.
\end{table}

The FOMC decisions come with the Board's schedule of meetings for each year, which the Internet Archive captured before the year began, in captures from \nAnnFomcCaptureOne to \nAnnFomcCaptureFive. Each of the stream's \nAnnFomcEvaluatedAccepted evaluated decisions appears on the schedule of its year, and every capture predates the year's first decision. The Board's page states ``\nAnnFomcTentative,'' so each date was announced as a plan that the preceding meeting confirmed. Table~\ref{tab:schedules} prints each year's schedule with its capture.

Each row gives the capture's date and address, the page's date of last update, the year's first decision and the schedule as printed. Every page was read on \nAnnReadDate.

\begin{table}[H]
\caption{The Federal Reserve Board's schedule of meetings for each year of the FOMC stream, as the Internet Archive captured its calendars page before the year began. Each page's last update precedes its year's first decision, and an asterisk marks a meeting with a Summary of Economic Projections.}
\label{tab:schedules}
\centering
\scriptsize
\setlength{\tabcolsep}{3pt}
\begin{tabular}{l>{\raggedright\arraybackslash}p{\schcolA}cc>{\raggedright\arraybackslash}p{\schcolB}}
\toprule
year & the capture and its address & \begin{tabular}[c]{@{}c@{}}the page's\\last update\end{tabular} & \begin{tabular}[c]{@{}c@{}}first\\decision\end{tabular} & the schedule as the page prints it \\
\midrule
\nAnnFomcYearOne & \nAnnFomcCaptureOne, \nAnnFomcAddressOne & \nAnnFomcLastUpdateOne & \nAnnFomcFirstDecisionOne & ``\nAnnFomcPassageOne'' \\
\nAnnFomcYearTwo & \nAnnFomcCaptureTwo, \nAnnFomcAddressTwo & \nAnnFomcLastUpdateTwo & \nAnnFomcFirstDecisionTwo & ``\nAnnFomcPassageTwo'' \\
\nAnnFomcYearThree & \nAnnFomcCaptureThree, \nAnnFomcAddressThree & \nAnnFomcLastUpdateThree & \nAnnFomcFirstDecisionThree & ``\nAnnFomcPassageThree'' \\
\nAnnFomcYearFour & \nAnnFomcCaptureFour, \nAnnFomcAddressFour & \nAnnFomcLastUpdateFour & \nAnnFomcFirstDecisionFour & ``\nAnnFomcPassageFour'' \\
\nAnnFomcYearFive & \nAnnFomcCaptureFive, \nAnnFomcAddressFive & \nAnnFomcLastUpdateFive & \nAnnFomcFirstDecisionFive & ``\nAnnFomcPassageFive'' \\
\bottomrule
\end{tabular}
\par\raggedright FOMC is the Federal Open Market Committee.
\end{table}

Each break's quantile shift is in Table~\ref{tab:library}, with whether it has a gain its window can detect, and \nLibEnteredWord break does.

The frozen quantile is the threshold frozen at the break. Its frequency gap over the days after the break is the distance from nominal coverage of the share of those days' scores at or below it, printed with the count of such scores beside the count of all of them. The minimum detectable difference is the coverage difference of restarted ACI against ACI that the window detects at \nFloorPowerPct{}\% power, and the gain is the synthetic curve's at the quantile shift, taken at shift zero below the grid. A break has a gain its window can detect as Section~\ref{sec:dose} states, and \nLibEntered of \nLibCases does.

\begin{table}[H]
\caption{Of the \nLibCases announced breaks, \nLibEnteredWord has a gain its window can detect. Rows give each break with its quantile shift, its minimum detectable difference, the synthetic curve's gain at its shift and the label of its marker in Figure~\ref{fig:dose}. Each shift at which gain meets detectable difference has as a subscript the shifts at which the gain plus and less one standard error meets it.}
\label{tab:library}
\centering
\scriptsize
\setlength{\tabcolsep}{2pt}
\begin{tabular}{l>{\raggedright\arraybackslash}p{\libcolA}lcccccccl}
\toprule
label & \begin{tabular}[c]{@{}c@{}}announced\\break\end{tabular} & date & \begin{tabular}[c]{@{}c@{}}quantile\\shift\end{tabular} & \begin{tabular}[c]{@{}c@{}}frozen\\quantile's\\frequency\\gap after\\the break\end{tabular} & \begin{tabular}[c]{@{}c@{}}scores at\\or below\end{tabular} & \begin{tabular}[c]{@{}c@{}}detectable\\difference\end{tabular} & \begin{tabular}[c]{@{}c@{}}gain at\\the shift\end{tabular} & \begin{tabular}[c]{@{}c@{}}shift where\\the gain\\meets it\end{tabular} & \begin{tabular}[c]{@{}c@{}}days after\\the break\end{tabular} & \begin{tabular}[c]{@{}c@{}}gain its\\window can\\detect\end{tabular} \\
\midrule
i & ERCOT offer cap & \nLibErcotCapDateIso & $\nLibErcotCapDose$ & \nLibErcotCapBstale & \nLibErcotCapStaleBelow of \nLibErcotCapStaleScores & \nLibErcotCapFloor & $\nLibErcotCapGain$ & \nLibErcotCapDoseEq\spread{$[\nLibErcotCapDoseEqLo,\, \nLibErcotCapDoseEqHi]$} & \nLibErcotCapPost & \nLibErcotCapRule \\
ii & first ERCOT reserve curve shift & \nLibOrdcOneDateIso & $\nLibOrdcOneDose$ & \nLibOrdcOneBstale & \nLibOrdcOneStaleBelow of \nLibOrdcOneStaleScores & \nLibOrdcOneFloor & $\nLibOrdcOneGain$ & \nLibOrdcOneDoseEq\spread{$[\nLibOrdcOneDoseEqLo,\, \nLibOrdcOneDoseEqHi]$} & \nLibOrdcOnePost & \nLibOrdcOneRule \\
iii & second ERCOT reserve curve shift & \nLibOrdcTwoDateIso & $\nLibOrdcTwoDose$ & \nLibOrdcTwoBstale & \nLibOrdcTwoStaleBelow of \nLibOrdcTwoStaleScores & \nLibOrdcTwoFloor & $\nLibOrdcTwoGain$ & \nLibOrdcTwoDoseEq\spread{$[\nLibOrdcTwoDoseEqLo,\, \nLibOrdcTwoDoseEqHi]$} & \nLibOrdcTwoPost & \nLibOrdcTwoRule \\
iv & ERCOT real-time co-optimization & \nLibErcotRtcbDateIso & \nLibErcotRtcbDose & \nLibErcotRtcbBstale & \nLibErcotRtcbStaleBelow of \nLibErcotRtcbStaleScores & \nLibErcotRtcbFloor & $\nLibErcotRtcbGain$ & \nLibErcotRtcbDoseEq\spread{$[\nLibErcotRtcbDoseEqLo,\, \nLibErcotRtcbDoseEqHi]$} & \nLibErcotRtcbPost & \nLibErcotRtcbRule \\
v & one FOMC decision & \nLibFomcSingleDateIso & \nLibFomcSingleDose & \nLibFomcSingleBstale & \nLibFomcSingleStaleBelow of \nLibFomcSingleStaleScores & \nLibFomcSingleFloor & $\nLibFomcSingleGain$ & \nLibFomcSingleDoseEq\spread{$[\nLibFomcSingleDoseEqLo,\, \nLibFomcSingleDoseEqHi]$} & \nLibFomcSinglePost & \nLibFomcSingleRule \\
vi & NEM five minute settlement & \nLibNemFiveDateIso & \nLibNemFiveDose & \nLibNemFiveBstale & \nLibNemFiveStaleBelow of \nLibNemFiveStaleScores & \nLibNemFiveFloor & \nLibNemFiveGain & \nLibNemFiveDoseEq\spread{$[\nLibNemFiveDoseEqLo,\, \nLibNemFiveDoseEqHi]$} & \nLibNemFivePost & \nLibNemFiveRule \\
vii & NEM administered price cap & \nLibNemApcDateIso & $\nLibNemApcDose$ & \nLibNemApcBstale & \nLibNemApcStaleBelow of \nLibNemApcStaleScores & \nLibNemApcFloor & $\nLibNemApcGain$ & \nLibNemApcDoseEq\spread{$[\nLibNemApcDoseEqLo,\, \nLibNemApcDoseEqHi]$} & \nLibNemApcPost & \nLibNemApcRule \\
viii & NEM market price cap & \nLibNemMpcDateIso & \nLibNemMpcDose & \nLibNemMpcBstale & \nLibNemMpcStaleBelow of \nLibNemMpcStaleScores & \nLibNemMpcFloor & $\nLibNemMpcGain$ & \nLibNemMpcDoseEq\spread{$[\nLibNemMpcDoseEqLo,\, \nLibNemMpcDoseEqHi]$} & \nLibNemMpcPost & \nLibNemMpcRule \\
ix & PJM reserve price formation & \nLibPjmRpfDateIso & $\nLibPjmRpfDose$ & \nLibPjmRpfBstale & \nLibPjmRpfStaleBelow of \nLibPjmRpfStaleScores & \nLibPjmRpfFloor & $\nLibPjmRpfGain$ & \nLibPjmRpfDoseEq\spread{$[\nLibPjmRpfDoseEqLo,\, \nLibPjmRpfDoseEqHi]$} & \nLibPjmRpfPost & \nLibPjmRpfRule \\
\bottomrule
\end{tabular}
\par\raggedright ERCOT is the Electric Reliability Council of Texas. FOMC is the Federal Open Market Committee. NEM is Australia's National Electricity Market. PJM is PJM Interconnection.
\end{table}

\paragraph{A simple rule of thumb often fails.} A rule of thumb predicts a restart's gain where the frozen quantile's frequency gap over the days after the break exceeds one over the square root of the number of forecasts since the break. At the break in the NEM, for example, the threshold frozen at the break lies \nLibNemFiveBstale from nominal coverage on the scores after it. Table~\ref{tab:predictions} sets the rule of thumb's prediction beside the realized winners, the methods that do best on each break. With the frozen quantile counted among the methods that use the announcement, ACI or a variant wins on \nLibPredHelpsFailedPrCf of the \nLibPredHelps breaks where the rule predicts a gain under the holdout reference, and every average over the cell gives the same count. Appendix~\ref{app:corrections} gives the count with the threshold forms of the variants in place of their official implementations. At the \nOrdFifty forecast the count is \nLibPredHelpsFailedFiftyReadersPrCf, or \nLibPredHelpsFailedFiftyOthersPrCf with the frozen quantile among the others. We chose this grouping after seeing the counts.

Winners are chosen among the reported methods. The first winner's coverage after the break is closest to nominal coverage, and its error is its frequency gap over the days after the break. The second has the smallest reliability gap on the holdout reference, averaged over the forecasts whose reference has at least \nLibMinReference scores, with an infinite threshold counted as a gap of the miscoverage level, and every method within one part in a trillion of it is named. The last four columns give that mean for the winner, the restarted quantile, restarted ACI and ACI. The Mondrian predictor is the plug-in construction of Appendix~\ref{app:mondrian}, whose threshold equals the restarted quantile's at every step. The released data include every method's values.

\begin{table}[H]
\caption{The rule of thumb's prediction beside what each break returned. The ratio is the frozen quantile's frequency gap over the days after the break, over one over the square root of the number of forecasts since the break, and the rule predicts a gain where it exceeds one. Each winner is a point estimate with no paired band.}
\label{tab:predictions}
\centering
\scriptsize
\setlength{\tabcolsep}{2pt}
\begin{tabular}{l>{\raggedright\arraybackslash}p{\libcolA}cc>{\raggedright\arraybackslash}p{\winnercol}c>{\raggedright\arraybackslash}p{\winnercol}c c c c}
\toprule
label & \begin{tabular}[c]{@{}c@{}}announced\\break\end{tabular} & ratio & prediction & \begin{tabular}[c]{@{}c@{}}least frequency gap\\after the break\end{tabular} & \begin{tabular}[c]{@{}c@{}}its\\frequency\\gap\end{tabular} & \begin{tabular}[c]{@{}c@{}}least reliability gap\\over the cell\end{tabular} & its gap & \begin{tabular}[c]{@{}c@{}}restarted\\quantile\end{tabular} & \begin{tabular}[c]{@{}c@{}}restarted\\ACI\end{tabular} & ACI \\
\midrule
i & ERCOT offer cap & \nLibErcotCapRatio & \nLibErcotCapPred & \nLibErcotCapWinnerCfCov & \nLibErcotCapWinnerCfCovErr & \nLibErcotCapWinnerCf & \nLibErcotCapWinnerCfValue & \nLibErcotCapCfCellSeg & \nLibErcotCapCfCellAbaci & \nLibErcotCapCfCellAci \\
ii & first ERCOT reserve curve shift & \nLibOrdcOneRatio & \nLibOrdcOnePred & \nLibOrdcOneWinnerCfCov & \nLibOrdcOneWinnerCfCovErr & \nLibOrdcOneWinnerCf & \nLibOrdcOneWinnerCfValue & \nLibOrdcOneCfCellSeg & \nLibOrdcOneCfCellAbaci & \nLibOrdcOneCfCellAci \\
iii & second ERCOT reserve curve shift & \nLibOrdcTwoRatio & \nLibOrdcTwoPred & \nLibOrdcTwoWinnerCfCov & \nLibOrdcTwoWinnerCfCovErr & \nLibOrdcTwoWinnerCf & \nLibOrdcTwoWinnerCfValue & \nLibOrdcTwoCfCellSeg & \nLibOrdcTwoCfCellAbaci & \nLibOrdcTwoCfCellAci \\
iv & ERCOT real-time co-optimization & \nLibErcotRtcbRatio & \nLibErcotRtcbPred & \nLibErcotRtcbWinnerCfCov & \nLibErcotRtcbWinnerCfCovErr & \nLibErcotRtcbWinnerCf & \nLibErcotRtcbWinnerCfValue & \nLibErcotRtcbCfCellSeg & \nLibErcotRtcbCfCellAbaci & \nLibErcotRtcbCfCellAci \\
v & one FOMC decision & \nLibFomcSingleRatio & \nLibFomcSinglePred & \nLibFomcSingleWinnerCfCov & \nLibFomcSingleWinnerCfCovErr & \nLibFomcSingleWinnerCf & \nLibFomcSingleWinnerCfValue & \nLibFomcSingleCfCellSeg & \nLibFomcSingleCfCellAbaci & \nLibFomcSingleCfCellAci \\
vi & NEM five minute settlement & \nLibNemFiveRatio & \nLibNemFivePred & \nLibNemFiveWinnerCfCov & \nLibNemFiveWinnerCfCovErr & \nLibNemFiveWinnerCf & \nLibNemFiveWinnerCfValue & \nLibNemFiveCfCellSeg & \nLibNemFiveCfCellAbaci & \nLibNemFiveCfCellAci \\
vii & NEM administered price cap & \nLibNemApcRatio & \nLibNemApcPred & \nLibNemApcWinnerCfCov & \nLibNemApcWinnerCfCovErr & \nLibNemApcWinnerCf & \nLibNemApcWinnerCfValue & \nLibNemApcCfCellSeg & \nLibNemApcCfCellAbaci & \nLibNemApcCfCellAci \\
viii & NEM market price cap & \nLibNemMpcRatio & \nLibNemMpcPred & \nLibNemMpcWinnerCfCov & \nLibNemMpcWinnerCfCovErr & \nLibNemMpcWinnerCf & \nLibNemMpcWinnerCfValue & \nLibNemMpcCfCellSeg & \nLibNemMpcCfCellAbaci & \nLibNemMpcCfCellAci \\
ix & PJM reserve price formation & \nLibPjmRpfRatio & \nLibPjmRpfPred & \nLibPjmRpfWinnerCfCov & \nLibPjmRpfWinnerCfCovErr & \nLibPjmRpfWinnerCf & \nLibPjmRpfWinnerCfValue & \nLibPjmRpfCfCellSeg & \nLibPjmRpfCfCellAbaci & \nLibPjmRpfCfCellAci \\
\bottomrule
\end{tabular}
\par\raggedright ACI is adaptive conformal inference. ERCOT is the Electric Reliability Council of Texas. Conformal PID is conformal proportional-integral-derivative control. DtACI is dynamically-tuned ACI. SAOCP is strongly adaptive online conformal prediction. AgACI is aggregated ACI. FOMC is the Federal Open Market Committee. NEM is Australia's National Electricity Market. PJM is PJM Interconnection.
\end{table}

\section{Compute and reproducibility}
\label{app:compute}

Every remote run ran on cloud CPU workers, in \nCompJobs jobs that Table~\ref{tab:compute} splits by experiment. The time the jobs spent running their own code projects to USD \nCompUsd, while the provider's own report posts USD \nCompPostedTotalUsd across the five applications that ran them. These include the runs of the gas transit break, which the experiments leave out, since those runs were spent. The \nCompLocalDeposits analyses run on our machine report \nCompLocalSeconds seconds. The provider posts USD \nCompPostedUsdFull for the application that ran the main experiments on \nCompPostedDate, against a projection of USD \nSpendUsd for the same runs, since the projection counts only the time the jobs spent running their own code. Read on \nCompBillingReadDate, its report posts USD \nCompRerunPostedUsd for the reruns of the FOMC stream and the announced breaks, USD \nCompSourcesPostedUsd for the variants' runs, USD \nCompRefitPostedUsd for the runs that fit the forecaster of Appendix~\ref{app:ercot} and USD \nCompArchPostedUsd for the second SAOCP run.

The account includes the runs of the gas transit break, since those runs were spent. The last row is the same product over every row's unrounded seconds. The projection leaves out attempts that failed before their body ran, container start and idle time, image builds and every run on our machine.

\begin{table}[H]
\caption{Compute per experiment on cloud CPU workers of \nCoreRequest core and \nMemMib mebibytes each. Wall seconds are the time the jobs spent running their own code, as the jobs reported it, and a row's price is its unrounded seconds times USD \nCompRatePerSecond a second, rounded once for printing.}
\label{tab:compute}
\centering
\scriptsize
\setlength{\tabcolsep}{4pt}
\begin{tabular}{>{\raggedright\arraybackslash}p{\compcolA} c c c}
\toprule
experiment & jobs & wall seconds & projected USD \\
\midrule
the synthetic family under the methods and the threshold forms & \nCompSynJobs & \nCompSynWall & \nCompSynUsd \\
the synthetic family under the four variants' official implementations, with the check of two machines & \nCompSynSrcJobs & \nCompSynSrcWall & \nCompSynSrcUsd \\
SAOCP's official implementation on the synthetic family, run again on the workers beside our machine & \nCompArchJobs & \nCompArchWall & \nCompArchUsd \\
the grid of shifts & \nCompDoseJobs & \nCompDoseWall & \nCompDoseUsd \\
the misdated break, the stress checks and the simulation without a break & \nCompCsixJobs & \nCompCsixWall & \nCompCsixUsd \\
the mixing check and its dependence measurement & \nCompMixJobs & \nCompMixWall & \nCompMixUsd \\
the ERCOT window, its reruns and its two replays & \nCompErcotJobs & \nCompErcotWall & \nCompErcotUsd \\
the FOMC stream, as run, under the convention and under the variants & \nCompFomcJobs & \nCompFomcWall & \nCompFomcUsd \\
the announced breaks with their reruns and the gas transit break & \nCompLibJobs & \nCompLibWall & \nCompLibUsd \\
the runs that fit and use the forecaster of Appendix~\ref{app:ercot} and its realigned controls & \nCompRefitJobs & \nCompRefitWall & \nCompRefitUsd \\
the numerical checks of the proofs & \nCompProofJobs & \nCompProofWall & \nCompProofUsd \\
\midrule
all experiments & \nCompJobs & \nCompWallSeconds & \nCompUsd \\
\bottomrule
\end{tabular}
\par\raggedright SAOCP is strongly adaptive online conformal prediction. ERCOT is the Electric Reliability Council of Texas. FOMC is the Federal Open Market Committee.
\end{table}

\paragraph{SAOCP on two machines.}
The official implementation of SAOCP returns different values on two machines over the synthetic family. Our machine is an \nArchMacCpu at \nArchMacMachine under \nArchMacOs, with Python \nArchMacPython and numpy \nArchMacNumpy. The workers run \nArchWorkersMachine processors of two generations of instruction set under Linux, with Python \nArchWorkersPython and numpy \nArchWorkersNumpy. The older generation has the Advanced Vector Extensions \nArchIsaOld alone and the newer \nArchIsaNew as well, and the sandbox reports each processor's model as unknown. Of the workers' \nArchWorkersChunks batches of replications, \nArchWorkersAvxTwoOnly ran on the older generation and \nArchWorkersAvxFiveTwelve on the newer. Both machines ran all \nArchReps replications, and every batch confirmed the implementation unmodified before its first stream. The workers reproduce the reported values on every replication and every reported quantity on both generations, so every SAOCP value in this paper is theirs. The two machines compute each stream's seasonal sine differently in its last bit on every replication, and SAOCP's radius begins to differ within a few steps of the first difference it sees. Every other method agrees with its reported values on the \nArchControlReps replications our machine reran. The difference appears at every count of every configuration, as Table~\ref{tab:arch} reports.

The next column counts the replications whose gaps differ, and the last two give the same difference and count for the frequency gap.

\begin{table}[H]
\caption{SAOCP's official implementation on the synthetic family on two machines. The first three value columns give its mean reliability gap at each count over \nReplayPairs replications on our machine and on the workers, and then our machine's mean less the workers' with its band.}
\label{tab:arch}
\centering
\scriptsize
\setlength{\tabcolsep}{3pt}
\begin{tabular}{llcccccc}
\toprule
configuration & count & \begin{tabular}[c]{@{}c@{}}our\\machine\end{tabular} & workers & difference & differing & \begin{tabular}[c]{@{}c@{}}frequency\\difference\end{tabular} & differing \\
\midrule
no break & \nCountTen & \nArchNoBreakPtTenMac & \nArchNoBreakPtTenWorkers & \nArchNoBreakPtTenDiff $[\nArchNoBreakPtTenLo,\, \nArchNoBreakPtTenHi]$ & \nArchNoBreakPtTenDiffering & \nArchNoBreakFreqTenDiff $[\nArchNoBreakFreqTenLo,\, \nArchNoBreakFreqTenHi]$ & \nArchNoBreakFreqTenDiffering \\
$(\nSynDeltaNoBreak,\, \nSynSigmaNoBreak)$ & \nCountTwentyFive & \nArchNoBreakPtTwentyFiveMac & \nArchNoBreakPtTwentyFiveWorkers & \nArchNoBreakPtTwentyFiveDiff $[\nArchNoBreakPtTwentyFiveLo,\, \nArchNoBreakPtTwentyFiveHi]$ & \nArchNoBreakPtTwentyFiveDiffering & \nArchNoBreakFreqTwentyFiveDiff $[\nArchNoBreakFreqTwentyFiveLo,\, \nArchNoBreakFreqTwentyFiveHi]$ & \nArchNoBreakFreqTwentyFiveDiffering \\
 & \nCountFifty & \nArchNoBreakPtFiftyMac & \nArchNoBreakPtFiftyWorkers & \nArchNoBreakPtFiftyDiff $[\nArchNoBreakPtFiftyLo,\, \nArchNoBreakPtFiftyHi]$ & \nArchNoBreakPtFiftyDiffering & $\nArchNoBreakFreqFiftyDiff$ $[\nArchNoBreakFreqFiftyLo,\, \nArchNoBreakFreqFiftyHi]$ & \nArchNoBreakFreqFiftyDiffering \\
 & \nCountTwoHundred & \nArchNoBreakPtTwoHundredMac & \nArchNoBreakPtTwoHundredWorkers & $\nArchNoBreakPtTwoHundredDiff$ $[\nArchNoBreakPtTwoHundredLo,\, \nArchNoBreakPtTwoHundredHi]$ & \nArchNoBreakPtTwoHundredDiffering & \nArchNoBreakFreqTwoHundredDiff $[\nArchNoBreakFreqTwoHundredLo,\, \nArchNoBreakFreqTwoHundredHi]$ & \nArchNoBreakFreqTwoHundredDiffering \\
 & \nCountThousand & \nArchNoBreakPtThousandMac & \nArchNoBreakPtThousandWorkers & \nArchNoBreakPtThousandDiff $[\nArchNoBreakPtThousandLo,\, \nArchNoBreakPtThousandHi]$ & \nArchNoBreakPtThousandDiffering & \nArchNoBreakFreqThousandDiff $[\nArchNoBreakFreqThousandLo,\, \nArchNoBreakFreqThousandHi]$ & \nArchNoBreakFreqThousandDiffering \\
 & \nCountFiveThousand & \nArchNoBreakPtFiveThousandMac & \nArchNoBreakPtFiveThousandWorkers & $\nArchNoBreakPtFiveThousandDiff$ $[\nArchNoBreakPtFiveThousandLo,\, \nArchNoBreakPtFiveThousandHi]$ & \nArchNoBreakPtFiveThousandDiffering & \nArchNoBreakFreqFiveThousandDiff $[\nArchNoBreakFreqFiveThousandLo,\, \nArchNoBreakFreqFiveThousandHi]$ & \nArchNoBreakFreqFiveThousandDiffering \\
\midrule
mean shift & \nCountTen & \nArchMeanShiftPtTenMac & \nArchMeanShiftPtTenWorkers & \nArchMeanShiftPtTenDiff $[\nArchMeanShiftPtTenLo,\, \nArchMeanShiftPtTenHi]$ & \nArchMeanShiftPtTenDiffering & \nArchMeanShiftFreqTenDiff $[\nArchMeanShiftFreqTenLo,\, \nArchMeanShiftFreqTenHi]$ & \nArchMeanShiftFreqTenDiffering \\
$(\nSynDeltaMeanShift,\, \nSynSigmaMeanShift)$ & \nCountTwentyFive & \nArchMeanShiftPtTwentyFiveMac & \nArchMeanShiftPtTwentyFiveWorkers & \nArchMeanShiftPtTwentyFiveDiff $[\nArchMeanShiftPtTwentyFiveLo,\, \nArchMeanShiftPtTwentyFiveHi]$ & \nArchMeanShiftPtTwentyFiveDiffering & \nArchMeanShiftFreqTwentyFiveDiff $[\nArchMeanShiftFreqTwentyFiveLo,\, \nArchMeanShiftFreqTwentyFiveHi]$ & \nArchMeanShiftFreqTwentyFiveDiffering \\
 & \nCountFifty & \nArchMeanShiftPtFiftyMac & \nArchMeanShiftPtFiftyWorkers & $\nArchMeanShiftPtFiftyDiff$ $[\nArchMeanShiftPtFiftyLo,\, \nArchMeanShiftPtFiftyHi]$ & \nArchMeanShiftPtFiftyDiffering & \nArchMeanShiftFreqFiftyDiff $[\nArchMeanShiftFreqFiftyLo,\, \nArchMeanShiftFreqFiftyHi]$ & \nArchMeanShiftFreqFiftyDiffering \\
 & \nCountTwoHundred & \nArchMeanShiftPtTwoHundredMac & \nArchMeanShiftPtTwoHundredWorkers & $\nArchMeanShiftPtTwoHundredDiff$ $[\nArchMeanShiftPtTwoHundredLo,\, \nArchMeanShiftPtTwoHundredHi]$ & \nArchMeanShiftPtTwoHundredDiffering & $\nArchMeanShiftFreqTwoHundredDiff$ $[\nArchMeanShiftFreqTwoHundredLo,\, \nArchMeanShiftFreqTwoHundredHi]$ & \nArchMeanShiftFreqTwoHundredDiffering \\
 & \nCountThousand & \nArchMeanShiftPtThousandMac & \nArchMeanShiftPtThousandWorkers & \nArchMeanShiftPtThousandDiff $[\nArchMeanShiftPtThousandLo,\, \nArchMeanShiftPtThousandHi]$ & \nArchMeanShiftPtThousandDiffering & \nArchMeanShiftFreqThousandDiff $[\nArchMeanShiftFreqThousandLo,\, \nArchMeanShiftFreqThousandHi]$ & \nArchMeanShiftFreqThousandDiffering \\
 & \nCountFiveThousand & \nArchMeanShiftPtFiveThousandMac & \nArchMeanShiftPtFiveThousandWorkers & \nArchMeanShiftPtFiveThousandDiff $[\nArchMeanShiftPtFiveThousandLo,\, \nArchMeanShiftPtFiveThousandHi]$ & \nArchMeanShiftPtFiveThousandDiffering & \nArchMeanShiftFreqFiveThousandDiff $[\nArchMeanShiftFreqFiveThousandLo,\, \nArchMeanShiftFreqFiveThousandHi]$ & \nArchMeanShiftFreqFiveThousandDiffering \\
\midrule
variance shift & \nCountTen & \nArchVarShiftPtTenMac & \nArchVarShiftPtTenWorkers & \nArchVarShiftPtTenDiff $[\nArchVarShiftPtTenLo,\, \nArchVarShiftPtTenHi]$ & \nArchVarShiftPtTenDiffering & \nArchVarShiftFreqTenDiff $[\nArchVarShiftFreqTenLo,\, \nArchVarShiftFreqTenHi]$ & \nArchVarShiftFreqTenDiffering \\
$(\nSynDeltaVarShift,\, \nSynSigmaVarShift)$ & \nCountTwentyFive & \nArchVarShiftPtTwentyFiveMac & \nArchVarShiftPtTwentyFiveWorkers & \nArchVarShiftPtTwentyFiveDiff $[\nArchVarShiftPtTwentyFiveLo,\, \nArchVarShiftPtTwentyFiveHi]$ & \nArchVarShiftPtTwentyFiveDiffering & \nArchVarShiftFreqTwentyFiveDiff $[\nArchVarShiftFreqTwentyFiveLo,\, \nArchVarShiftFreqTwentyFiveHi]$ & \nArchVarShiftFreqTwentyFiveDiffering \\
 & \nCountFifty & \nArchVarShiftPtFiftyMac & \nArchVarShiftPtFiftyWorkers & \nArchVarShiftPtFiftyDiff $[\nArchVarShiftPtFiftyLo,\, \nArchVarShiftPtFiftyHi]$ & \nArchVarShiftPtFiftyDiffering & \nArchVarShiftFreqFiftyDiff $[\nArchVarShiftFreqFiftyLo,\, \nArchVarShiftFreqFiftyHi]$ & \nArchVarShiftFreqFiftyDiffering \\
 & \nCountTwoHundred & \nArchVarShiftPtTwoHundredMac & \nArchVarShiftPtTwoHundredWorkers & $\nArchVarShiftPtTwoHundredDiff$ $[\nArchVarShiftPtTwoHundredLo,\, \nArchVarShiftPtTwoHundredHi]$ & \nArchVarShiftPtTwoHundredDiffering & \nArchVarShiftFreqTwoHundredDiff $[\nArchVarShiftFreqTwoHundredLo,\, \nArchVarShiftFreqTwoHundredHi]$ & \nArchVarShiftFreqTwoHundredDiffering \\
 & \nCountThousand & \nArchVarShiftPtThousandMac & \nArchVarShiftPtThousandWorkers & \nArchVarShiftPtThousandDiff $[\nArchVarShiftPtThousandLo,\, \nArchVarShiftPtThousandHi]$ & \nArchVarShiftPtThousandDiffering & $\nArchVarShiftFreqThousandDiff$ $[\nArchVarShiftFreqThousandLo,\, \nArchVarShiftFreqThousandHi]$ & \nArchVarShiftFreqThousandDiffering \\
 & \nCountFiveThousand & \nArchVarShiftPtFiveThousandMac & \nArchVarShiftPtFiveThousandWorkers & \nArchVarShiftPtFiveThousandDiff $[\nArchVarShiftPtFiveThousandLo,\, \nArchVarShiftPtFiveThousandHi]$ & \nArchVarShiftPtFiveThousandDiffering & \nArchVarShiftFreqFiveThousandDiff $[\nArchVarShiftFreqFiveThousandLo,\, \nArchVarShiftFreqFiveThousandHi]$ & \nArchVarShiftFreqFiveThousandDiffering \\
\midrule
both shifts & \nCountTen & \nArchBothShiftPtTenMac & \nArchBothShiftPtTenWorkers & \nArchBothShiftPtTenDiff $[\nArchBothShiftPtTenLo,\, \nArchBothShiftPtTenHi]$ & \nArchBothShiftPtTenDiffering & $\nArchBothShiftFreqTenDiff$ $[\nArchBothShiftFreqTenLo,\, \nArchBothShiftFreqTenHi]$ & \nArchBothShiftFreqTenDiffering \\
$(\nSynDeltaBothShift,\, \nSynSigmaBothShift)$ & \nCountTwentyFive & \nArchBothShiftPtTwentyFiveMac & \nArchBothShiftPtTwentyFiveWorkers & $\nArchBothShiftPtTwentyFiveDiff$ $[\nArchBothShiftPtTwentyFiveLo,\, \nArchBothShiftPtTwentyFiveHi]$ & \nArchBothShiftPtTwentyFiveDiffering & \nArchBothShiftFreqTwentyFiveDiff $[\nArchBothShiftFreqTwentyFiveLo,\, \nArchBothShiftFreqTwentyFiveHi]$ & \nArchBothShiftFreqTwentyFiveDiffering \\
 & \nCountFifty & \nArchBothShiftPtFiftyMac & \nArchBothShiftPtFiftyWorkers & \nArchBothShiftPtFiftyDiff $[\nArchBothShiftPtFiftyLo,\, \nArchBothShiftPtFiftyHi]$ & \nArchBothShiftPtFiftyDiffering & \nArchBothShiftFreqFiftyDiff $[\nArchBothShiftFreqFiftyLo,\, \nArchBothShiftFreqFiftyHi]$ & \nArchBothShiftFreqFiftyDiffering \\
 & \nCountTwoHundred & \nArchBothShiftPtTwoHundredMac & \nArchBothShiftPtTwoHundredWorkers & \nArchBothShiftPtTwoHundredDiff $[\nArchBothShiftPtTwoHundredLo,\, \nArchBothShiftPtTwoHundredHi]$ & \nArchBothShiftPtTwoHundredDiffering & \nArchBothShiftFreqTwoHundredDiff $[\nArchBothShiftFreqTwoHundredLo,\, \nArchBothShiftFreqTwoHundredHi]$ & \nArchBothShiftFreqTwoHundredDiffering \\
 & \nCountThousand & \nArchBothShiftPtThousandMac & \nArchBothShiftPtThousandWorkers & \nArchBothShiftPtThousandDiff $[\nArchBothShiftPtThousandLo,\, \nArchBothShiftPtThousandHi]$ & \nArchBothShiftPtThousandDiffering & \nArchBothShiftFreqThousandDiff $[\nArchBothShiftFreqThousandLo,\, \nArchBothShiftFreqThousandHi]$ & \nArchBothShiftFreqThousandDiffering \\
 & \nCountFiveThousand & \nArchBothShiftPtFiveThousandMac & \nArchBothShiftPtFiveThousandWorkers & \nArchBothShiftPtFiveThousandDiff $[\nArchBothShiftPtFiveThousandLo,\, \nArchBothShiftPtFiveThousandHi]$ & \nArchBothShiftPtFiveThousandDiffering & \nArchBothShiftFreqFiveThousandDiff $[\nArchBothShiftFreqFiveThousandLo,\, \nArchBothShiftFreqFiveThousandHi]$ & \nArchBothShiftFreqFiveThousandDiffering \\
\midrule
large mean shift & \nCountTen & \nArchLargeMeanPtTenMac & \nArchLargeMeanPtTenWorkers & \nArchLargeMeanPtTenDiff $[\nArchLargeMeanPtTenLo,\, \nArchLargeMeanPtTenHi]$ & \nArchLargeMeanPtTenDiffering & $\nArchLargeMeanFreqTenDiff$ $[\nArchLargeMeanFreqTenLo,\, \nArchLargeMeanFreqTenHi]$ & \nArchLargeMeanFreqTenDiffering \\
$(\nSynDeltaLargeMean,\, \nSynSigmaLargeMean)$ & \nCountTwentyFive & \nArchLargeMeanPtTwentyFiveMac & \nArchLargeMeanPtTwentyFiveWorkers & \nArchLargeMeanPtTwentyFiveDiff $[\nArchLargeMeanPtTwentyFiveLo,\, \nArchLargeMeanPtTwentyFiveHi]$ & \nArchLargeMeanPtTwentyFiveDiffering & \nArchLargeMeanFreqTwentyFiveDiff $[\nArchLargeMeanFreqTwentyFiveLo,\, \nArchLargeMeanFreqTwentyFiveHi]$ & \nArchLargeMeanFreqTwentyFiveDiffering \\
 & \nCountFifty & \nArchLargeMeanPtFiftyMac & \nArchLargeMeanPtFiftyWorkers & $\nArchLargeMeanPtFiftyDiff$ $[\nArchLargeMeanPtFiftyLo,\, \nArchLargeMeanPtFiftyHi]$ & \nArchLargeMeanPtFiftyDiffering & \nArchLargeMeanFreqFiftyDiff $[\nArchLargeMeanFreqFiftyLo,\, \nArchLargeMeanFreqFiftyHi]$ & \nArchLargeMeanFreqFiftyDiffering \\
 & \nCountTwoHundred & \nArchLargeMeanPtTwoHundredMac & \nArchLargeMeanPtTwoHundredWorkers & $\nArchLargeMeanPtTwoHundredDiff$ $[\nArchLargeMeanPtTwoHundredLo,\, \nArchLargeMeanPtTwoHundredHi]$ & \nArchLargeMeanPtTwoHundredDiffering & \nArchLargeMeanFreqTwoHundredDiff $[\nArchLargeMeanFreqTwoHundredLo,\, \nArchLargeMeanFreqTwoHundredHi]$ & \nArchLargeMeanFreqTwoHundredDiffering \\
 & \nCountThousand & \nArchLargeMeanPtThousandMac & \nArchLargeMeanPtThousandWorkers & \nArchLargeMeanPtThousandDiff $[\nArchLargeMeanPtThousandLo,\, \nArchLargeMeanPtThousandHi]$ & \nArchLargeMeanPtThousandDiffering & \nArchLargeMeanFreqThousandDiff $[\nArchLargeMeanFreqThousandLo,\, \nArchLargeMeanFreqThousandHi]$ & \nArchLargeMeanFreqThousandDiffering \\
 & \nCountFiveThousand & \nArchLargeMeanPtFiveThousandMac & \nArchLargeMeanPtFiveThousandWorkers & \nArchLargeMeanPtFiveThousandDiff $[\nArchLargeMeanPtFiveThousandLo,\, \nArchLargeMeanPtFiveThousandHi]$ & \nArchLargeMeanPtFiveThousandDiffering & \nArchLargeMeanFreqFiveThousandDiff $[\nArchLargeMeanFreqFiveThousandLo,\, \nArchLargeMeanFreqFiveThousandHi]$ & \nArchLargeMeanFreqFiveThousandDiffering \\
\bottomrule
\end{tabular}
\par\raggedright SAOCP is strongly adaptive online conformal prediction.
\end{table}

Every band of a comparison with SAOCP keeps its side of zero on both machines, as Table~\ref{tab:arch-bands} shows, and a replay of every band of Tables~\ref{tab:arch} and~\ref{tab:arch-bands} on the \nReplayPairs replication pairs of each configuration agrees with the reported band in all \nReplayAgree cases. The machines reorder SAOCP among the five baselines on two results, moving its rank from \nArchLargeMeanFreqTwoHundredPlaceWorkers to \nArchLargeMeanFreqTwoHundredPlaceMac on the large mean shift's frequency gap over the first \nCountTwoHundred forecasts and from \nArchNoBreakFreqThousandPlaceWorkers to \nArchNoBreakFreqThousandPlaceMac without a break over the first \nCountThousand.

\begin{table}[H]
\caption{Every reported comparison with SAOCP on the synthetic family, as a mean difference over \nReplayPairs paired replications with its band, under the workers' values and under our machine's. Both columns use the same resamples as the paper's bands, so they differ only in SAOCP's values, and a negative difference favours the method named first.}
\label{tab:arch-bands}
\centering
\scriptsize
\setlength{\tabcolsep}{4pt}
\begin{tabular}{l>{\raggedright\arraybackslash}p{\archcolA}cc}
\toprule
configuration & comparison & workers & our machine \\
\midrule
no break & restarted quantile less SAOCP, reliability, \nOrdTen & \nArchBandNoBreakResetPtTenWorkersDiff $[\nArchBandNoBreakResetPtTenWorkersLo,\, \nArchBandNoBreakResetPtTenWorkersHi]$ & \nArchBandNoBreakResetPtTenMacDiff $[\nArchBandNoBreakResetPtTenMacLo,\, \nArchBandNoBreakResetPtTenMacHi]$ \\
$(\nSynDeltaNoBreak,\, \nSynSigmaNoBreak)$ & restarted quantile less SAOCP, reliability, \nOrdFifty & $\nArchBandNoBreakResetPtFiftyWorkersDiff$ $[\nArchBandNoBreakResetPtFiftyWorkersLo,\, \nArchBandNoBreakResetPtFiftyWorkersHi]$ & $\nArchBandNoBreakResetPtFiftyMacDiff$ $[\nArchBandNoBreakResetPtFiftyMacLo,\, \nArchBandNoBreakResetPtFiftyMacHi]$ \\
 & the boosted form less SAOCP, frequency, first \nCountFifty & $\nArchBandNoBreakFormAFreqFiftyWorkersDiff$ $[\nArchBandNoBreakFormAFreqFiftyWorkersLo,\, \nArchBandNoBreakFormAFreqFiftyWorkersHi]$ & $\nArchBandNoBreakFormAFreqFiftyMacDiff$ $[\nArchBandNoBreakFormAFreqFiftyMacLo,\, \nArchBandNoBreakFormAFreqFiftyMacHi]$ \\
 & SAOCP less its threshold form, reliability, \nOrdTen & \nArchBandNoBreakTwinPtTenWorkersDiff $[\nArchBandNoBreakTwinPtTenWorkersLo,\, \nArchBandNoBreakTwinPtTenWorkersHi]$ & \nArchBandNoBreakTwinPtTenMacDiff $[\nArchBandNoBreakTwinPtTenMacLo,\, \nArchBandNoBreakTwinPtTenMacHi]$ \\
 & SAOCP less its threshold form, reliability, \nOrdFifty & \nArchBandNoBreakTwinPtFiftyWorkersDiff $[\nArchBandNoBreakTwinPtFiftyWorkersLo,\, \nArchBandNoBreakTwinPtFiftyWorkersHi]$ & \nArchBandNoBreakTwinPtFiftyMacDiff $[\nArchBandNoBreakTwinPtFiftyMacLo,\, \nArchBandNoBreakTwinPtFiftyMacHi]$ \\
\midrule
mean shift & restarted quantile less SAOCP, reliability, \nOrdTen & $\nArchBandMeanShiftResetPtTenWorkersDiff$ $[\nArchBandMeanShiftResetPtTenWorkersLo,\, \nArchBandMeanShiftResetPtTenWorkersHi]$ & $\nArchBandMeanShiftResetPtTenMacDiff$ $[\nArchBandMeanShiftResetPtTenMacLo,\, \nArchBandMeanShiftResetPtTenMacHi]$ \\
$(\nSynDeltaMeanShift,\, \nSynSigmaMeanShift)$ & restarted quantile less SAOCP, reliability, \nOrdFifty & $\nArchBandMeanShiftResetPtFiftyWorkersDiff$ $[\nArchBandMeanShiftResetPtFiftyWorkersLo,\, \nArchBandMeanShiftResetPtFiftyWorkersHi]$ & $\nArchBandMeanShiftResetPtFiftyMacDiff$ $[\nArchBandMeanShiftResetPtFiftyMacLo,\, \nArchBandMeanShiftResetPtFiftyMacHi]$ \\
 & the boosted form less SAOCP, frequency, first \nCountFifty & $\nArchBandMeanShiftFormAFreqFiftyWorkersDiff$ $[\nArchBandMeanShiftFormAFreqFiftyWorkersLo,\, \nArchBandMeanShiftFormAFreqFiftyWorkersHi]$ & $\nArchBandMeanShiftFormAFreqFiftyMacDiff$ $[\nArchBandMeanShiftFormAFreqFiftyMacLo,\, \nArchBandMeanShiftFormAFreqFiftyMacHi]$ \\
 & SAOCP less its threshold form, reliability, \nOrdTen & $\nArchBandMeanShiftTwinPtTenWorkersDiff$ $[\nArchBandMeanShiftTwinPtTenWorkersLo,\, \nArchBandMeanShiftTwinPtTenWorkersHi]$ & $\nArchBandMeanShiftTwinPtTenMacDiff$ $[\nArchBandMeanShiftTwinPtTenMacLo,\, \nArchBandMeanShiftTwinPtTenMacHi]$ \\
 & SAOCP less its threshold form, reliability, \nOrdFifty & $\nArchBandMeanShiftTwinPtFiftyWorkersDiff$ $[\nArchBandMeanShiftTwinPtFiftyWorkersLo,\, \nArchBandMeanShiftTwinPtFiftyWorkersHi]$ & $\nArchBandMeanShiftTwinPtFiftyMacDiff$ $[\nArchBandMeanShiftTwinPtFiftyMacLo,\, \nArchBandMeanShiftTwinPtFiftyMacHi]$ \\
\midrule
variance shift & restarted quantile less SAOCP, reliability, \nOrdTen & $\nArchBandVarShiftResetPtTenWorkersDiff$ $[\nArchBandVarShiftResetPtTenWorkersLo,\, \nArchBandVarShiftResetPtTenWorkersHi]$ & $\nArchBandVarShiftResetPtTenMacDiff$ $[\nArchBandVarShiftResetPtTenMacLo,\, \nArchBandVarShiftResetPtTenMacHi]$ \\
$(\nSynDeltaVarShift,\, \nSynSigmaVarShift)$ & restarted quantile less SAOCP, reliability, \nOrdFifty & $\nArchBandVarShiftResetPtFiftyWorkersDiff$ $[\nArchBandVarShiftResetPtFiftyWorkersLo,\, \nArchBandVarShiftResetPtFiftyWorkersHi]$ & $\nArchBandVarShiftResetPtFiftyMacDiff$ $[\nArchBandVarShiftResetPtFiftyMacLo,\, \nArchBandVarShiftResetPtFiftyMacHi]$ \\
 & the boosted form less SAOCP, frequency, first \nCountFifty & $\nArchBandVarShiftFormAFreqFiftyWorkersDiff$ $[\nArchBandVarShiftFormAFreqFiftyWorkersLo,\, \nArchBandVarShiftFormAFreqFiftyWorkersHi]$ & $\nArchBandVarShiftFormAFreqFiftyMacDiff$ $[\nArchBandVarShiftFormAFreqFiftyMacLo,\, \nArchBandVarShiftFormAFreqFiftyMacHi]$ \\
 & SAOCP less its threshold form, reliability, \nOrdTen & $\nArchBandVarShiftTwinPtTenWorkersDiff$ $[\nArchBandVarShiftTwinPtTenWorkersLo,\, \nArchBandVarShiftTwinPtTenWorkersHi]$ & $\nArchBandVarShiftTwinPtTenMacDiff$ $[\nArchBandVarShiftTwinPtTenMacLo,\, \nArchBandVarShiftTwinPtTenMacHi]$ \\
 & SAOCP less its threshold form, reliability, \nOrdFifty & $\nArchBandVarShiftTwinPtFiftyWorkersDiff$ $[\nArchBandVarShiftTwinPtFiftyWorkersLo,\, \nArchBandVarShiftTwinPtFiftyWorkersHi]$ & $\nArchBandVarShiftTwinPtFiftyMacDiff$ $[\nArchBandVarShiftTwinPtFiftyMacLo,\, \nArchBandVarShiftTwinPtFiftyMacHi]$ \\
\midrule
both shifts & restarted quantile less SAOCP, reliability, \nOrdTen & $\nArchBandBothShiftResetPtTenWorkersDiff$ $[\nArchBandBothShiftResetPtTenWorkersLo,\, \nArchBandBothShiftResetPtTenWorkersHi]$ & $\nArchBandBothShiftResetPtTenMacDiff$ $[\nArchBandBothShiftResetPtTenMacLo,\, \nArchBandBothShiftResetPtTenMacHi]$ \\
$(\nSynDeltaBothShift,\, \nSynSigmaBothShift)$ & restarted quantile less SAOCP, reliability, \nOrdFifty & $\nArchBandBothShiftResetPtFiftyWorkersDiff$ $[\nArchBandBothShiftResetPtFiftyWorkersLo,\, \nArchBandBothShiftResetPtFiftyWorkersHi]$ & $\nArchBandBothShiftResetPtFiftyMacDiff$ $[\nArchBandBothShiftResetPtFiftyMacLo,\, \nArchBandBothShiftResetPtFiftyMacHi]$ \\
 & the boosted form less SAOCP, frequency, first \nCountFifty & $\nArchBandBothShiftFormAFreqFiftyWorkersDiff$ $[\nArchBandBothShiftFormAFreqFiftyWorkersLo,\, \nArchBandBothShiftFormAFreqFiftyWorkersHi]$ & $\nArchBandBothShiftFormAFreqFiftyMacDiff$ $[\nArchBandBothShiftFormAFreqFiftyMacLo,\, \nArchBandBothShiftFormAFreqFiftyMacHi]$ \\
 & SAOCP less its threshold form, reliability, \nOrdTen & $\nArchBandBothShiftTwinPtTenWorkersDiff$ $[\nArchBandBothShiftTwinPtTenWorkersLo,\, \nArchBandBothShiftTwinPtTenWorkersHi]$ & $\nArchBandBothShiftTwinPtTenMacDiff$ $[\nArchBandBothShiftTwinPtTenMacLo,\, \nArchBandBothShiftTwinPtTenMacHi]$ \\
 & SAOCP less its threshold form, reliability, \nOrdFifty & $\nArchBandBothShiftTwinPtFiftyWorkersDiff$ $[\nArchBandBothShiftTwinPtFiftyWorkersLo,\, \nArchBandBothShiftTwinPtFiftyWorkersHi]$ & $\nArchBandBothShiftTwinPtFiftyMacDiff$ $[\nArchBandBothShiftTwinPtFiftyMacLo,\, \nArchBandBothShiftTwinPtFiftyMacHi]$ \\
\midrule
large mean shift & restarted quantile less SAOCP, reliability, \nOrdTen & $\nArchBandLargeMeanResetPtTenWorkersDiff$ $[\nArchBandLargeMeanResetPtTenWorkersLo,\, \nArchBandLargeMeanResetPtTenWorkersHi]$ & $\nArchBandLargeMeanResetPtTenMacDiff$ $[\nArchBandLargeMeanResetPtTenMacLo,\, \nArchBandLargeMeanResetPtTenMacHi]$ \\
$(\nSynDeltaLargeMean,\, \nSynSigmaLargeMean)$ & restarted quantile less SAOCP, reliability, \nOrdFifty & $\nArchBandLargeMeanResetPtFiftyWorkersDiff$ $[\nArchBandLargeMeanResetPtFiftyWorkersLo,\, \nArchBandLargeMeanResetPtFiftyWorkersHi]$ & $\nArchBandLargeMeanResetPtFiftyMacDiff$ $[\nArchBandLargeMeanResetPtFiftyMacLo,\, \nArchBandLargeMeanResetPtFiftyMacHi]$ \\
 & the boosted form less SAOCP, frequency, first \nCountFifty & $\nArchBandLargeMeanFormAFreqFiftyWorkersDiff$ $[\nArchBandLargeMeanFormAFreqFiftyWorkersLo,\, \nArchBandLargeMeanFormAFreqFiftyWorkersHi]$ & $\nArchBandLargeMeanFormAFreqFiftyMacDiff$ $[\nArchBandLargeMeanFormAFreqFiftyMacLo,\, \nArchBandLargeMeanFormAFreqFiftyMacHi]$ \\
 & SAOCP less its threshold form, reliability, \nOrdTen & $\nArchBandLargeMeanTwinPtTenWorkersDiff$ $[\nArchBandLargeMeanTwinPtTenWorkersLo,\, \nArchBandLargeMeanTwinPtTenWorkersHi]$ & $\nArchBandLargeMeanTwinPtTenMacDiff$ $[\nArchBandLargeMeanTwinPtTenMacLo,\, \nArchBandLargeMeanTwinPtTenMacHi]$ \\
 & SAOCP less its threshold form, reliability, \nOrdFifty & $\nArchBandLargeMeanTwinPtFiftyWorkersDiff$ $[\nArchBandLargeMeanTwinPtFiftyWorkersLo,\, \nArchBandLargeMeanTwinPtFiftyWorkersHi]$ & $\nArchBandLargeMeanTwinPtFiftyMacDiff$ $[\nArchBandLargeMeanTwinPtFiftyMacLo,\, \nArchBandLargeMeanTwinPtFiftyMacHi]$ \\
\bottomrule
\end{tabular}
\par\raggedright SAOCP is strongly adaptive online conformal prediction.
\end{table}

\paragraph{Software and assets.}
Over the ERCOT window's \nSaocpArchErcotSteps evaluated days the two machines return the same values. AgACI ran under R \nVerR with opera \nVerOpera. The workers ran Python \nVerPython for the main experiments and the reruns of the FOMC stream and the announced breaks, and Python \nVerPythonLater for the variants' runs, the runs that fit the forecaster and the second SAOCP run. Our machine ran Python \nArchMacPython. All three run numpy \nVerNumpy and pandas \nVerPandas, with scipy \nVerScipy and scikit-learn \nVerSklearn for fitting and LightGBM \nVerLightgbm for the forecasters. Table~\ref{tab:assets} names every asset the runs read, with how each was obtained and the terms its page states. The code and the data are released at \codehost, with the code of every run, the builders of every figure and table, the analyses over the runs' rows and the results of every run, from which every number and exhibit of this paper rebuilds. The released results cover the \nLibCases announced breaks the experiments run.

A scripted client that gave no name or address fetched each page, and the two pages of PJM Interconnection's data service were read on \nAssetPjmPageDate.

\begin{table}[H]
\caption{Every asset the runs read, with how each was obtained and the terms its page states. The Australian Energy Market Operator (AEMO) publishes the NEM's trading prices, Federal Reserve Economic Data (FRED) is the service the Treasury yield came from, and the opera package is released under the GNU Lesser General Public License (LGPL). Quoted terms are printed as each page gave them, and every other term paraphrases its page's own words.}
\label{tab:assets}
\centering
\scriptsize
\setlength{\tabcolsep}{4pt}
\begin{tabular}{>{\raggedright\arraybackslash}p{\assetcolA}>{\raggedright\arraybackslash}p{\assetcolB}>{\raggedright\arraybackslash}p{\assetcolC}}
\toprule
asset & read by & stated terms \\
\midrule
ERCOT's historical real-time settlement point prices at its hubs, pulled from its public report by the gridstatus reader's automated download & the ERCOT window and the four ERCOT breaks & The report is classified public, and ERCOT's terms let raw data from the public portions of its website be used, reproduced and redistributed in compilations, charts and analyses. \\
AEMO's thirty minute NEM trading prices, pulled from its public archive by the nemosis reader's automated download & the three NEM breaks & Its copyright page refused the scripted client, and the Internet Archive's capture of the page of \nPermAemoCaptureDate states ``In addition to the uses permitted under copyright laws, AEMO confirms its general permission for anyone to use AEMO Material for any purpose, but only with accurate and appropriate attribution of the relevant AEMO Material and AEMO as its author.'' It adds ``You do not need to obtain specific permission to use AEMO Material in this way.'' The released data name AEMO as the author of every NEM value they contain. \\
PJM Interconnection's hourly day-ahead zonal prices, pulled month by month from its Data Miner service by the gridstatus reader's automated requests & the reserve price formation break & Its Data Miner page states ``Information and data contained in Data Miner is for internal use only and redistribution of information and or data contained in or derived from Data Miner is strictly prohibited without an active PJM Membership.'' Its terms of use state ``Subject to Section 2 below, you may use the Data and the Tools for your internal business use or commercial purposes, including publishing and making derivatives of the Data, Members may republish data, however non-members are specifically prohibited from republishing data.'' Our position is that these data were used for academic research alone. \\
FRED's daily series of the ten-year constant maturity Treasury yield, downloaded from FRED before the filing & the FOMC stream and the single decision break & FRED marks the series public domain with citation requested, and its terms allow use without permission barring a prohibited use, with the source cited. \\
the Federal Reserve Board's FOMC meeting calendars, read off the Board's pages & the FOMC stream & The Board's information is in the public domain unless otherwise indicated, with the Board cited as the source. \\
the code of DtACI, SAOCP, conformal PID and AgACI, cloned from each variant's repository & the four variants' runs & SAOCP's tree states the Apache licence and conformal PID's and AgACI's the MIT licence. DtACI's states none, so each tree is read and run outside the paper's code and never redistributed. \\
the opera package for R, with which AgACI's code aggregates its experts, fetched from its archive & AgACI's runs & The LGPL. \\
numpy, pandas, scipy, scikit-learn and gridstatus, installed from their published packages & every run & Each has a three-clause BSD licence. \\
LightGBM, installed from its published package & the forecasters & The MIT licence. \\
pyarrow, installed from its published package & the parquet streams & The Apache licence. \\
matplotlib, installed from its published package & the figures & The matplotlib licence agreement. \\
nemosis, installed from its published package & the pull of the NEM prices & The GNU General Public License in its third version. \\
\bottomrule
\end{tabular}
\end{table}

\subsection{The implementations, the forecaster and the band}
\label{app:corrections}

Table~\ref{tab:twins} sets each variant's official implementation against the threshold form of the same name from Appendix~\ref{app:synthetic}.

Synthetic bands resample the \nSrcSynReps shared replications \nSrcSynDraws times. Real bands use the paired moving block bootstrap with \nErcotDraws resamples, in blocks of \nErcotBlock days on the ERCOT window's \nErcotPost days after the reform under the forecaster of Appendix~\ref{app:ercot} and in blocks of \nFomcBlock days on the FOMC stream's \nFomcEval evaluation days, and each real difference's $p$ is printed beside it. Each configuration's name gives its mean shift and its noise scale after the break, $(\delta,\sigma_2)$, in noise standard deviations before the break.

\begin{table}[H]
\caption{Each variant's official implementation less the threshold form of the same name from Appendix~\ref{app:synthetic}, so a negative gap difference favours the official implementation. The upper panel covers the synthetic family and the lower panel the two real streams. Each subscript is the band of that difference.}
\label{tab:twins}
\centering
\footnotesize
\setlength{\tabcolsep}{5pt}
\begin{tabular}{lllll}
\toprule
 & \multicolumn{2}{c}{reliability gap on the large mean shift $(\nSynDeltaLargeMean,\, \nSynSigmaLargeMean)$} & \multicolumn{1}{c}{frequency gap} &  \\
\cmidrule(lr){2-3} \cmidrule(lr){4-4}
method & \nOrdTen forecast & \nOrdFifty forecast & \begin{tabular}[c]{@{}c@{}}mean shift $(\nSynDeltaMeanShift,\, \nSynSigmaMeanShift)$\\first \nCountFifty forecasts\end{tabular} &  \\
\midrule
DtACI & \nTwinLargeMeanDtaciSrcTenPtDiff\spread{$[\nTwinLargeMeanDtaciSrcTenPtLo,\, \nTwinLargeMeanDtaciSrcTenPtHi]$} & $\nTwinLargeMeanDtaciSrcFiftyPtDiff$\spread{$[\nTwinLargeMeanDtaciSrcFiftyPtLo,\, \nTwinLargeMeanDtaciSrcFiftyPtHi]$} & \nTwinMeanShiftDtaciSrcFiftyFreqDiff\spread{$[\nTwinMeanShiftDtaciSrcFiftyFreqLo,\, \nTwinMeanShiftDtaciSrcFiftyFreqHi]$} &  \\
SAOCP & $\nTwinLargeMeanSaocpSrcTenPtDiff$\spread{$[\nTwinLargeMeanSaocpSrcTenPtLo,\, \nTwinLargeMeanSaocpSrcTenPtHi]$} & $\nTwinLargeMeanSaocpSrcFiftyPtDiff$\spread{$[\nTwinLargeMeanSaocpSrcFiftyPtLo,\, \nTwinLargeMeanSaocpSrcFiftyPtHi]$} & $\nTwinMeanShiftSaocpSrcFiftyFreqDiff$\spread{$[\nTwinMeanShiftSaocpSrcFiftyFreqLo,\, \nTwinMeanShiftSaocpSrcFiftyFreqHi]$} &  \\
conformal PID & $\nTwinLargeMeanPidSrcTenPtDiff$\spread{$[\nTwinLargeMeanPidSrcTenPtLo,\, \nTwinLargeMeanPidSrcTenPtHi]$} & $\nTwinLargeMeanPidSrcFiftyPtDiff$\spread{$[\nTwinLargeMeanPidSrcFiftyPtLo,\, \nTwinLargeMeanPidSrcFiftyPtHi]$} & $\nTwinMeanShiftPidSrcFiftyFreqDiff$\spread{$[\nTwinMeanShiftPidSrcFiftyFreqLo,\, \nTwinMeanShiftPidSrcFiftyFreqHi]$} &  \\
AgACI & \nTwinLargeMeanAgaciSrcTenPtDiff\spread{$[\nTwinLargeMeanAgaciSrcTenPtLo,\, \nTwinLargeMeanAgaciSrcTenPtHi]$} & $\nTwinLargeMeanAgaciSrcFiftyPtDiff$\spread{$[\nTwinLargeMeanAgaciSrcFiftyPtLo,\, \nTwinLargeMeanAgaciSrcFiftyPtHi]$} & \nTwinMeanShiftAgaciSrcFiftyFreqDiff\spread{$[\nTwinMeanShiftAgaciSrcFiftyFreqLo,\, \nTwinMeanShiftAgaciSrcFiftyFreqHi]$} &  \\
\midrule
 & \begin{tabular}[c]{@{}c@{}}ERCOT window\\coverage after the reform\end{tabular} & $p$ & \begin{tabular}[c]{@{}c@{}}FOMC stream\\coverage\end{tabular} & $p$ \\
\midrule
DtACI & \nErcotTwinDtaciSrcDiff\spread{$[\nErcotTwinDtaciSrcLo,\, \nErcotTwinDtaciSrcHi]$} & \nErcotTwinDtaciSrcP & $\nFomcTwinDtaciSrcDiff$\spread{$[\nFomcTwinDtaciSrcLo,\, \nFomcTwinDtaciSrcHi]$} & \nFomcTwinDtaciSrcP \\
SAOCP & \nErcotTwinSaocpSrcDiff\spread{$[\nErcotTwinSaocpSrcLo,\, \nErcotTwinSaocpSrcHi]$} & \nErcotTwinSaocpSrcP & \nFomcTwinSaocpSrcDiff\spread{$[\nFomcTwinSaocpSrcLo,\, \nFomcTwinSaocpSrcHi]$} & \nFomcTwinSaocpSrcP \\
conformal PID & \nErcotTwinPidSrcDiff\spread{$[\nErcotTwinPidSrcLo,\, \nErcotTwinPidSrcHi]$} & \nErcotTwinPidSrcP & \nFomcTwinPidSrcDiff\spread{$[\nFomcTwinPidSrcLo,\, \nFomcTwinPidSrcHi]$} & \nFomcTwinPidSrcP \\
AgACI & \nErcotTwinAgaciSrcDiff\spread{$[\nErcotTwinAgaciSrcLo,\, \nErcotTwinAgaciSrcHi]$} & \nErcotTwinAgaciSrcP & $\nFomcTwinAgaciSrcDiff$\spread{$[\nFomcTwinAgaciSrcLo,\, \nFomcTwinAgaciSrcHi]$} & \nFomcTwinAgaciSrcP \\
\bottomrule
\end{tabular}
\end{table}

At LightGBM's default minimum of \nFitFiledMinChild rows per leaf, a forecaster of \nLgbEstimators trees fitted on \nFitFiledPairs pairs, each matching a day's price with that day's own features, makes no split, since a split needs \nFitRowsSplitNeeds rows. It returns \nFitFiledConstant USD per MWh on every evaluated day, which differs from the mean training price by \nFitFiledConstantRelDiff of its value. Pairing each feature row with the next day's price gives \nFitRealignedPairs pairs and, under the same default, again one value, \nFitRealignedConstant. These default settings return one value on all \nFcFiledStreams streams the forecaster serves, the ERCOT window and the announced breaks in power markets, while the forecaster of Appendix~\ref{app:ercot} returns at least \nRefitLevelDistinctMin distinct evaluated forecasts on each.

Table~\ref{tab:masked} computes each method's coverage after the reform on the ERCOT window five ways, from a run on a daily summary that lacked the last \nErcotMaskedDays days, from \nErcotMaskedFirstDateIso, to the forecaster of Appendix~\ref{app:ercot}. A replay with those days missing returns that run's values. Restoring them moves threshold SAOCP's coverage from \nErcotMaskedCovSaocpFiled to \nErcotFiledCovSaocpFiled on the aggregate formed after the step's experts spawn, and measuring the interval held before the outcome moves it to \nErcotFiledCovSaocp.

The replay sets the tail days from \nErcotMaskedFirstDateIso to missing and counts them as misses. The realigned forecaster is fitted on each day's features and the next day's price. The rerun and the realigned forecaster each return one value on every evaluated day. The threshold rows are the threshold forms of Appendix~\ref{app:synthetic}. Threshold SAOCP takes two rows, the aggregate formed after the step's new experts spawn and the interval held before the outcome.

\begin{table}[H]
\caption{Coverage after the reform on the ERCOT window over \nErcotPost days, computed five ways. The columns give a run on a daily summary that lacked the last \nErcotMaskedDays days, its replay with those days missing, a rerun of its forecaster on all \nErcotRows priced days, the realigned forecaster and the forecaster of Appendix~\ref{app:ercot}.}
\label{tab:masked}
\centering
\scriptsize
\setlength{\tabcolsep}{5pt}
\begin{tabular}{lccccc}
\toprule
method & short summary & \begin{tabular}[c]{@{}c@{}}masked\\replay\end{tabular} & \begin{tabular}[c]{@{}c@{}}rerun of the\\default fit\end{tabular} & \begin{tabular}[c]{@{}c@{}}realigned\\fit\end{tabular} & \begin{tabular}[c]{@{}c@{}}forecaster of\\Appendix~\ref{app:ercot}\end{tabular} \\
\midrule
ACI & \nFiledErcotCovAci & \nErcotMaskedCovAci & \nErcotFiledCovAci & \nErcotRealignedCovAci & \nErcotCovAci \\
restarted ACI & \nFiledErcotCovAbaci & \nErcotMaskedCovAbaci & \nErcotFiledCovAbaci & \nErcotRealignedCovAbaci & \nErcotCovAbaci \\
restarted quantile & \nFiledErcotCovSeg & \nErcotMaskedCovSeg & \nErcotFiledCovSeg & \nErcotRealignedCovSeg & \nErcotCovSeg \\
threshold AgACI & \nFiledErcotCovAgaci & \nErcotMaskedCovAgaci & \nErcotFiledCovAgaci & \nErcotRealignedCovAgaci & \nErcotCovAgaci \\
threshold DtACI & \nFiledErcotCovDtaci & \nErcotMaskedCovDtaci & \nErcotFiledCovDtaci & \nErcotRealignedCovDtaci & \nErcotCovDtaci \\
naive & \nFiledErcotCovNaive & \nErcotMaskedCovNaive & \nErcotFiledCovNaive & \nErcotRealignedCovNaive & \nErcotCovNaive \\
threshold conformal PID & \nFiledErcotCovPid & \nErcotMaskedCovPid & \nErcotFiledCovPid & \nErcotRealignedCovPid & \nErcotCovPid \\
threshold SAOCP after the spawn & \nFiledErcotCovSaocp & \nErcotMaskedCovSaocpFiled & \nErcotFiledCovSaocpFiled & \nErcotRealignedCovSaocpFiled & \nErcotCovSaocpFiled \\
threshold SAOCP held before the outcome & not reported & \nErcotMaskedCovSaocp & \nErcotFiledCovSaocp & \nErcotRealignedCovSaocp & \nErcotCovSaocp \\
\bottomrule
\end{tabular}
\end{table}

Table~\ref{tab:fomc-vintages} sets the FOMC band of the burn-in form against ACI under four combinations of resampling settings and score set handling. A rerun under the convention of Section~\ref{sec:setup} returns all \nFomcRerunAgree reported values unchanged, with the burn-in form's $p$ against ACI at \nFomcRerunP.

The first row's band carries no record of its resampling settings. The replay runs a block length of \nFomcFiledScriptBlock, \nFomcFiledScriptDraws draws and a seed of \nFomcFiledScriptSeed, the settings that run's resampling fixed, on its two coverage series. The first rerun runs our code on a form that clears its score set one update early, and the second runs the burn-in form under the convention of Section~\ref{sec:setup}.

\begin{table}[H]
\caption{Band of the burn-in form against ACI over the FOMC stream's \nFomcEval evaluation days under four combinations of resampling settings and score set handling, with differences taken as the burn-in form's coverage less ACI's. Each row pairs the two series under one moving block resample, and every later row differs from the first row's band.}
\label{tab:fomc-vintages}
\centering
\scriptsize
\setlength{\tabcolsep}{2.5pt}
\begin{tabular}{>{\raggedright\arraybackslash}p{\vintcolA}>{\raggedright\arraybackslash}p{\vintcolB} c c c c c c}
\toprule
run & form & block & draws & seed & difference & band & $p$ \\
\midrule
a band with no recorded settings & the burn-in form that clears its score set one update early & none recorded & none recorded & none recorded & not reported & $[\nFiledFomcCiLo,\, \nFiledFomcCiHi]$ & not reported \\
a replay at that band's settings & the burn-in form that clears its score set one update early & \nVintReplayBlock & \nVintReplayDraws & \nVintReplaySeed & $\nVintReplayDiff$ & $[\nVintReplayLo,\, \nVintReplayHi]$ & \nVintReplayP \\
the first rerun & the burn-in form that clears its score set one update early & \nFomcVintWaveBlock & \nFomcVintWaveDraws & \nFomcVintWaveSeed & $\nFomcVintWaveDiff$ & $[\nFomcVintWaveLo,\, \nFomcVintWaveHi]$ & \nFomcVintWaveP \\
the rerun under the convention & the burn-in form under Section~\ref{sec:setup} & \nFomcVintRerunBlock & \nFomcVintRerunDraws & \nFomcVintRerunSeed & $\nFomcVintRerunDiff$ & $[\nFomcVintRerunLo,\, \nFomcVintRerunHi]$ & \nFomcVintRerunP \\
\bottomrule
\end{tabular}
\par\raggedright ACI is adaptive conformal inference. FOMC is the Federal Open Market Committee.
\end{table}

A burn-in form that also puts the boundary's own score in its score set covers \nExtHeavyAbaciFiledPost and \nExtTwoAbaciFiledPost on the two checks of Table~\ref{tab:stress}.

Under the in-sample reference, and with the threshold forms of the variants in place of their official implementations, a method that uses no announcement wins on \nLibPredHelpsFailed of the \nLibPredHelps announced breaks where the rule of thumb of Appendix~\ref{app:library} predicts a gain.

\FloatBarrier
\newpage

\end{document}